\documentclass[letterpaper]{article}
\usepackage[margin=1in,dvips]{geometry}
\usepackage{graphicx,psfrag,amsmath,amsthm,amssymb}
\usepackage[numbers,sort&compress]{natbib}
\usepackage{url,color,booktabs}
\usepackage[bookmarksnumbered=true,bookmarksopen=true]{hyperref}

\definecolor{DarkRed}{rgb}{0.545,0,0}
\hypersetup{
  breaklinks=true, colorlinks=true, linkcolor=black,
  citecolor=black, urlcolor=DarkRed % blue prints well in B&W printers
}

\def \ifempty#1{\def\temp{#1} \ifx\temp\empty }

\newcommand{\F}{\ensuremath{\mathbf{F}}}

\newcommand{\Q}{\ensuremath{\mathbf{Q}}}

\newcommand{\W}{\ensuremath{\mathbf{W}}}

\newcommand{\f}{\ensuremath{\mathbf{f}}}
\newcommand{\g}{\ensuremath{\mathbf{g}}}

\newcommand{\w}{\ensuremath{\mathbf{w}}}
\newcommand{\x}{\ensuremath{\mathbf{x}}}
\newcommand{\y}{\ensuremath{\mathbf{y}}}
\newcommand{\z}{\ensuremath{\mathbf{z}}}
\newcommand{\0}{\ensuremath{\mathbf{0}}}

\newcommand{\bmu}{\ensuremath{\boldsymbol{\mu}}}

\newcommand{\btheta}{\ensuremath{\boldsymbol{\theta}}}

\newcommand{\bTheta}{\ensuremath{\boldsymbol{\Theta}}}

\newcommand{\bbR}{\ensuremath{\mathbb{R}}}

\newcommand{\calL}{\ensuremath{\mathcal{L}}}

\newcommand{\calN}{\ensuremath{\mathcal{N}}}

\newcommand{\calR}{\ensuremath{\mathcal{R}}}
\newcommand{\calS}{\ensuremath{\mathcal{S}}}

\newcommand{\abs}[2][]{%
  \ifempty{#1} {\left\lvert#2\right\rvert} \else {#1\lvert#2#1\rvert} \fi}
\newcommand{\norm}[2][]{%
  \ifempty{#1} {\left\lVert#2\right\rVert} \else {#1\lVert#2#1\rVert} \fi}

\newcommand{\caja}[4][1]{{%
    \renewcommand{\arraystretch}{#1}%
    \begin{tabular}[#2]{@{}#3@{}}%
      #4%
    \end{tabular}%
    }}

{%
\begin{list}{#1}{
\vspace{-\topsep}
\vspace{-\partopsep}
\setlength{\itemindent}{0cm}
\setlength{\rightmargin}{0cm}
\setlength{\listparindent}{0cm}
\settowidth{\labelwidth}{#1}
\setlength{\leftmargin}{\labelwidth}
\addtolength{\leftmargin}{\labelsep}
\setlength{\itemsep}{0cm}
}%
}%
{%
\end{list}
\vspace{-\topsep}
\vspace{-\partopsep}
}

{\begin{enumerate}%
}%
{\end{enumerate}}

\theoremstyle{plain}% default
\newtheorem{thm}{Theorem}[section]

\newtheorem*{lemma*}{Lemma}

\newtheorem*{prop*}{Proposition}

\theoremstyle{definition}

\newtheorem*{defn*}{Definition}

\newtheorem*{exmp*}{Example}

\newtheorem*{conj*}{Conjecture}

\theoremstyle{remark}

\newtheorem*{rmk*}{Remark}

\graphicspath{{grf/},{grf-miguel/}}

\AtBeginDvi{}

\title{Sparse Oblique Decision Trees: \\ A Tool to Understand and Manipulate Neural Net Features}
\author{
  Suryabhan Singh Hada \hspace{5ex} Miguel \'A.\ Carreira-Perpi\~n\'an \hspace{5ex} Arman Zharmagambetov \\
  Dept.\ of Computer Science \& Engineering, University of California, Merced \\
  {\url{http://eecs.ucmerced.edu}}
}
\date{February 20, 2022}

\begin{document}

\maketitle

\begin{abstract}

  The widespread deployment of deep nets in practical applications has lead to a growing desire to understand how and why such black-box methods perform prediction. Much work has focused on understanding what part of the input pattern (an image, say) is responsible for a particular class being predicted, and how the input may be manipulated to predict a different class. We focus instead on understanding which of the internal features computed by the neural net are responsible for a particular class. We achieve this by mimicking part of the neural net with an oblique decision tree having sparse weight vectors at the decision nodes. Using the recently proposed Tree Alternating Optimization (TAO) algorithm, we are able to learn trees that are both highly accurate and interpretable. Such trees can faithfully mimic the part of the neural net they replaced, and hence they can provide insights into the deep net black box. Further, we show we can easily manipulate the neural net features in order to make the net predict, or not predict, a given class, thus showing that it is possible to carry out adversarial attacks at the level of the features. These insights and manipulations apply globally to the entire training and test set, not just at a local (single-instance) level. We demonstrate this robustly in the MNIST and ImageNet datasets with LeNet5 and VGG networks.

\end{abstract}

\section{Introduction}
\label{s:intro}

Deep neural nets are accurate black-box models. They are highly successful in terms of predictive performance (say, classifying an input image) but remarkably difficult to understand in terms of how exactly they come up with a prediction for an input. Both of these issues have been known to researchers and practitioners for many years, but it is in the 2010s that deep learning has achieved a wild, unexpected success that has attracted widespread attention beyond computer science. In only a few years, neural nets have become the workhorse model in a number of practical problems, in computer vision, speech and language processing, games, self-driving cars and other engineering applications; legal, financial and medical applications; and many others. Neural nets now underlie intelligent processing in desktops, cloud computing and IoT devices. Yet, the way neural nets are defined and optimized, and the sheer size and complexity of state-of-the-art deep nets, make them very hard to understand in explanatory terms. This is also true of other machine learning models, but deep nets sit at the far end in terms of opaqueness. Deep neural nets are generally not based on mechanistic models that involve physical entities in a causal way. They purely learn a correspondence between complex high-dimensional inputs and outputs by means of function approximation techniques based on using many adjustable building blocks (layers, neurons, weights and various nonlinear transformations). Such models can potentially approximate many possible correspondences with appropriate choices for these parameters, and finding a good choice is possible via a numerical optimization algorithm that minimizes a prediction loss over a large, labeled dataset. The resulting net can make highly accurate predictions for test inputs, but leaves many questions unanswered. We do not know what a given neuron or weight, or group of them, codes for at a level that a human can understand; or what would happen if we remove or alter a given set of neurons or weights; or what should we change in the input instance to change the prediction in a certain way; or what should we change in the trained net to correct a wrong prediction in some specific input instance. Further, for reasons not well understood, the class predicted by a deep net can be very sensitive to minute alterations of the input in ways that can be used adversarially.

Most of these questions are not new \citep{Guidot_18a,Rudin19a}, but they have become urgent due to the widespread deployment of deep nets in sensitive applications. Indeed, this is leading to law changes regarding the use of AI systems and data, such as the EU General Data Protection Regulation. Erring occasionally in a movie recommendation system is not serious, but crafting an adversarial attack so it systematically recommends or does not recommend certain movies is. Much more serious are errors or attacks in legal, financial or medical applications. An example of all these three at once is the processing of medical insurance claims, an area traditionally fraught with more or less legal attempts to affect the payment outcomes. The use of deep nets for automatic claim decision making introduces further opportunities for manipulation that are more creative and difficult to detect \citep{Finlay_19a}. More generally, there is a need to understand AI systems experimentally, and different perspectives (including but not limited to computer science) will likely be necessary \citep{Rahwan_19a}.

Our paper has two contributions that can improve our ability to explain and manipulate trained deep nets. Firstly, we propose sparse oblique decision trees as a tool to understand deep nets. Using decision trees is by itself not a new idea. What is new is the specific, novel type of tree we use, and how we apply it to a given deep net. Traditional tree learning algorithms typically construct trees where each decision node thresholds a single input feature (axis-aligned trees). Although such trees are considered among the most interpretable machine learning models, this is only true if the tree is relatively small; it is very hard to interpret an axis-aligned tree with thousands of nodes. More importantly, the axis-aligned decisions are ill-suited to handle datasets with many, correlated features. In practice, axis-aligned trees usually achieve too low an accuracy, and are wholly inadequate for high-dimensional complex inputs such as pixels of an image or neural net features. We capitalize on a recently proposed \emph{Tree Alternating Optimization (TAO)} algorithm which can learn far more accurate trees that remain small and very interpretable, because each decision node operates on a small, learnable subset of features.

Second, we apply the tree to an internal layer of the deep net, hence mimicking its remaining (classifier) layers, rather than attempting to mimic the entire deep net. This allows us to probe the relation between deep net features and classes. As a subproduct, inspection of the tree allows us to construct a new kind of adversarial attacks where we manipulate the deep net features via a mask to block a specific set of neurons. This gives us surprising control on what class the deep net will output. Among other possibilities, we can make it output the same, desired class for all dataset instances; or make it never output a given class; or make it misclassify certain pairs of classes.

Next, we review related work (section~\ref{s:related}) and the TAO algorithm (section~\ref{s:TAO}), describe how we use trees to understand and manipulate deep net features (section~\ref{s:tool}--\ref{s:tool:attacks}), and demonstrate this in MNIST and ImageNet data with LeNet5 and VGG16 deep neural nets (section~\ref{s:expts}). A short version of this paper appeared in \cite{Hada_21b}.

\section{Related work}
\label{s:related}

The last few years have seen much work in the area of interpreting or understanding, in some way, the internal workings of a trained neural network. We describe the most relevant work, organized in several categories.

\subsection{Feature inversion or activation maximization}

The idea here is to find what input feature vectors (e.g.\ what images, for a VGG16 network) produce a certain output under the neural network. This can be done for individual neurons, with the goal of understanding what ``concept'' a neuron may encode, or for an entire layer of neurons. Mathematically, this is essentially a problem of inverting the neural net function.

One way to do this is to formulate the inversion problem as an optimization: to minimize the Euclidean distance between the target outputs (at a neuron or layer of neurons) and the outputs generated by the input feature vector sought. In its basic form, this idea goes back decades \citep{KinderLinden90a,Jensen_99a}, and has been revisited by various papers recently. With images, naive application of this procedure will generate noisy or unrealistic images. Various approaches have been proposed to mitigate this, such as regularizing the optimization problem using total variation \citep{MahendVedald16a} or data-driven patch priors \citep{Wei_15a}, or learning these regularizers by training a deep neural network to generate images from the features \citep{DosovitBrox16a}.

Another way is to seek an input pattern that will maximize the activation (output) of a given neuron, analogously to seeking the ``receptive field'' of the neuron. Such pattern would represent the preferred stimulus to which that neuron responds, and might give a clue to what that neuron encodes. This is again an optimization problem over the input space of the network, which can be solved in various ways (e.g.\ \citep{Simony_14a}). Other works seek to provide multiple patterns rather than a single one, in order to find a better characterization of a given neuron \citep{Nguyen_16a,Nguyen_17a,HadaCarreir19a}. Again, this can be combined with regularizers or generative adversarial networks to obtain realistic images.

Finally, there also exist approaches that are not based on optimization \citep{Bau_17a,MuAndreas20a}.

\subsection{Local, instance-level explanations}

This line of work seeks to explain the neural net prediction for a given input instance (or group of instances). For example: what part of a given input image was mostly responsible for the network to classify it as a certain class? This is often referred to as a \emph{saliency map} of the image. An intuitive way to do this is via sensitivity analysis, such as computing or approximating the gradient of the output score with respect to the input image \citep{Simony_14a,ZeilerFergus14a}. With ReLU activation functions, which have a discontinuous gradient, this can produce artifacts \citep{Sundar_17a,Shrikum_17a}. Other approaches and variations exist \citep{Bach_15a,Montav_16a,Montav_18a,Shrikum_17a,Zhou_16c,Selvar_17a,FongVedald17a,Qi_20a,Sundar_17a}. Although saliency maps are visually appealing and can sometimes agree with human intuition, it is not clear how robust and consistent they are, and they can actually be misleading depending on the case \citep{FongVedald17a,Adebay_18a,Ghorban_19a,Rudin19a}. For example, the saliency map for a given image may be very similar even when it is computed for different target classes.

Another way to seek input features that are particularly important in predicting a given instance's class are \emph{Shapley values}, originally proposed to attribute reward among players of a cooperative game. Calculating Shapley values is NP-hard, so they are approximated in practice \citep{StrumbKononen14a,Datta_16a,MerricTaly20a,LundberLee17a}. The ability of Shapley values to provide explanations that are useful for humans has also been criticized \citep{Kumar_20a}.

Explanation by examples is another way to provide instance-level explanations, where we seek which instances in the training set (on which the neural net was trained) are most responsible for a given instance to be classified as a certain class. Naively, this would involve retraining the neural net without each particular instance, to see what the effect of that instance would have been. This is computationally very costly and is approximated in various ways \citep{KohLiang17a,Yeh_18a,Pruthi_20a}.

Finally, another line of work for local explanation is to replace the neural network function locally around the given instance with a simpler model that can be interpreted. One of these methods is LIME \citep{Ribeir_16a}, which describes a somewhat involved procedure to fit a sparse linear model using a sample of instances near the given instance. The nonzero coefficients in this linear model can be used to gauge the importance of input features in the prediction for the instance. This has been extended to decision rules \citep{Ribeir_18a} instead of a linear model. Some of these local explanation methods can be seen from a common point of view of explaining the network's output as a weighted sum of the input features \citep{LundberLee17a}. Other works \citep{Singh_19a,Zhang_19a} are based on constructing an agglomerative clustering tree over the neurons or input features. This is done by defining a similarity measure for the latter in terms of their ability to predict the local instance. The clustering tree provides a hierarchical arrangement of the input features and can be inspected by a human to look for groups of input features that are influential in the original instance's prediction. (Although the clustering tree is called a ``decision tree'' in \citep{Zhang_19a}, it is not a decision tree in the usual sense of classification or regression.) Because of the multiple approximations involved and the lack of a clear criterion of what the proxy model is supposed to explain, these approaches are somewhat ad-hoc.

\subsection{Global explanation via an interpretable mimic of the neural network}

The goal of these types of methods is to mimic the entire deep neural net via a more interpretable model such as decision trees or decision rules. This then provides a \emph{global explanation}, applicable to any input instance, unlike the previous, local explanation methods, which are only valid near a given instance.

The topic of extracting sets of rules from neural nets was actively researched in the 1990s \citep{Andrew_95a,Mccorm_13a,Guidot_18a}. Two basic approaches were used: in rule extraction as search \citep{Fu94a,TowellShavlik93a}, a specialized heuristic search over possible rules was based on the neurons' connectivity pattern, but this assumed binary activations and did not scale beyond small nets. In rule extraction as learning, or the teacher-student approach \citep{CravenShavlik94a,CravenShavlik96a,Doming98a}, one trains a decision tree to mimic the neural net by using the latter's predictions on the training set (possibly augmented with random instances). Crucially, the success of this idea relies on the faithfulness of the mimic. Although some of these papers \citep{TowellShavlik93a,CravenShavlik96a,Baesen_03a} claimed that the extracted rules closely approximate the original neural net, this was based on very small problems and networks (single-layer). In such problems, training an axis-aligned decision tree directly was not far in accuracy from the neural net in the first place, and could produce a relatively small tree and a correspondingly small set of rules.

The fundamental problem with this is that traditional algorithms to learn decision trees or decision rules, such as CART \citep{Breiman_84a} or C4.5 \citep{Quinlan93a}, based on axis-aligned trees, are unable to learn accurate enough trees to be useful mimics of a neural net except in very small, low-dimensional problems. They fall far short of handling the large, deep neural nets that are used in current computer vision applications, for example.

\subsection{Our work in context}

Our work belongs to the category of global explanation via an interpretable mimic that is a decision tree, with two important differences. First, we use a special type of decision tree, a \emph{sparse oblique tree}. This can be trained to be much more accurate than axis-aligned, CART-type trees, which makes it more likely that one can obtain a faithful mimic. At the same time, the sparse oblique tree remains interpretable. This is because the tree size is far smaller than that of an axis-aligned tree (in depth and number of nodes), and because it uses relatively few features in each decision node. As we show in our experiments, inspection (manual or automatic) of the nonzero weights in the decision nodes leads to important insights about the tree (and about the neural net), and shows how to manipulate the neural net features to alter the classification result in a desired way. (Further in terms of interpretability, counterfactual explanations can be solved exactly and efficiently for oblique trees \cite{CarreirHada21a,HadaCarreir21b}, although we do not use this here.)

A second difference is that we do not aim at replacing the entire neural net, but at replacing the classifier portion of a deep net (globally over all instances). \emph{This allows us to study the relation between the deep net features (neuron activations at a certain internal layer) and the output classes}---note that those features were specifically learned by the neural net during training with the goal of predicting the classes optimally. This is unlike much of the work cited earlier, which studies the relation between input features (e.g.\ pixels) and neuron activations at a certain layer. (In \cite{ZharmagCarreir21b}, we also train a sparse oblique tree on features obtained from multiple pretrained deep neural networks, but the goal there is not to interpret a neural net, but to achieve a model with higher accuracy than the pretrained networks.)

\section{Learning sparse oblique trees with the Tree Alternating Optimization (TAO) algorithm}
\label{s:TAO}

We briefly explain the Tree Alternating Optimization (TAO) algorithm; the original references give more details \citep{Carreir22a,CarreirTavall18a}. Among other types of trees, TAO can learn \emph{sparse oblique trees}. These have a constant label at each leaf and a linear decision function at each decision node but, crucially, the decision function only uses a typically small, learned subset of features (see fig.~\ref{f:VGG16-tree1}). TAO achieves this by optimizing an objective function which is the sum of the classification loss and an $\ell_1$ penalty (with hyperparameter $\lambda \ge 0$) on the weight vectors of the decision nodes (similar to a LASSO \citep{Hastie_15a} but on each decision node). Each TAO iteration decreases this objective and consists of optimizing over groups of non-descendant nodes (such as all the nodes at the same depth), where the objective can be shown to separate over the nodes in the group. The optimization over each node can be shown to be equivalent to a simpler, ``reduced'' problem taking the form of a majority classifier at a leaf, and a binary classifier at a decision node. In the latter, each instance reaching the node is assigned a ``pseudolabel'' indicating the child that gives the lowest loss under the current tree. TAO uses a fixed tree structure while iterating, which is automatically pruned because subtrees may eventually receive training instances of the same class, or not receive instances at all; this is heavily promoted by the $\ell_1$ penalty on the decision nodes, which can drive all weights in a decision node to zero, thus making it redundant.

In more detail, consider a tree $T$ of a given structure (usually, complete of depth $\Delta$) with nodes in a set $\calN$ and parameters $\bTheta = \{\btheta_i\}_{i \in \calN}$. For a decision node $i$, $\btheta_i$ consists of the weights and bias of the hyperplane. For a leaf $i$, $\theta_i \in \{1,\dots,K\}$ is a class label. Now,  to train $T$ on a dataset $\{\x_n,y_n\}^N_{n=1} \subset \bbR^D \times \{1,\dots,K\}$ of input instances and their labels, we optimize the following objective function:
\begin{equation}
\label{e:objfcn}
\textcolor{black} { \min_{\bTheta} E(\bTheta) = \sum_{n=1}^N \calL(y_n, T(\x_n;\bTheta)) + \lambda \sum_{i \in \calN} \phi_i(\btheta_i)}
\end{equation}
\textcolor{black}{where $\calL(.,.)$ is the cross-entropy loss, $\phi_i(\btheta_i)$ is a regularization term with hyperparameter $\lambda \ge 0$ (we will use $\ell_1$ regularization on the weight vectors of the decision nodes). TAO relies on two theorems: a separability condition, and a reduced problem over each node (see details and proofs in \cite{Carreir22a,CarreirTavall18a}). Here, we describe them briefly.
\begin{thm}[separability condition]
\label{th:separability}
  Consider any pair of nodes $i$ and $j$. Assume the parameters of all other nodes ($\bTheta_{\text{rest}}$) are fixed. If nodes $i$ and $j$ are not descendants of each other, then $E(\bTheta)$ can be rewritten as:
  \begin{equation}
    E(\bTheta) = E(\btheta_i) + E(\btheta_j) + E(\bTheta_{\text{rest}}).
  \end{equation}
\end{thm}
In other words, the separability condition states that any set of non-descendant nodes of a tree can be optimized independently. Note that $E(\bTheta_{\text{rest}})$ can be treated as a constant since we fix $\bTheta_{\text{rest}}$.}
\textcolor{black}{
\begin{thm}[reduced problem]
For a single node $i$, optimizing over its parameters simplifies to a well-defined reduced problem over the instances that currently reach node $i$ (the reduced set $\calR_i \subset \{1,\dots,N\}$).
\end{thm}
\begin{itemize}
\item For a decision node $i$, the reduced problem can be written as a 0/1 loss binary classification problem, where the target class refers to the left or right subtree.  For this we assign a ``pseudolabel" ($\overline{y} \in \{\texttt{left,right}\}$) indicating the child that gives the lowest loss under the current tree. Thus, the reduced problem takes the form:
  \begin{equation}
    E(\btheta_i) = \sum_{n \in \calR_i} \smash{\overline{L}}_n (\overline{y}_n,(f(\x_i);\btheta_i)) + \lambda \phi_i(\btheta_i)
    \label{e:binary-class}
  \end{equation}
where $\overline{L}_n$ is the 0/1 loss and $f_i\mathpunct{:}\ \bbR^D \rightarrow \{\texttt{left,right}\}$ is the decision function at node $i$ with parameters $\btheta_i$ (which, for an oblique tree, is defined by a linear classifier). This is an NP-hard problem, but it can be approximated by using a convex surrogate loss.  In this work, we use $\ell_1$ regularized logistic regression ($\phi_i = \norm{\cdot}_1$) and optimize it using LIBLINEAR \citep{Fan_08a}.
\item For a leaf node $i$, the reduced problem can be shown to have an exact solution, given by setting the leaf label $\theta_i$ to the majority class among all the instances in its reduced set $\calR_i$.
\end{itemize}}
The above theorems mean that we can monotonically reduce the objective function~\eqref{e:objfcn} by cycling over the nodes in the tree in some order and optimizing each node's parameters by solving its reduced problem. We can optimize any subset of non-descendant nodes in parallel, e.g.\ all nodes at the same depth. As for the tree structure (and initial parameters), this will depend on the dataset, but we find that using a complete tree of large enough depth $\Delta$ and random (Gaussian 0,1) initial parameters often works well.

In a sense, TAO operates similarly to how a neural net (or other machine learning models) are trained: by fixing the model structure and optimizing a desired loss function iteratively. However, instead of using gradients (which are not available for a decision tree), it uses alternating optimization. Model selection over the tree structure happens automatically by using a large enough tree structure and letting TAO prune it via the $\ell_1$ penalty (as has also been done to prune weights and neurons in neural nets, e.g.\ \citep{CarreirIdelbay18a}). \textcolor{black}{We describe the entire tree training process in fig.~\ref{f:pseudocode}. As described in \cite{Carreir22a,CarreirTavall18a}, the computational complexity of one TAO iteration is upper bounded by the depth of the tree times the cost of solving a logistic regression problem on the entire dataset. This is because the reduced sets of all the nodes at the same depth form a partition of the entire training set.}

We can control the tradeoff between accuracy and sparsity (in terms of the size of the tree and the number of nonzero weights at the nodes), and hence control the amount of feature selection and interpretability, similarly as with the LASSO regularization path \citep{Hastie_15a}. We simply trace a family of trees of decreasing accuracy and increasing sparsity as the hyperparameter $\lambda$ goes from 0 to $\infty$.

TAO considerably improves \citep{Zharmag_21b} over the traditional, widely used algorithms (such as CART \citep{Breiman_84a} or C4.5 \citep{Quinlan93a}) that are based on greedy recursive partitioning based on a purity measure. This is because these algorithms have no useful guarantees concerning the classification loss to start with, and are only moderately effective with axis-aligned trees. The same approach can be used to train oblique trees \cite{Breiman_84a,Murthy_94a}, but it results in large, nonsparse and highly suboptimal trees that often do not improve over axis-aligned ones. These are the fundamental reasons why axis-aligned trees are the only type of tree that is widespread in practice, at least until now. TAO also improves tree ensembles (forests) considerably: if using TAO instead of a CART-type algorithm (as done in random forests \citep{Breiman01a} and XGBoost \citep{ChenGuestr16a}), the resulting forest contains fewer, shallower trees but is consistently more accurate, whether using bagging or boosting to ensemble the trees \citep{CarreirZharmag20a,ZharmagCarreir20a,Zharmag_21a,Zharmag_21c}. Finally, TAO can also be used to train novel forms of tree models \citep{ZharmagCarreir21a,Zharmag_21d}.

\begin{figure}[t]
  \centering
  \setlength{\fboxsep}{1ex}
  \framebox{%
    \begin{minipage}[c]{0.80\linewidth}
      \begin{tabbing}
        n \= n \= n \= n \= n \= \kill
        \underline{\textbf{input}} \caja{t}{l}{training set $\{(\x_n,y_n)\}^N_{n=1} \subset \bbR^D \times \{1,\dots,K\}$ \\ initial tree $T$} \\
        \underline{\textbf{repeat}} \+ \\
        \underline{\textbf{for}} $i \in$ nodes of $T$, visited in reverse BFS \+ \\
        \underline{\textbf{if}} $i$ is a leaf \underline{\textbf{then}} \+   \\
        $\theta_i \gets$ majority-class label in the reduced set $\calR_i$ \+ \- \- \\
        \underline{\textbf{else}} \+ \\
        generate pseudolabels $\overline{y}_n$ for each instance $n \in \calR_i$ \\
        $\btheta_i \gets$ minimizer of the reduced problem (eq.~\eqref{e:binary-class}) \- \- \- \\
        \underline{\textbf{until}} stop \\
        postprocess $T$: remove dead branches \& pure subtrees \\
        \underline{\textbf{return}} $T$
      \end{tabbing}
    \end{minipage}
  }
  \caption{\textcolor{black}{Pseudocode for the tree alternating optimization (TAO) algorithm~\citep{Carreir22a,CarreirTavall18a}. Visiting each node in reverse breadth-first search (BFS) order means scanning depths from depth($T$) down to 1, and at each depth processing (in parallel, if so desired) all nodes at that depth. ``stop'' occurs when either the objective function decreases less than a set tolerance or the number of iterations reaches a set limit.}}
  \label{f:pseudocode}
\end{figure}

\section{Sparse oblique trees: a microscope to observe a deep neural net}
\label{s:tool}

\begin{figure}[t]
  \centering
  \scriptsize
  \begin{tabular}{@{}c@{}}
    \psfrag{x}[t][t]{\caja{c}{c}{\x \\ $D$ input \\ features}}
    \psfrag{y}[b][b]{\caja{c}{c}{\y \\ $K$ output \\ classes}}
    \psfrag{f}[][B]{\colorbox{white}{\caja{c}{c}{$\y = \f(\x)$, \\ entire neural net}}}
    \psfrag{F}[][B]{\colorbox{white}{\caja{c}{c}{$\z = \F(\x)$, \\ $F$ neural net features}}}
    \psfrag{g}[][B]{\colorbox{white}{\caja{c}{c}{$\y = \g(\z)$, \\ classifier part \\ (mimicked \\ by tree)}}}
    \includegraphics*[width=\linewidth]{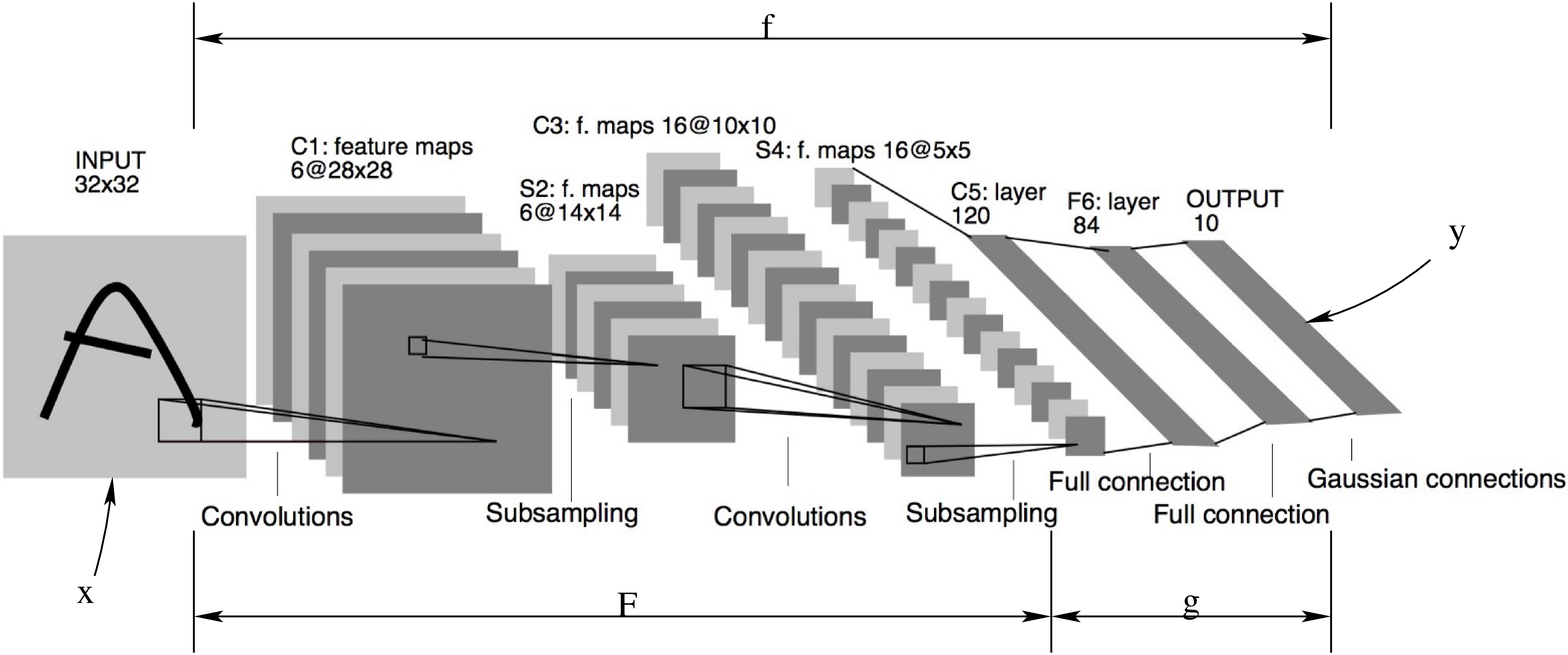}
  \end{tabular}
  \caption{Mimicking part of a neural net with a decision tree. The figure shows the neural net $\y = \f(\x) = \g(\F(\x))$, considered as the composition of a feature extraction part $\z = \F(\x)$ and a classifier part $\y = \g(\z)$. For example, for the LeNet5 neural net of \citep{Lecun_98a} in the diagram, this corresponds to the first 4 layers (convolutional and subsampling) followed by the last 2, fully-connected layers, respectively. The ``neural net feature'' vector \z\ consists of the activations (outputs) of $F$ neurons, and can be considered as features extracted by the neural net from the original features \x\ (pixel values, for LeNet5). We use a sparse oblique tree to mimic the classifier part $\y = \g(\z)$, by training the tree using as input the neural net features \z\ and as output the corresponding ground-truth labels.}
  \label{f:tree-mimic}
\end{figure}

Our overall approach is as follows (see fig.~\ref{f:tree-mimic}). Assume we have a trained deep net $\y = \f(\x)$ for classification of an input instance $\x \in \bbR^D$ into $K$ classes, so \y\ is a vector of $K$ softmax values encoding (an approximation to) the class distribution given \x. We write the net $\f(\x) = \g(\F(\x))$ as the composition of a feature-extraction layer \F, so $\z = \F(\x) \in \bbR^F$ are the deep net features (neuron outputs at that layer), and a classifier layer $\y = \g(\z)$ consisting of the rest of the net%
\footnote{This is an operational definition, since the original net was trained without an explicit construction of feature extraction and classifier, and indeed such distinction is blurred in some architectures such as ResNets \citep{He_16a}. Other architectures, such as LeNet5 \citep{Lecun_98a} and VGG \citep{SimonyZisser15a} do have a clear separation into feature extraction (based on convolutional, pooling and subsampling layers), and classification (fully-connected MLP).}.
This includes as particular cases of features the raw inputs \x\ (where \F\ is the identity) and the class label or softmax outputs (where \g\ is the identity), but we will usually be more interested in features at an intermediate layer. Each neuron at that layer can be considered as a feature detector which encodes some property or concept of the input pattern \x, which may be useful (in combination with other neurons' concepts) to provide information for or against one or more classes.

Assume we have a dataset (usually the one used to train the net) $\{(\x_n,y_n)\}^N_{n=1} \subset \bbR^D \times \{1,\dots,K\}$ of input instances and their labels. Then:
\begin{enumerate}
\item Train a sparse oblique tree $y = T(\z)$ with TAO on the training set $\{(\F(\x_n),y_n)\}^N_{n=1} \subset \bbR^F \times \{1,\dots,K\}$. Explore the interpretability-accuracy tradeoff over a useful a range of the sparsity hyperparameter $\lambda \in [0,\infty)$ and pick a final tree. Usually this will be a tree with close to highest validation accuracy and as sparse as possible.
\item Inspect the tree to find interesting patterns about the deep net.
\end{enumerate}
Our goal is to achieve a tree that both mimics well the deep net and is as simple as possible. We achieve this by training the tree on the same training set as the net (using the latter's features but the ground-truth labels%
\footnote{This is equivalent to using the deep net predictions as labels (teacher-student approach), because our deep nets achieve nearly zero training error.}).

Step 2 is purposely vague. There is probably a wealth of information in the tree regarding the features' meaning and effect on the classification, both at the level of a specific input instance or more globally. In this paper we focus on one specific pattern described in the next section.

\section{Manipulating the features of a deep net to alter its classification behavior}
\label{s:tool:attacks}

\begin{figure}[p]
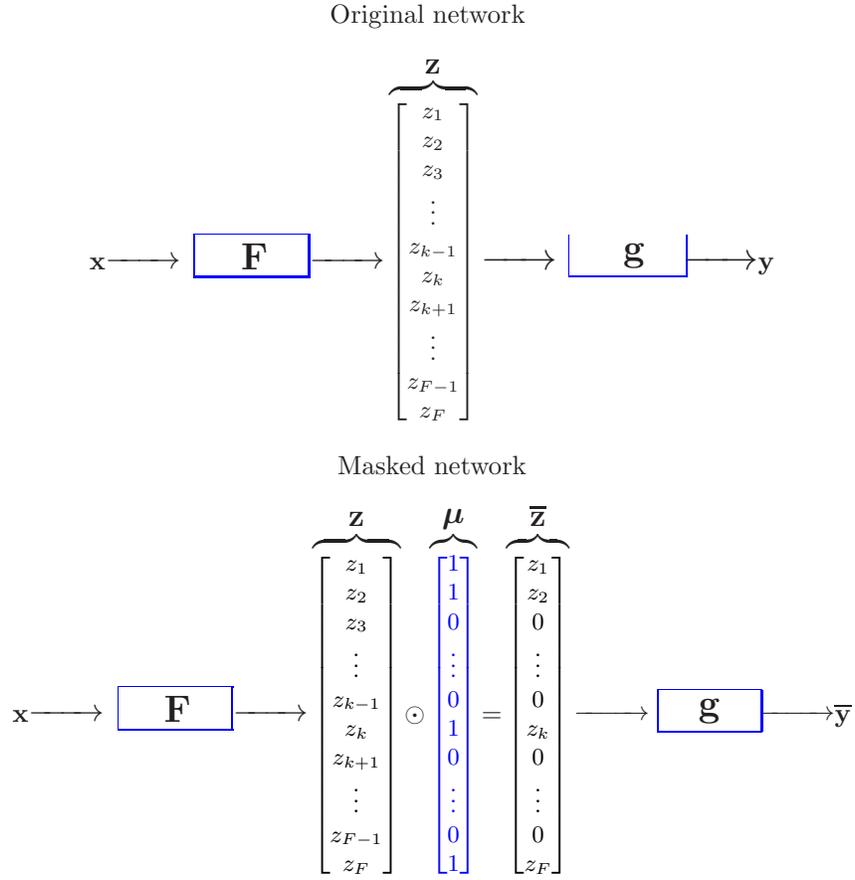

  \centering
  {Original network }
  \begin{align*}
  \large \x \parbox{1cm}{\rightarrowfill} \hspace{1ex} \fcolorbox{blue}{white}{\hspace{2ex} \Large \F \hspace{2ex}} \parbox{1cm}{\rightarrowfill} &
  \overbrace{\small   
   \begin{bmatrix}
            z_{1} \\
           z_{2} \\
           z_{3} \\
           \vdots \\
           z_{k-1} \\
           z_{k} \\
           z_{k+1} \\
           \vdots \\
           z_{F-1} \\
           z_{F}
         \end{bmatrix}}^\text{{\large $\z$}}\Large{ \parbox{1cm}{\rightarrowfill}  \raisebox{0.5ex}{  \fcolorbox{blue}{white} {\hspace{2ex} \Large \g \hspace{2ex}} } \hspace{-1ex}\parbox{1cm}{\rightarrowfill} \large \y}
  \end{align*}

{Masked network}
  \begin{align*}
   \large \x \parbox{1cm}{\rightarrowfill} \hspace{1ex} \fcolorbox{blue}{white}{\hspace{2ex} \Large \F \hspace{2ex}} \parbox{1cm}{\rightarrowfill} &
 \overbrace{\small   
  \textcolor{black}{ \begin{bmatrix}
            z_{1} \\
           z_{2} \\
           z_{3} \\
           \vdots \\
           z_{k-1} \\
           z_{k} \\
           z_{k+1} \\
           \vdots \\
           z_{F-1} \\
           z_{F}
         \end{bmatrix}} }^\text{\large $\z$}   \odot     
\overbrace{\small \textcolor{blue}{ \begin{bmatrix}
            1 \\
           1 \\
           0 \\
           \vdots \\
           0 \\
           1\\
           0 \\
           \vdots \\
           0 \\
           1
         \end{bmatrix}}}^\text{{\large $\bmu$}}                
          {=} 
          \overbrace{\small \textcolor{black}{ \begin{bmatrix}
            z_{1} \\
           z_{2} \\
           0 \\
           \vdots \\
           0 \\
           z_{k} \\
           0 \\
           \vdots \\
           0 \\
           z_{F}
         \end{bmatrix}}}^\text{{\large $\overline{\z}$}}     
          \parbox{1cm}{\rightarrowfill}  \raisebox{0.5ex}{  \fcolorbox{blue}{white} {\hspace{2ex} \Large \g \hspace{2ex}} } \hspace{-1ex}\parbox{1cm}{\rightarrowfill} \large \overline{\y}
  \end{align*}
  \caption{\textcolor{black}{\emph{Top:} original network. \emph{Bottom:} masking operation in the network. The symbols' meaning is as follows: input $\x$, feature extraction part of the network $\F$, original features $\z$, binary mask created using the tree $\bmu$, modified features $\overline{\z}$, classifier part of the network \g, original output \y, and modified output $\overline{\y}$ .}}
  \label{f:masking-oper}
\end{figure}

Our overall objective is to manipulate the value of the deep net features $\z \in \bbR^F$ to alter in a controlled way the class predicted by the net. We will not alter the weights of the net, i.e., \F\ and \g\ remain the same. We just alter \z\ into a masked $\overline{\z} = \bmu(\z) = \bmu^{\times} \odot \z + \bmu^{+}$ via a \emph{multiplicative and an additive mask} $\smash{\bmu^{\times},\bmu^{+} \in \bbR^F}$, respectively (where ``$\odot$'' means elementwise multiplication). Specifically, we have:
\begin{align}
 \label{e:orig-mask}
 \text{Original net: } & \y = \f(\x) = \g(\F(\x)) \\
 \text{Original features: } & \z = \F(\x) \\
 \text{Masked net: } & \overline{\y} = \overline{\f}(\x) = \g(\bmu(\F(\x))) \\
 \text{Masked features: } & \overline{\z} = \bmu(\F(\x)) = \bmu(\z).
\end{align}
\textcolor{black}{We demonstrate the masking operation in fig.~\ref{f:masking-oper}. }In the simplest, most intuitive version of the mask, we just need a binary multiplicative mask $\overline{\z} = \bmu^{\times} \odot \z$ where $\smash{\bmu^{\times} \in \{0,1\}^F}$. Using an additive mask and real-valued masks makes the manipulation's effect more robust and harder to detect.

We will construct a mask by inspecting the tree, specifically by observing the weight of each feature in each decision node of the tree. \emph{By selectively zeroing some features we can guarantee that any instance will follow a specific child in a given node and hence direct instances as desired towards a target leaf}. Under some assumptions, we will be able to guarantee a desired effect if using the tree, i.e., in the classifier $y = T(\bmu(\F(\x)))$. Then we will apply the mask to the deep net as $\overline{\y} = \g(\bmu(\F(\x)))$. While we cannot guarantee anything in the masked net, \emph{we can reasonably expect similar results if the tree is a good mimic of the classifier \g, and indeed our experiments show that the masked net behaves like the masked tree most of the times}.

At a decision node $i$, the decision rule is ``if $\w^T_i \x + b_i \ge 0$ then go to right child, else go to left child'', where $\w_i \in \bbR^F$ is the weight vector and $b_i \in \bbR$ the bias. We will assume the following (throughout we use elementwise notation as needed, as in ``$\z \ge \0$''):
\begin{itemize}
\item The deep net features are nonnegative: $\z = \F(\x) \ge \0$. This is true for ReLUs, which are used in most deep nets at present.
\item The bias at each decision node $i$ of the tree is zero: $b_i = 0$. This holds very well in the trees we trained, specifically $\abs{b_i} \ll \norm{\w_i}$ at each decision node $i$.
\end{itemize}
If these assumptions do not hold, it is still possible to design masks that work reliably in some cases. We mention some of them but generally leave such details out of this paper.

\subsection{Diverting all instances to one child}

We give a basic procedure ``$\bmu^{\times},\bmu^{+} \gets$ \textsc{Node-Mask}$(i,c)$'' that underlies our mask construction. Assume an instance \z\ reaches a decision node $i$ in the tree. \textsc{Node-Mask} produces a mask that guarantees that $\overline{\z} = \bmu^{\times} \odot \z + \bmu^{+}$ goes left (right) if $c =$ left (right) child, for any instance \z\ in the training set. Essentially, \textsc{Node-Mask} allows us to ``cut'' a subtree, so by applying it as needed we can cut subtrees to leave only a certain path in the tree for any instance to follow and hence effect a desired classification.

\textsc{Node-Mask} works as follows. Call the weight vector \w\ and bias $b$ (assumed zero anyway) at node $i$. Write them w.l.o.g.\ as $\w = (\w_0\ \w_-\ \w_+)$ and $\z = (\z_0\ \z_-\ \z_+)$ where $\w_0 = \0$, $\w_- < \0$ and $\w_+ > \0$ contain the zero, negative and positive weights in \w, and $\z \ge \0$ is reordered according to that. Call $\calS_0$, $\calS_-$ and $\calS_+$ the corresponding sets of indices in \w. Then $\w^T \z + b = \w^T_- \z_- + \w^T_+ \z_+$ with $\smash{\w^T_-} \z_- \le 0$ and $\smash{\w^T}_+ \z_+ \ge 0$. So if $\z_- = \0$ then $\smash{\w^T} \z + b \ge 0$ and \z\ would go to the right child and if $\z_+ = \0$ then $\smash{\w^T \z} + b \le 0$ and \z\ would go to the left child. Hence, our masks are as follows: to go left, $\bmu^{\times} \in \{0,1\}^F$ is a binary vector containing ones at $\calS_-$, zeros at $\calS_+$ and $\ast$ (meaning any value) at $\calS_0$; and $\bmu^{+} \ge \0$ is a vector containing small positive values at $\calS_-$ and zero elsewhere. To go right, exchange ``$-$'' and ``$+$'' in the procedure. The additive mask $\bmu^{+}$ is necessary only if the features in $\calS_-$ all happen to be zero (which would produce $\smash{\w^T \overline{\z}} + b = 0$ and be on the decision boundary). This is unlikely to happen unless $\calS_-$ contains very few features, but still the additive mask is useful to push $\w^T \overline{\z}$ away from the boundary and hence make it more likely that the masked deep net will perform as desired%
\footnote{In fact, just setting $\z = \bmu^+$ (i.e., replacing the features with the additive mask without even using the multiplicative mask) would work in the tree. This boils down to replacing any incoming feature vector with a fixed feature vector of known classification under \g. But this makes the attack very obvious: the softmax values outputted by the net are the same for every instance.}.

\subsection{Masks}

We now show how to construct masks that effect a certain class outcome. For each case, we state the desired goal and the corresponding mask. In the manipulations below we may use \textsc{Node-Mask} repeatedly over several nodes to construct the mask (which is applied to the feature vector and hence applies globally to each node). In that case, we will only use the multiplicative mask produced by \textsc{Node-Mask} at each node, and create the additive mask at the end given the final multiplicative mask.
\begin{itemize}
\item \textsc{All class $k_1$ to class $k_2$}: \emph{let $k_1 \neq k_2 \in \{1,\dots,K\}$. For any instance \x\ originally classified as $k_1$, classify it as $k_2$. For any other instance, do not alter its classification}. This case only works if the classes $k_1$ and $k_2$ are leaf siblings (have the same parent). Class $k_2$ may be represented by multiple leaves since we only need to deal with one of them (the sibling of $k_1$). \emph{Mask}: simply apply \textsc{Node-Mask} to the parent of the leaves of $k_1$ and $k_2$.
\item \textsc{None to class $k$}: \emph{let $k \in \{1,\dots,K\}$. For any instance \x\ originally classified as $k$, classify it as any other class. For any other instance, do not alter its classification}. \emph{Mask}: simply apply \textsc{Node-Mask} to the parent of each leaf of $k$ and combine the resulting multiplicative masks as extended-AND (defined below). Finally, add the additive mask. \\
  Strictly speaking, we can guarantee that class-$k$ instances are classified as some other class, but not that we do not alter the classification of other instances. This is because the features that are masked out may appear in other nodes and possibly affect the path of an instance. However, with our deep nets the number of features masked out is very small and the mask works well. If the features selected in a node only appear in that node, thein their effect is purely local, of course.
\item \textsc{All to class $k$}: \emph{let $k \in \{1,\dots,K\}$. Classify all instances \x\ as class $k$}. \emph{Mask}: find the path from the root to the leaf of class $k$. At each node $i$ in the path, apply \textsc{Node-Mask} (to divert instances along the path) and keep the multiplicative mask only. The final multiplicative mask, elementwise, has a 0 where any of the node masks has a 0, a 1 where all node masks have no 0s but at least one 1, and $\ast$ elsewhere. This masks out all the ``undesired'' features that might divert us from the path. Equivalently, this is the logical extended-AND of all the multiplicative masks along the path (where we extend AND to mean AND$(\ast,0) = 0$, AND$(\ast,1) = 1$ and AND$(\ast,\ast) = \ast$). \\
  This would not work if the multiplicative mask is zero at all features, but this is unlikely if the nodes have sparse weight vectors, as happens in our experiments.
  This also works if class $k$ is represented as multiple leaves. We simply take the union of the masks over each leaf.
\item \textsc{None to a subset of classes}: we can apply this in some cases depending on the tree. It simply follows by applying \textsc{Node-Mask} to a given decision node. The classes in the subtree that is cut out cannot be reached by any instance. Applying this to multiple nodes and creating the multiplicative mask by extended-AND removes the corresponding subtrees and their classes.
\end{itemize}

\subsection{Hiding the adversarial attack}

Our motivation for manipulating the neural net features was to illustrate how the sparse oblique trees are able to gain information about how the network works internally. However, such manipulations can also be seen as adversarial attacks at the feature level (which may or may not be practically feasible). Applying the attack is very simple, as it needs no optimization (which is the case with many pixel-level attacks).

We can also make the attack less obvious. In practice with our trees (which have sparse weight vectors), the above masks only require setting to 0 or 1, always in the same place, a small number of features ($\approx$ 10--40) out of the total (hundreds or thousands), so they would be easily detectable to an observer of either the masked features or the softmax values at the output. We can easily randomize the above masks (and make them continuous rather than binary) so that they still work as intended but vary for each instance. First, the wildcard indices $\ast$ in the multiplicative masks can take any real value (positive, negative or zero). Second, the additive mask can take any positive values as long as they are small enough. Third, there typically is a subset of features which are not used in any node of the tree, so they can also take any real value.

Changing some additional features may of course have some unintended effects in the deep net. However, this can be reduced by applying the above randomization to a small, itself random number of features in the mask. Also, since the deep net features are the output of a ReLU, some of them are 0 to start with, so the mask has no effect there anyway.

Some of our manipulations are less detectable than others by their nature. Two classes that are siblings (such as `4' and `9' in the MNIST tree; fig.~\ref{f:LeNet5-tree}) are likely more similar than if they are far apart in the tree, so misclassifying them (as one of our masks does) would not raise much suspicion.

\section{Experiments}
\label{s:expts}

We have evaluated our trees and masks thoroughly on two deep nets:
\begin{itemize}
\item VGG16 \citep{SimonyZisser15a} in a subset of 16 classes of ImageNet \citep{Deng_09a}, for which we select the $F =$ 8\,192 neurons from its last convolutional layer. Table~\ref{t:VGG16} gives the network architecture.
\item LeNet5 in MNIST on 10 digit classes \citep{Lecun_98a}, for which we select the $F =$ 800 neurons at layer conv2 as features. Table~\ref{t:LeNet5} gives the network architecture.
\end{itemize}
For both of them, we can train trees that accurately mimic the deep net classifier \g. The trees give remarkable insight in the relation of deep net features to classes and allow us to construct masks that indeed work as intended in the deep net for most instances. We describe this in detail next.

\subsection{Results on VGG16 (subset of ImageNet dataset)}
\label{s:VGG}

\begin{table}[t]
  \centering
  \begin{tabular}[c]{@{}cc@{}}
    \hline 
    Label id & Class \\
    \hline 
    0 &	goldfish \\
    1 & bald eagle\\
    2 &	goose\\
    3 &	killer whale\\
    4 &	Siberian husky\\
    5 &	white wolf\\
    6 &	tiger cat \\
    7 &	lion \\
    8 &	airliner \\
    9 &	container ship \\
    A &	fire engine \\
    B &	school bus\\
    C &	speedboat \\
    D &	sports car \\
    E &	warplane \\
    F &	coral reef \\
    \hline 
  \end{tabular}
  \caption{Classes in our ImageNet subset and their id (for reference in other figures).}
  \label{t:ImageNet-subset-classes}
\end{table}

\begin{figure}[p]
  \centering
  \begin{tabular}{@{}c@{}}
    \psfrag{error}[][]{Error (VGG16)}
    \psfrag{c}[][B]{$\log_{10}\lambda$}
    \psfrag{net(train)}[l][B][0.9]{\hspace{-2ex}net (train)}
    \psfrag{net(test)}[c][][0.9]{net (test)}
    \psfrag{tree(train)}[l][][0.9]{\hspace{-3ex}tree (train)}
    \psfrag{tree(test)}[l][][0.9]{\hspace{-3ex}tree (test)}
    \psfrag{Selected tree}[c][][0.9]{\hspace{9ex}Selected tree}
    \includegraphics*[width=.54\linewidth]{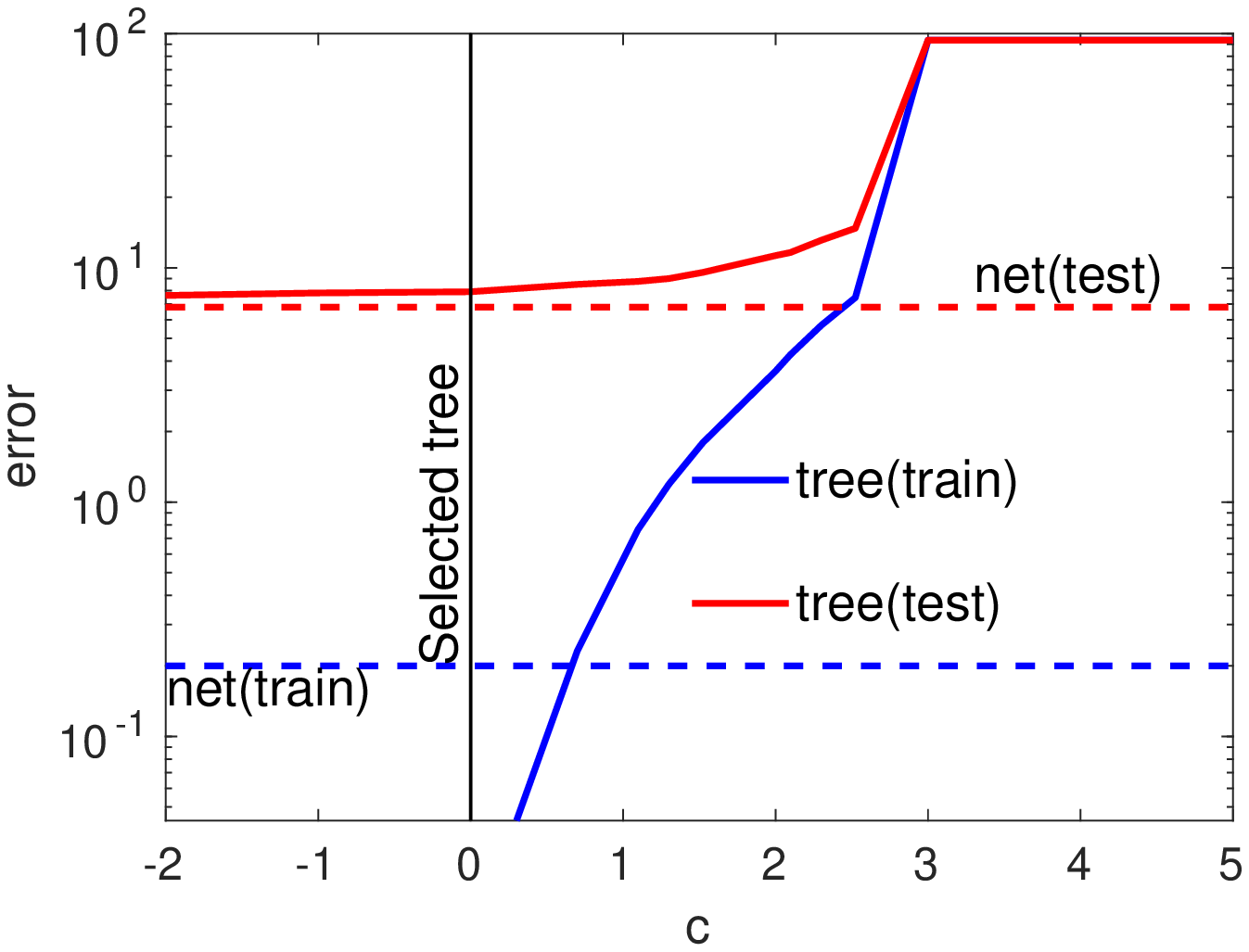} \\[1ex]
    \psfrag{nodes}[][]{}
    \psfrag{\# nodes}[c][][0.9]{\hspace{4ex}\# nodes}
    \psfrag{depth}[c][][0.9]{\hspace{4ex}depth}
    \psfrag{Selected tree}[c][][0.9]{\hspace{4ex}Selected tree}
    \psfrag{depths}[][]{}
    \psfrag{c}[][B]{$\log_{10}\lambda$}
    \hspace{5ex}\includegraphics*[width=.57\linewidth]{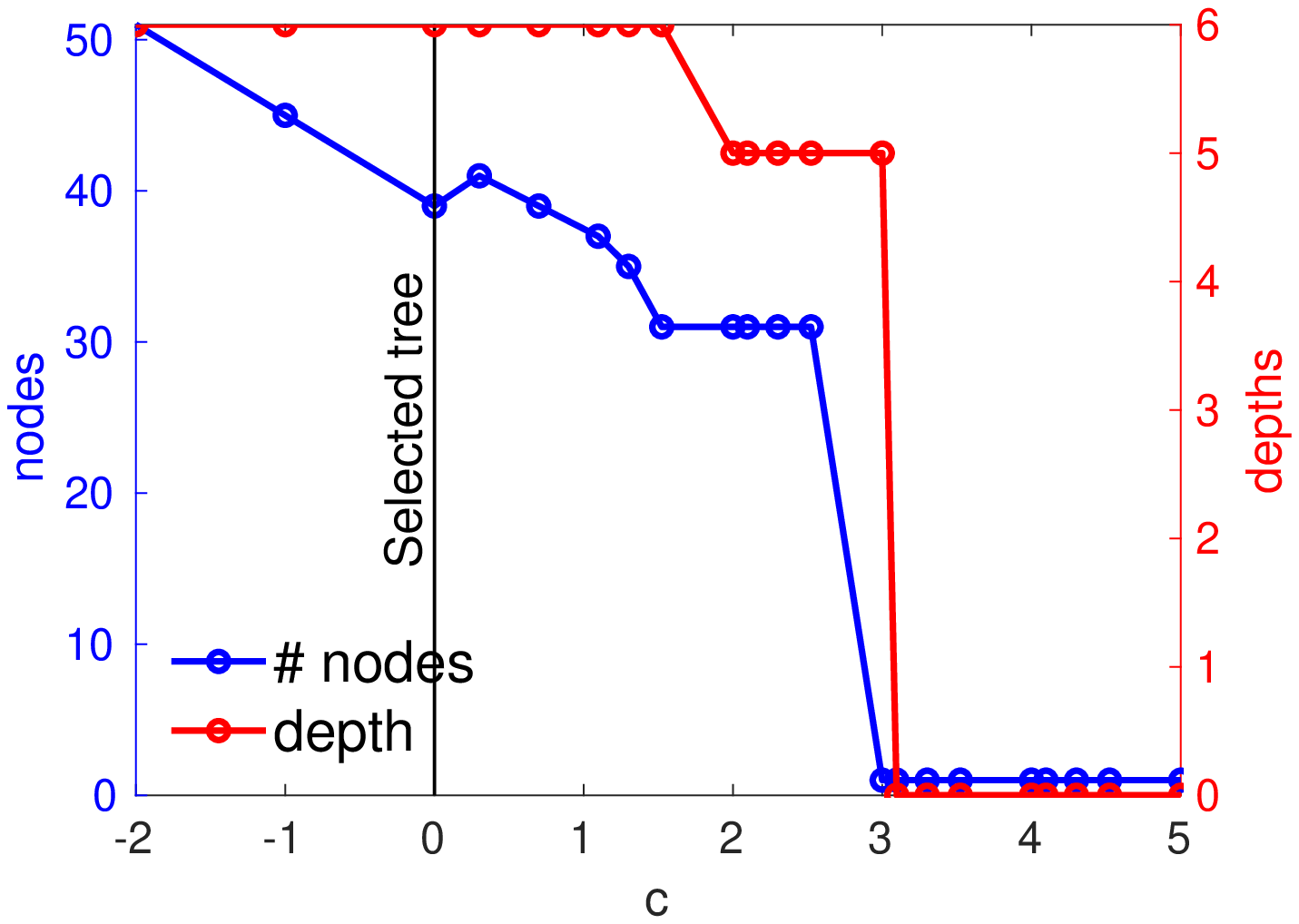} \\[1ex]
    \psfrag{sparse}[][][0.9]{\hspace{-2ex}\# nonzero weights ($\times$1000)}
    \psfrag{c}[][B]{$\log_{10}\lambda$}
    \psfrag{Selected tree}[c][][0.9]{\hspace{4ex}Selected tree}
    \includegraphics*[width=.52\linewidth]{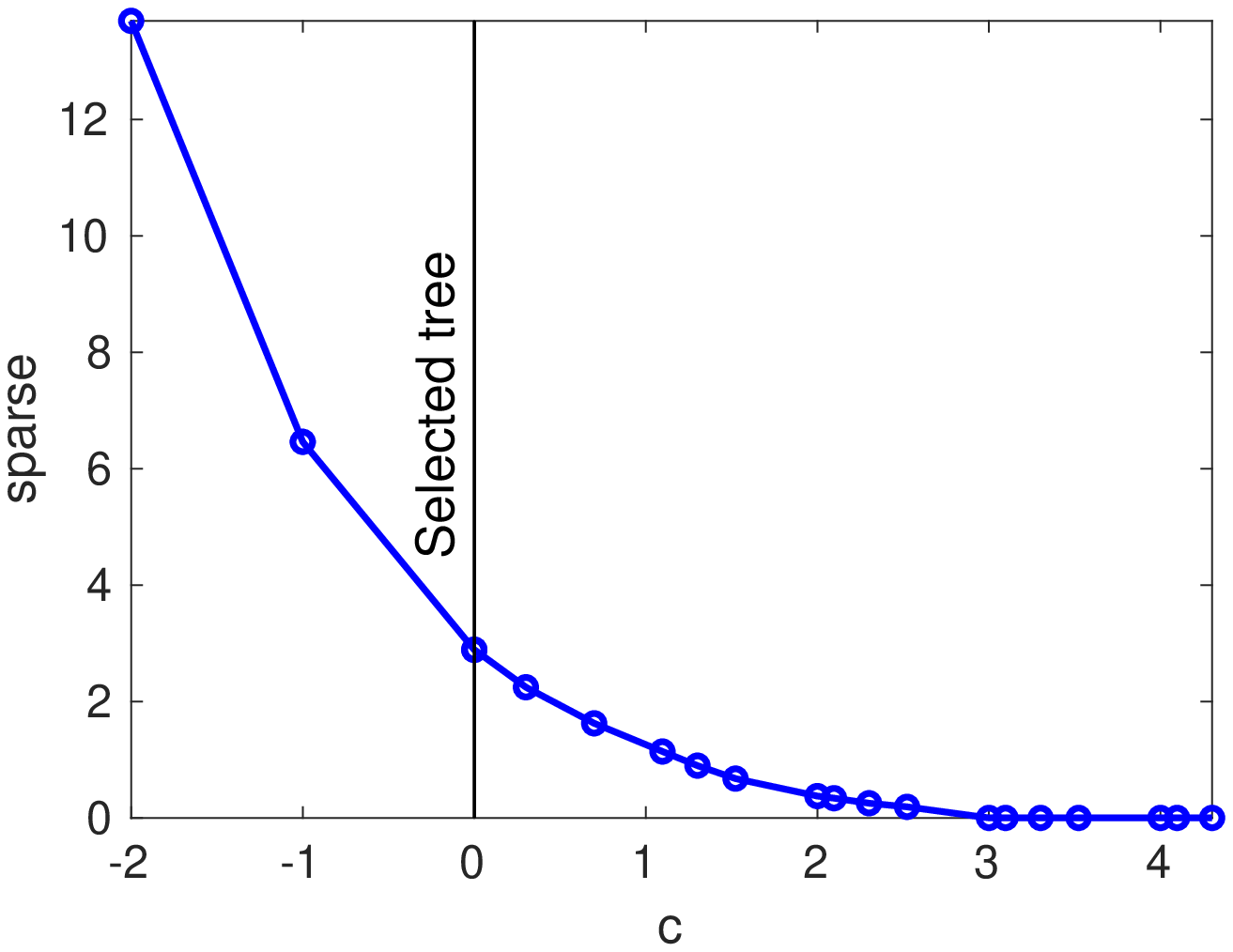}
  \end{tabular}
  \caption{Classification error (training and test) and number of nodes and of nonzero weights of the trees as a function of $\lambda$ for VGG16. The vertical line indicates the tree we selected as mimic ($\lambda = 1$).}
  \label{f:error}
\end{figure}

We selected a subset of 16 classes from the ImageNet object classification dataset \citep{Deng_09a}; they are listed in table~\ref{t:ImageNet-subset-classes}. For each class we split the available images into 200 for test, 100 for validation and 1\,000 for training (total 20\,800 images). We used a VGG16 deep net \citep{SimonyZisser15a} with the architecture shown in table~\ref{t:VGG16}, which takes as input color images of 64 $\times$ 64 pixels. We fine-tuned a pretrained VGG16 for our ImageNet subset of 16 classes. We train the network using Nesterov's accelerated gradient method for 20 epochs using minibatches of size 32, learning rate 0.02  and momentum rate 0.9. Our resulting VGG16 net achieves an error of 0.2\% (training) and 6.79\% (test). We select the $F =$ 8\,192 neurons from VGG16's last convolutional layer as features on which to train the tree.

We trained sparse oblique trees on these features with the TAO algorithm, using our own Python implementation of TAO. We used as initial tree structure for TAO a complete tree of depth 6 (total 127 nodes), which we found sufficient to produce small abut accurate trees in this case, and random initial values for the weights at the nodes. The decision nodes are hyperplanes and each leaf contains a single class label. We constructed a collection of trees over a range of the sparsity hyperparameter $\lambda \in [0,\infty)$ (the regularization path). We ran TAO for 40 iterations (when it approximately converged) to learn each tree. Fig.~\ref{f:error} shows, for each tree, its error (training and test), size (depth and number of nodes) and number of nonzero weights as a function of the sparsity hyperparameter $\lambda \in [0,\infty)$.

\begin{figure}[p]
  \centering
  \begin{tabular}{@{}c@{}}
    \includegraphics*[width=1\linewidth]{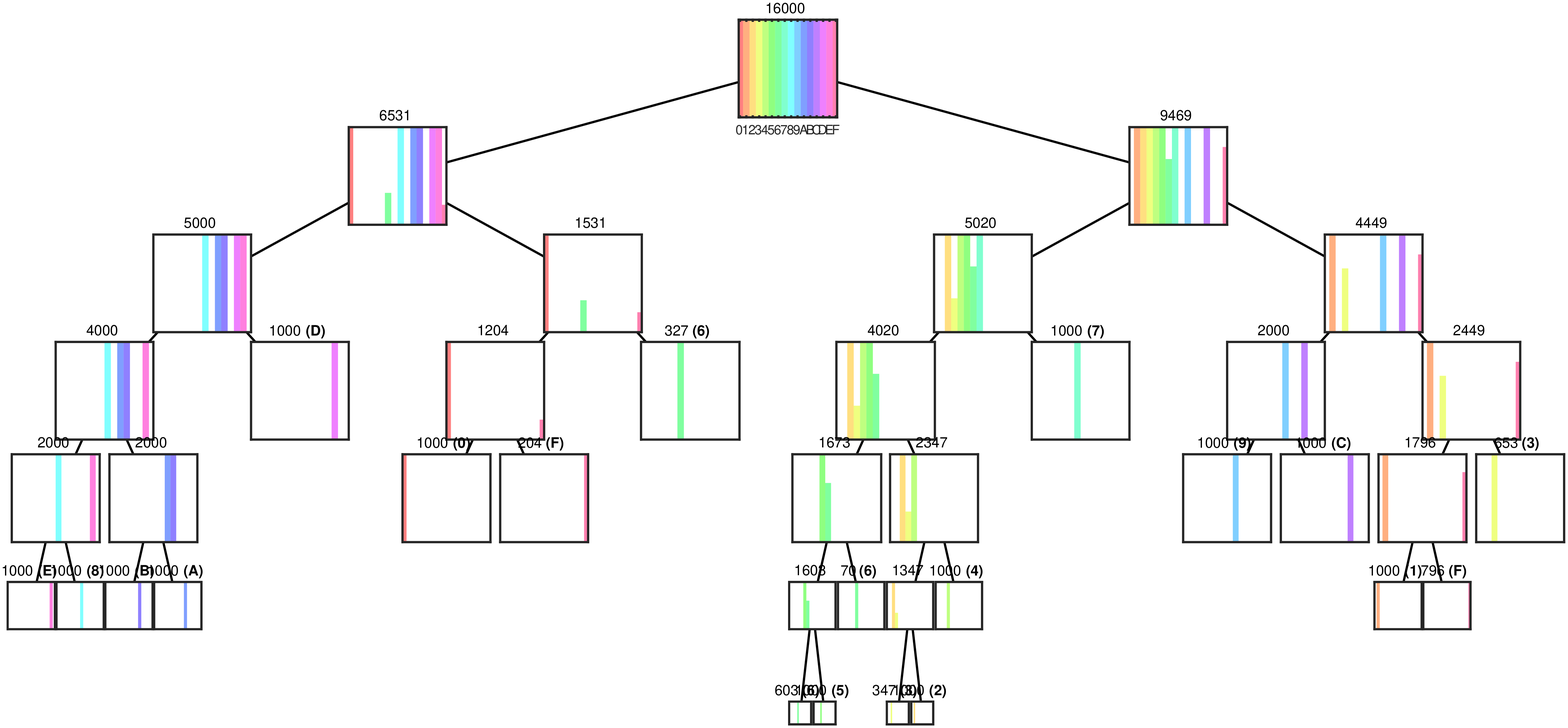} \\[5ex]
    \psfrag{p0}[l][][.6][315]{\textsf{goldfish}}
    \psfrag{p1}[l][][.6][315]{\textsf{bald eagle}}
    \psfrag{p2}[l][c][.6][315]{\textsf{goose}}
    \psfrag{p21}[c][r][.3][0]{}
    \psfrag{p3}[l][][.6][315]{\textsf{killer whale}}
    \psfrag{p31}[t][r][.3][0]{}
    \psfrag{p32}[l][l][.6][315]{\textsf{killer whale}}
    \psfrag{p4}[l][][.6][315]{\textsf{Siberian husky}}
    \psfrag{p5}[l][c][.6][315]{\textsf{white wolf}}
    \psfrag{p6}[l][][.6][315]{\textsf{tiger cat}}
    \psfrag{p61}[l][l][.6][315]{\textsf{tiger cat}}
    \psfrag{p62}[l][c][.6][315]{\textsf{tiger cat}}
    \psfrag{p7}[l][][.6][315]{\textsf{lion}}
    \psfrag{p8}[l][][.6][315]{\textsf{airliner}}
    \psfrag{p9}[l][][.6][315]{\textsf{container ship}}
    \psfrag{p10}[l][][.6][315]{\textsf{fire engine}}
    \psfrag{p11}[l][][.6][315]{\textsf{school bus}}
    \psfrag{p12}[l][][.6][315]{\textsf{speedboat}}
    \psfrag{p13}[l][][.6][315]{\textsf{sports car}}
    \psfrag{p14}[l][][.6][315]{\textsf{warplane}}
    \psfrag{p15}[l][][.6][315]{\textsf{coral reef}}
    \includegraphics*[width=1\linewidth]{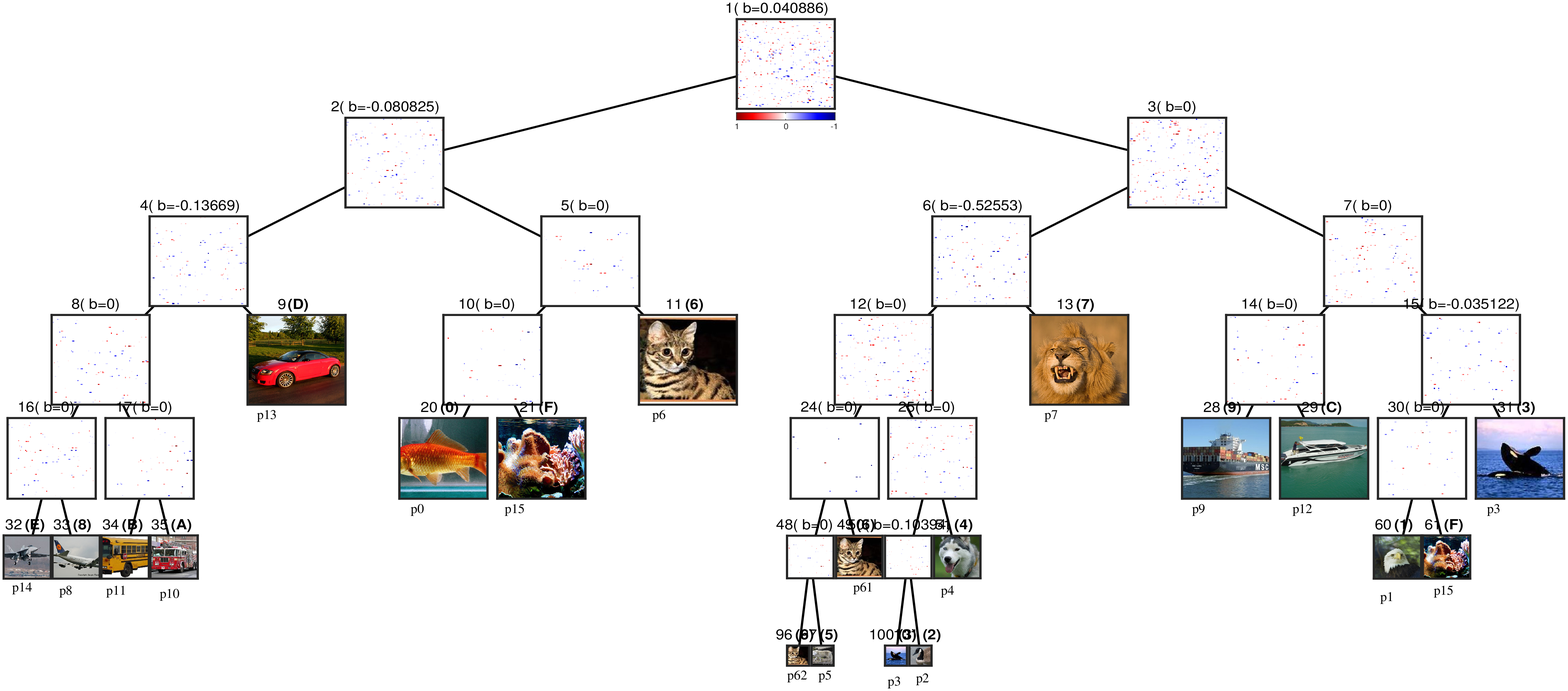} \\[5ex]
  \end{tabular}
  \caption{Tree selected as mimic for VGG16 features ($\lambda = 1$), with a training error of 0\% and a test error of 7.90\%. \emph{Top}: class histograms; we show the number of training instances reaching the node and, for leaves, their label. \emph{Bottom}: weight vector at each decision node and an image from their class at each leaf; we show the node index, bias (always zero) and, for leaves, their label. We plot the weight vector, of dimension 8\,192, as a 91$\times$91 square (the last pixels are unused), with features in the original order in VGG16 (which is determined during training and arbitrary, hence the random aspect of the images), and colored according to their sign and magnitude (positive, negative and zero values are blue, red and white, respectively). You may need to zoom in the plot.}
  \label{f:VGG16-tree1}
\end{figure}

\begin{figure}[p]
  \centering
  \begin{tabular}{@{}c@{}}
    \includegraphics*[width=1\linewidth]{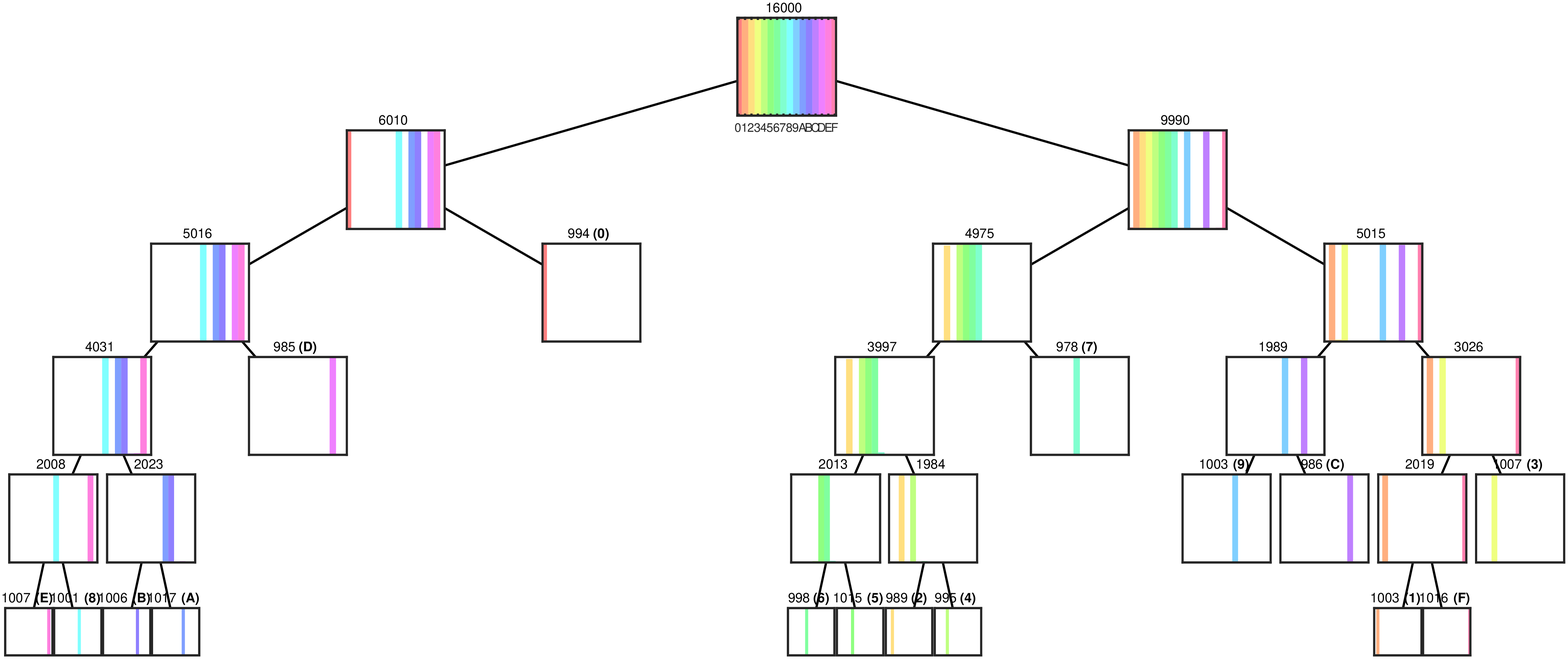} \\[5ex]
    \psfrag{p0}[l][][.6][315]{\textsf{goldfish}}
    \psfrag{p1}[l][][.6][315]{\textsf{bald eagle}}
    \psfrag{p2}[l][][.6][315]{\textsf{goose}}
    \psfrag{p3}[l][][.6][315]{\textsf{killer whale}}
    \psfrag{p4}[l][][.6][315]{\textsf{Siberian husky}}
    \psfrag{p5}[l][][.6][315]{\textsf{white wolf}}
    \psfrag{p6}[l][][.6][315]{\textsf{tiger cat}}
    \psfrag{p7}[l][][.6][315]{\textsf{lion}}
    \psfrag{p8}[l][][.6][315]{\textsf{airliner}}
    \psfrag{p9}[l][][.6][315]{\textsf{container ship}}
    \psfrag{p10}[l][][.6][315]{\textsf{fire engine}}
    \psfrag{p11}[l][][.6][315]{\textsf{school bus}}
    \psfrag{p12}[l][l][.6][315]{\textsf{speedboat}}
    \psfrag{p13}[l][][.6][315]{\textsf{sports car}}
    \psfrag{p14}[l][][.6][315]{\textsf{warplane}}
    \psfrag{p15}[l][][.6][315]{\textsf{coral reef}}
    \includegraphics*[width=1\linewidth]{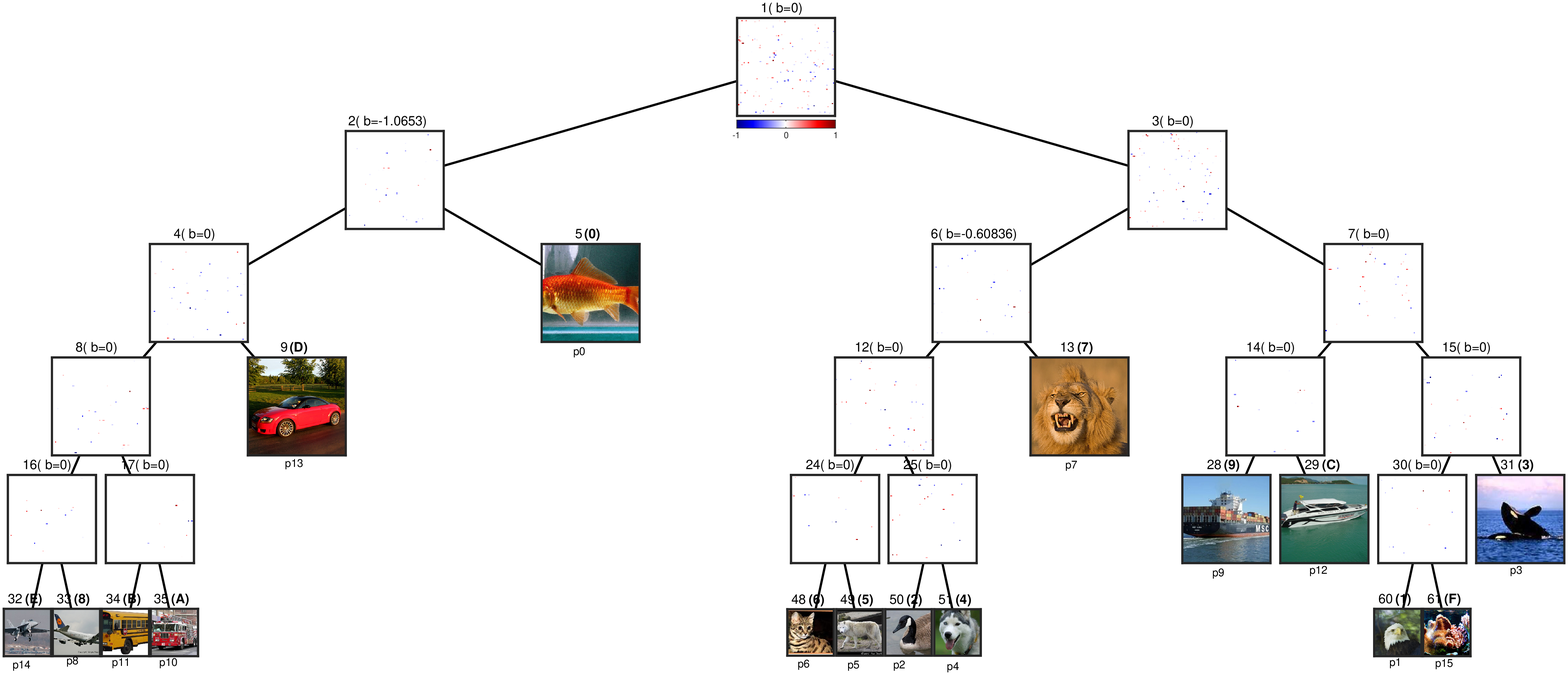} \\[3ex]
  \end{tabular}
  \caption{Like fig.~\ref{f:VGG16-tree1} but for $\lambda = 33$. This tree has exactly one leaf per class (total 16 classes), and a slightly higher error (training error 1.79\%, test error 9.56\%).}
  \label{f:VGG16-tree2}
\end{figure}

The tree with the lowest error over the values of $\lambda$ we considered occurred for $\lambda =$ 0.01. It has depth 6 and 51 nodes, and an error or 0\% (training) and 7.62\% (test). It uses only 4\,423 features out of the total 8\,192. We did not use this tree, instead we selected as mimic the tree for $\lambda =$ 1, which is quite smaller (depth 6 and only 39 nodes) but has nearly the same error (0\% training, 7.90\% test). It uses just 1\,366 features (17\% of the total 8\,192). Its error is very close to that of VGG16, so we expect the tree to be a good mimic of the net. We normalize the final tree so each node weight vector has norm 1. This tree is shown in fig.~\ref{f:VGG16-tree1}. We also discuss another tree that is slightly less accurate but which has exactly one leaf per class and is even more interpretable (fig.~\ref{f:VGG16-tree2}). This tree ($\lambda = 33$) has an error of 1.79\% (training) and 9.56\% (test); it has 31 nodes and uses just 408 features (5\% of the total 8\,192).

\begin{figure}[p]
 \centering 

\begin{tabular}{@{}c@{}c@{}c@{}c@{}c@{}c@{}c@{}c@{}}

   && \multicolumn{2}{c}{ground truth vs deep net}& 
    \multicolumn{3}{c}{features selected}&\\
    && \multicolumn{2}{c}{vs tree}& 
    \multicolumn{3}{c}{by the tree}&\\
&&\psfrag{Ground truth}[][][.8]{ground truth}
\psfrag{Original net}[][][.8]{original net}
\includegraphics*[width=.12\columnwidth]{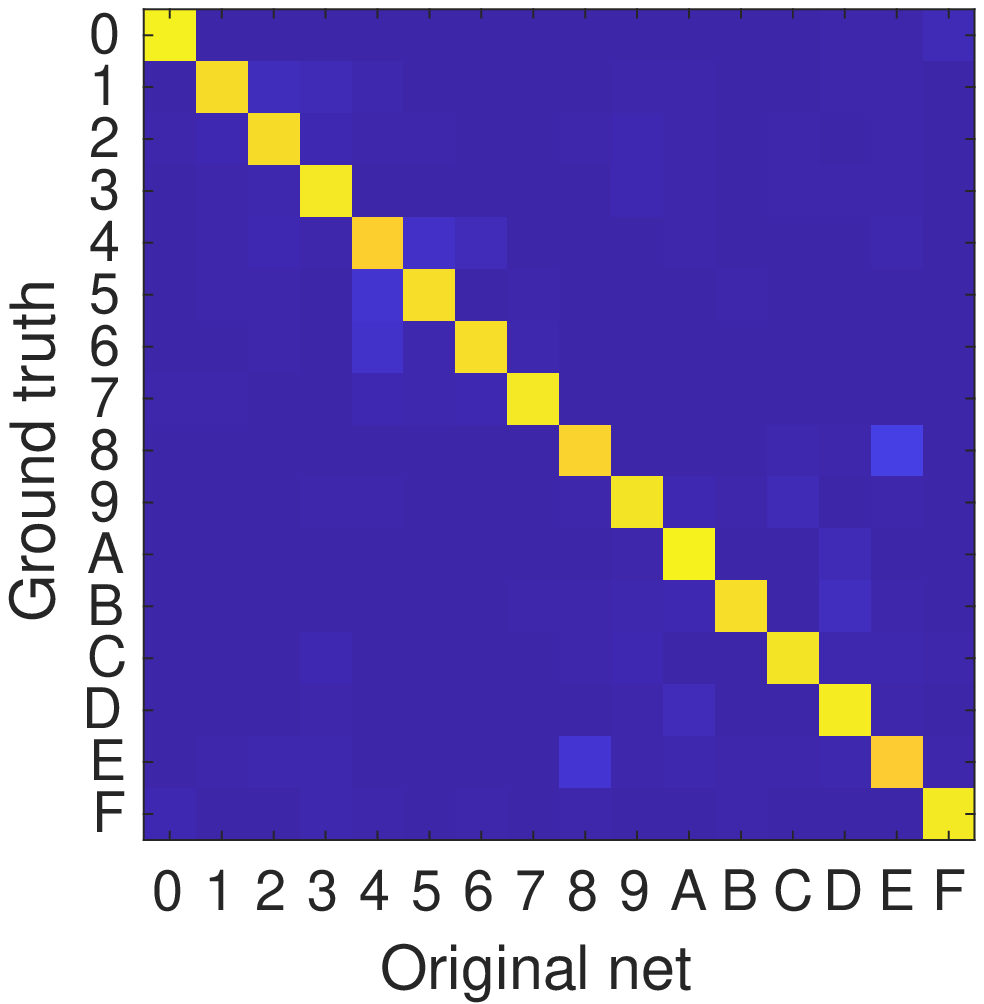}&
\psfrag{Original net}[][][.8]{original net}
\psfrag{Tree}[][][.8]{Tree}
\includegraphics*[width=.12\columnwidth]{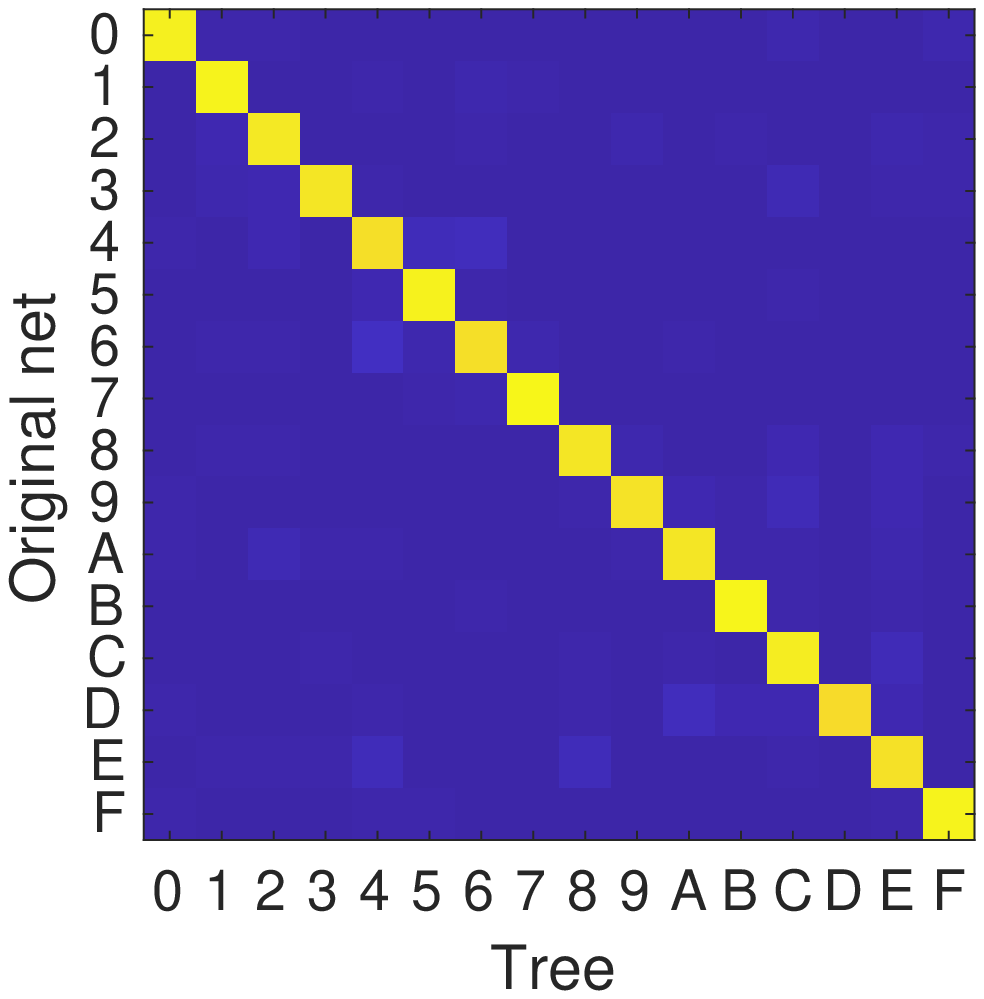}&

\multicolumn{3}{c}{
\psfrag{Original net}[][][.8]{original net}
\psfrag{Modified net}[][][.8]{masked net}
\includegraphics*[width=.12\columnwidth]{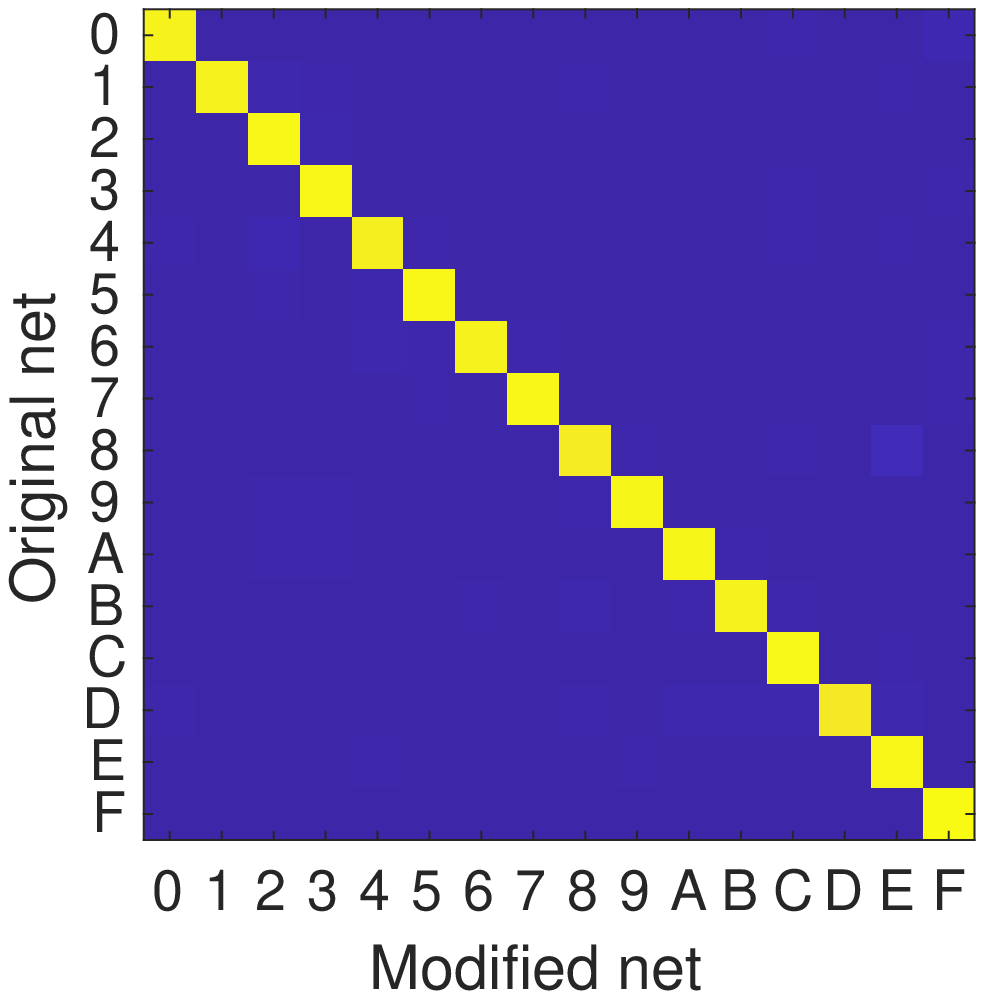}
}&

\hspace*{-33ex} \raisebox{1ex}{ \includegraphics*[height=.11\textwidth]{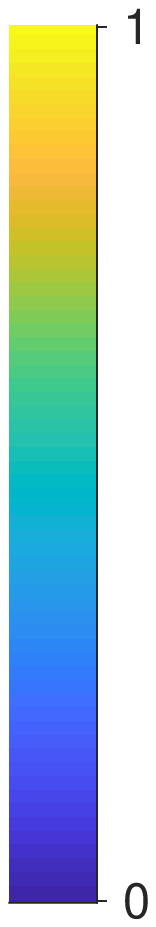} }\\[2ex]
& \multicolumn{6}{c}{\dotfill \textsc{All class $k_1$ to class $k_2$} \dotfill}&\\

&{$8\rightarrow14$}&{$14\rightarrow8$}&{$10\rightarrow11$}&{$11\rightarrow10$}&{$9\rightarrow12$}&{$12\rightarrow9$}&\\

&\psfrag{Original net}[][][.8]{original net}
\psfrag{Modified net}[][][.8]{masked net}
\includegraphics*[width=.12\columnwidth]{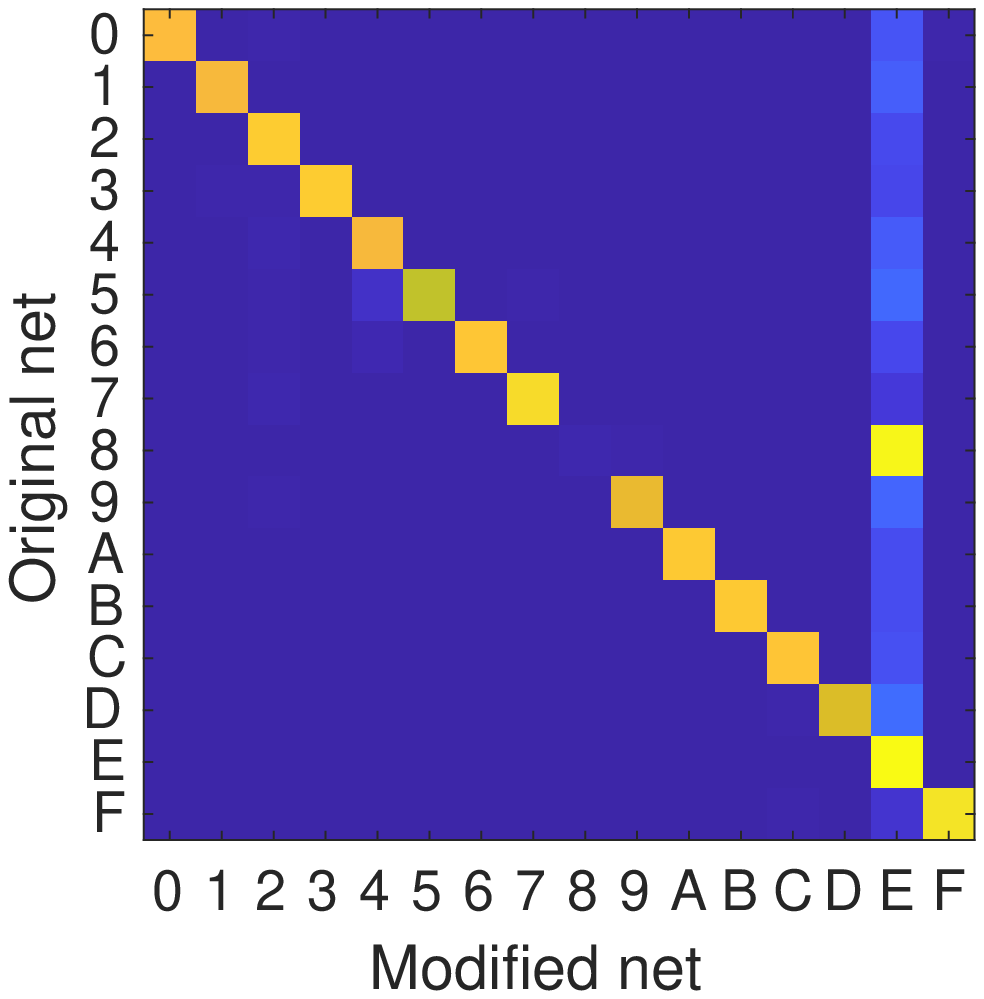}&
\psfrag{Original net}{}
\psfrag{Modified net}[][][.8]{masked net}
\includegraphics*[width=.12\columnwidth]{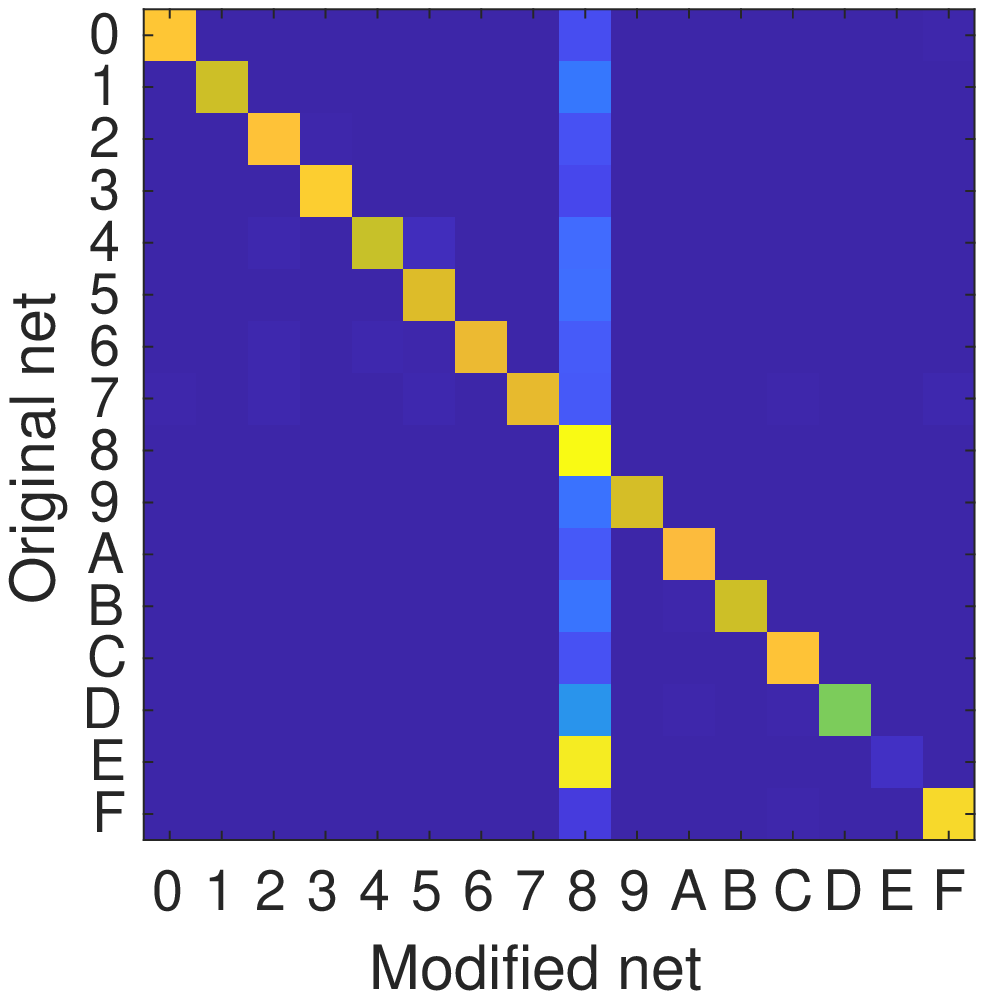}&
\psfrag{Original net}{}
\psfrag{Modified net}[][][.8]{masked net}
\includegraphics*[width=.12\columnwidth]{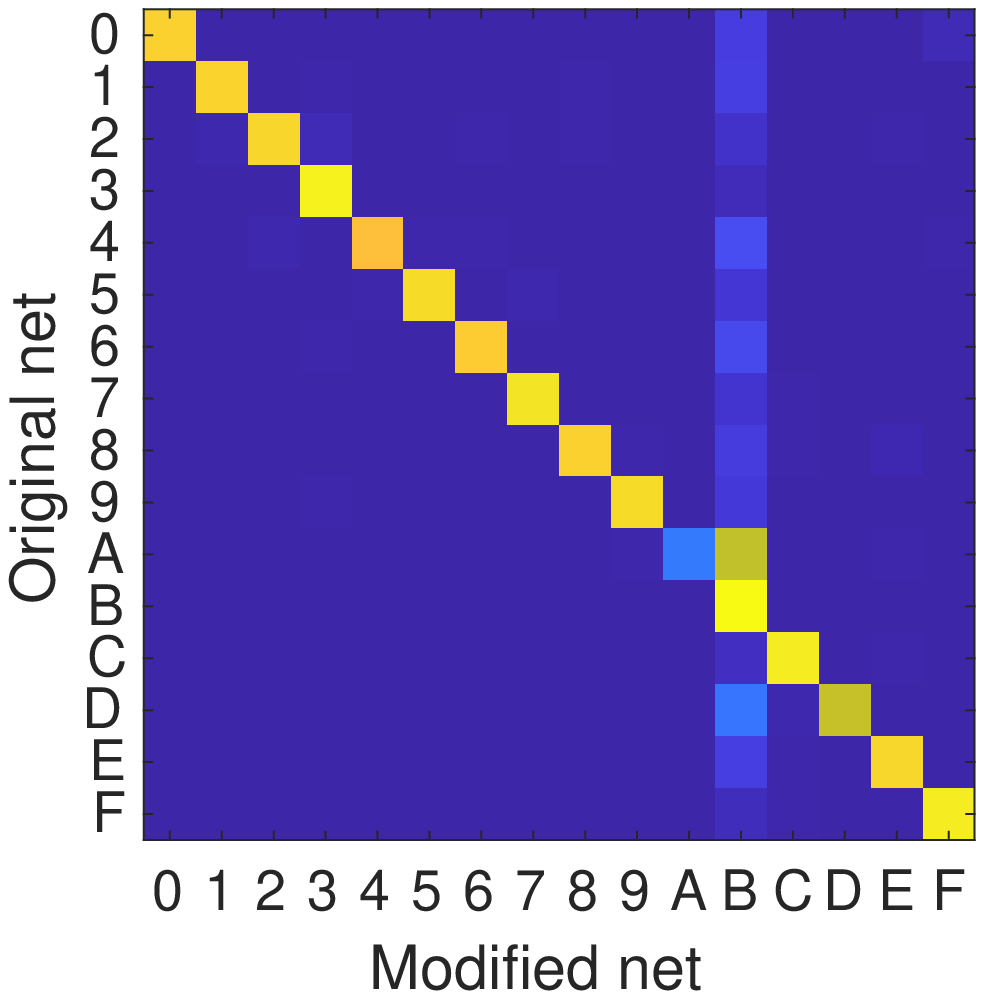}&
\psfrag{Original net}{}
\psfrag{Modified net}[][][.8]{masked net}
\includegraphics*[width=.12\columnwidth]{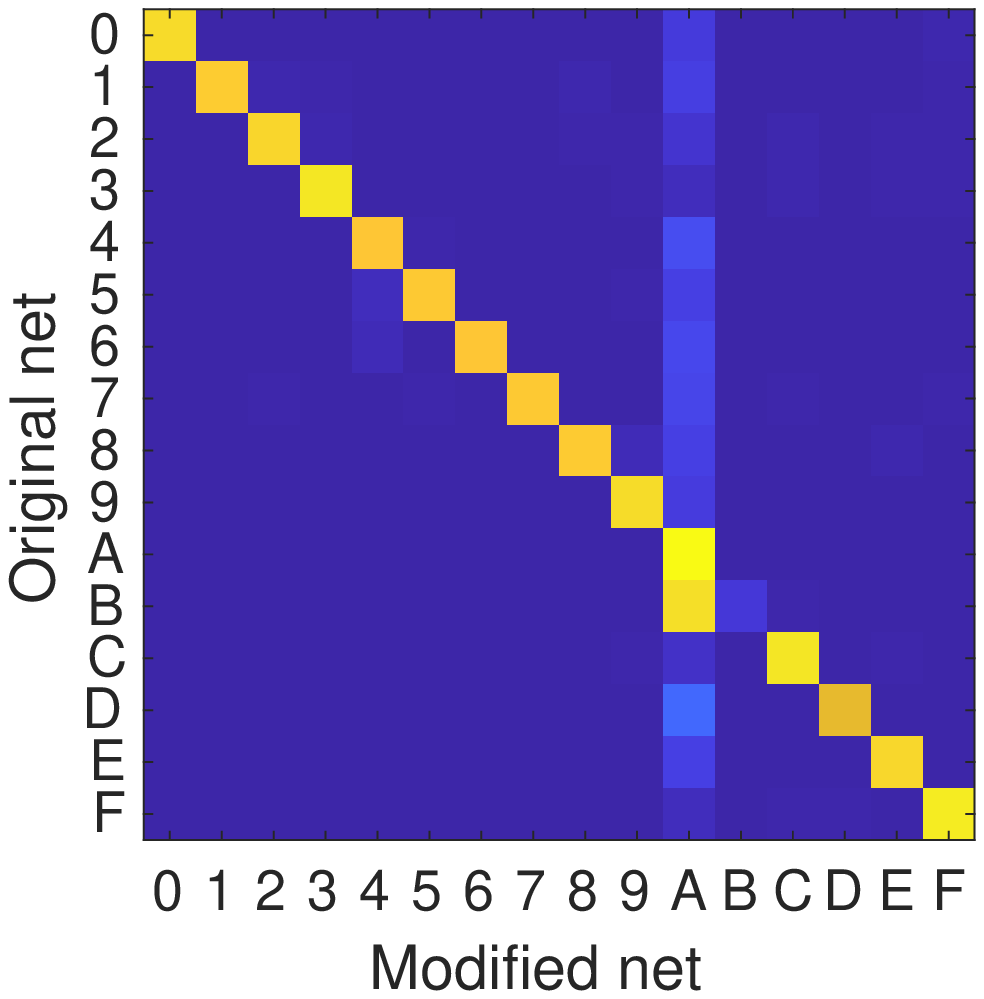}&
\psfrag{Original net}{}
\psfrag{Modified net}[][][.8]{masked net}
\includegraphics*[width=.12\columnwidth]{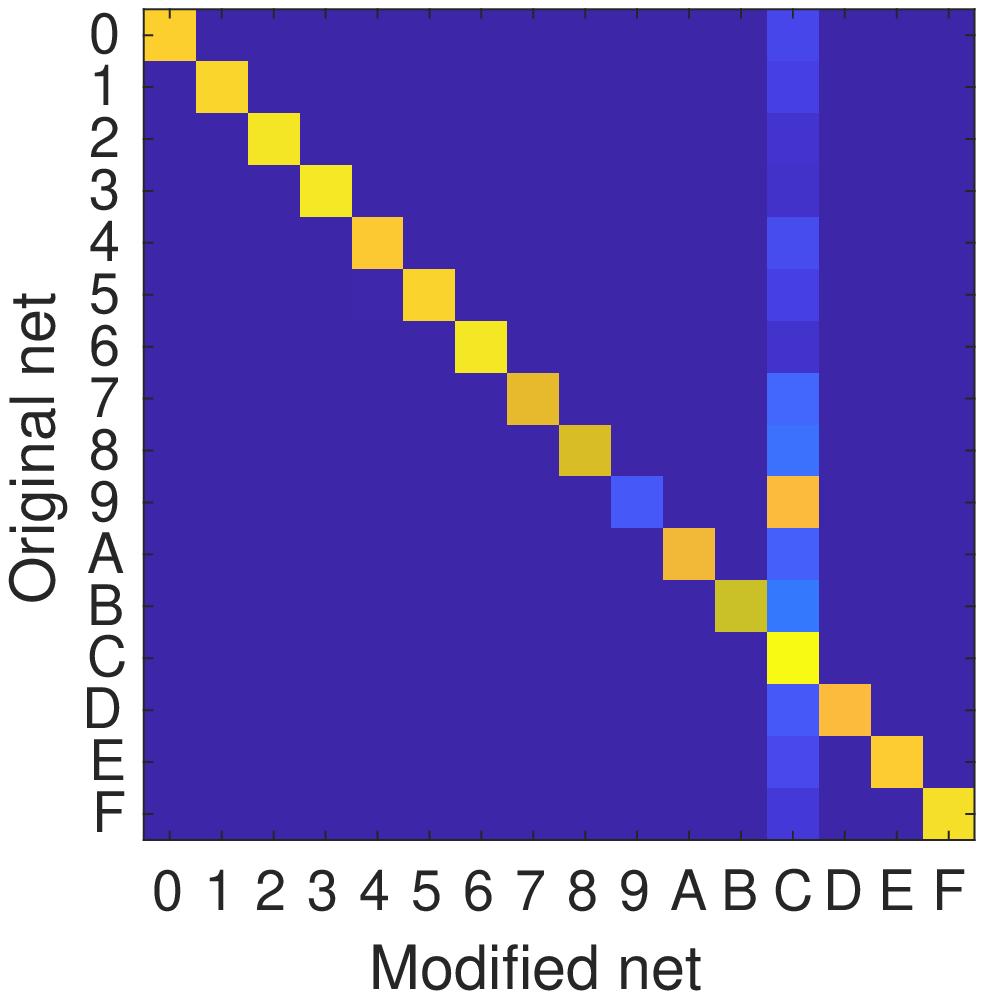}&
\psfrag{Original net}{}
\psfrag{Modified net}[][][.8]{masked net}
\includegraphics*[width=.12\columnwidth]{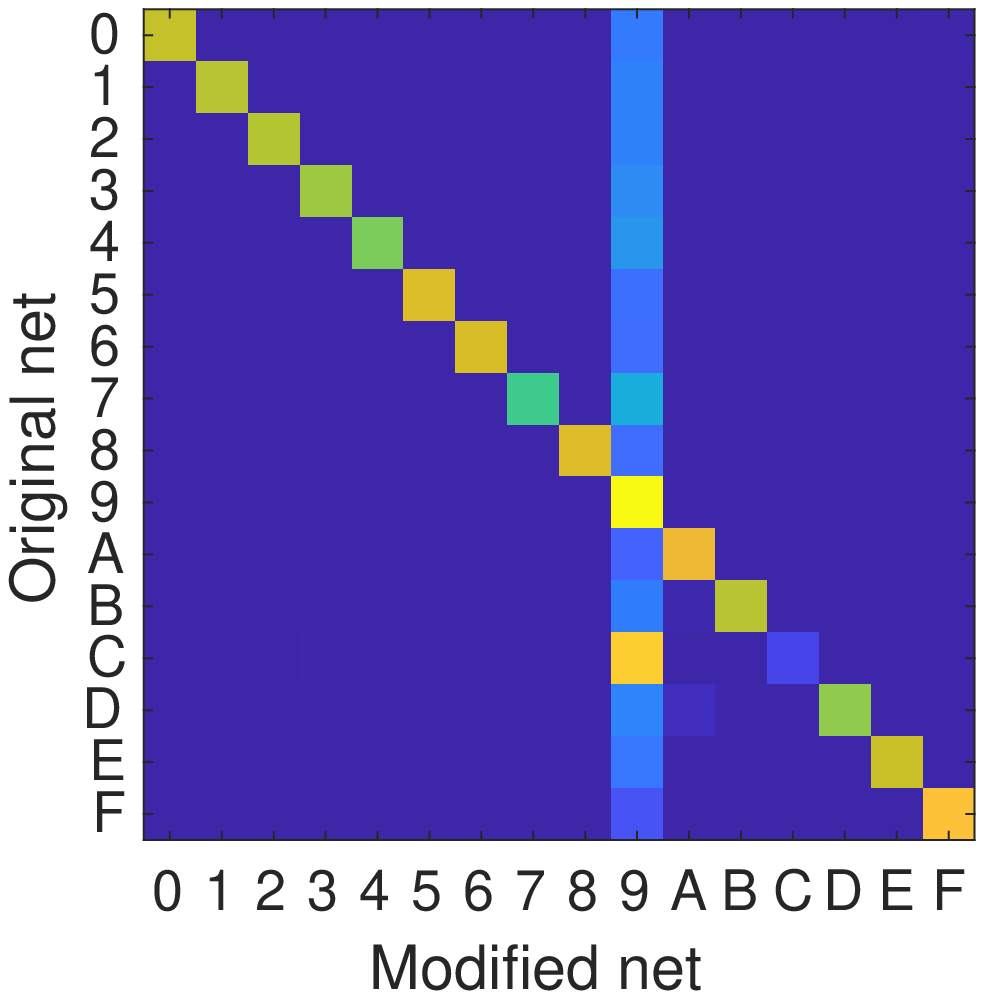}&\\[2ex]

\multicolumn{8}{c}{\dotfill \textsc{None to class $k$} \dotfill}\\
$k = 0$ & $k = 1$ & $k = 2$ & $k = 3$ & $k = 4$ & $k = 5$ & $k = 6$ & $k = 7$  \\
\psfrag{Original net}[][][.8]{original net}
\psfrag{Modified net}[][][.8]{masked net}
\includegraphics*[width=.12\columnwidth]{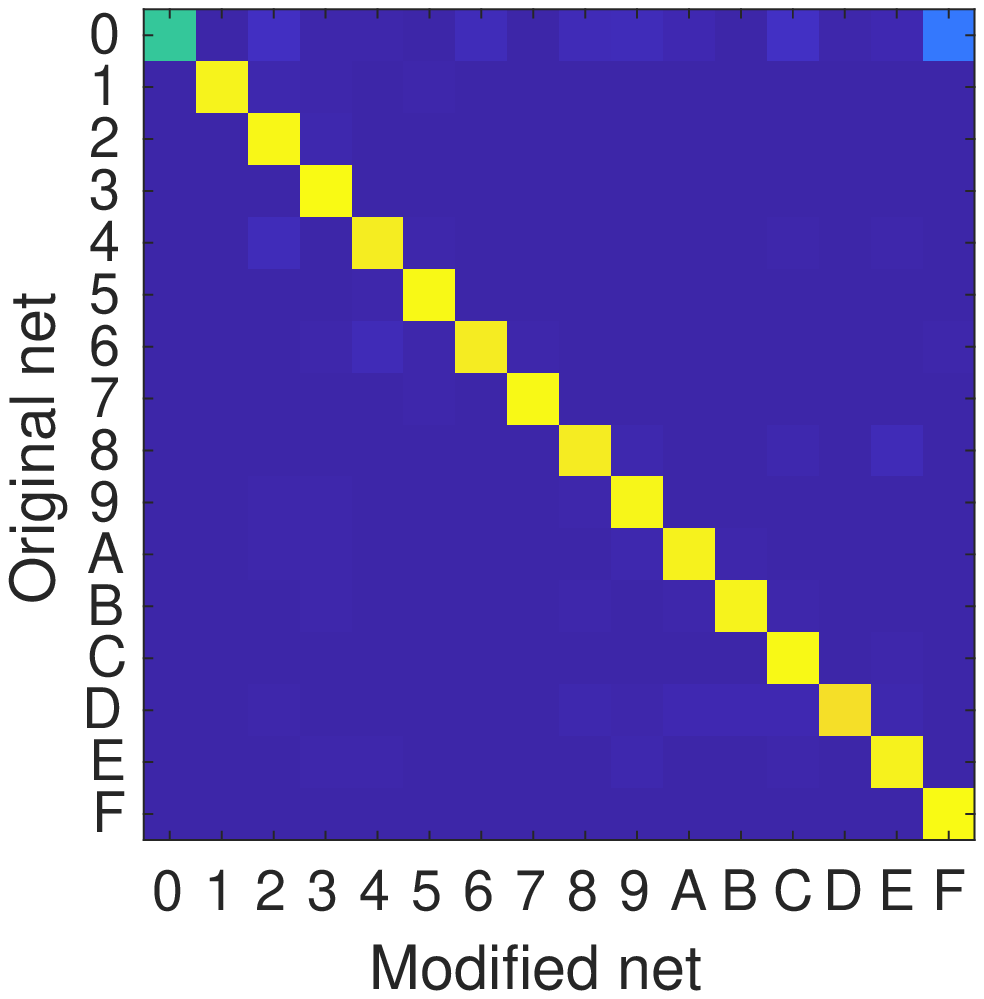}&
\psfrag{Original net}{}
\psfrag{Modified net}[][][.8]{masked net}
\includegraphics*[width=.12\columnwidth]{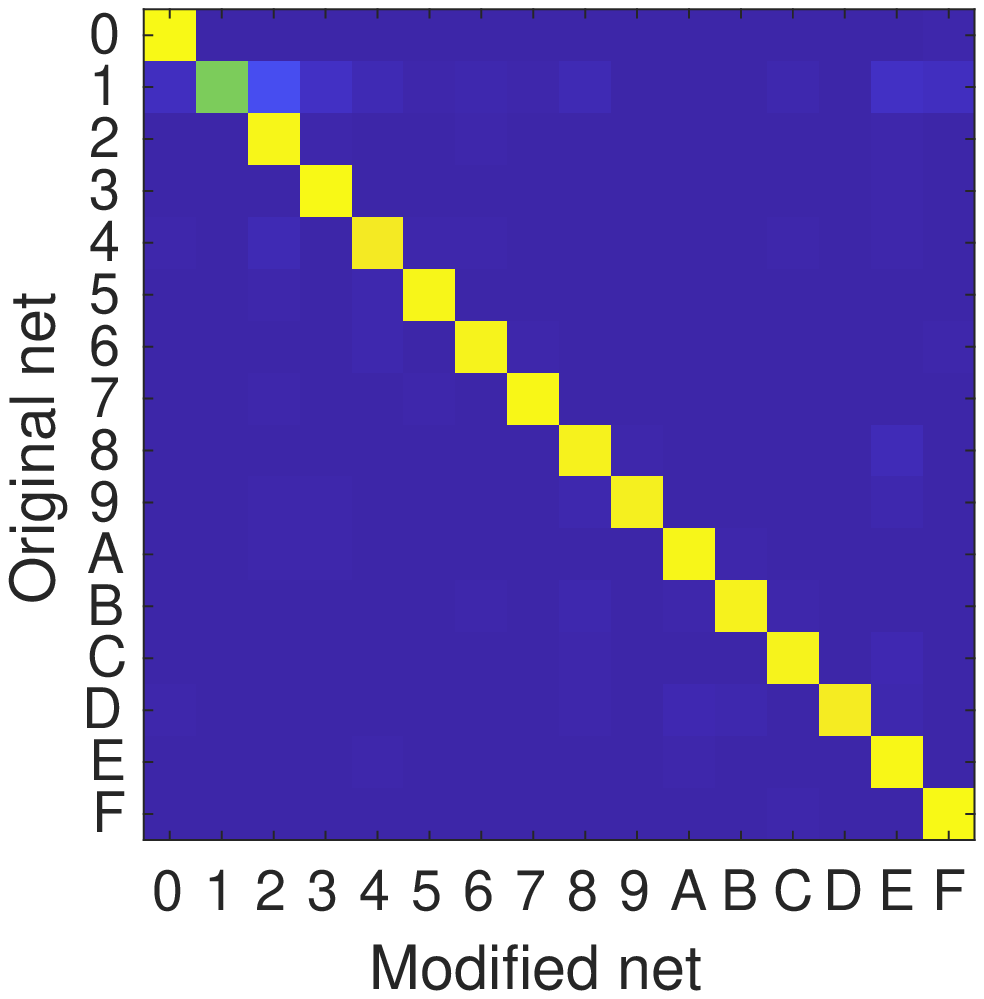}&
\psfrag{Original net}{}
\psfrag{Modified net}[][][.8]{masked net}
\includegraphics*[width=.12\columnwidth]{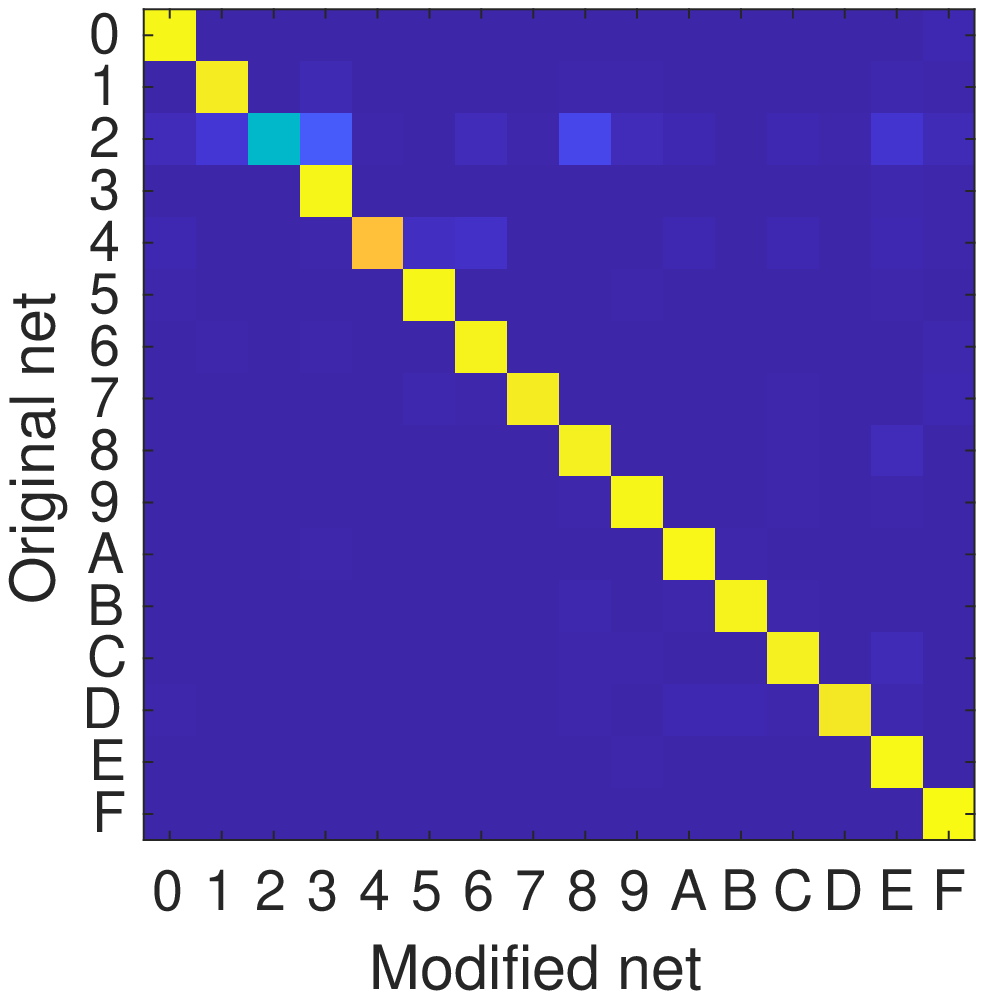}&
\psfrag{Original net}{}
\psfrag{Modified net}[][][.8]{masked net}
\includegraphics*[width=.12\columnwidth]{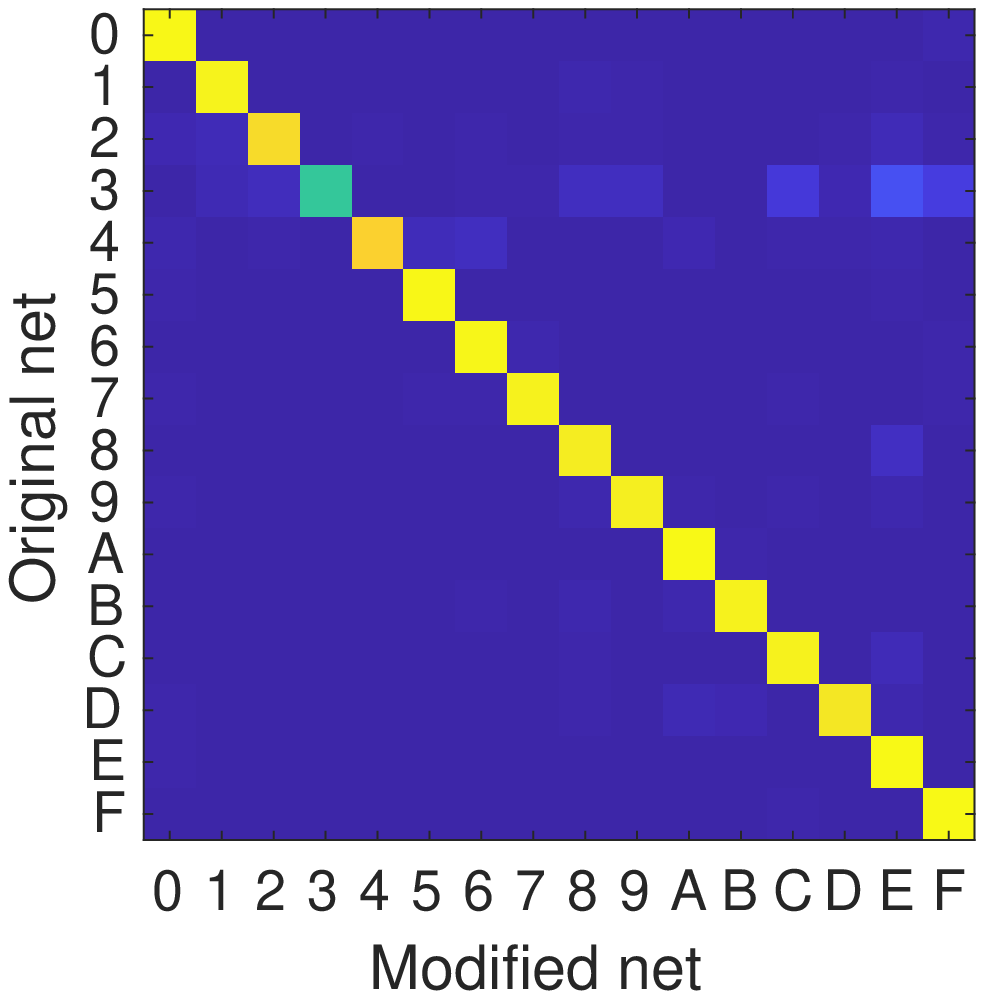}&
\psfrag{Original net}{}
\psfrag{Modified net}[][][.8]{masked net}
\includegraphics*[width=.12\columnwidth]{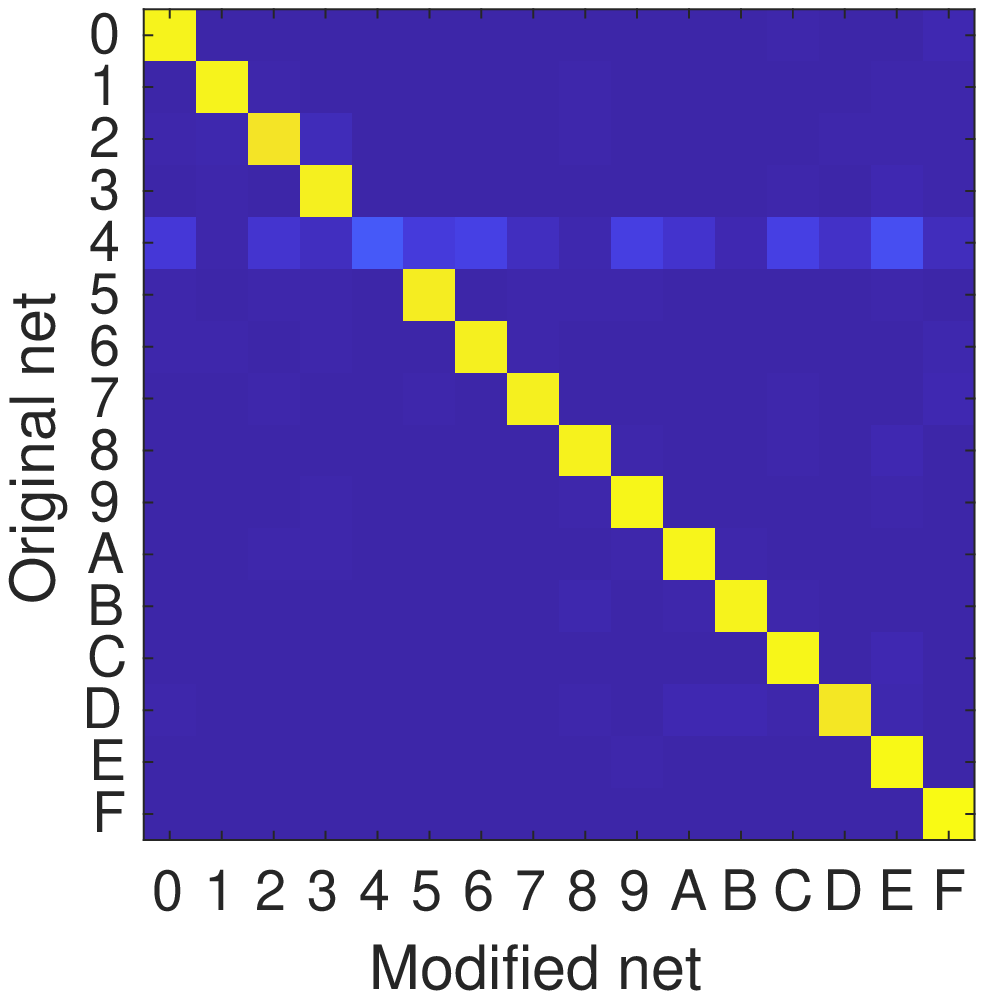}&
\psfrag{Original net}{}
\psfrag{Modified net}[][][.8]{masked net}
\includegraphics*[width=.12\columnwidth]{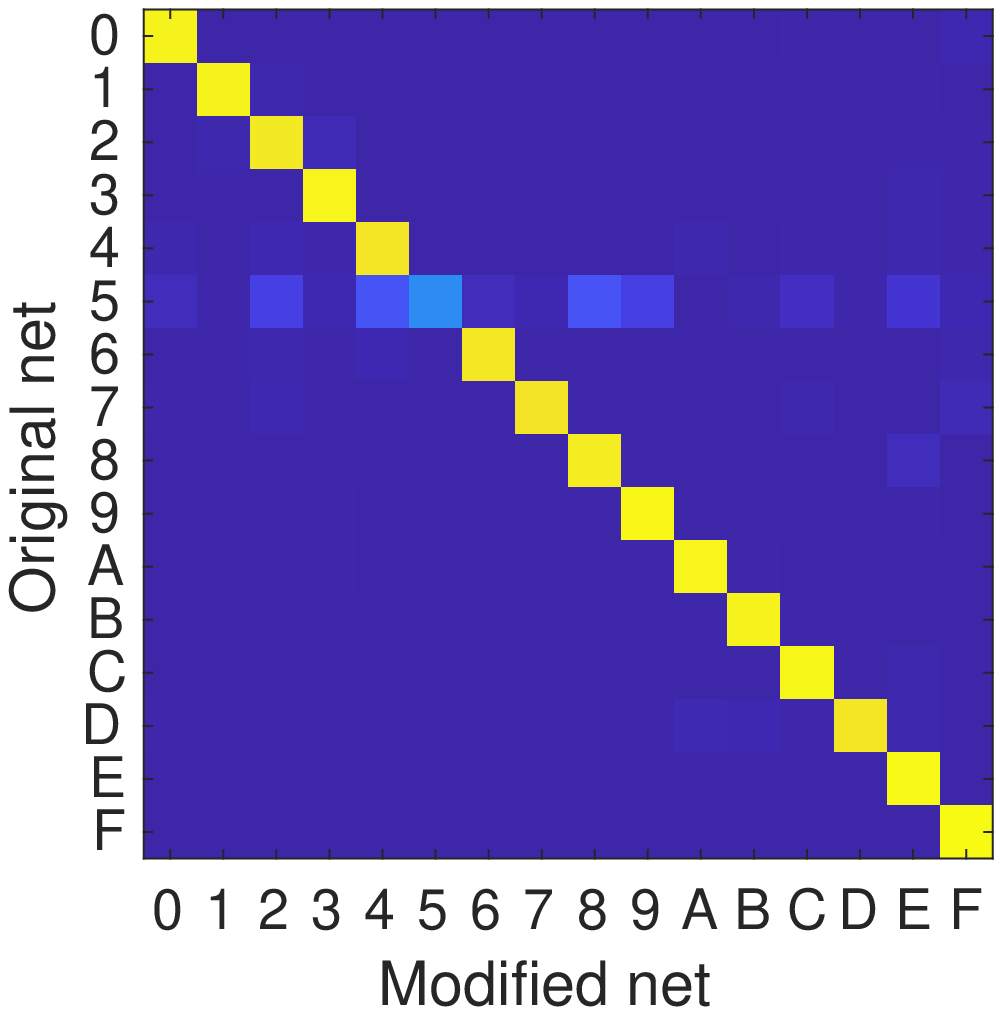}&
\psfrag{Original net}{}
\psfrag{Modified net}[][][.8]{masked net}
\includegraphics*[width=.12\columnwidth]{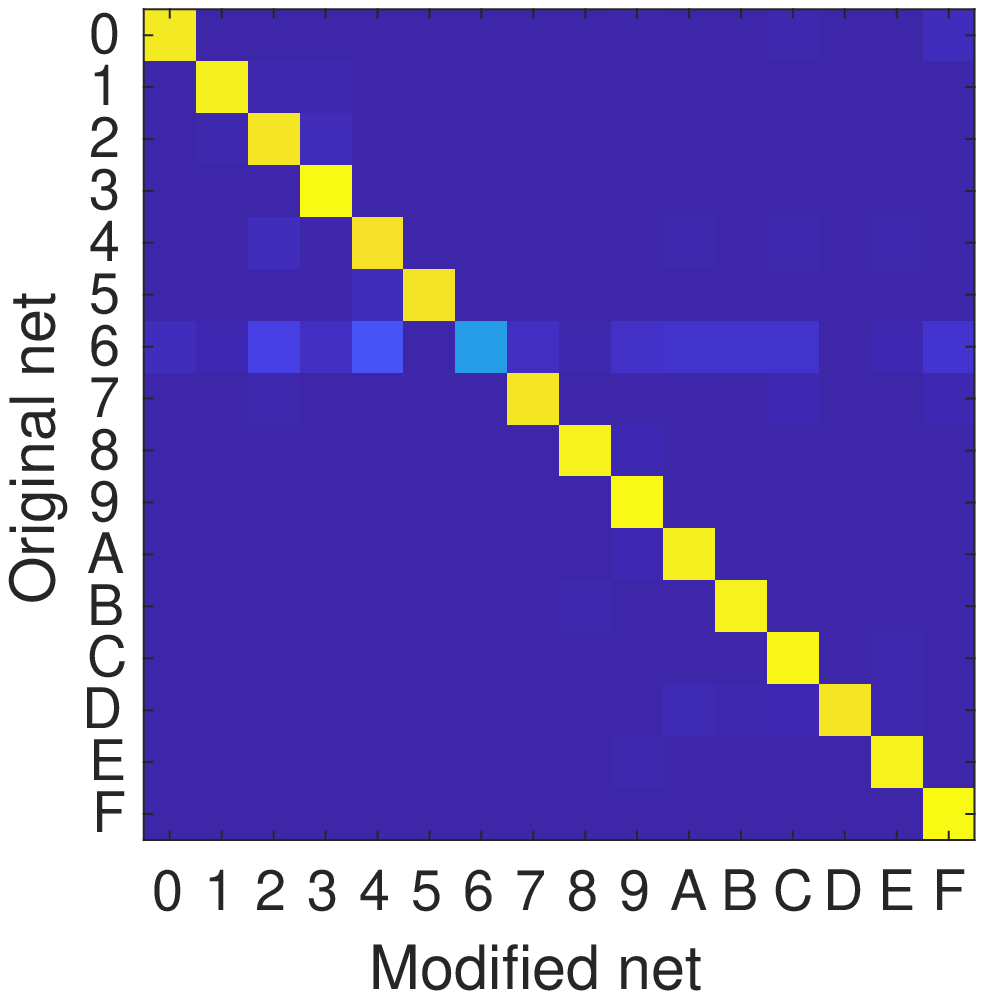}&
\psfrag{Original net}{}
\psfrag{Modified net}[][][.8]{masked net}
\includegraphics*[width=.12\columnwidth]{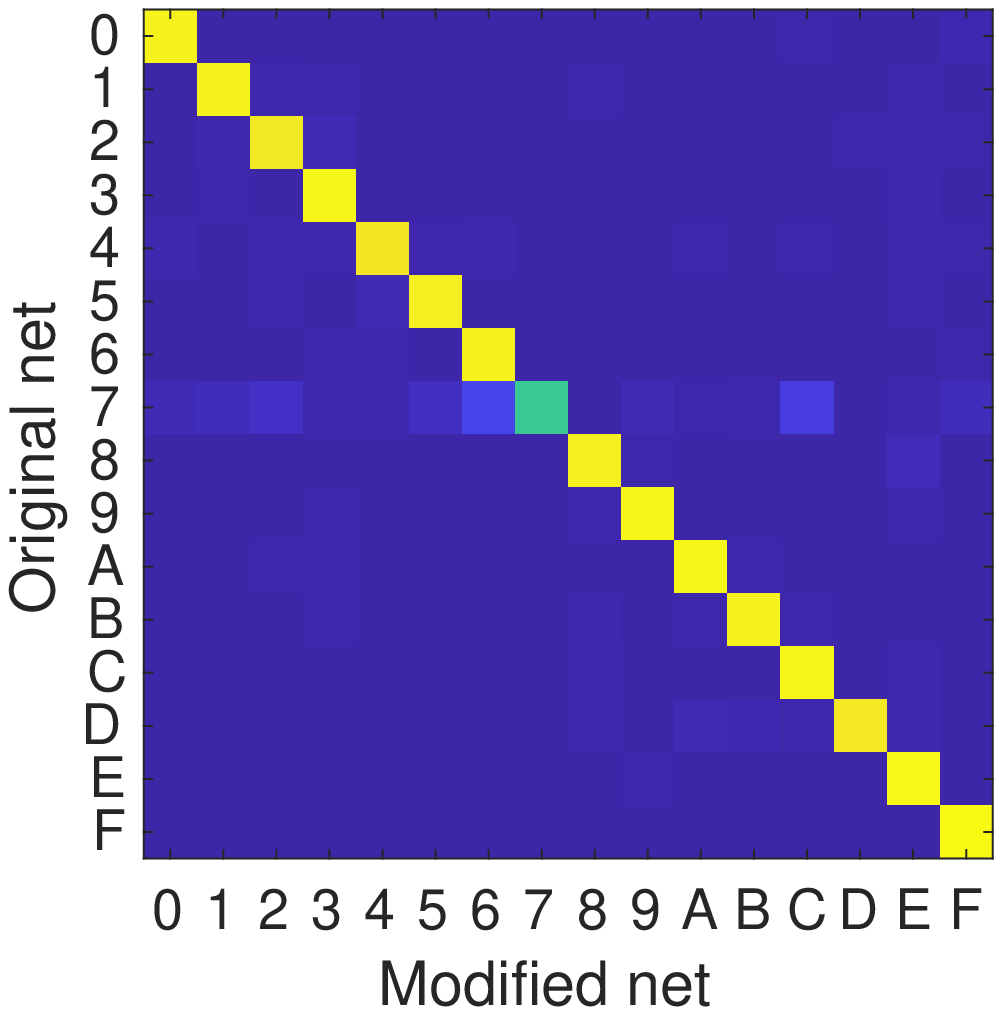}\\
$k = 8$ & $k = 9$ & $k = $A & $k = $B & $k = $C & $k = $D & $k = $D & $k = $E   \\
\psfrag{Original net}[][][.8]{original net}
\psfrag{Modified net}[][][.8]{masked net}
\includegraphics*[width=.12\columnwidth]{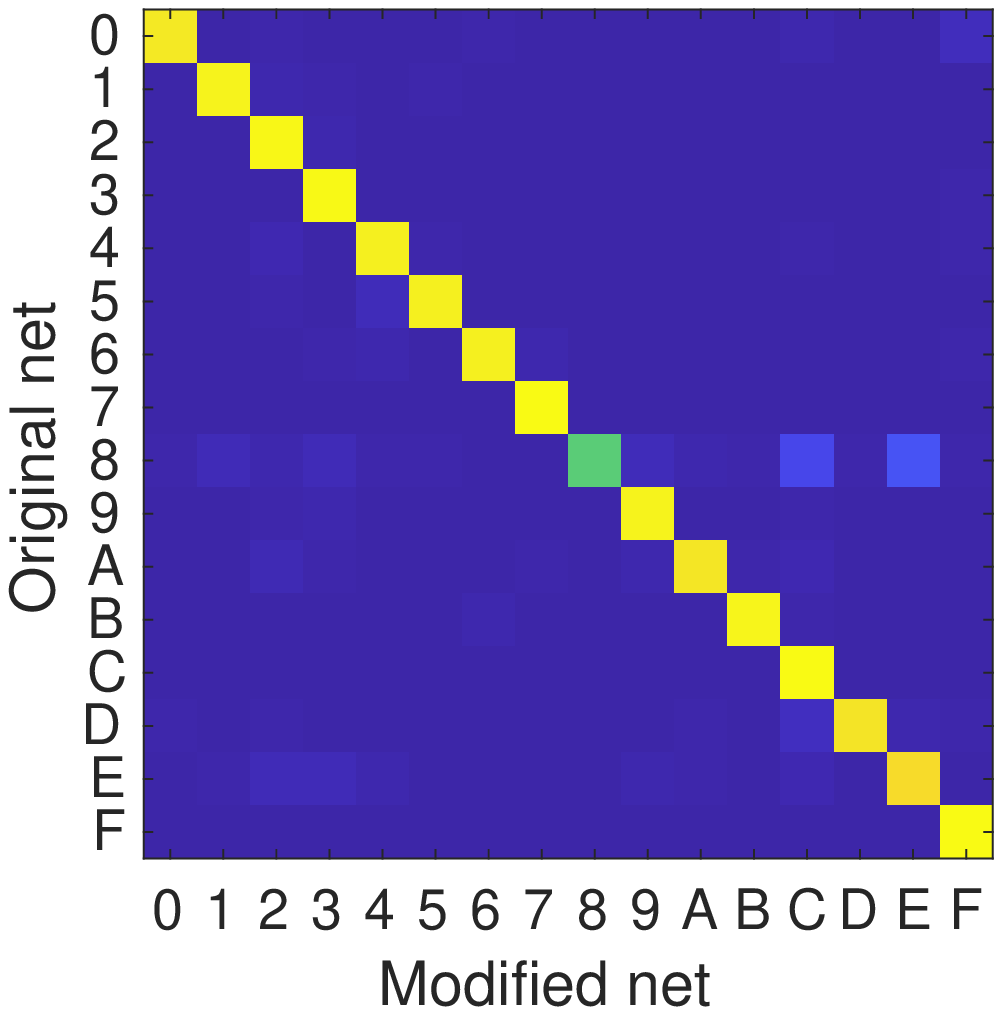}&
\psfrag{Original net}{}
\psfrag{Modified net}[][][.8]{masked net}
\includegraphics*[width=.12\columnwidth]{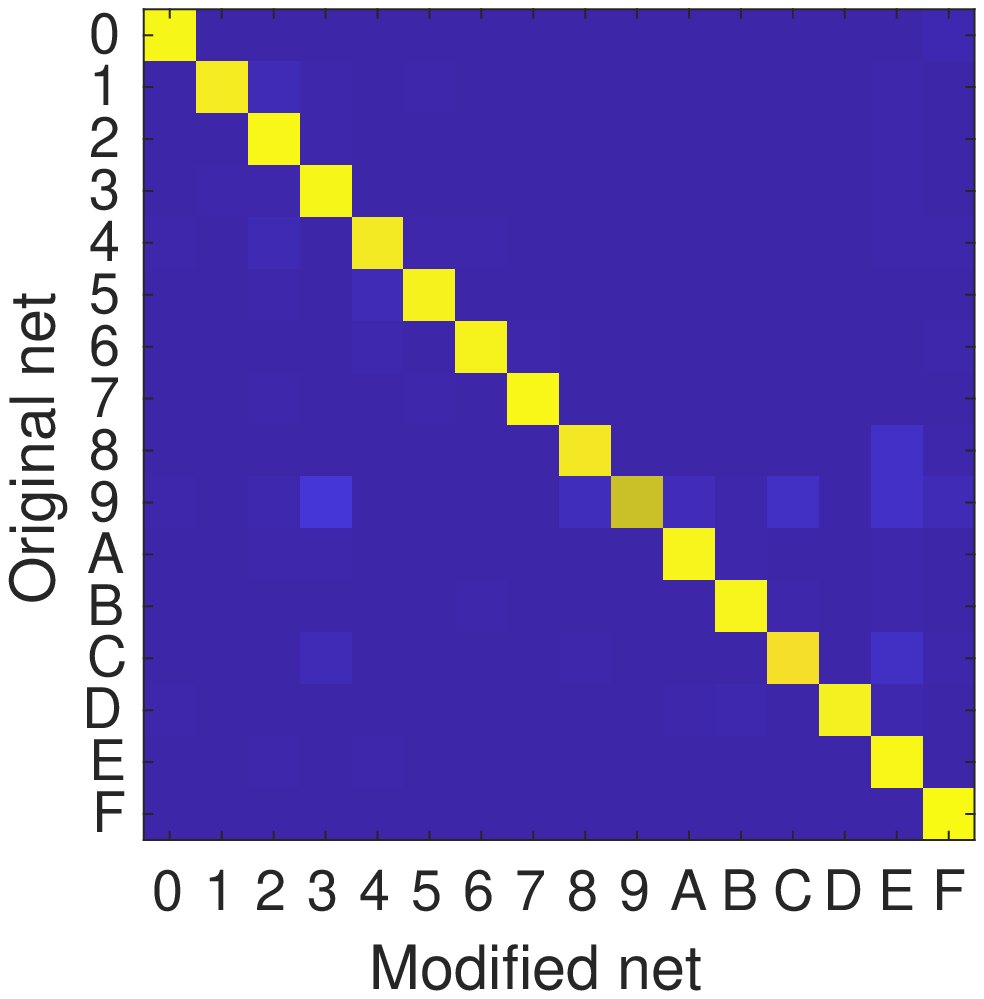}&
\psfrag{Original net}{}
\psfrag{Modified net}[][][.8]{masked net}
\includegraphics*[width=.12\columnwidth]{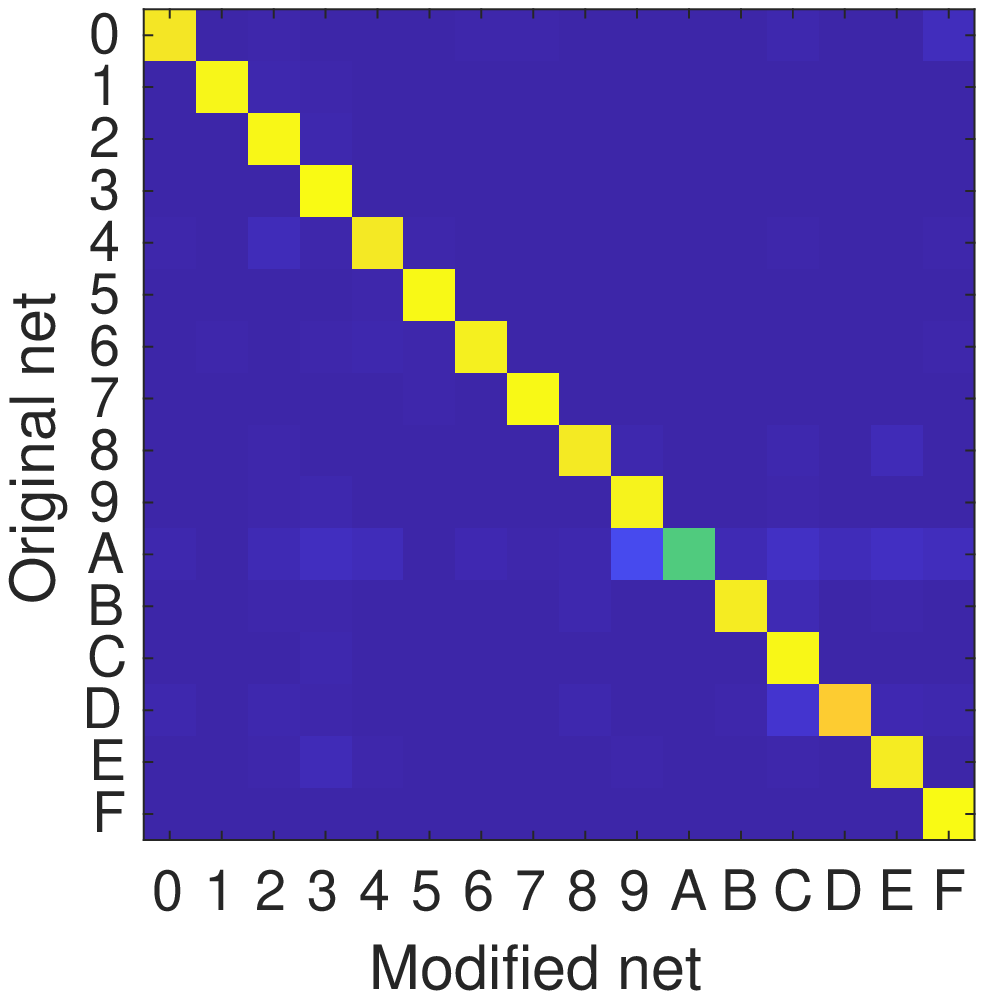}&
\psfrag{Original net}{}
\psfrag{Modified net}[][][.8]{masked net}
\includegraphics*[width=.12\columnwidth]{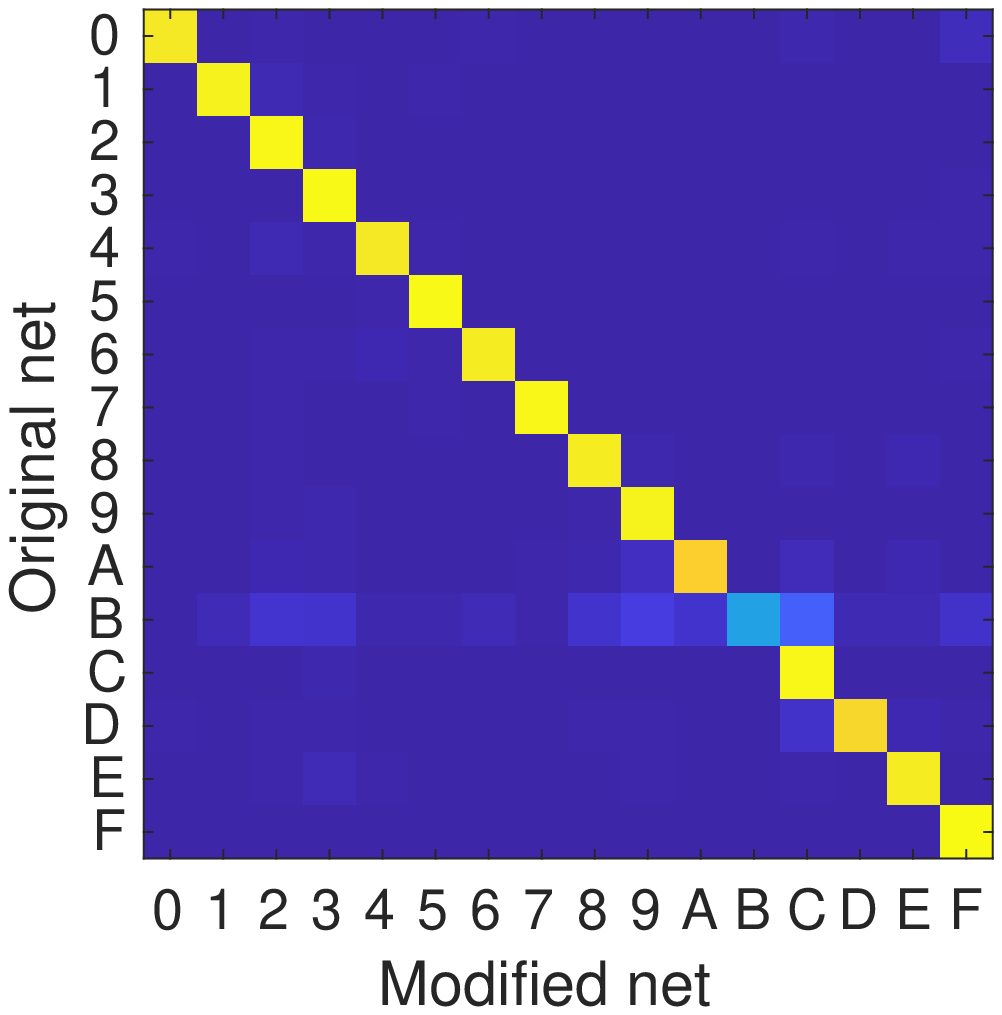}&
\psfrag{Original net}{}
\psfrag{Modified net}[][][.8]{masked net}
\includegraphics*[width=.12\columnwidth]{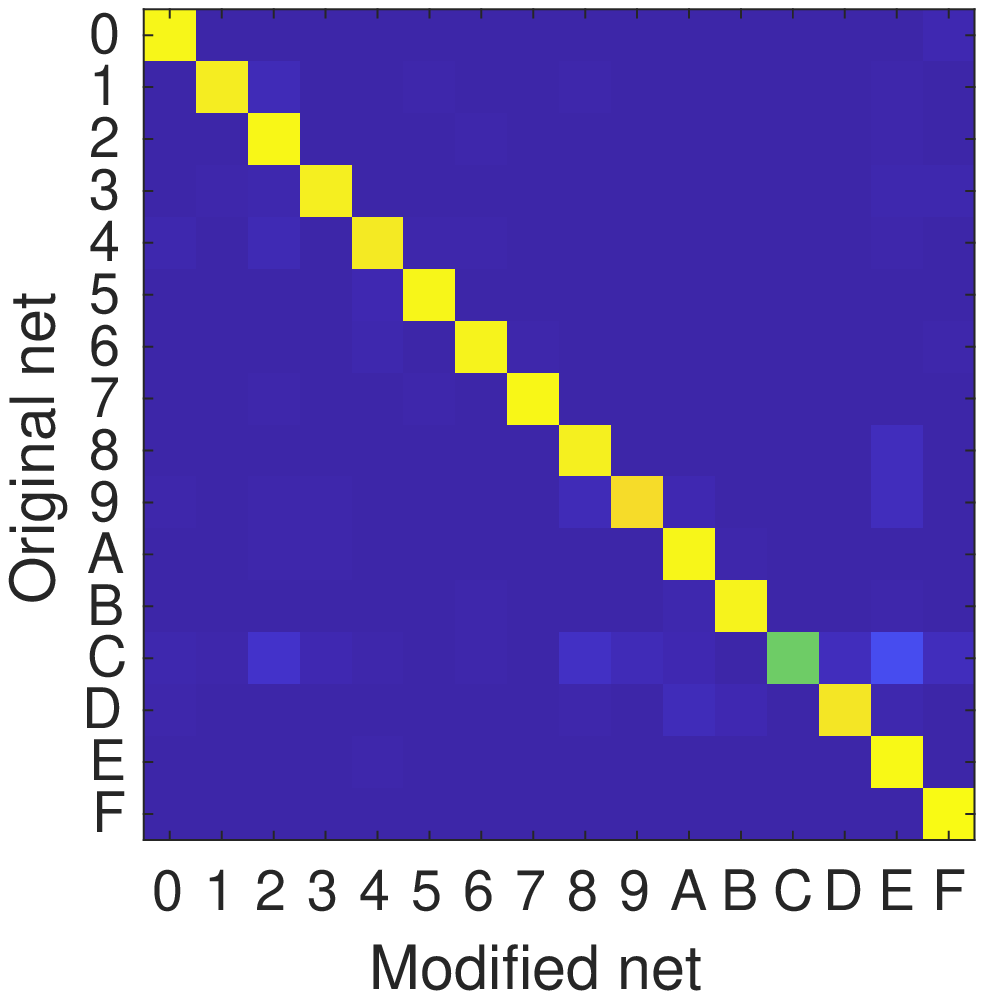}&
\psfrag{Original net}{}
\psfrag{Modified net}[][][.8]{masked net}
\includegraphics*[width=.12\columnwidth]{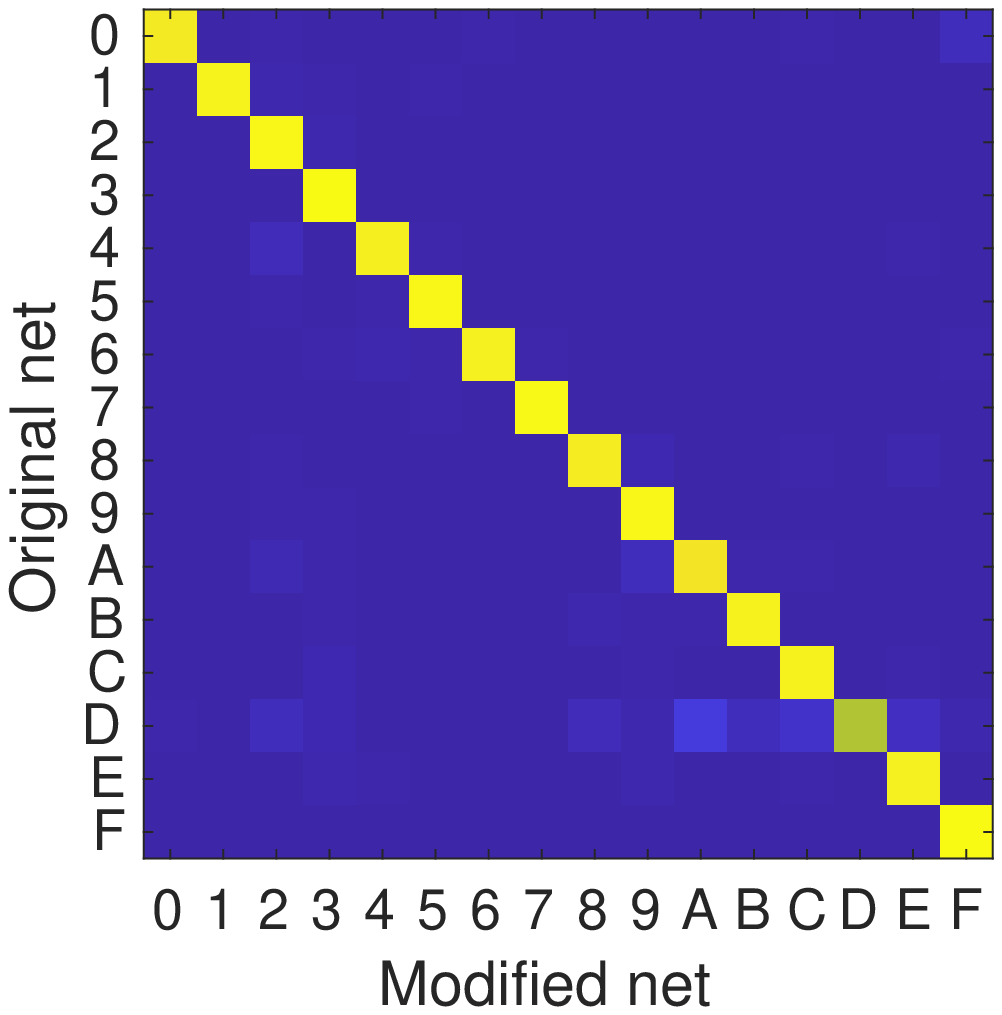}&
\psfrag{Original net}{}
\psfrag{Modified net}[][][.8]{masked net}
\includegraphics*[width=.12\columnwidth]{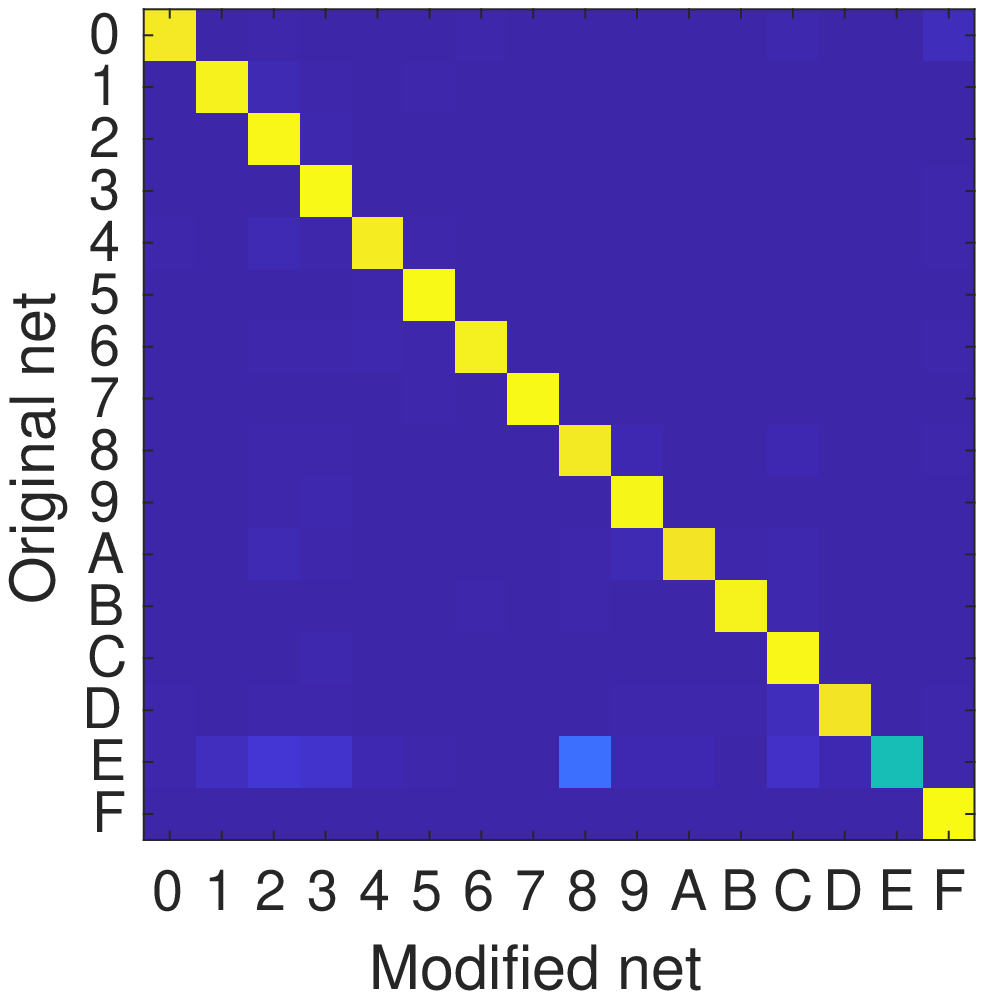}&
\psfrag{Original net}{}
\psfrag{Modified net}[][][.8]{masked net}
\includegraphics*[width=.12\columnwidth]{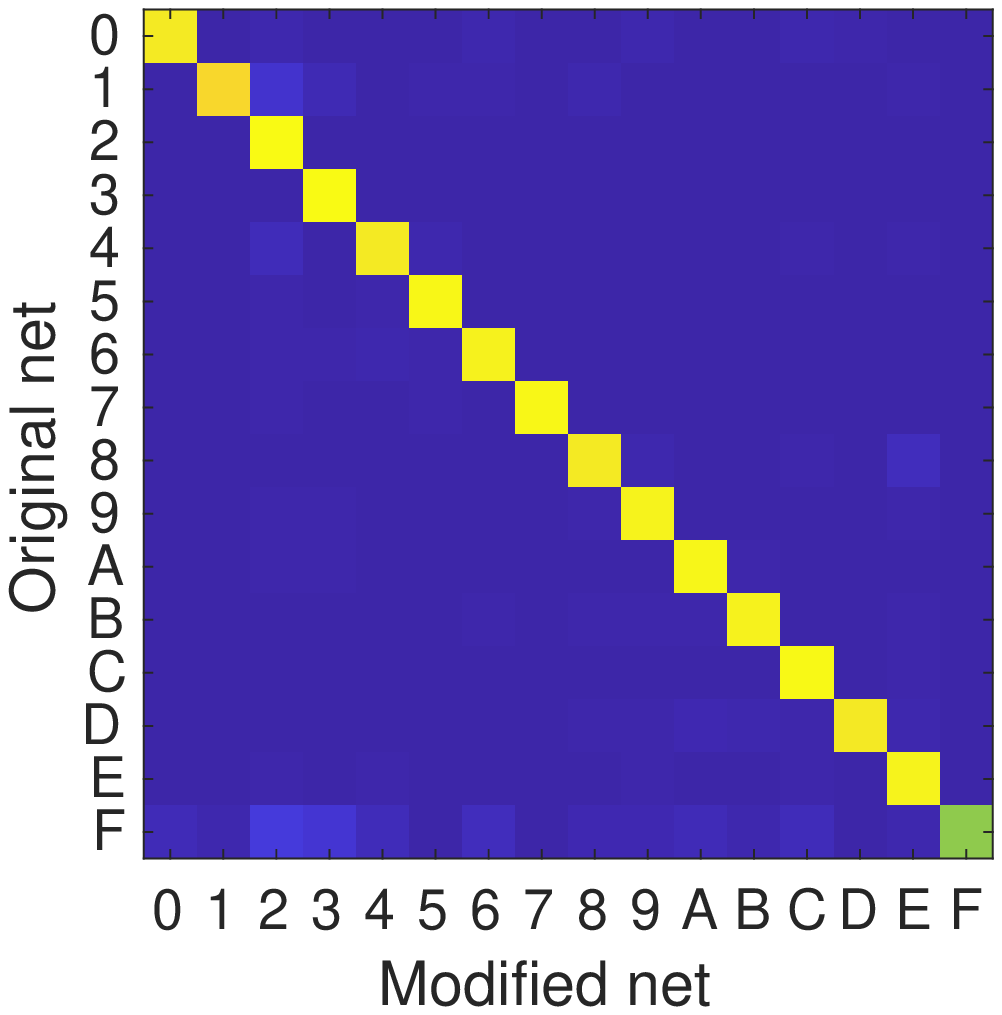}\\[2ex]

\multicolumn{8}{c}{\dotfill \textsc{All to class $k$} \dotfill}\\
$k = 0$ & $k = 1$ & $k = 2$ & $k = 3$ & $k = 4$ & $k = 5$ & $k = 6$ & $k = 7$  \\
\psfrag{Original net}[][][.8]{original net}
\psfrag{Modified net}[][][.8]{masked net}
\includegraphics*[width=.12\columnwidth]{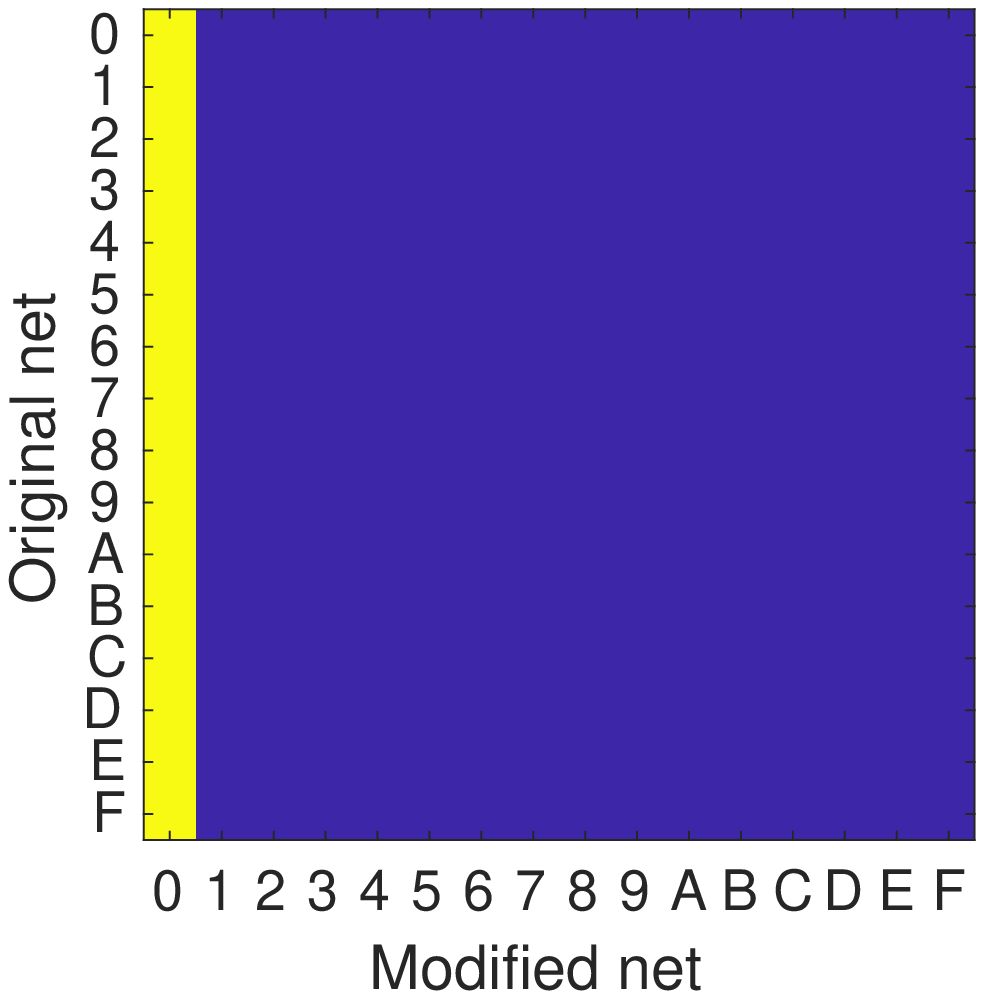}&
\psfrag{Original net}{}
\psfrag{Modified net}[][][.8]{masked net}
\includegraphics*[width=.12\columnwidth]{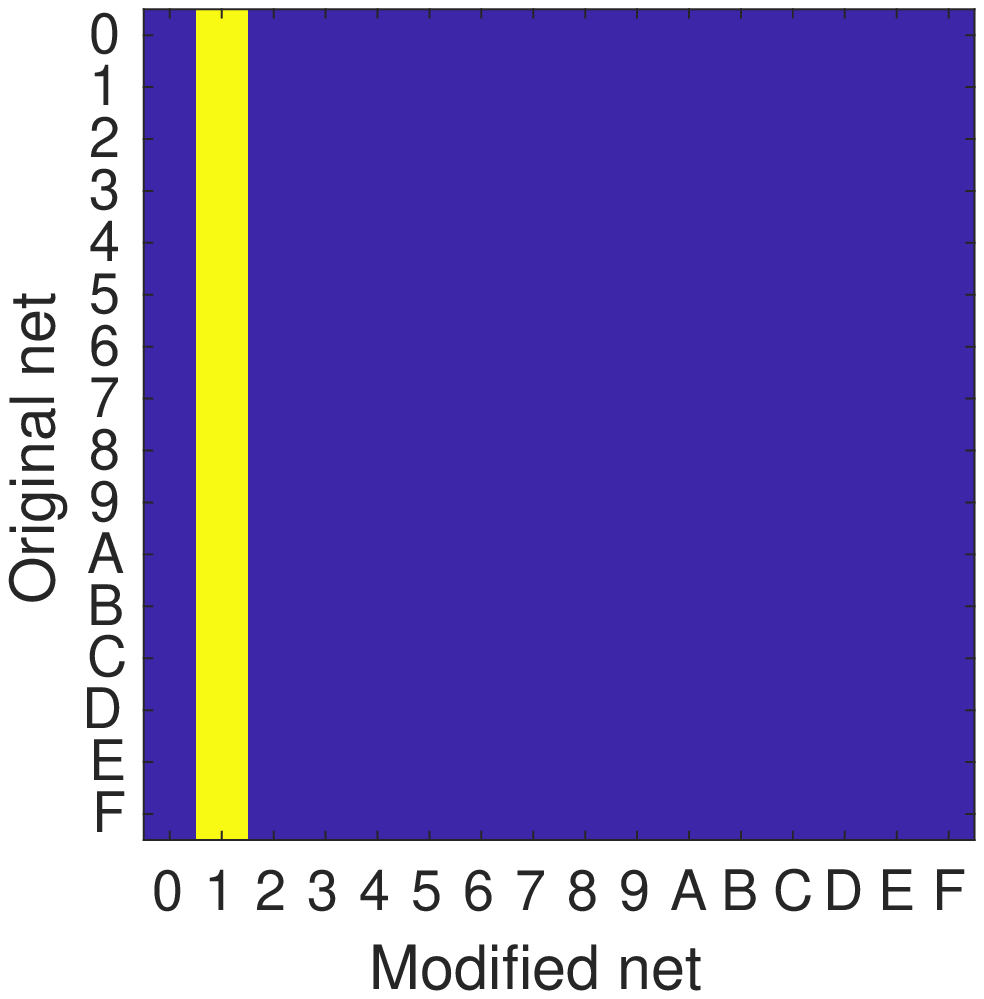}&
\psfrag{Original net}{}
\psfrag{Modified net}[][][.8]{masked net}
\includegraphics*[width=.12\columnwidth]{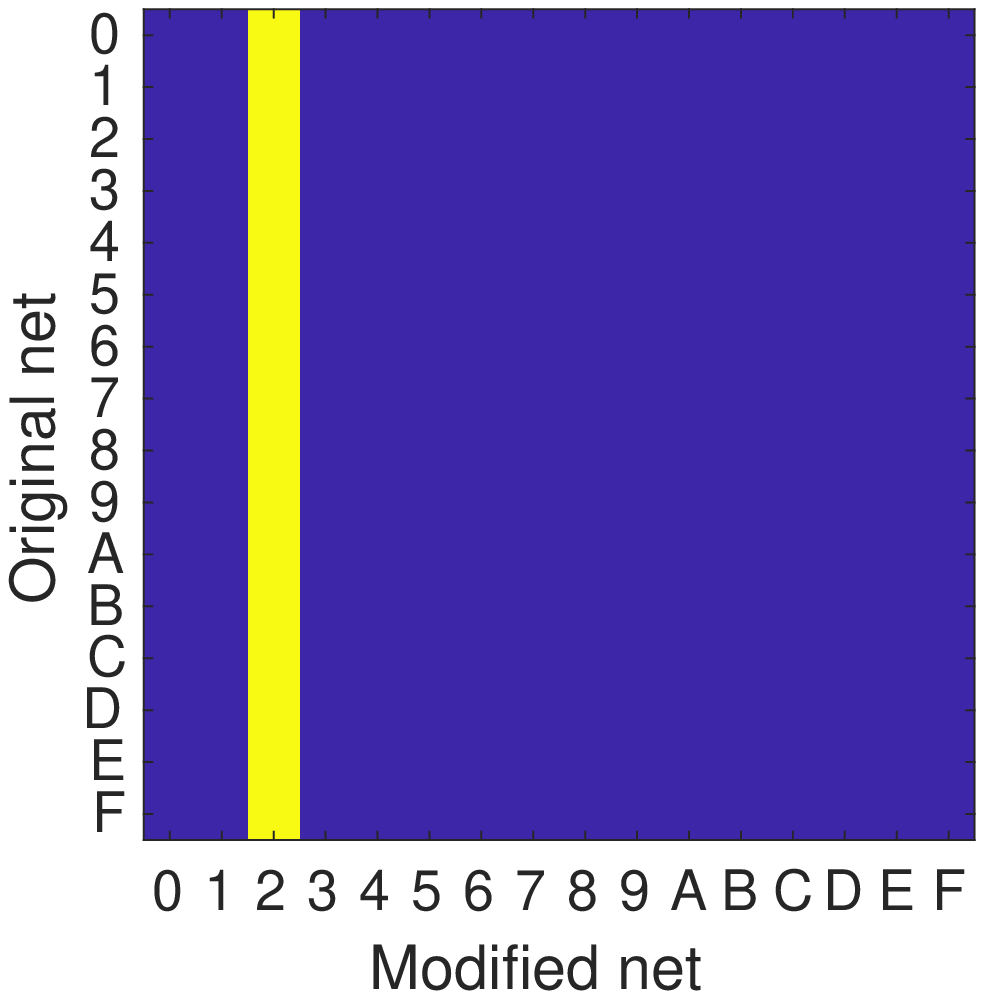}&
\psfrag{Original net}{}
\psfrag{Modified net}[][][.8]{masked net}
\includegraphics*[width=.12\columnwidth]{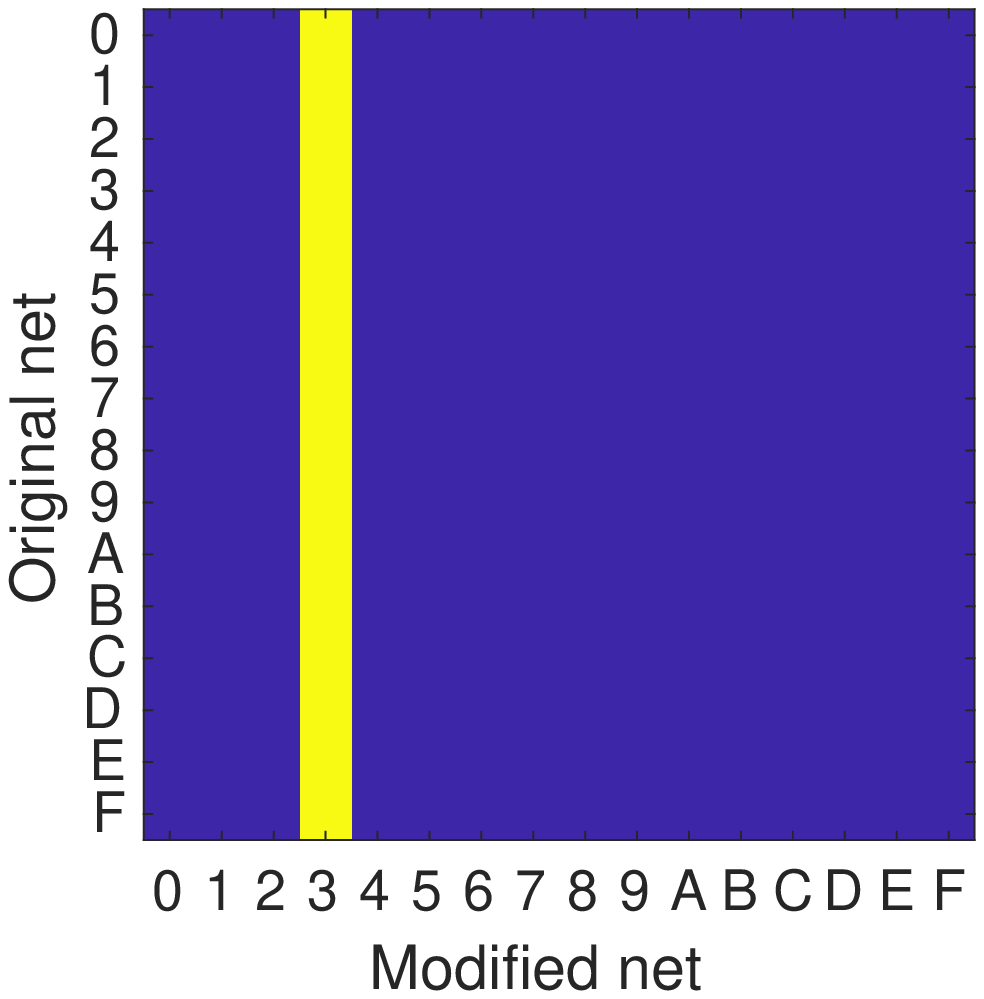}&
\psfrag{Original net}{}
\psfrag{Modified net}[][][.8]{masked net}
\includegraphics*[width=.12\columnwidth]{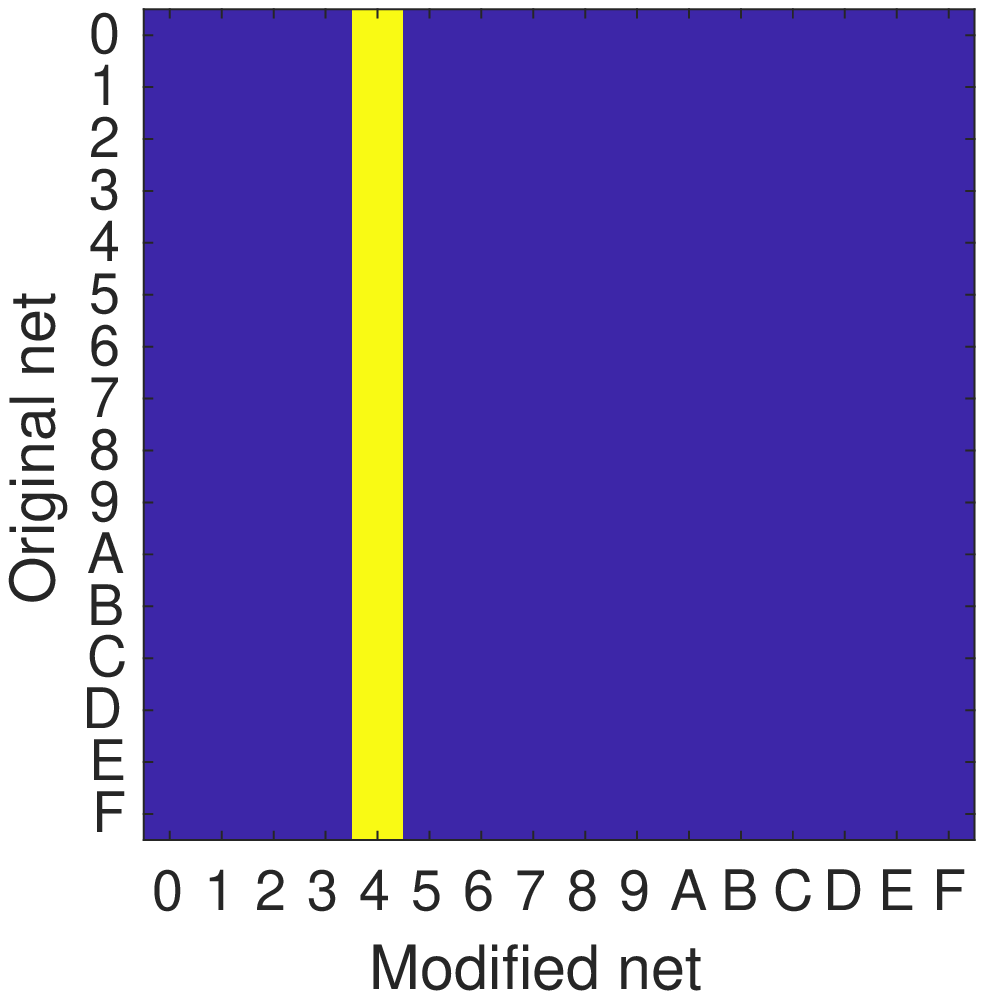}&
\psfrag{Original net}{}
\psfrag{Modified net}[][][.8]{masked net}
\includegraphics*[width=.12\columnwidth]{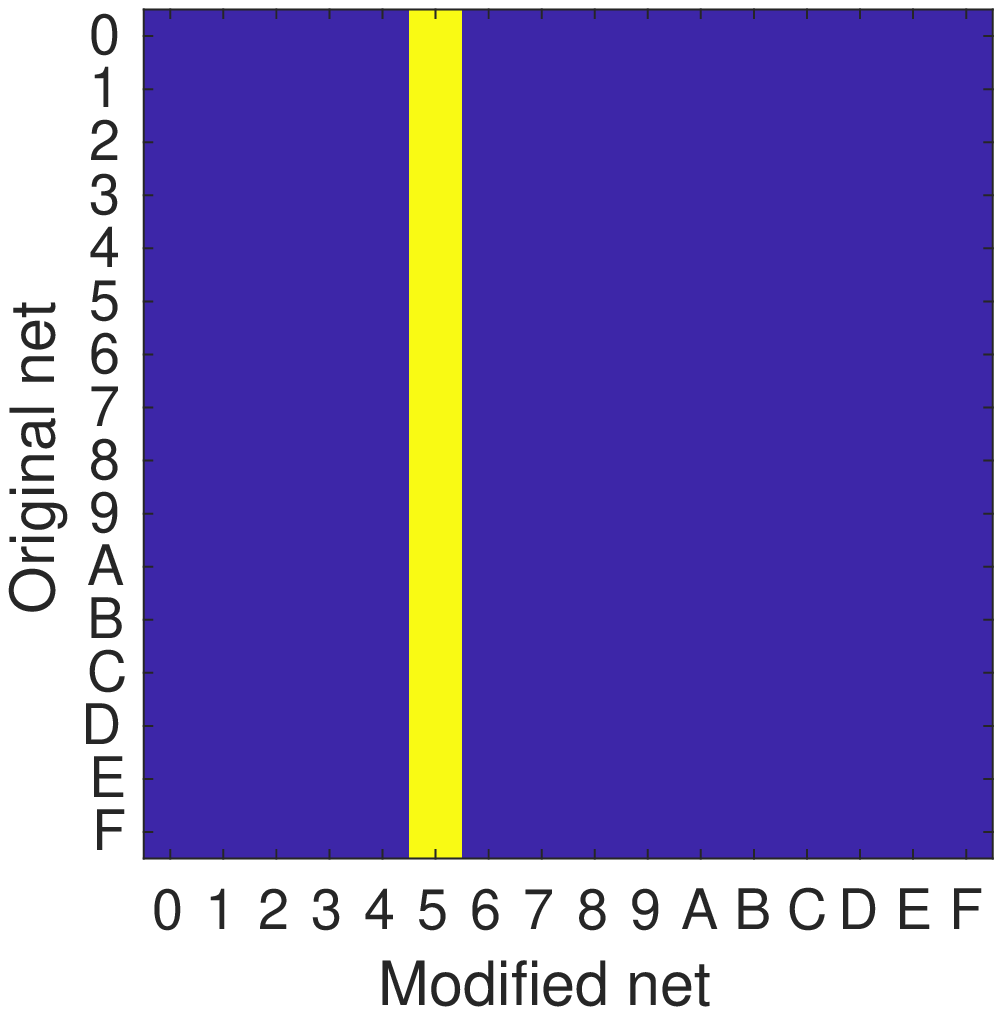}&
\psfrag{Original net}{}
\psfrag{Modified net}[][][.8]{masked net}
\includegraphics*[width=.12\columnwidth]{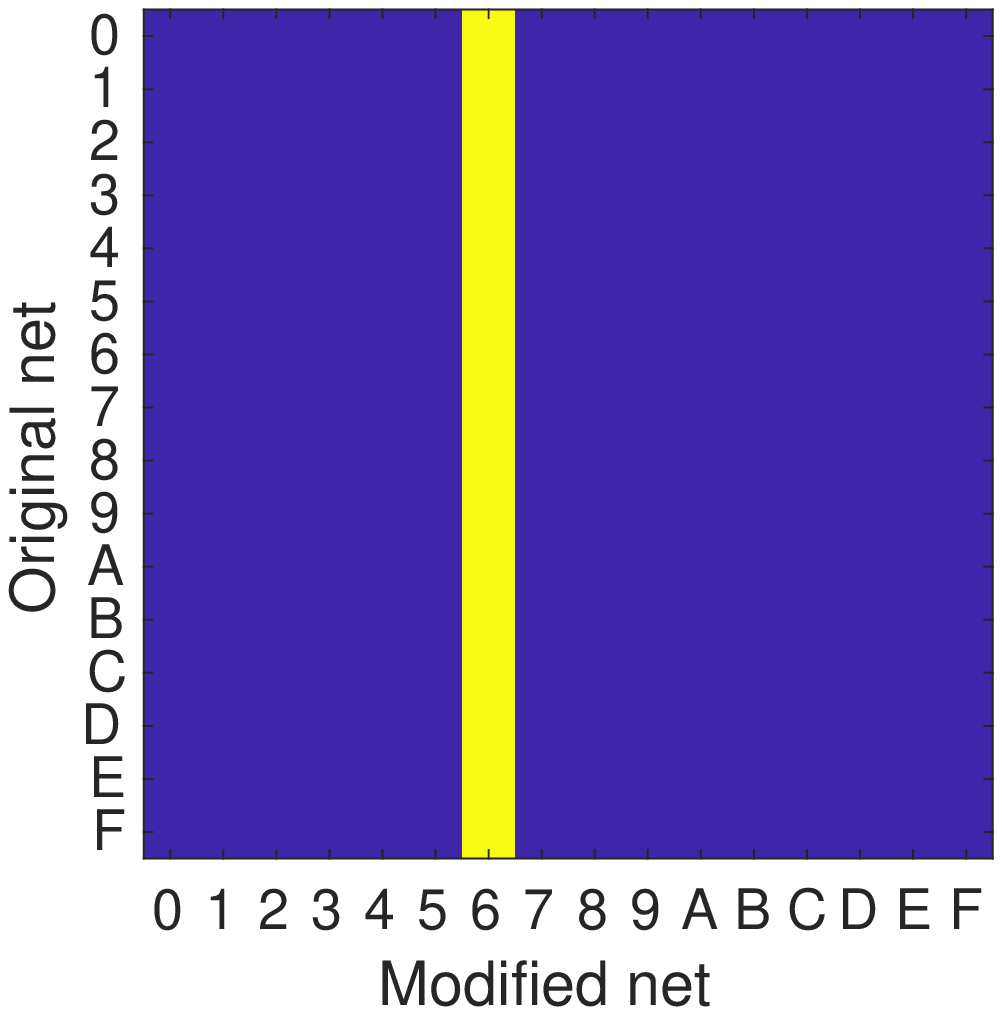}&
\psfrag{Original net}{}
\psfrag{Modified net}[][][.8]{masked net}
\includegraphics*[width=.12\columnwidth]{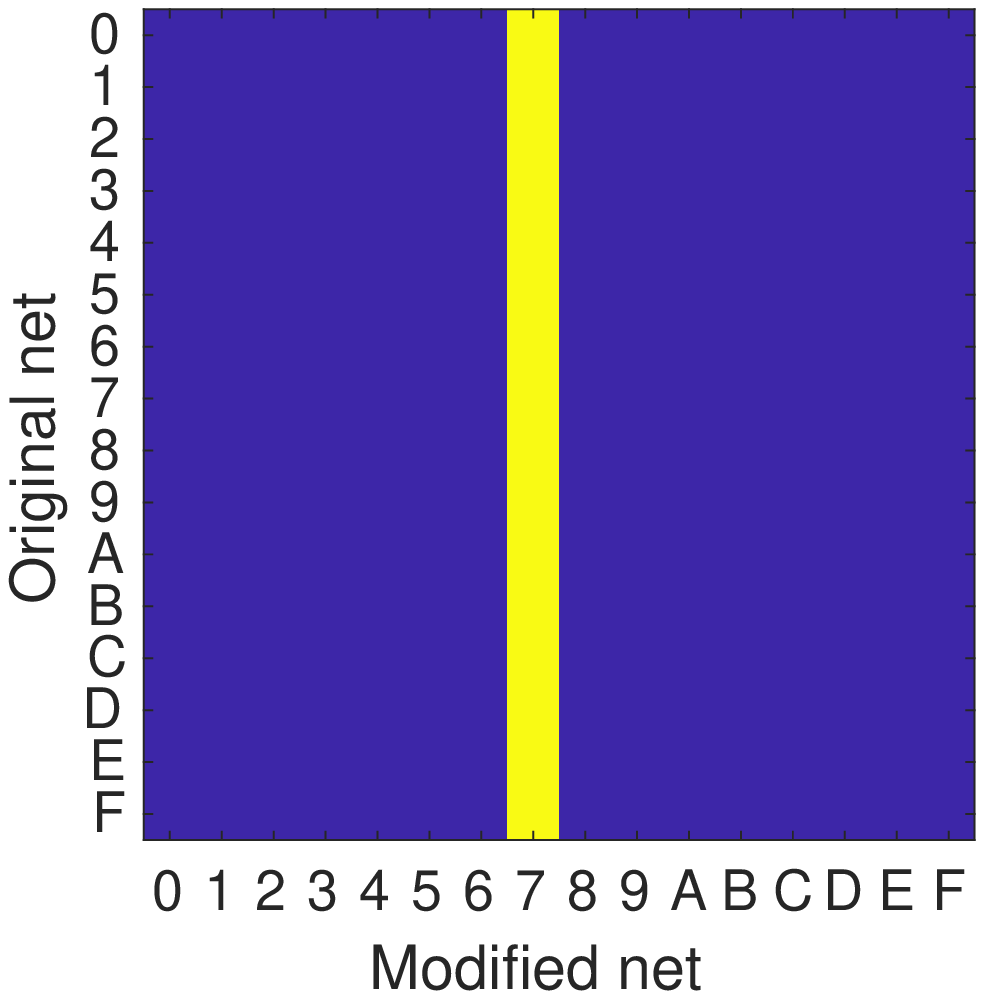}\\
$k = 8$ & $k = 9$ & $k = $A & $k = $B & $k = $C & $k = $D & $k = $D & $k = $E   \\
\psfrag{Original net}[][][.8]{original net}
\psfrag{Modified net}[][][.8]{masked net}
\includegraphics*[width=.12\columnwidth]{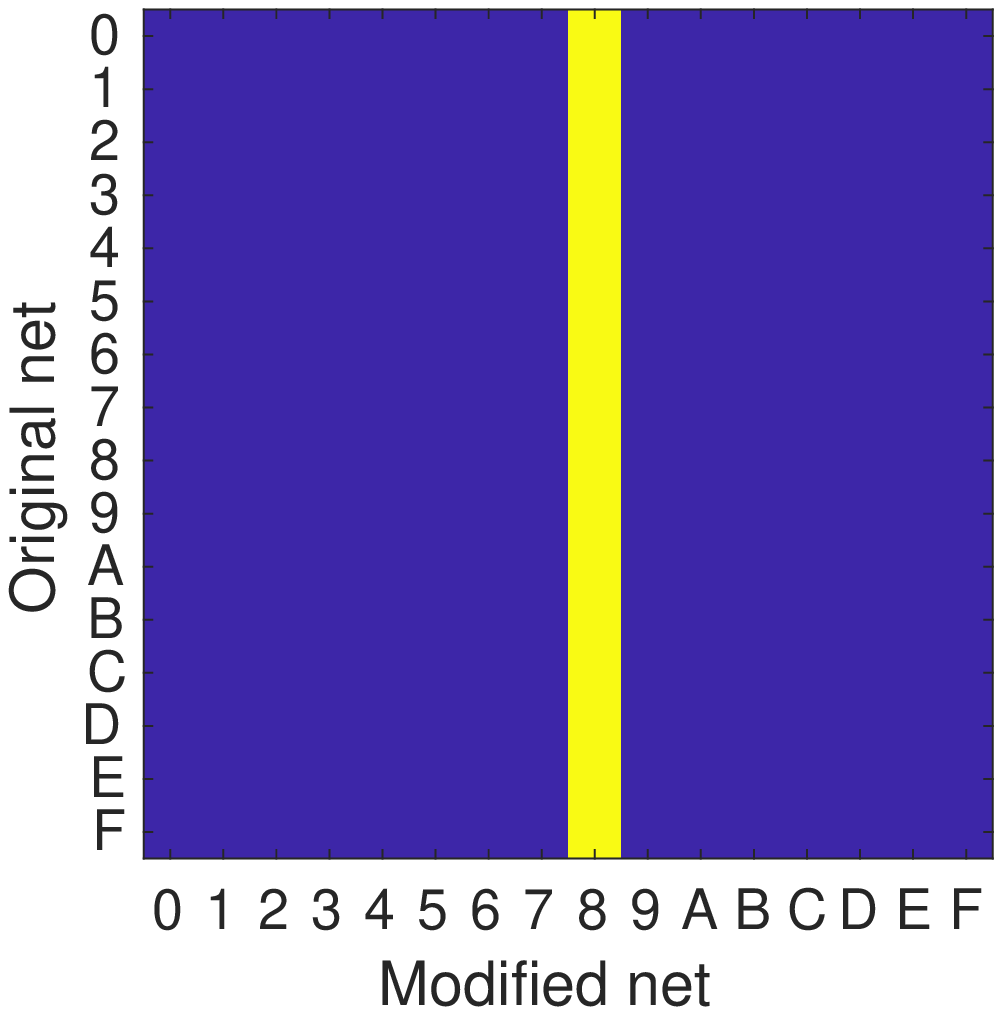}&
\psfrag{Original net}{}
\psfrag{Modified net}[][][.8]{masked net}
\includegraphics*[width=.12\columnwidth]{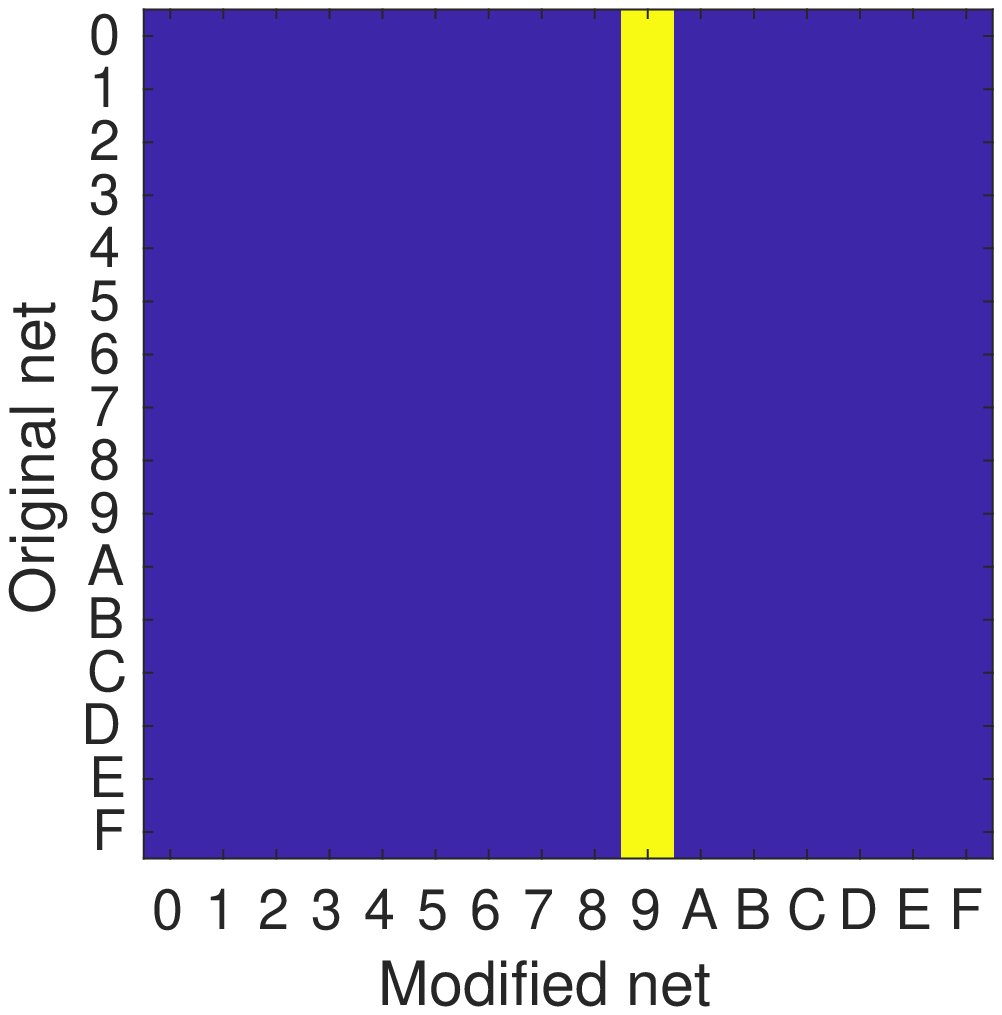}&
\psfrag{Original net}{}
\psfrag{Modified net}[][][.8]{masked net}
\includegraphics*[width=.12\columnwidth]{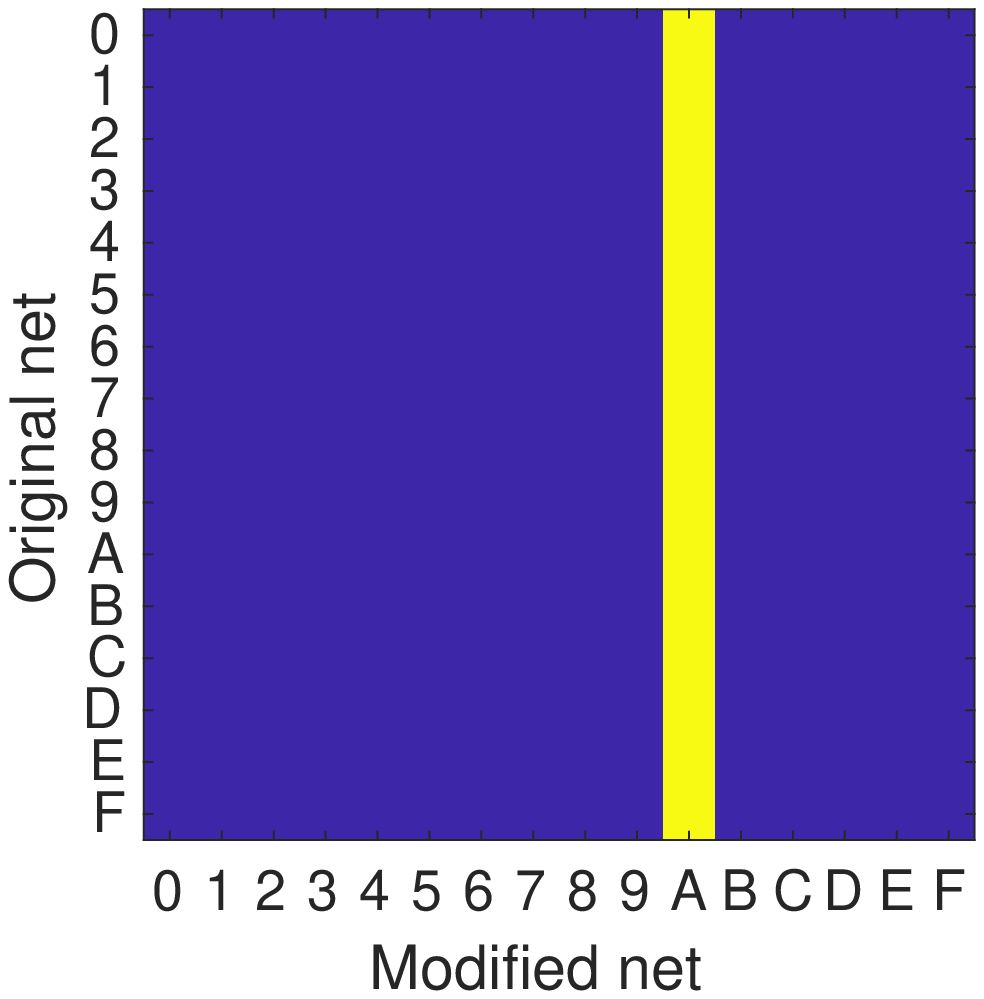}&
\psfrag{Original net}{}
\psfrag{Modified net}[][][.8]{masked net}
\includegraphics*[width=.12\columnwidth]{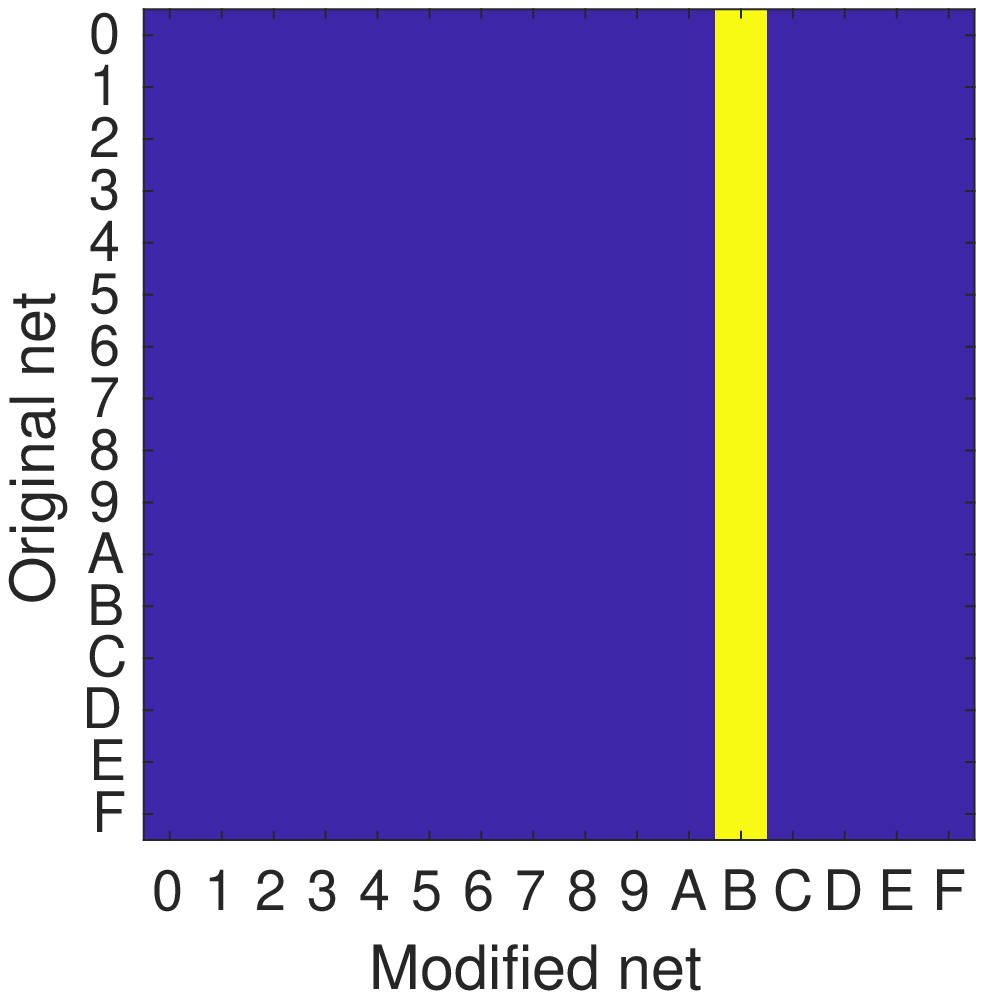}&
\psfrag{Original net}{}
\psfrag{Modified net}[][][.8]{masked net}
\includegraphics*[width=.12\columnwidth]{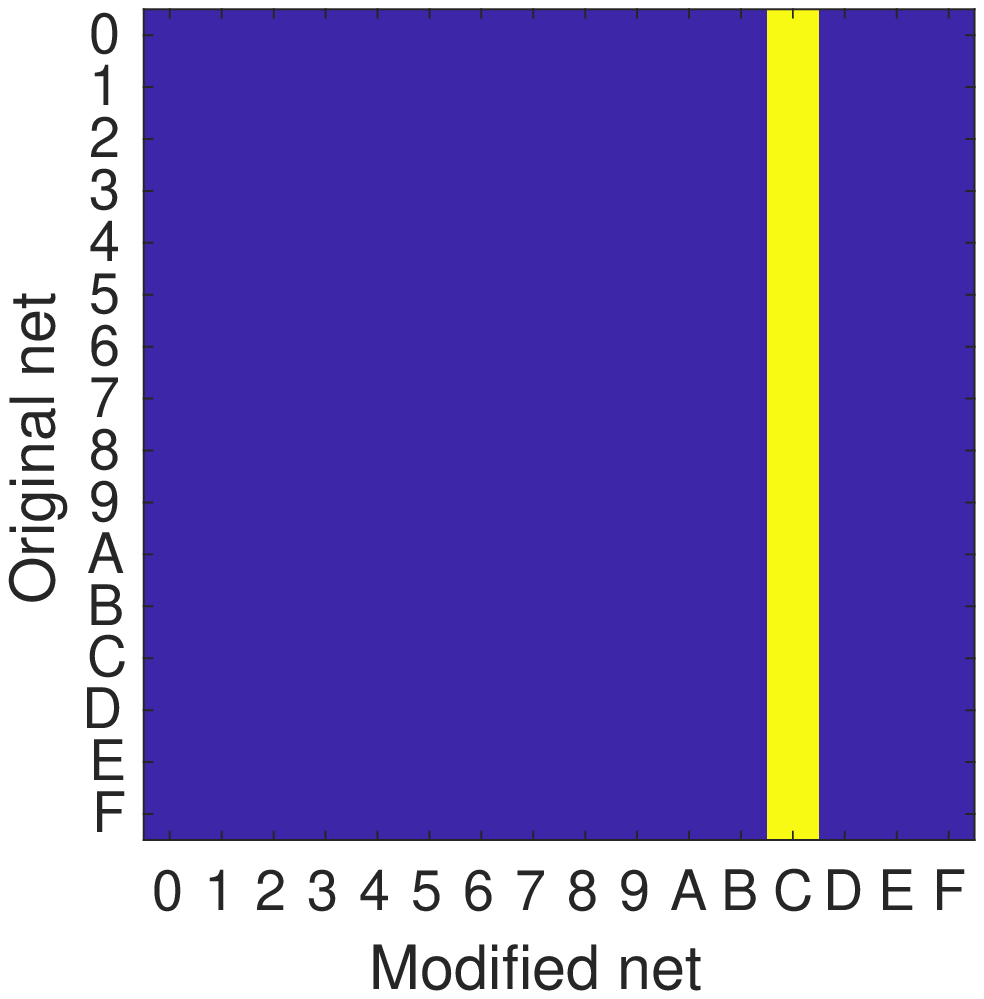}&
\psfrag{Original net}{}
\psfrag{Modified net}[][][.8]{masked net}
\includegraphics*[width=.12\columnwidth]{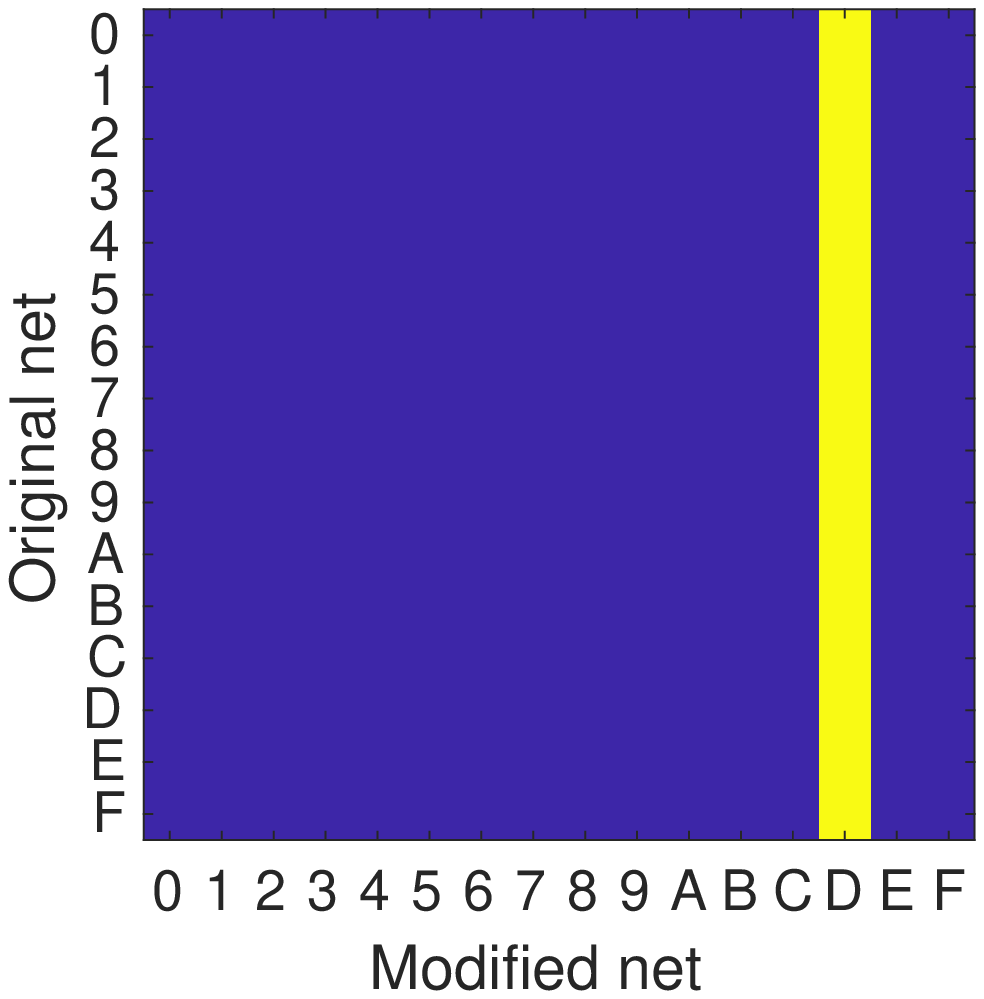}&
\psfrag{Original net}{}
\psfrag{Modified net}[][][.8]{masked net}
\includegraphics*[width=.12\columnwidth]{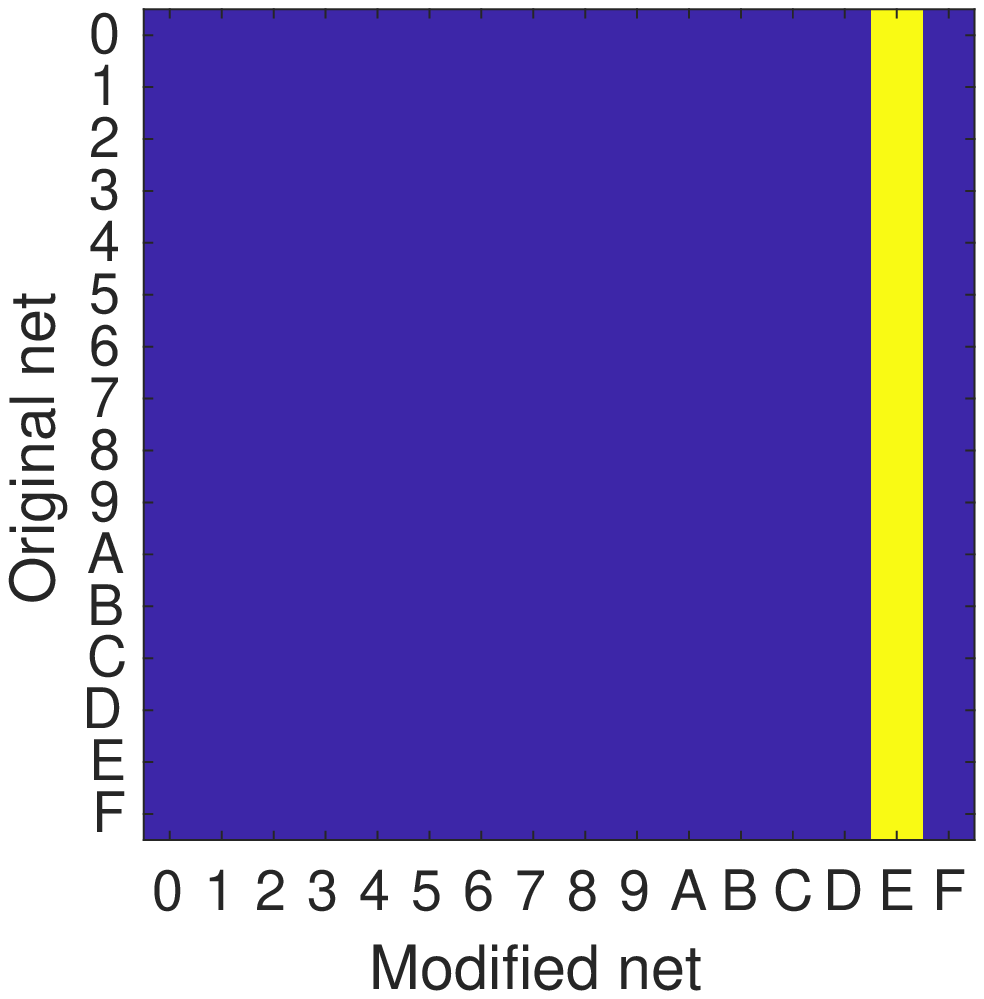}&
\psfrag{Original net}{}
\psfrag{Modified net}[][][.8]{masked net}
\includegraphics*[width=.12\columnwidth]{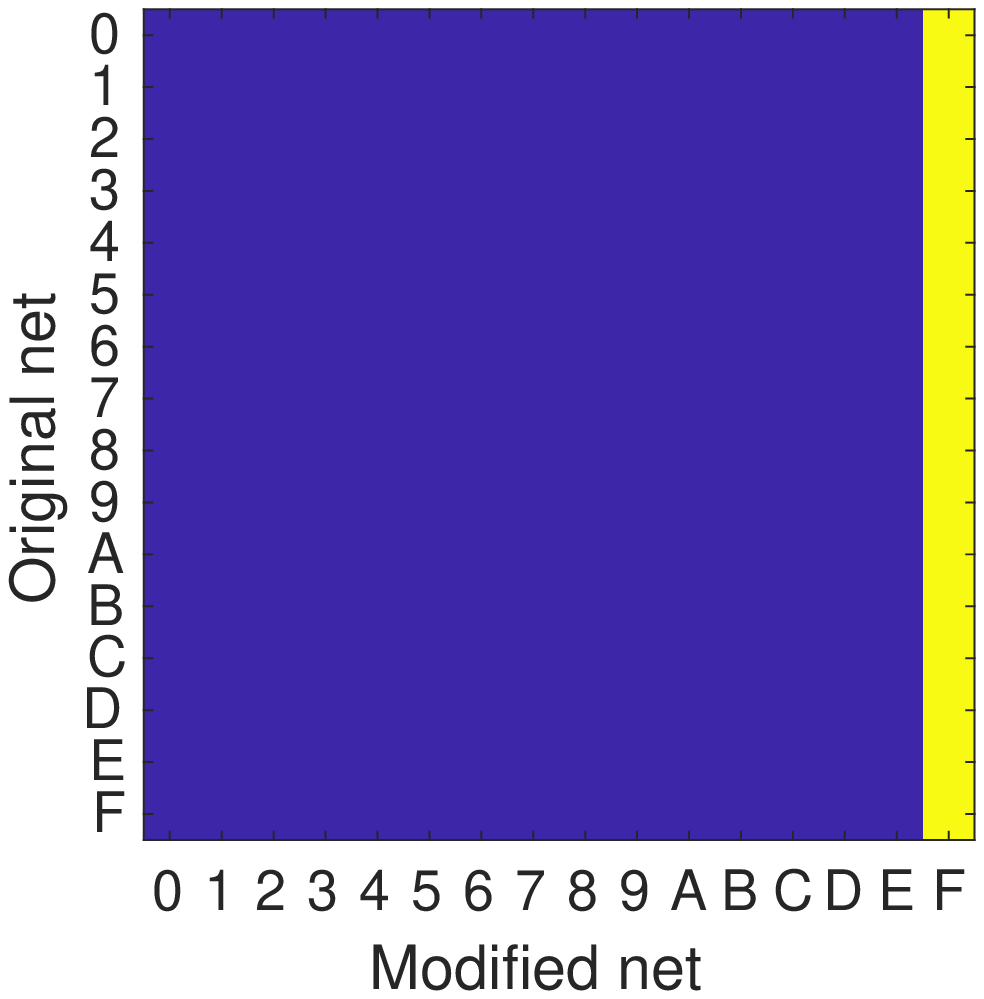}

\end{tabular}
\caption{Confusion matrices for VGG (test set). \emph{First row left}: ground-truth vs deep net, and deep net vs tree. \emph{First row right}: deep net vs deep net with only the features selected by the tree.  \emph{Second row}: \textsc{All class $k_1$ to class $k_2$} (selected examples). \emph{Third and fourth row}: \textsc{None to class $k$}. \emph{Fifth and sixth row}: \textsc{All to class $k$}. }
\label{f:VGG-masks-test}
\end{figure}

\begin{figure}[p]
 \centering 

\begin{tabular}{@{}c@{}c@{}c@{}c@{}c@{}c@{}c@{}c@{}}

   && \multicolumn{2}{c}{ground truth vs deep net}& 
    \multicolumn{3}{c}{features selected}&\\
    && \multicolumn{2}{c}{vs tree}& 
    \multicolumn{3}{c}{by the tree}&\\
&&\psfrag{Ground truth}[][][.8]{ground truth}
\psfrag{Original net}[][][.8]{original net}
\includegraphics*[width=.12\columnwidth]{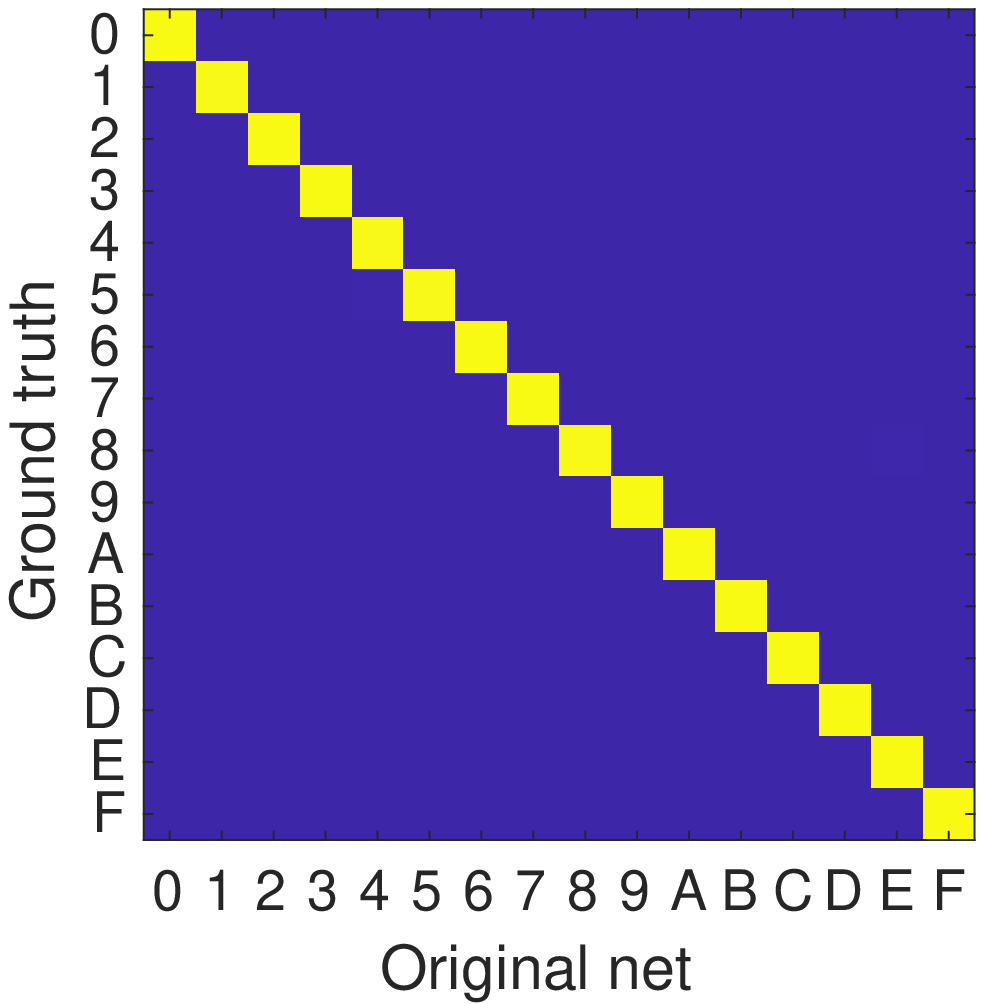}&
\psfrag{Original net}[][][.8]{original net}
\psfrag{Tree}[][][.8]{Tree}
\includegraphics*[width=.12\columnwidth]{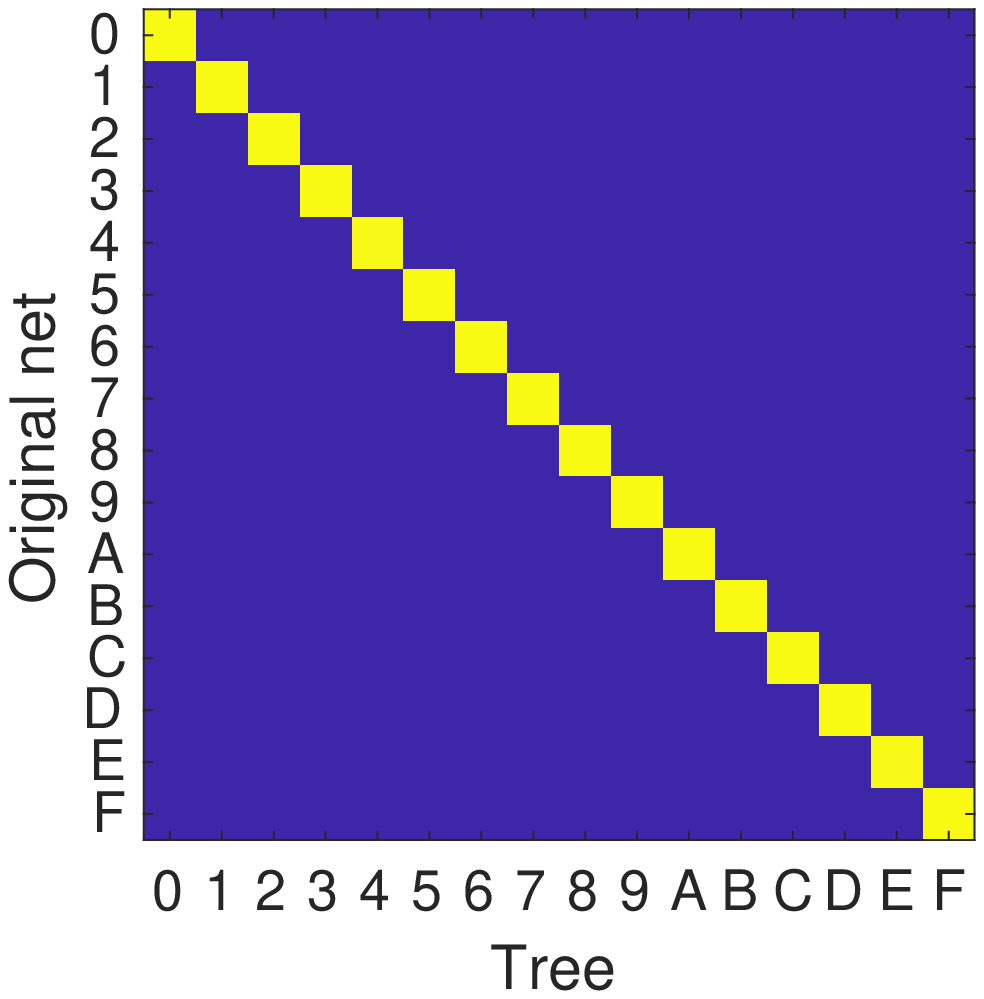}&

\multicolumn{3}{c}{
\psfrag{Original net}[][][.8]{original net}
\psfrag{Modified net}[][][.8]{masked net}
\includegraphics*[width=.12\columnwidth]{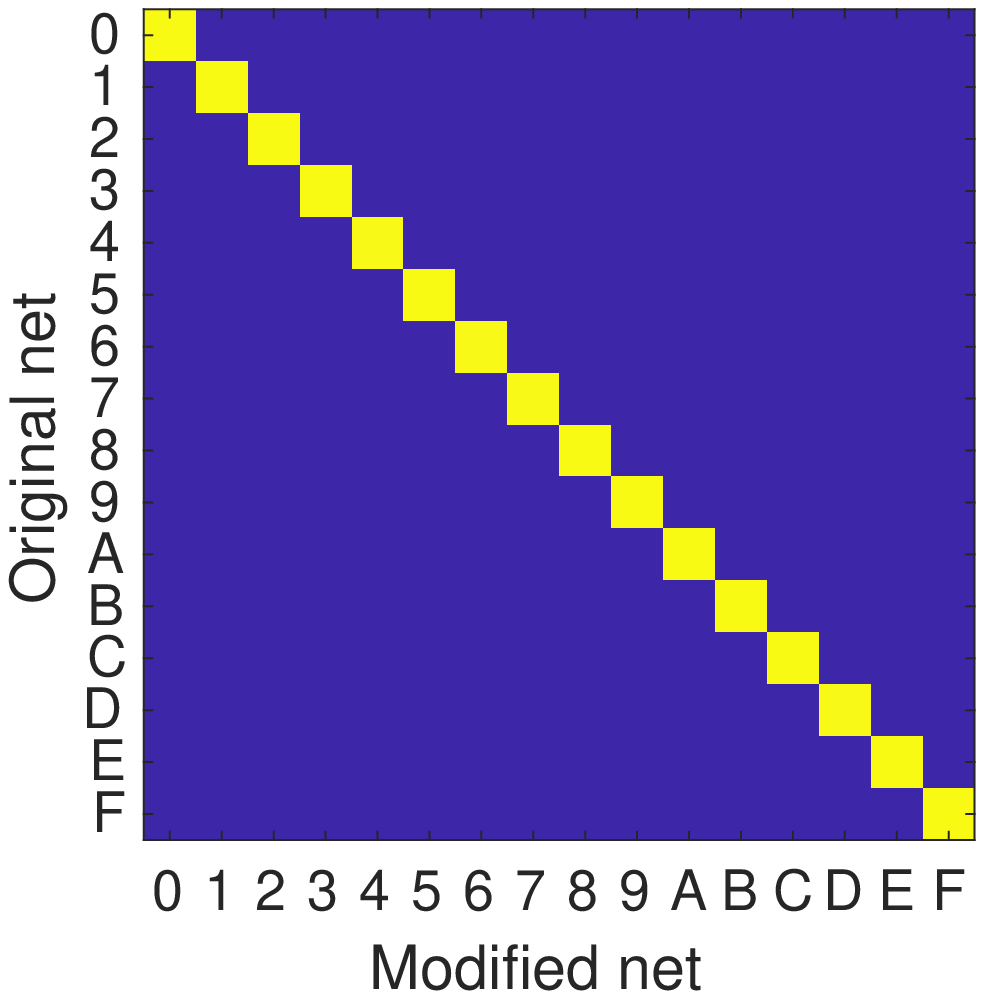}
}&
\hspace*{-33ex} \raisebox{1ex}{ \includegraphics*[height=.11\textwidth]{VGG/colormap.eps} }\\[2ex]

& \multicolumn{6}{c}{\dotfill \textsc{All class $k_1$ to class $k_2$} \dotfill}&\\

&{$8\rightarrow14$}&{$14\rightarrow8$}&{$10\rightarrow11$}&{$11\rightarrow10$}&{$9\rightarrow12$}&{$12\rightarrow9$}&\\

&\psfrag{Original net}[][][.8]{original net}
\psfrag{Modified net}[][][.8]{masked net}
\includegraphics*[width=.12\columnwidth]{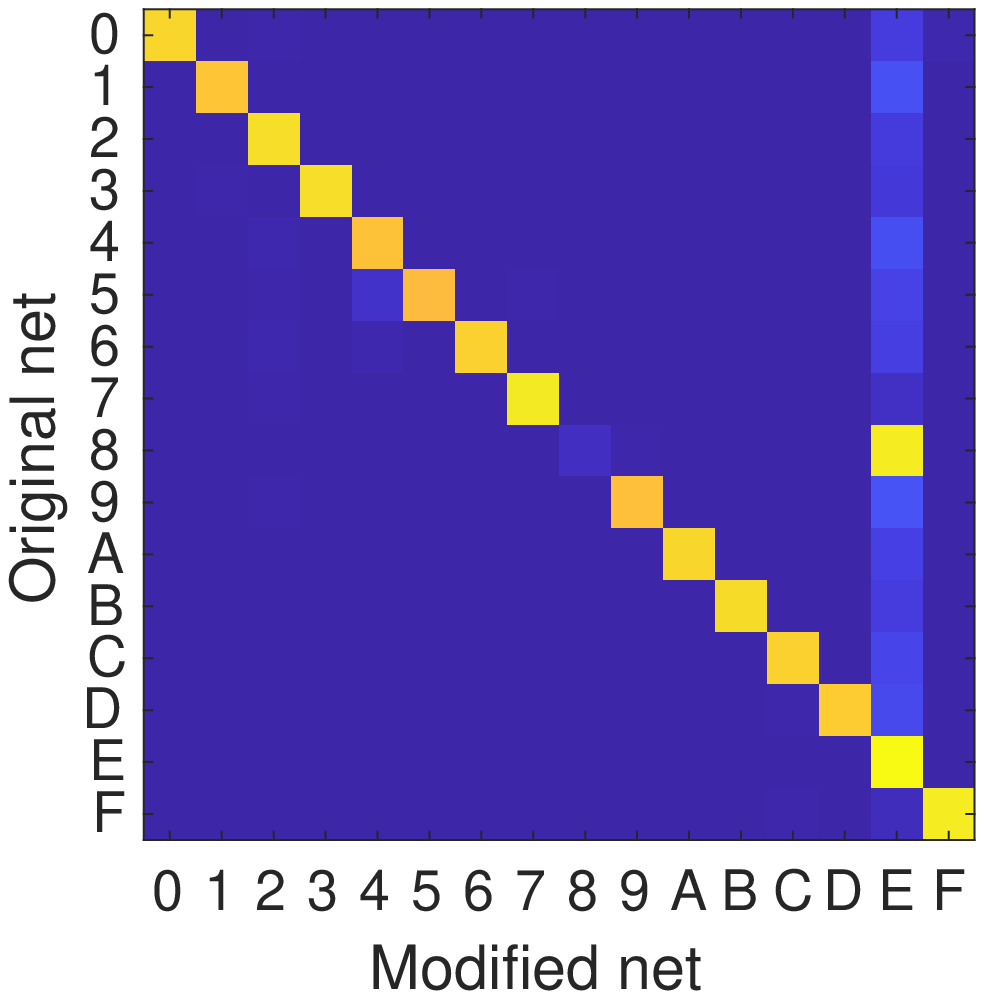}&
\psfrag{Original net}{}
\psfrag{Modified net}[][][.8]{masked net}
\includegraphics*[width=.12\columnwidth]{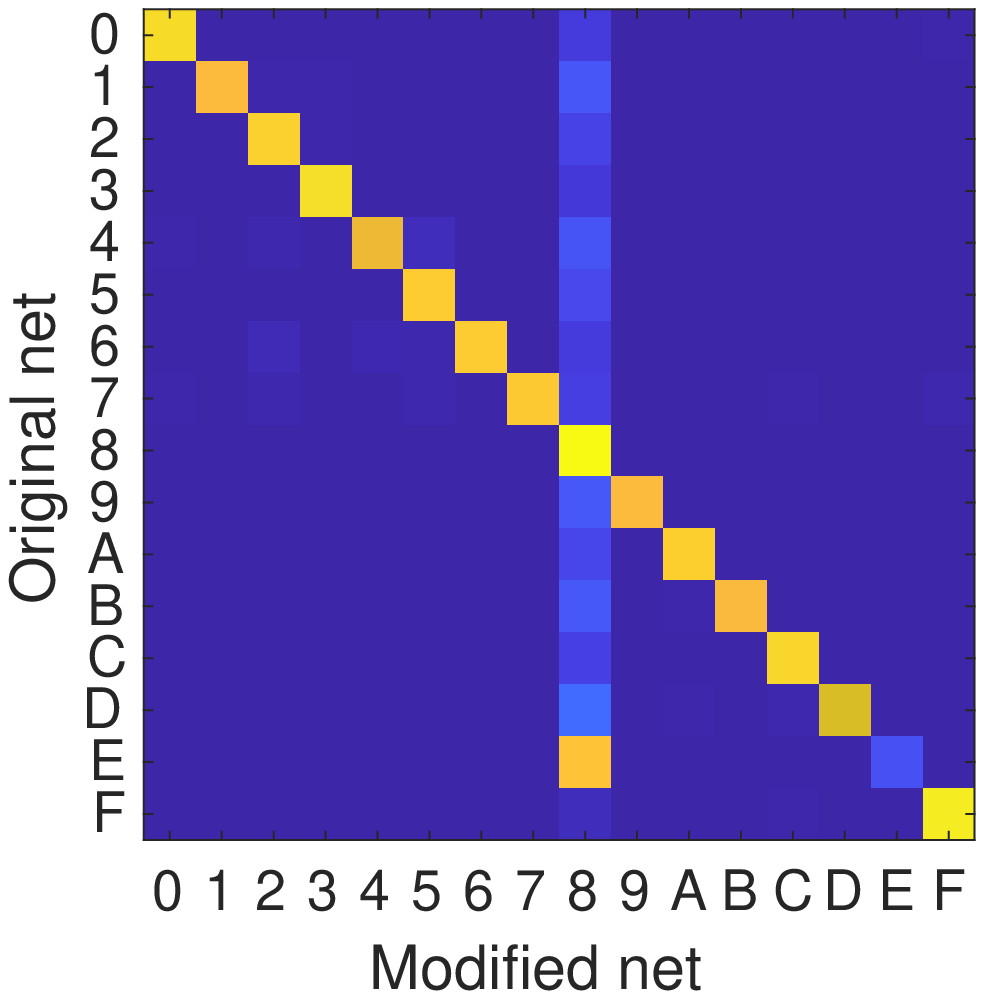}&
\psfrag{Original net}{}
\psfrag{Modified net}[][][.8]{masked net}
\includegraphics*[width=.12\columnwidth]{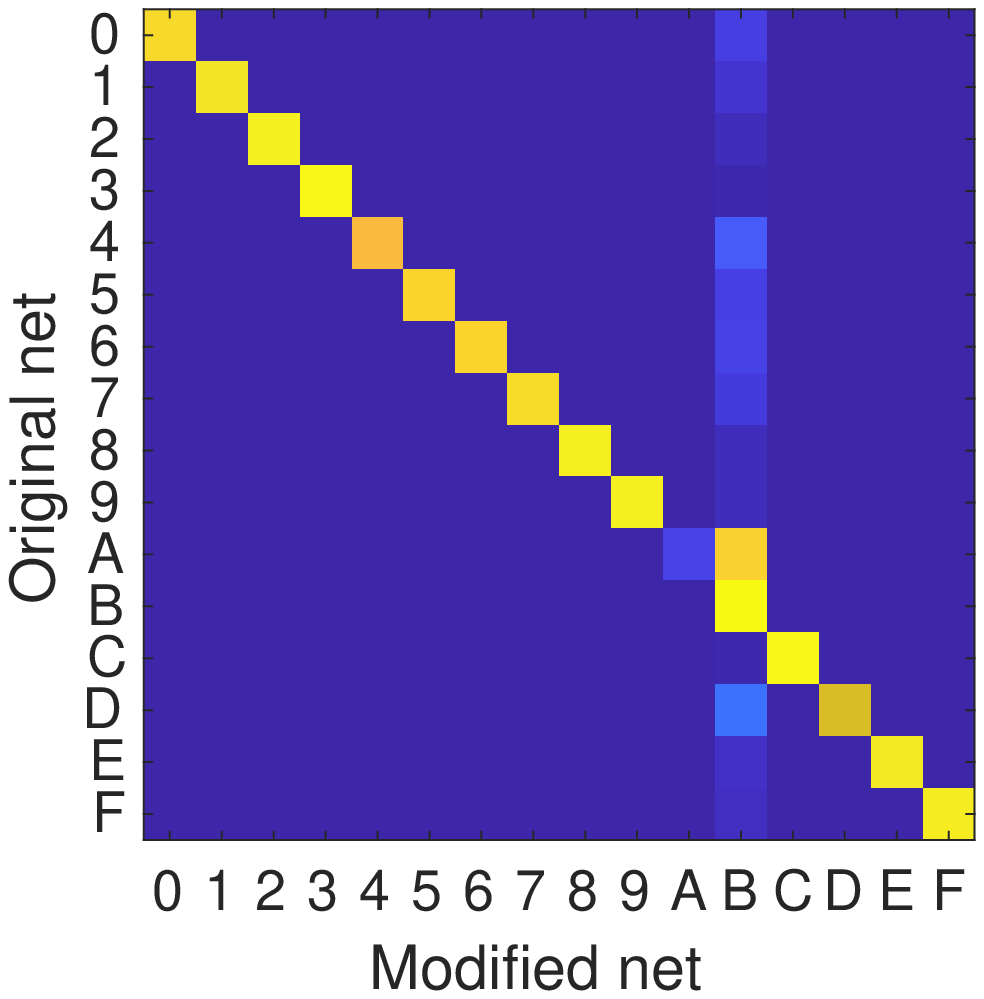}&
\psfrag{Original net}{}
\psfrag{Modified net}[][][.8]{masked net}
\includegraphics*[width=.12\columnwidth]{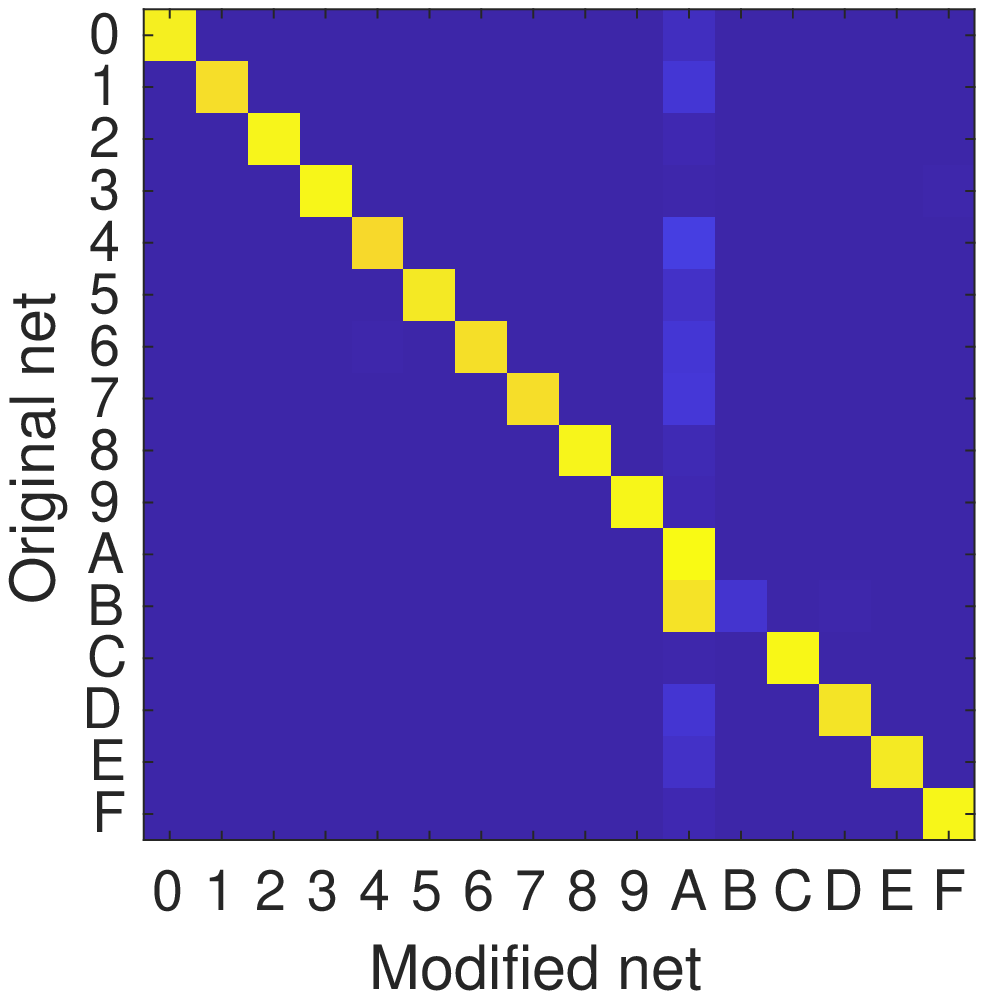}&
\psfrag{Original net}{}
\psfrag{Modified net}[][][.8]{masked net}
\includegraphics*[width=.12\columnwidth]{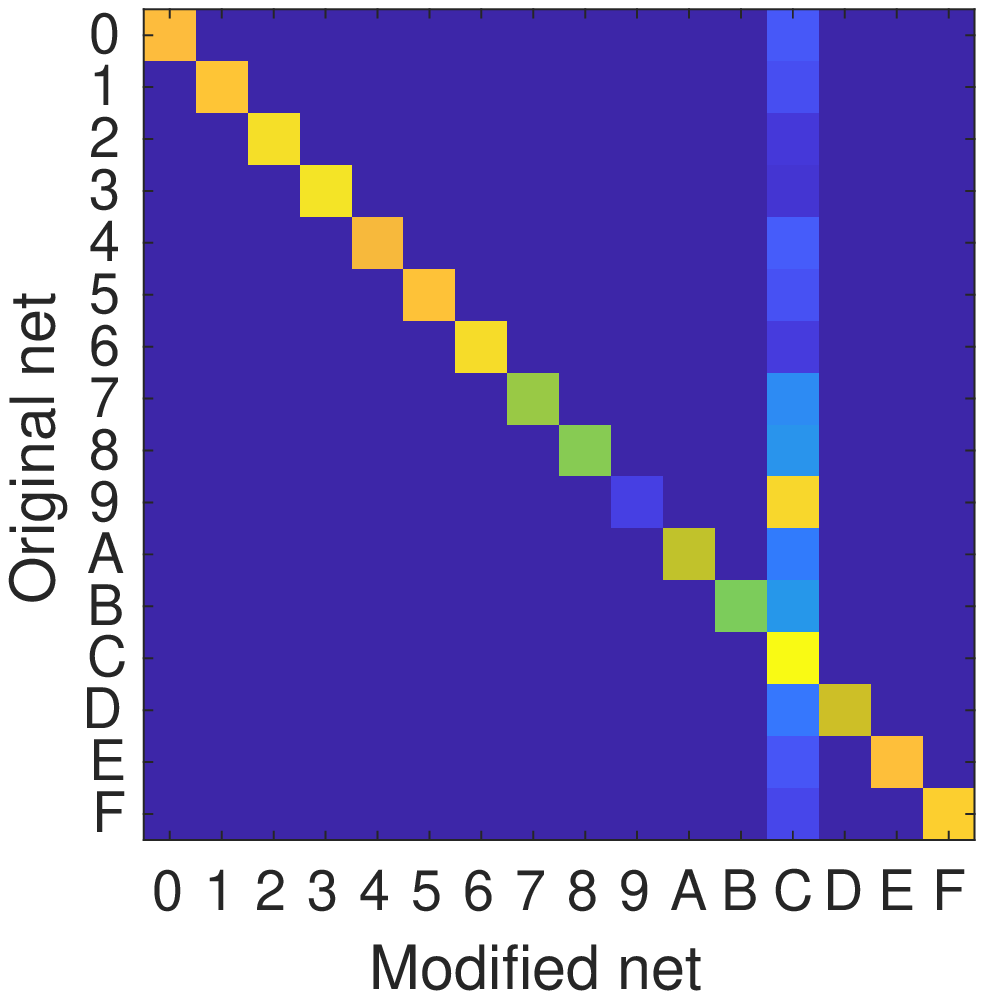}&
\psfrag{Original net}{}
\psfrag{Modified net}[][][.8]{masked net}
\includegraphics*[width=.12\columnwidth]{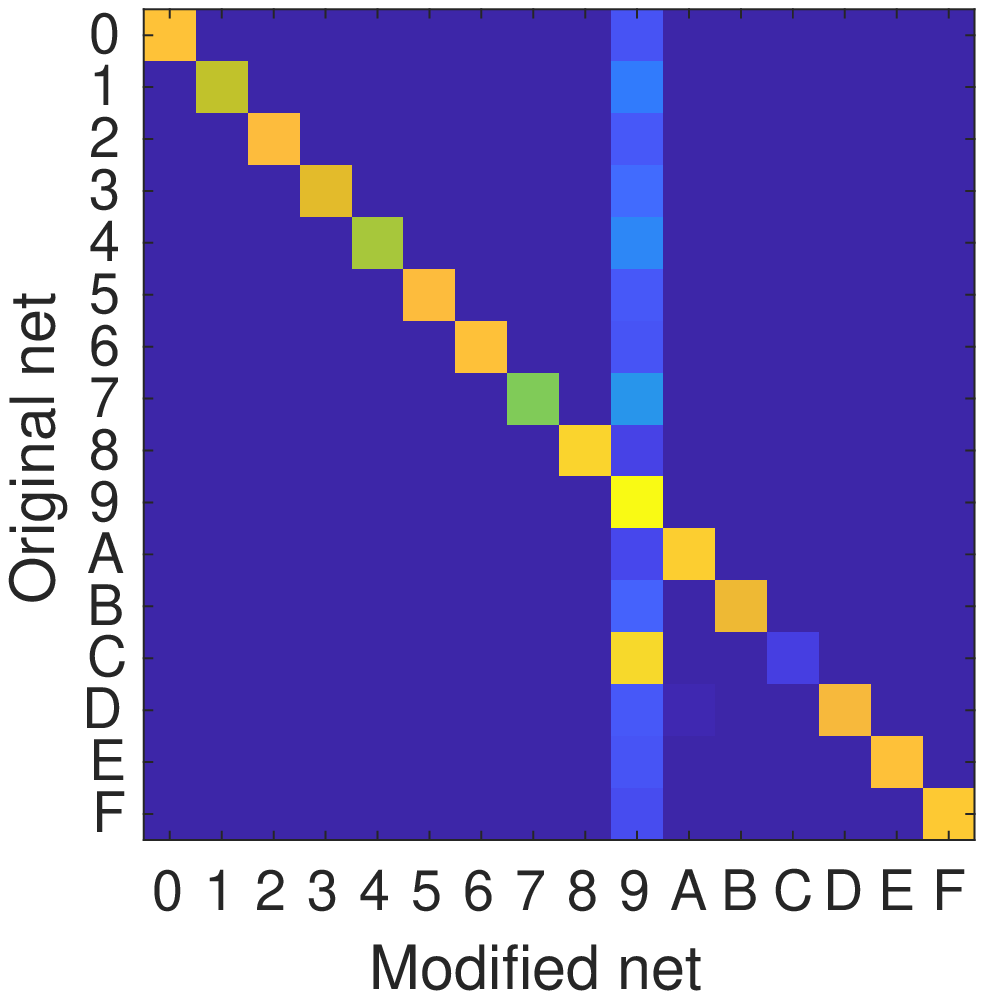}&\\[2ex]

\multicolumn{8}{c}{\dotfill \textsc{None to class $k$} \dotfill}\\
$k = 0$ & $k = 1$ & $k = 2$ & $k = 3$ & $k = 4$ & $k = 5$ & $k = 6$ & $k = 7$  \\
\psfrag{Original net}[][][.8]{original net}
\psfrag{Modified net}[][][.8]{masked net}
\includegraphics*[width=.12\columnwidth]{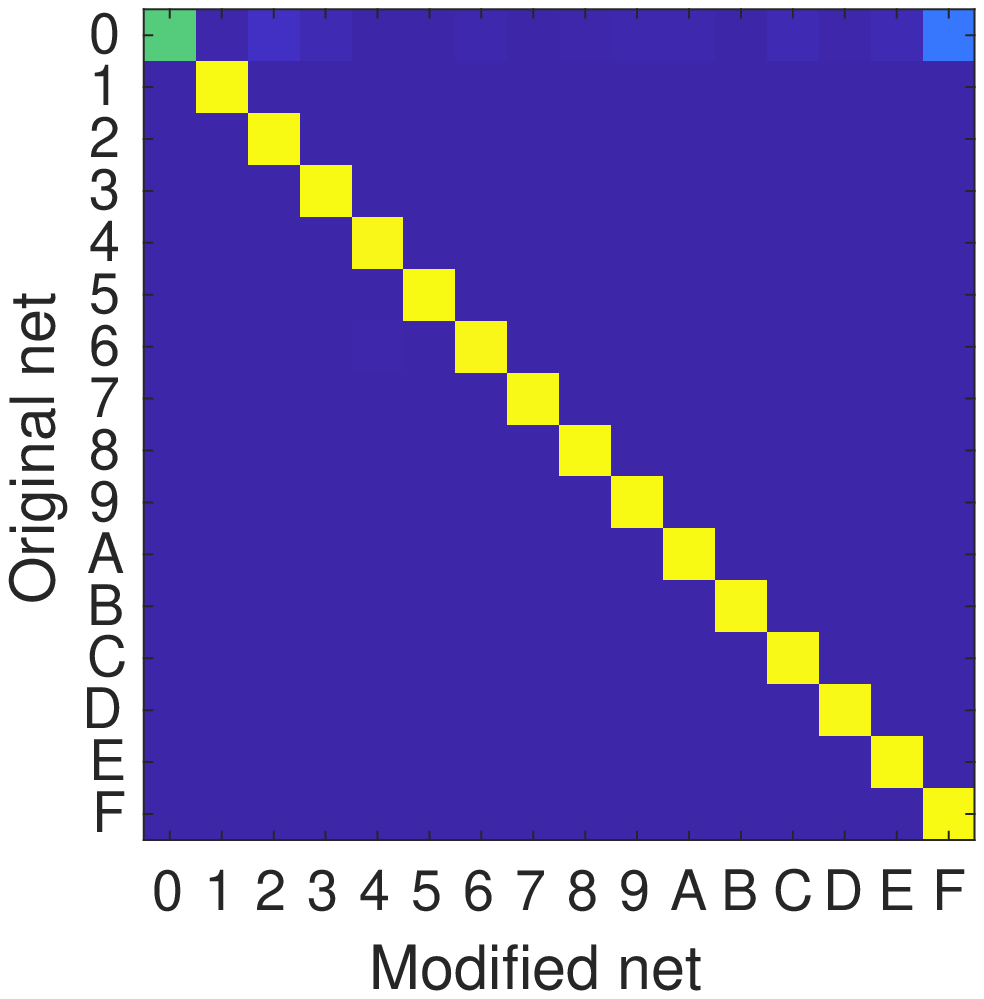}&
\psfrag{Original net}{}
\psfrag{Modified net}[][][.8]{masked net}
\includegraphics*[width=.12\columnwidth]{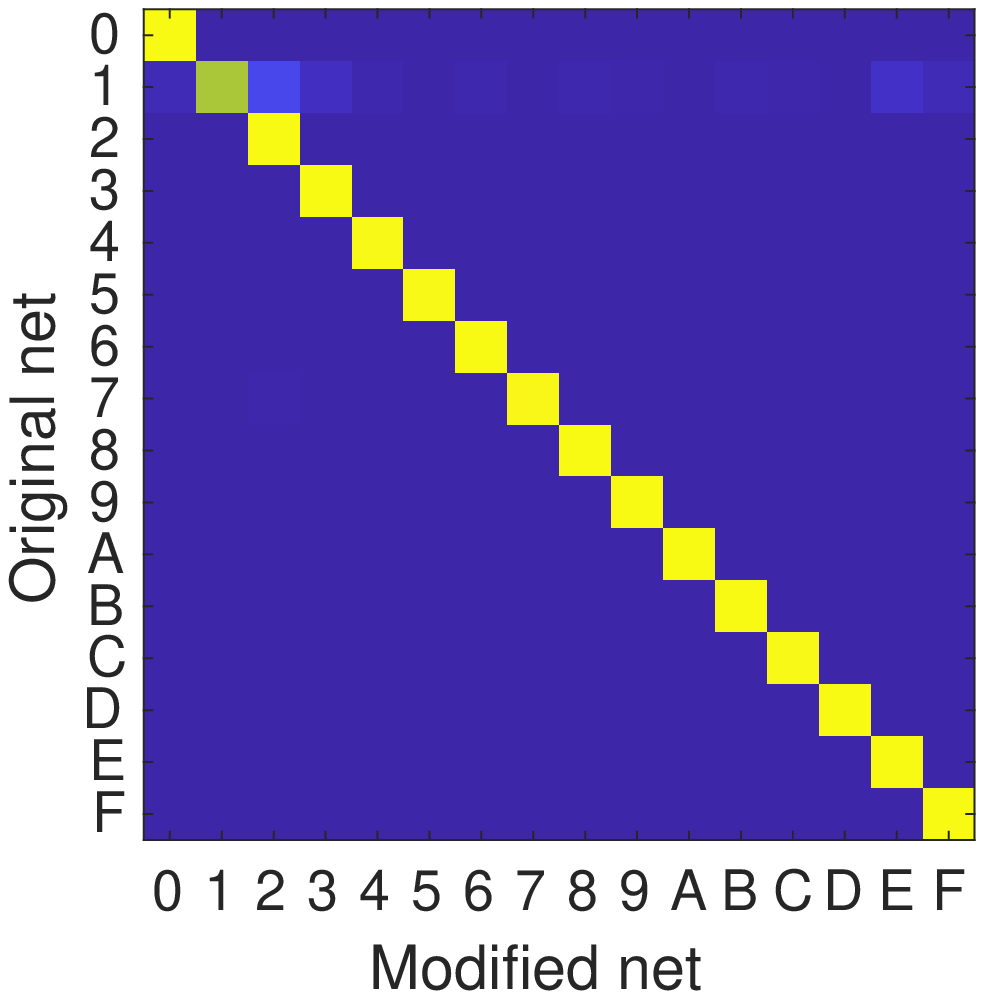}&
\psfrag{Original net}{}
\psfrag{Modified net}[][][.8]{masked net}
\includegraphics*[width=.12\columnwidth]{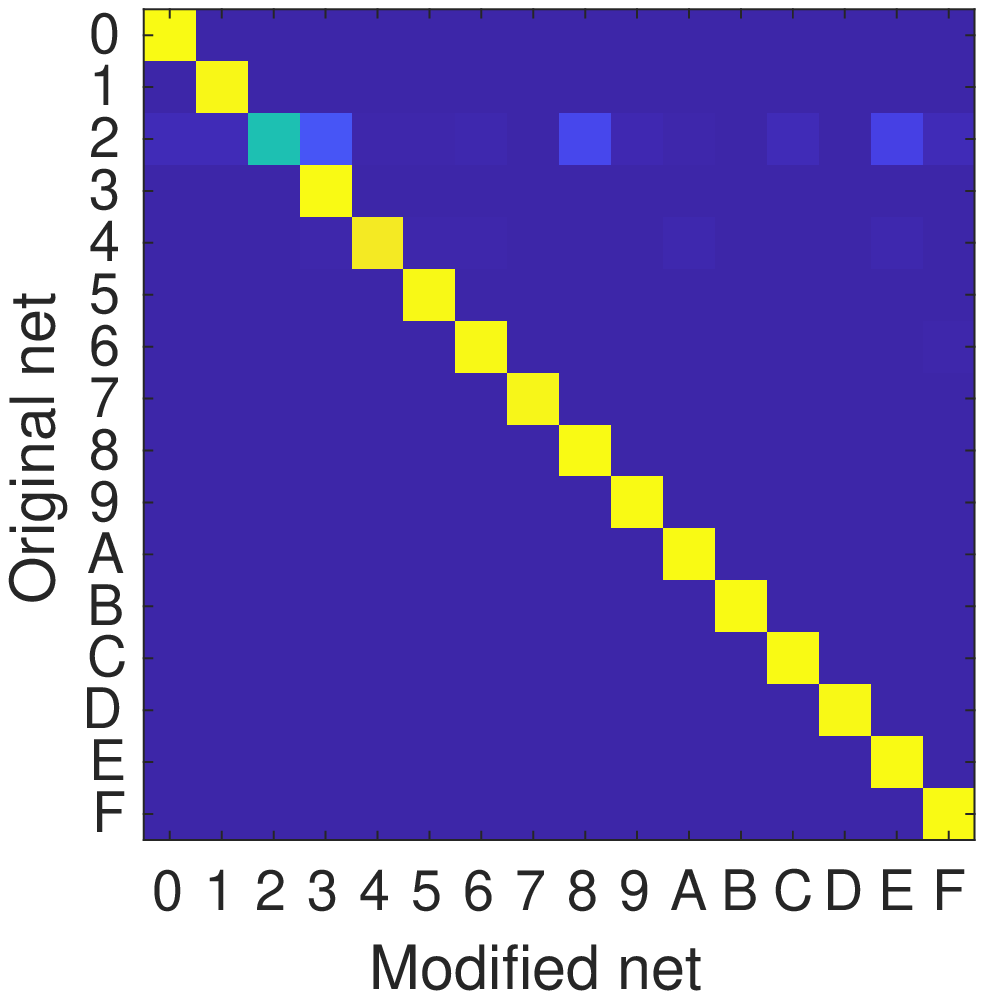}&
\psfrag{Original net}{}
\psfrag{Modified net}[][][.8]{masked net}
\includegraphics*[width=.12\columnwidth]{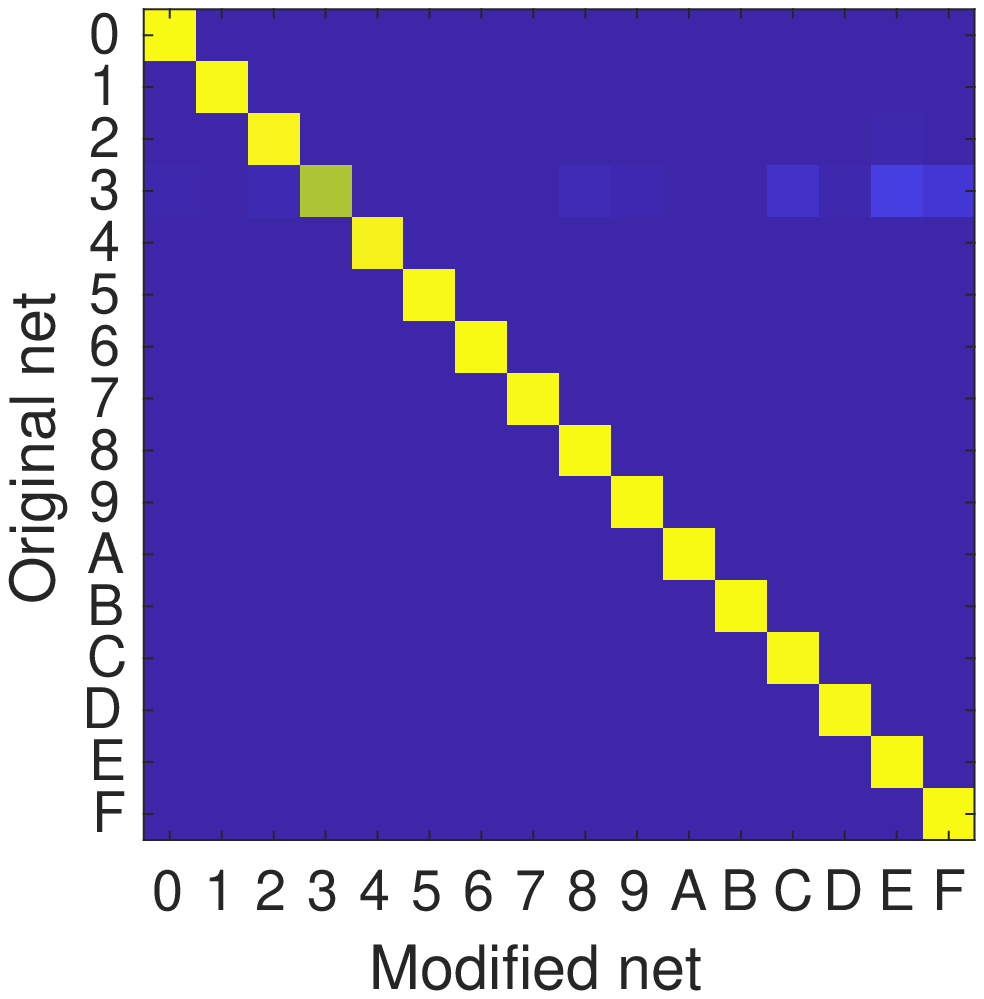}&
\psfrag{Original net}{}
\psfrag{Modified net}[][][.8]{masked net}
\includegraphics*[width=.12\columnwidth]{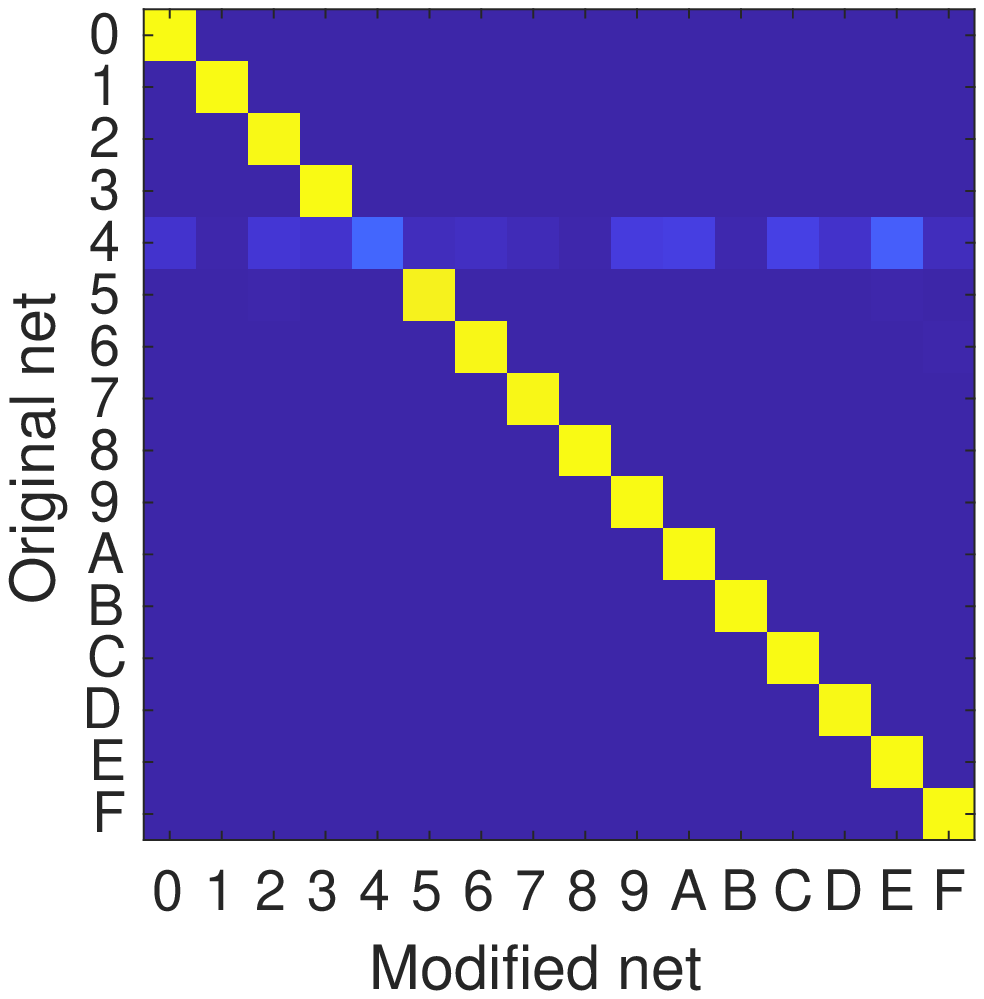}&
\psfrag{Original net}{}
\psfrag{Modified net}[][][.8]{masked net}
\includegraphics*[width=.12\columnwidth]{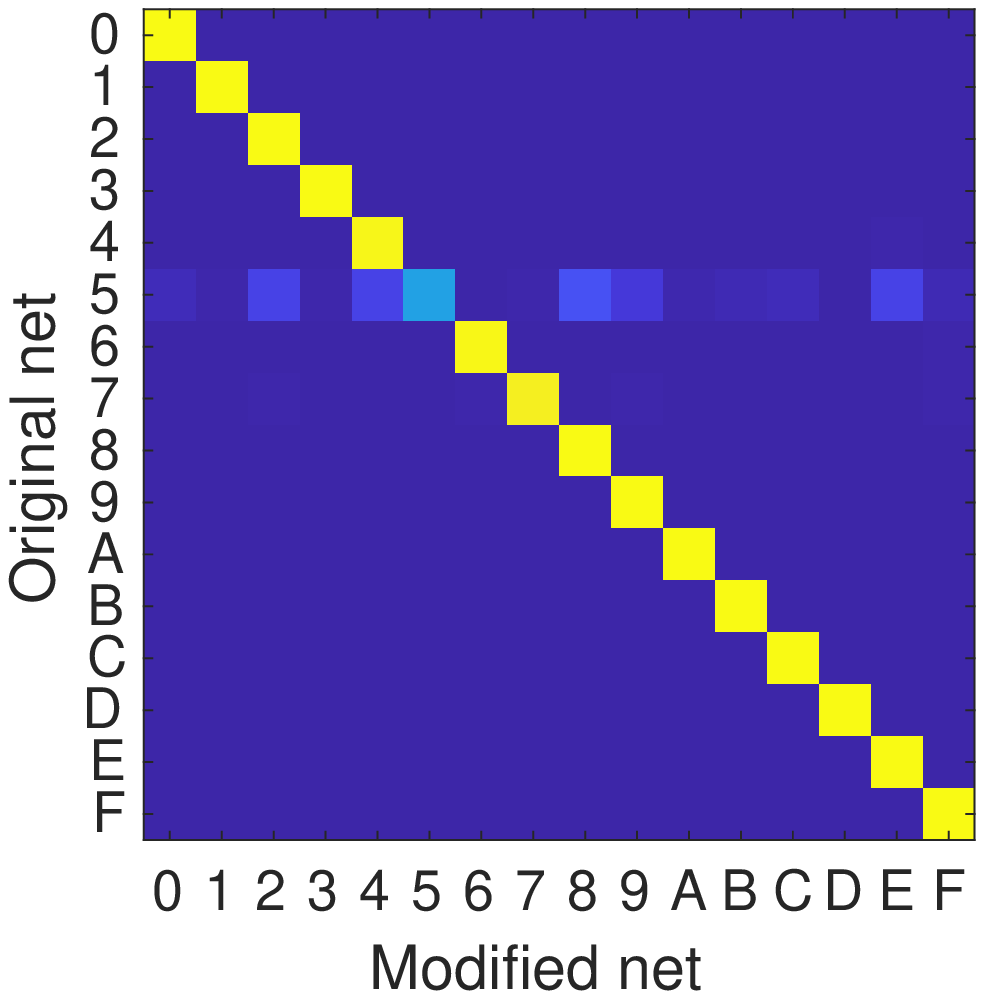}&
\psfrag{Original net}{}
\psfrag{Modified net}[][][.8]{masked net}
\includegraphics*[width=.12\columnwidth]{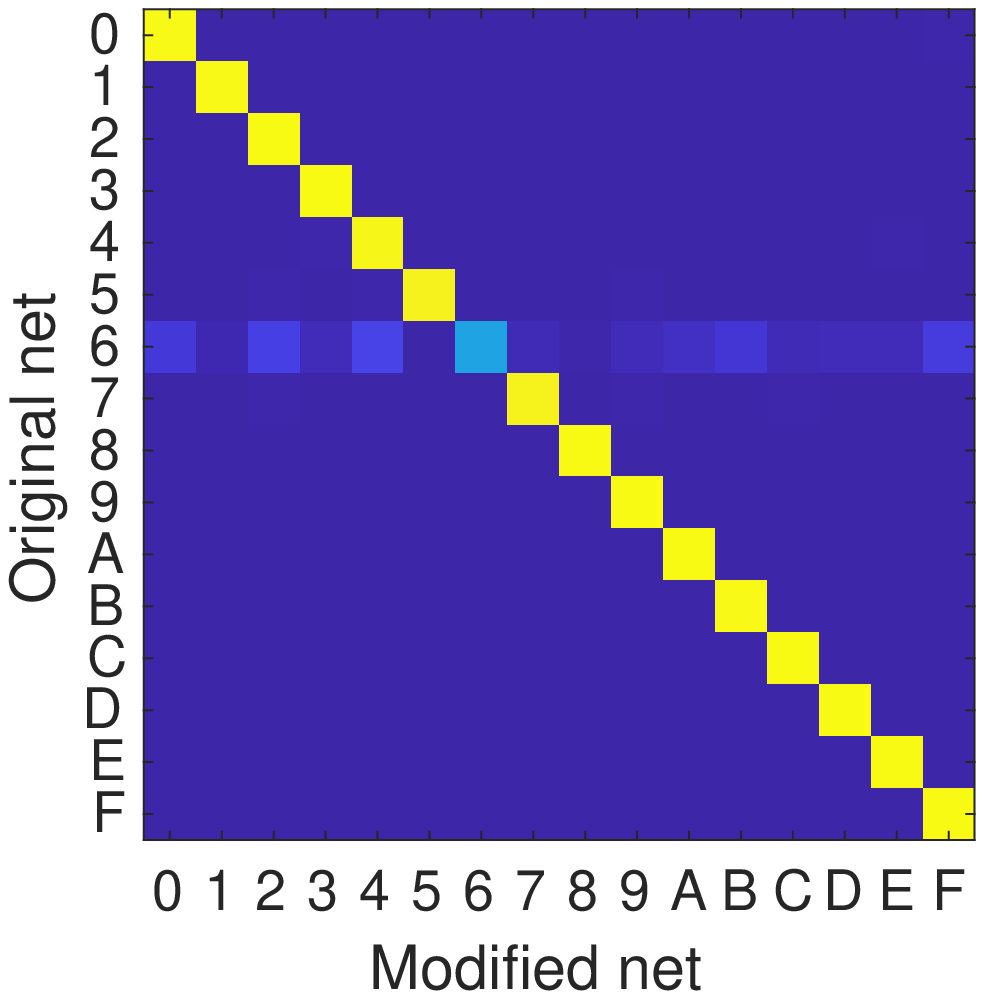}&
\psfrag{Original net}{}
\psfrag{Modified net}[][][.8]{masked net}
\includegraphics*[width=.12\columnwidth]{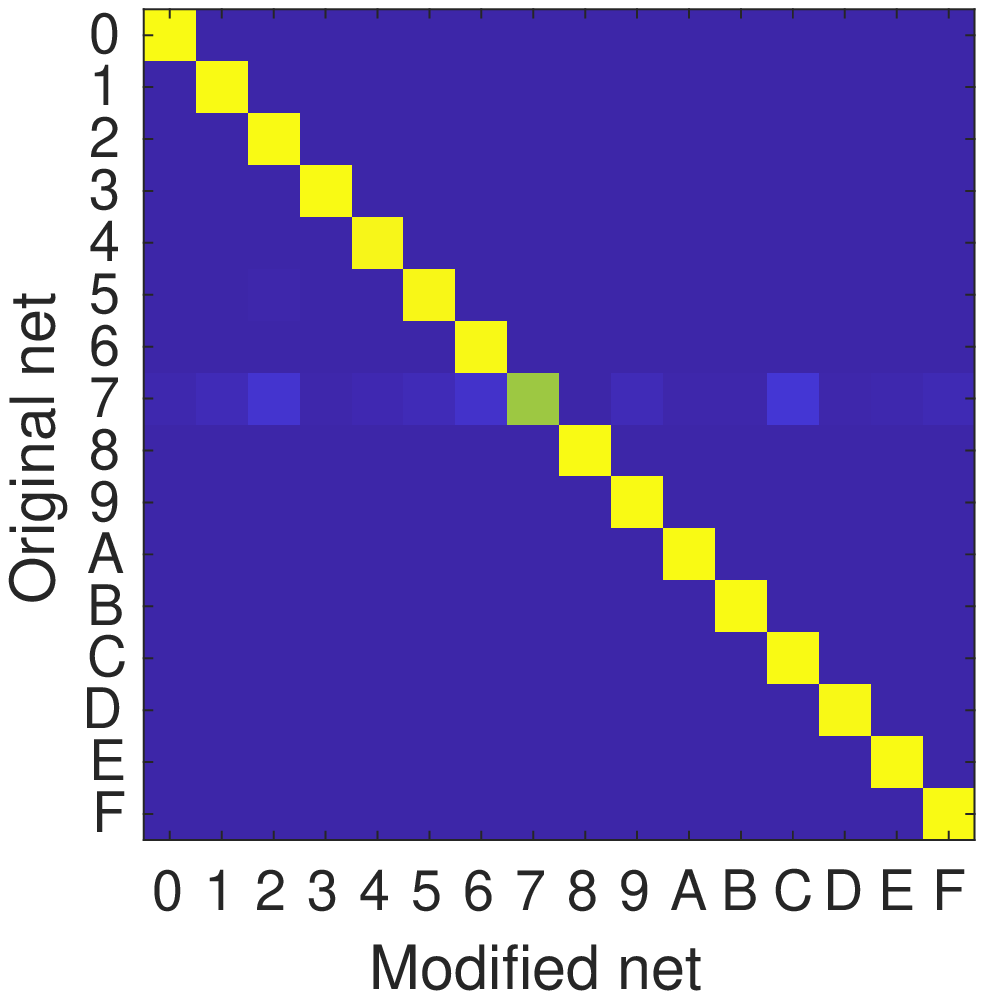}\\
$k = 8$ & $k = 9$ & $k = $A & $k = $B & $k = $C & $k = $D & $k = $D & $k = $E   \\
\psfrag{Original net}[][][.8]{original net}
\psfrag{Modified net}[][][.8]{masked net}
\includegraphics*[width=.12\columnwidth]{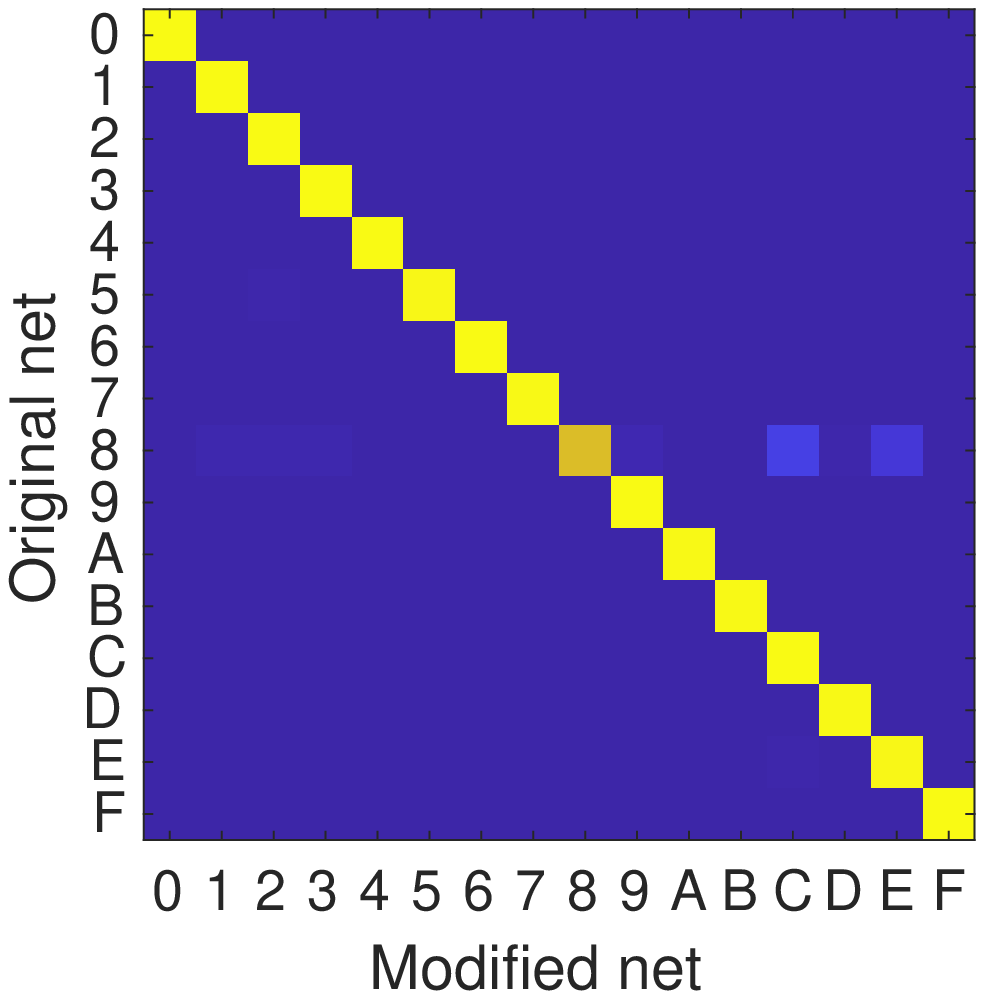}&
\psfrag{Original net}{}
\psfrag{Modified net}[][][.8]{masked net}
\includegraphics*[width=.12\columnwidth]{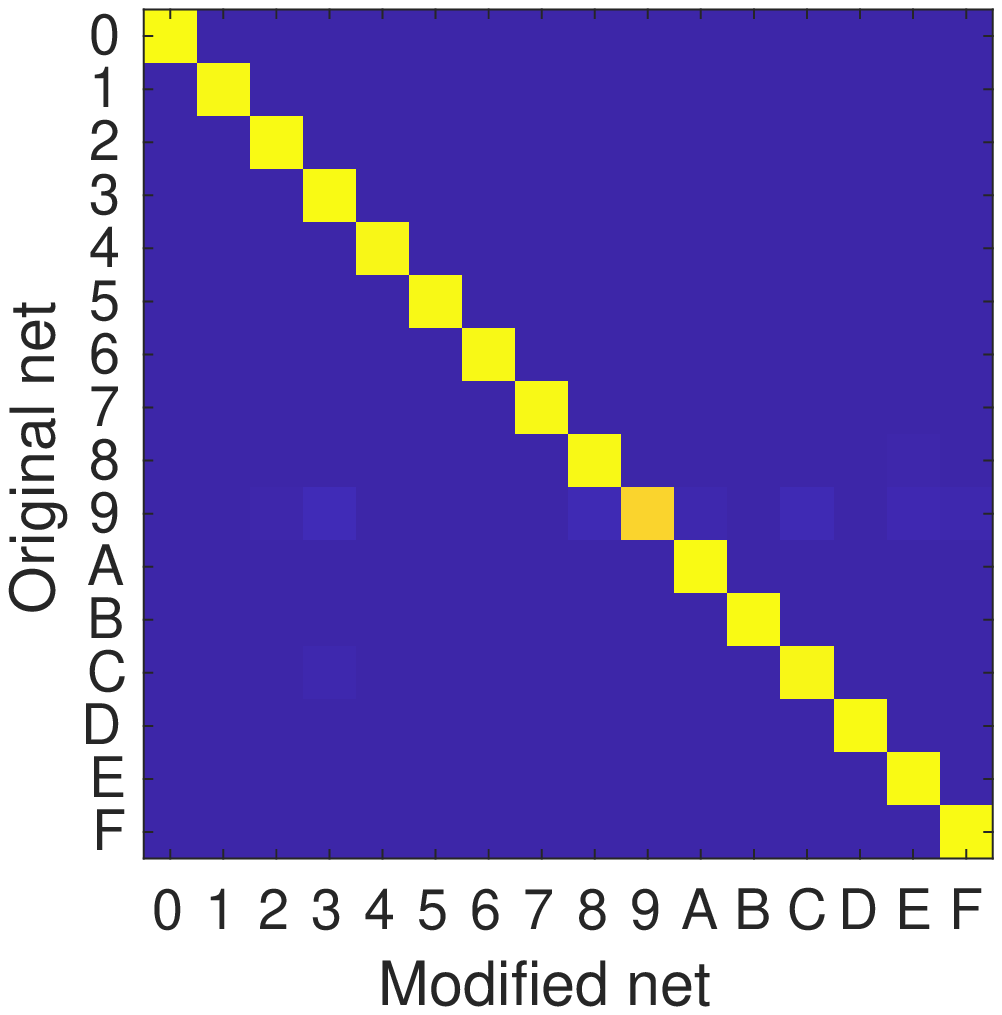}&
\psfrag{Original net}{}
\psfrag{Modified net}[][][.8]{masked net}
\includegraphics*[width=.12\columnwidth]{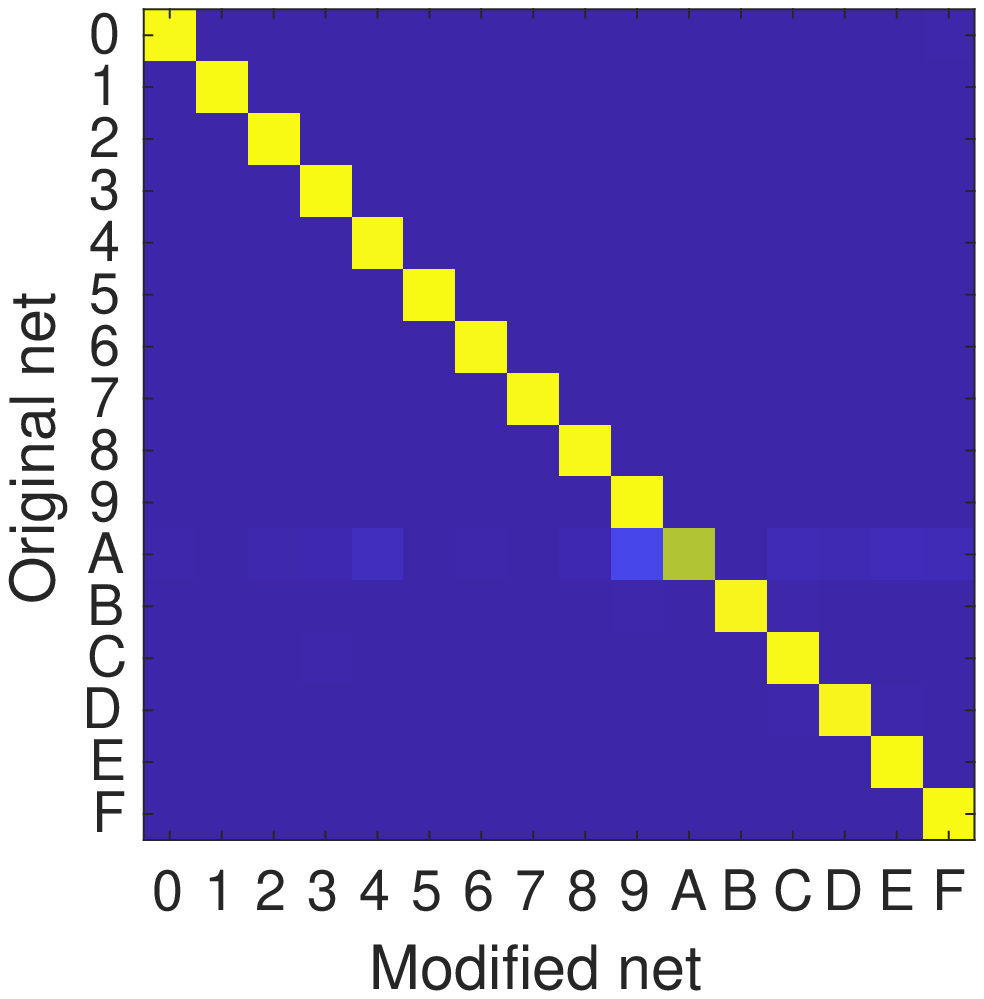}&
\psfrag{Original net}{}
\psfrag{Modified net}[][][.8]{masked net}
\includegraphics*[width=.12\columnwidth]{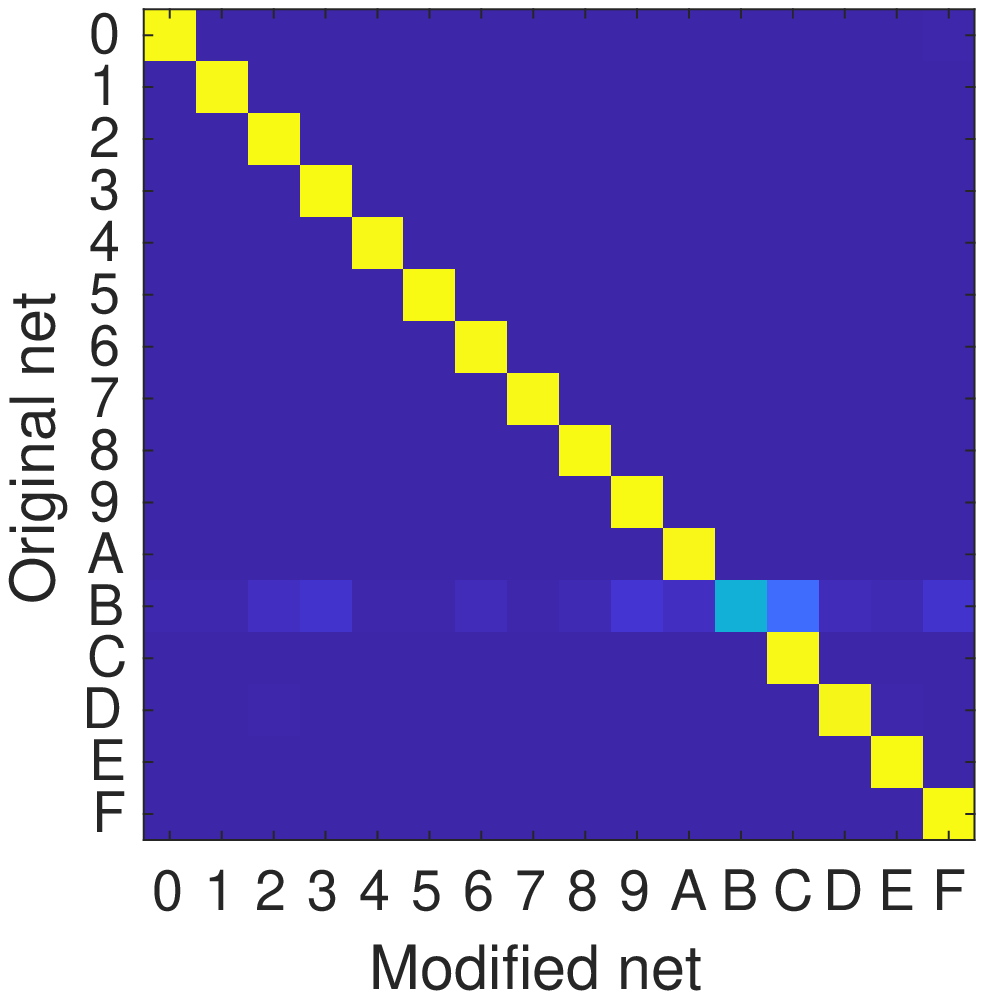}&
\psfrag{Original net}{}
\psfrag{Modified net}[][][.8]{masked net}
\includegraphics*[width=.12\columnwidth]{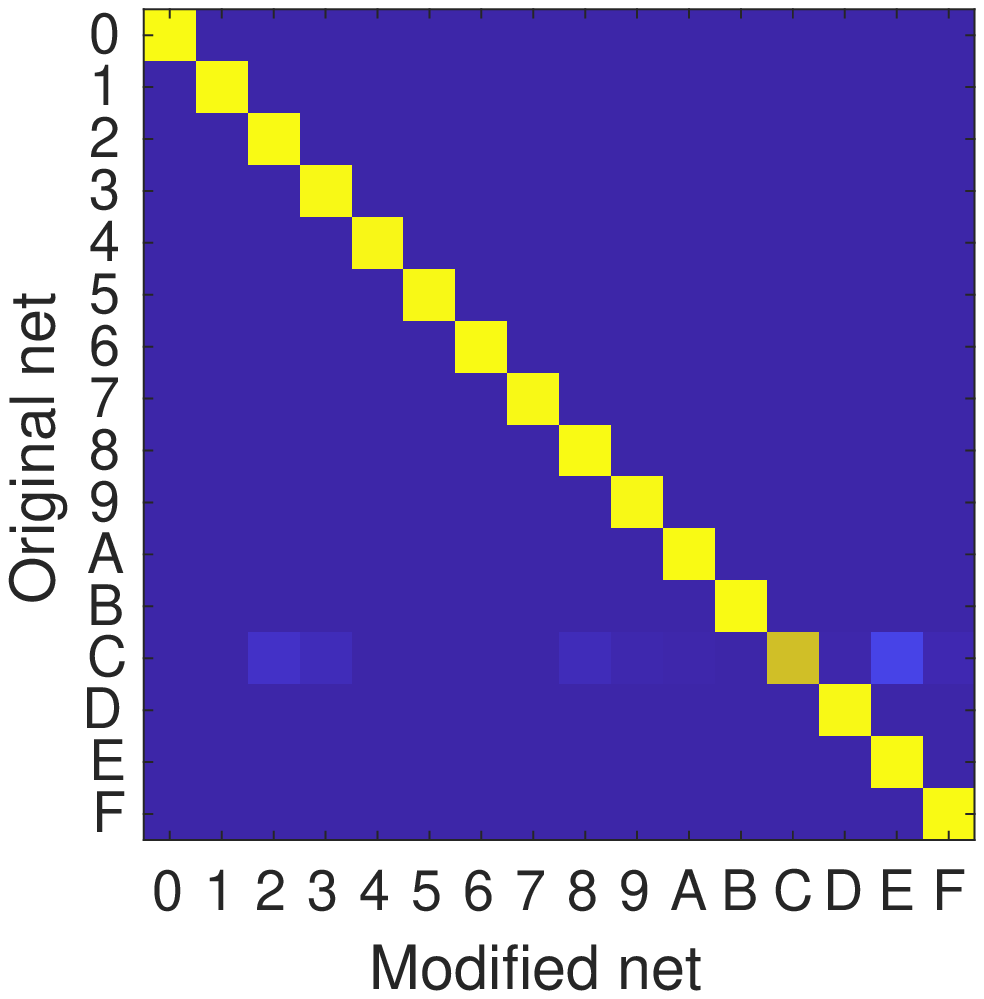}&
\psfrag{Original net}{}
\psfrag{Modified net}[][][.8]{masked net}
\includegraphics*[width=.12\columnwidth]{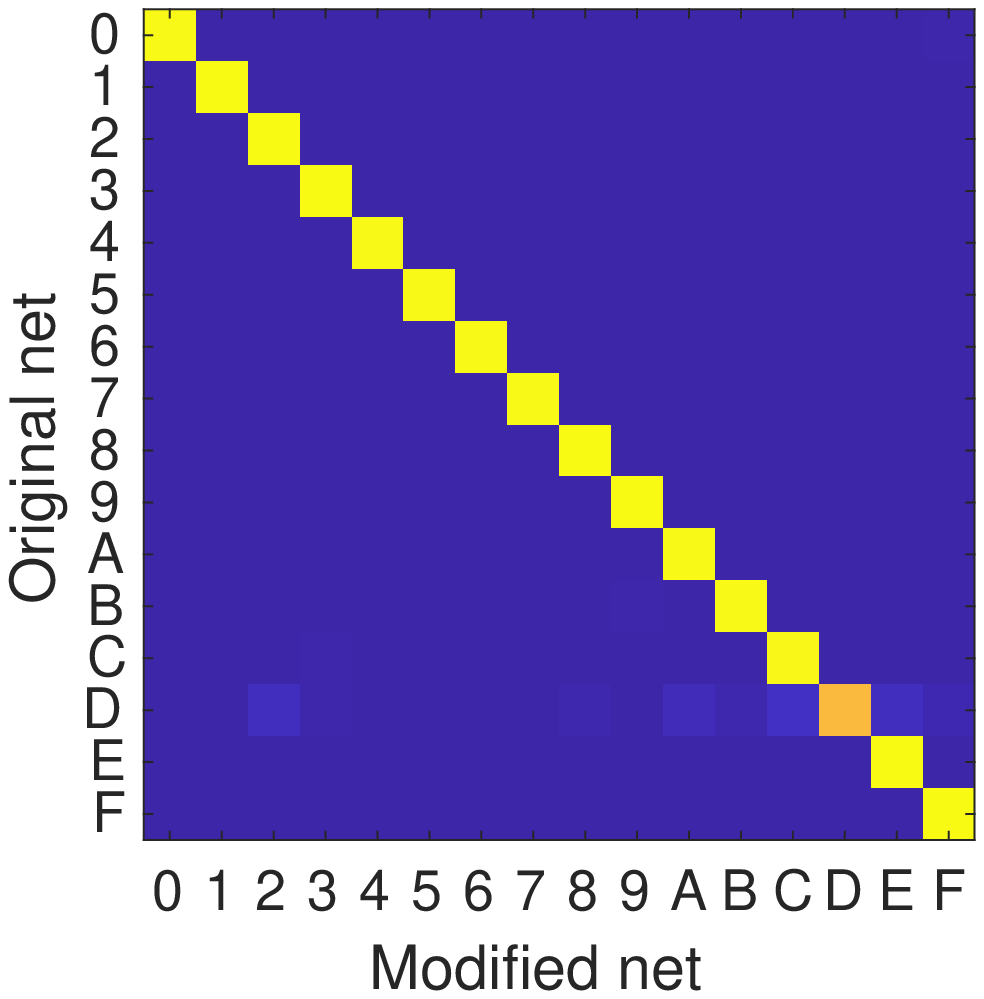}&
\psfrag{Original net}{}
\psfrag{Modified net}[][][.8]{masked net}
\includegraphics*[width=.12\columnwidth]{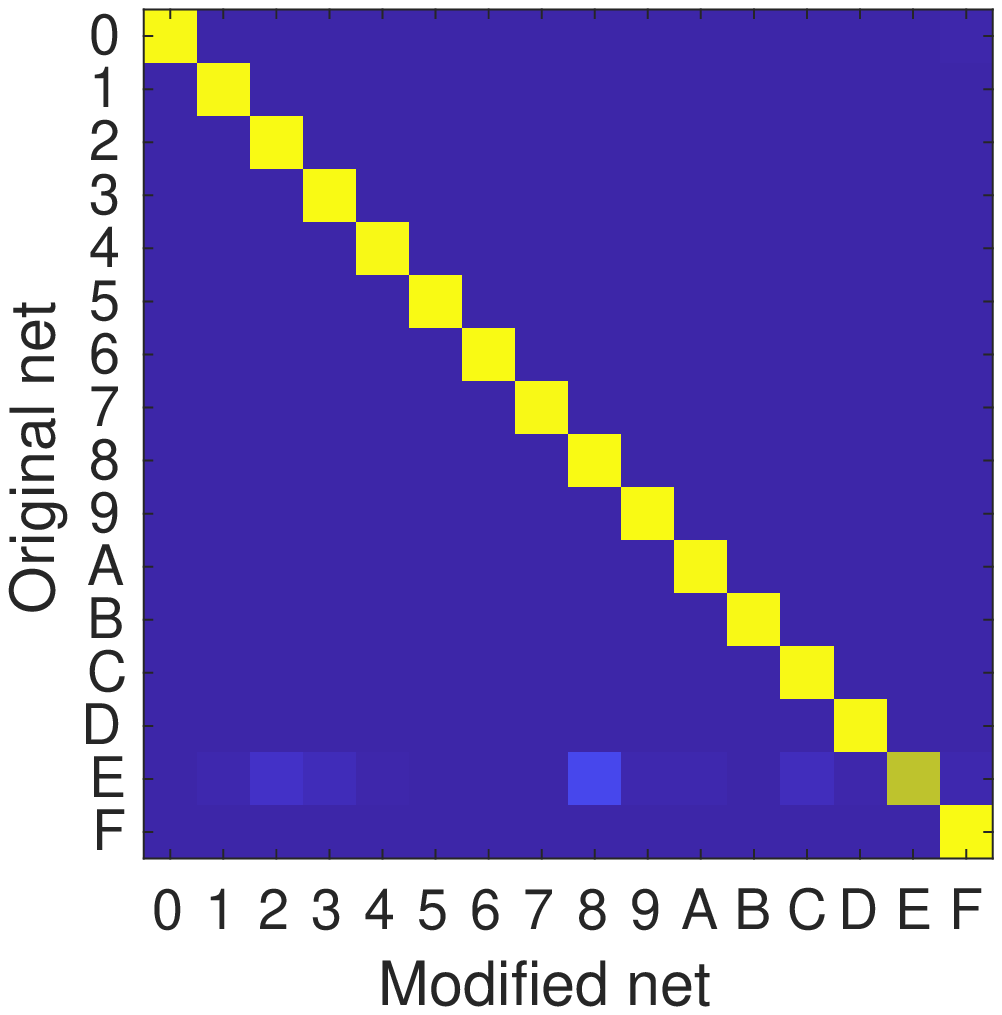}&
\psfrag{Original net}{}
\psfrag{Modified net}[][][.8]{masked net}
\includegraphics*[width=.12\columnwidth]{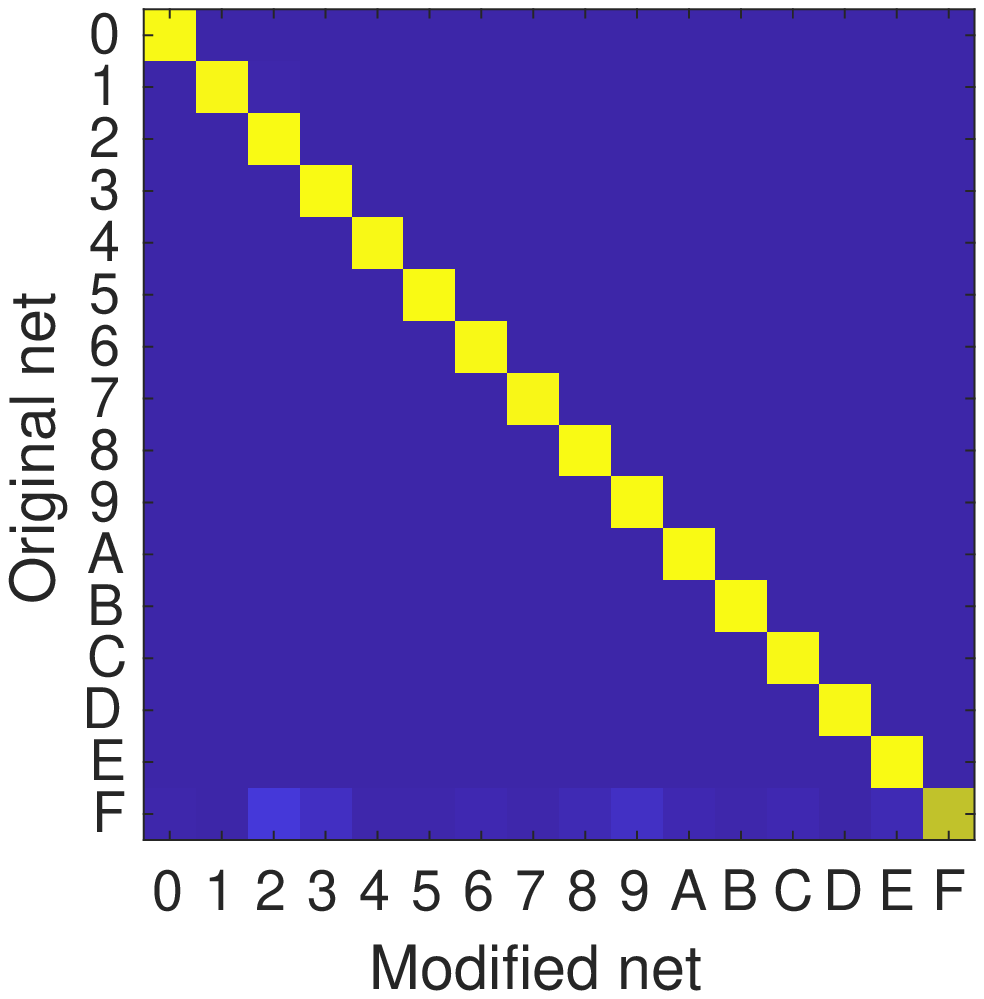}\\[2ex]

\multicolumn{8}{c}{\dotfill \textsc{All to class $k$} \dotfill}\\
$k = 0$ & $k = 1$ & $k = 2$ & $k = 3$ & $k = 4$ & $k = 5$ & $k = 6$ & $k = 7$  \\
\psfrag{Original net}[][][.8]{original net}
\psfrag{Modified net}[][][.8]{masked net}
\includegraphics*[width=.12\columnwidth]{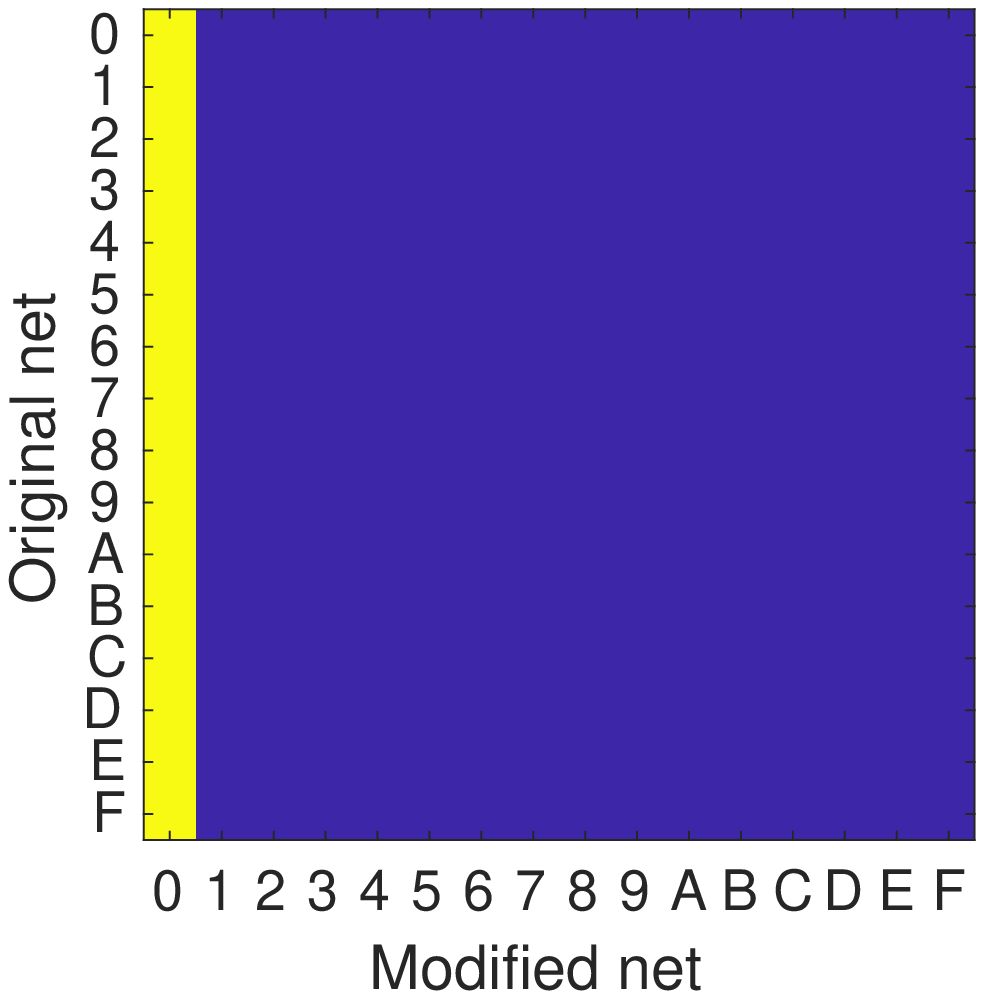}&
\psfrag{Original net}{}
\psfrag{Modified net}[][][.8]{masked net}
\includegraphics*[width=.12\columnwidth]{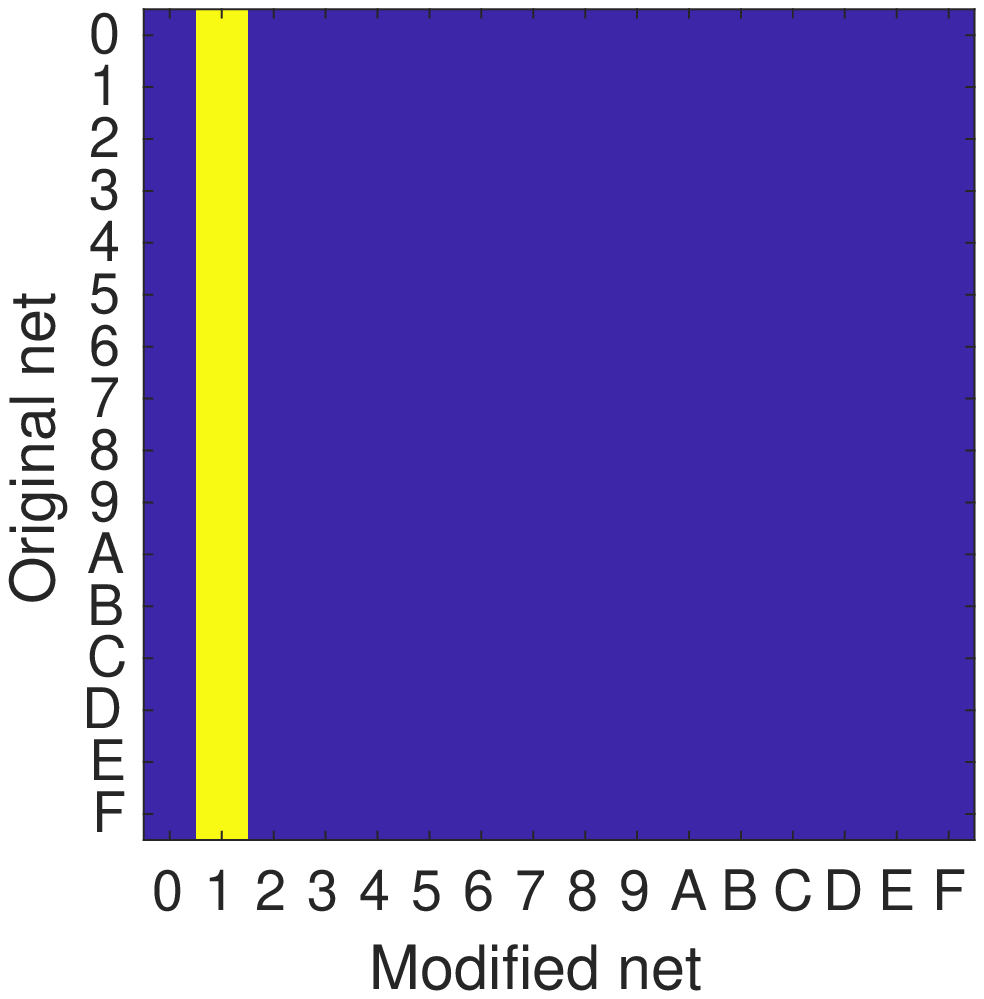}&
\psfrag{Original net}{}
\psfrag{Modified net}[][][.8]{masked net}
\includegraphics*[width=.12\columnwidth]{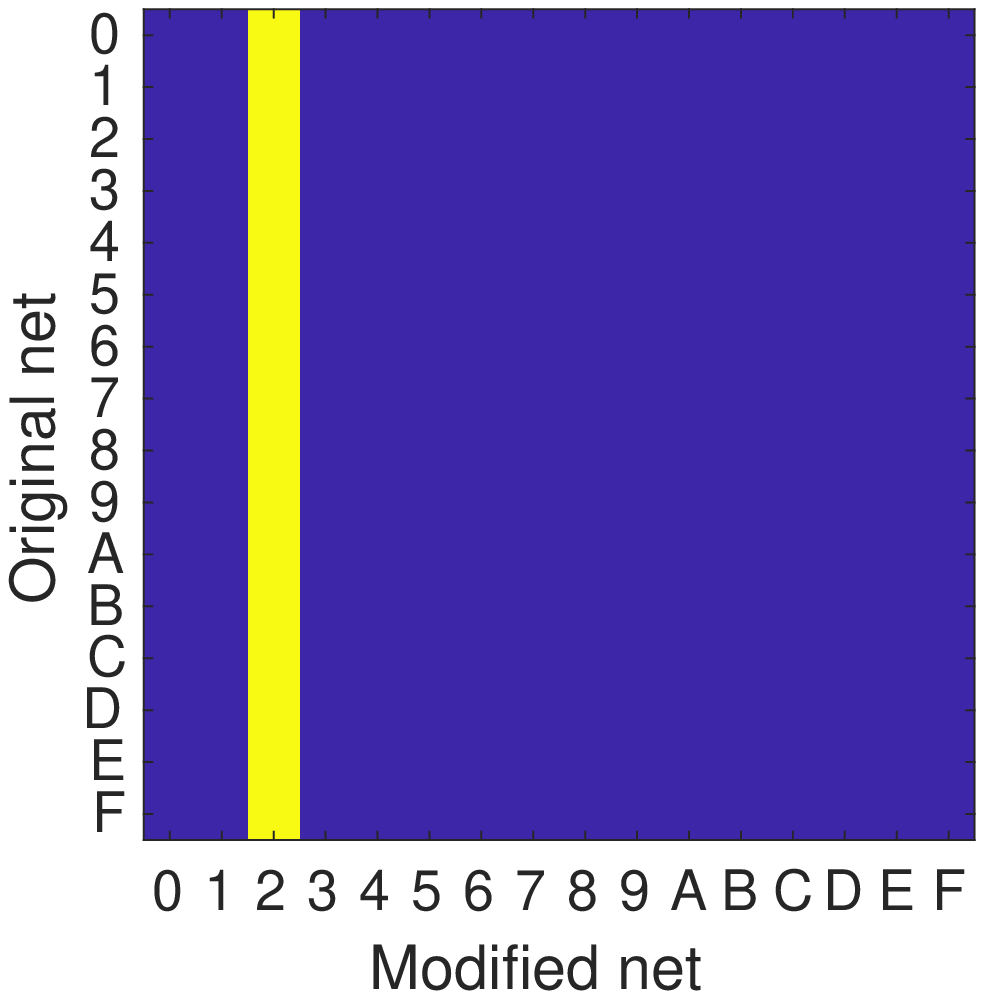}&
\psfrag{Original net}{}
\psfrag{Modified net}[][][.8]{masked net}
\includegraphics*[width=.12\columnwidth]{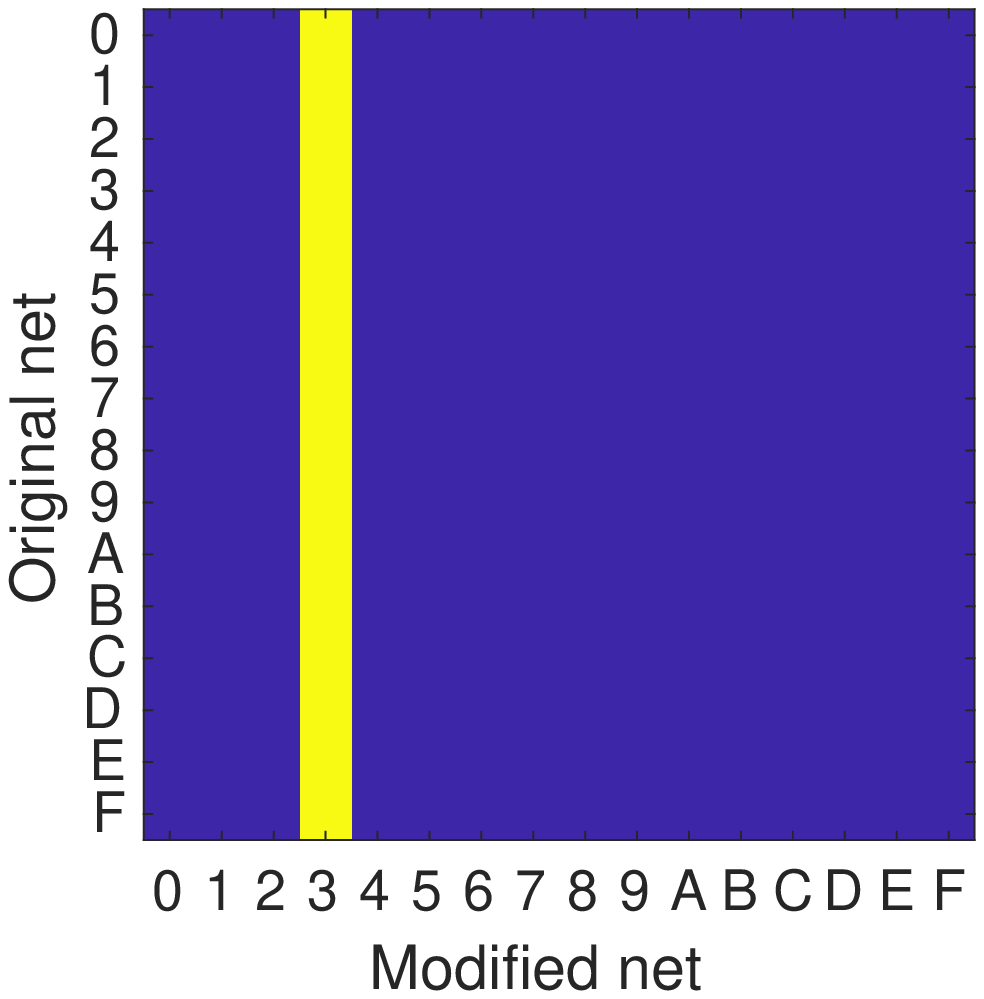}&
\psfrag{Original net}{}
\psfrag{Modified net}[][][.8]{masked net}
\includegraphics*[width=.12\columnwidth]{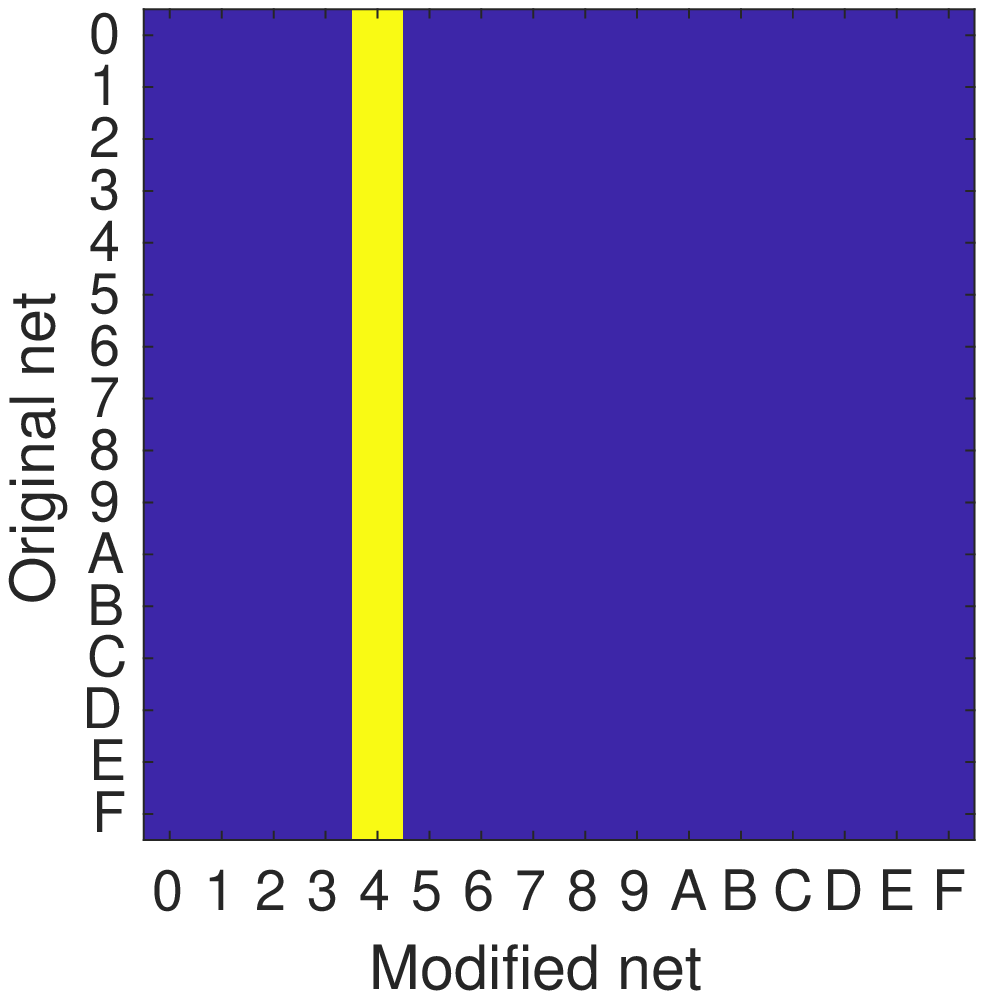}&
\psfrag{Original net}{}
\psfrag{Modified net}[][][.8]{masked net}
\includegraphics*[width=.12\columnwidth]{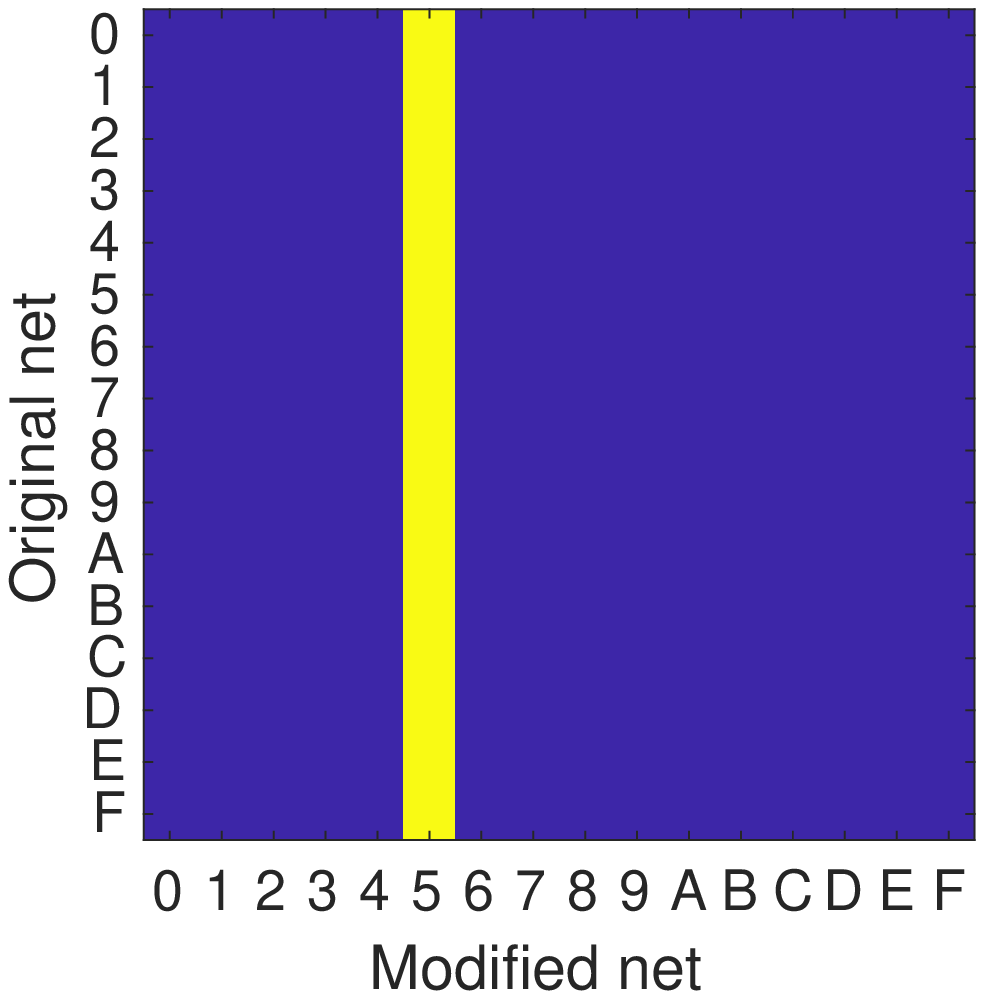}&
\psfrag{Original net}{}
\psfrag{Modified net}[][][.8]{masked net}
\includegraphics*[width=.12\columnwidth]{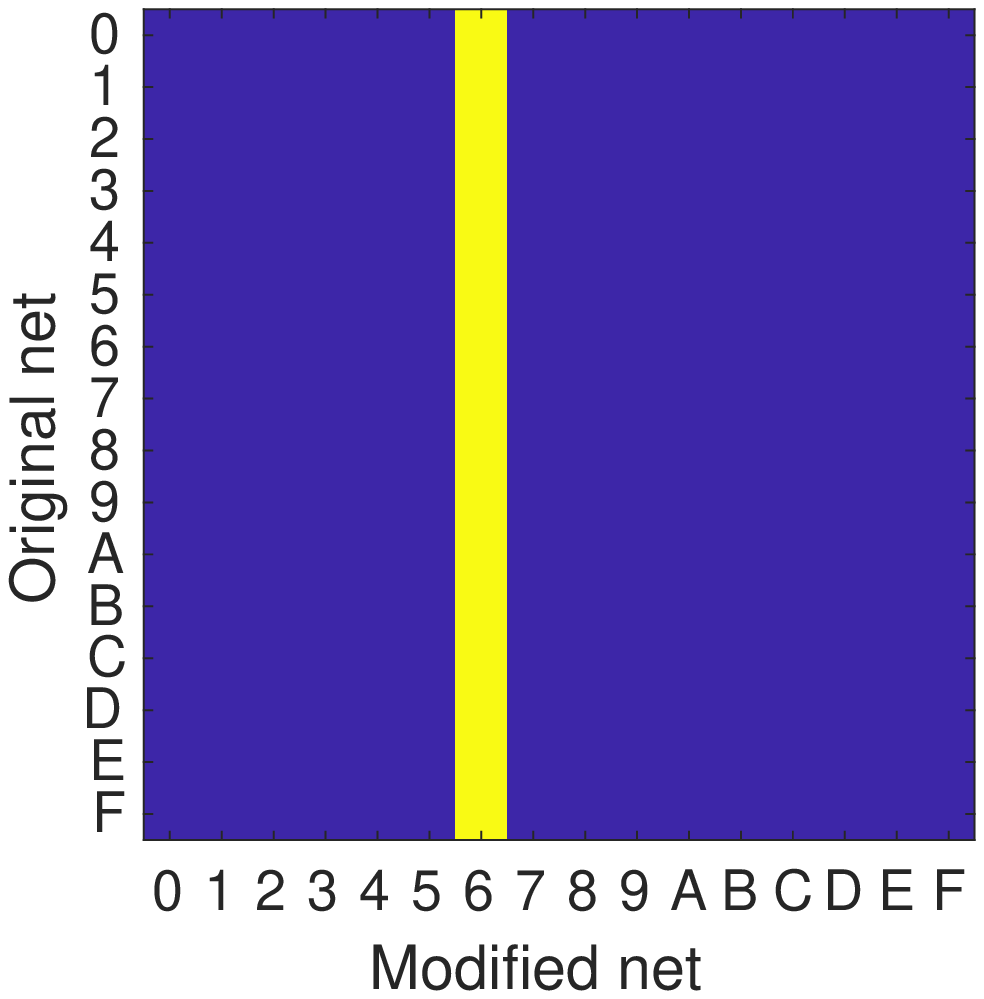}&
\psfrag{Original net}{}
\psfrag{Modified net}[][][.8]{masked net}
\includegraphics*[width=.12\columnwidth]{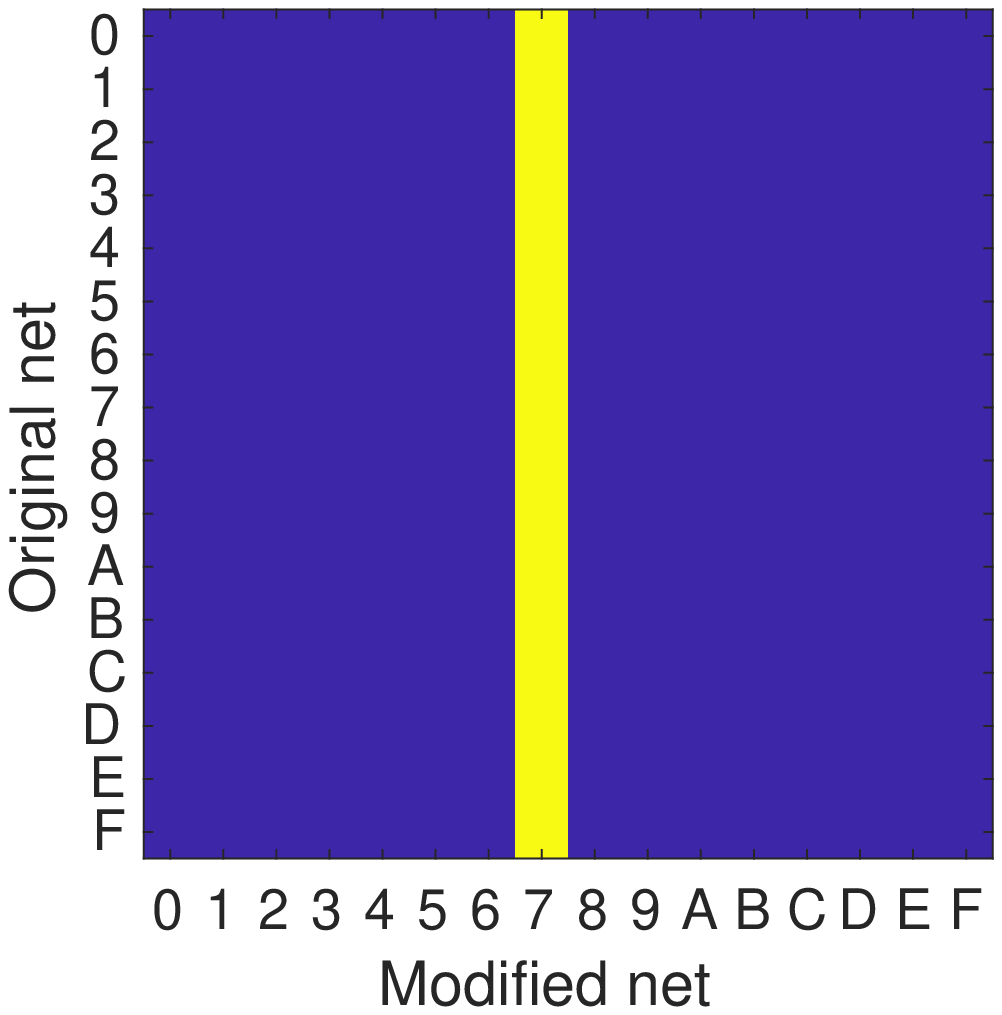}\\
$k = 8$ & $k = 9$ & $k = $A & $k = $B & $k = $C & $k = $D & $k = $D & $k = $E   \\
\psfrag{Original net}[][][.8]{original net}
\psfrag{Modified net}[][][.8]{masked net}
\includegraphics*[width=.12\columnwidth]{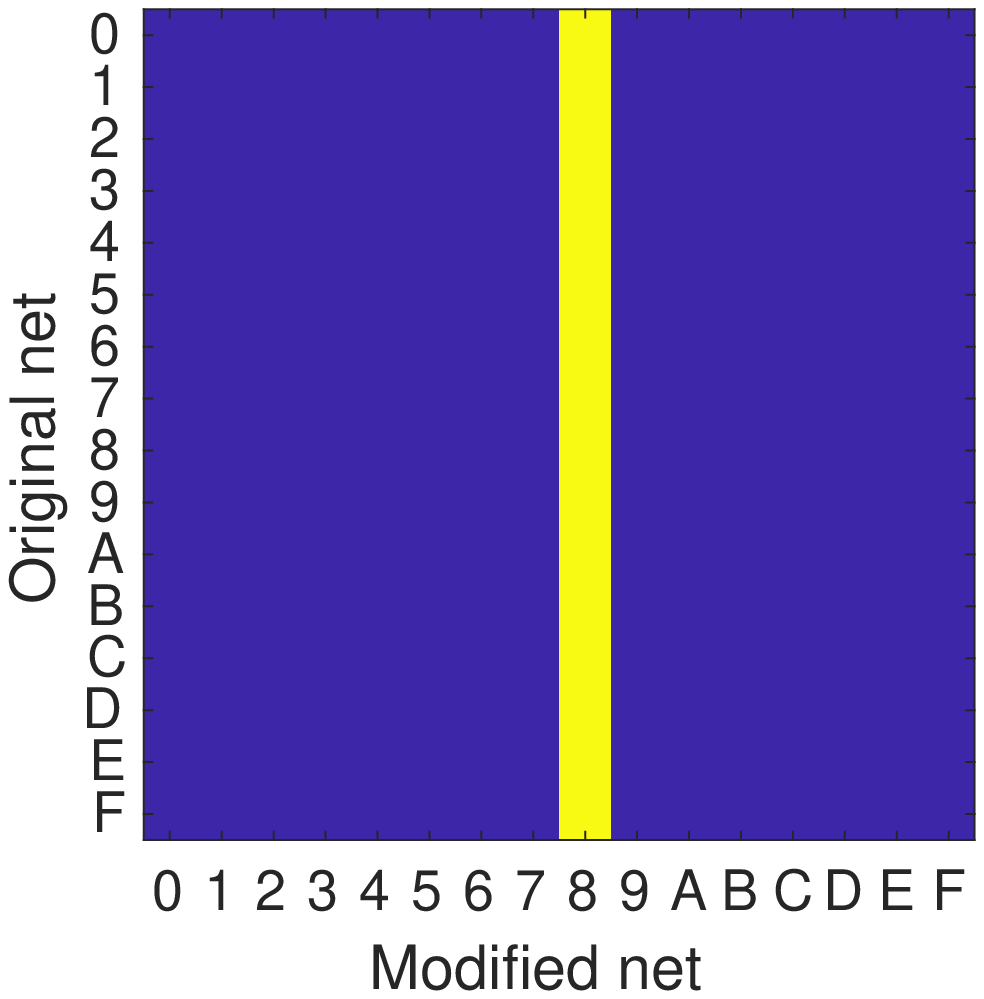}&
\psfrag{Original net}{}
\psfrag{Modified net}[][][.8]{masked net}
\includegraphics*[width=.12\columnwidth]{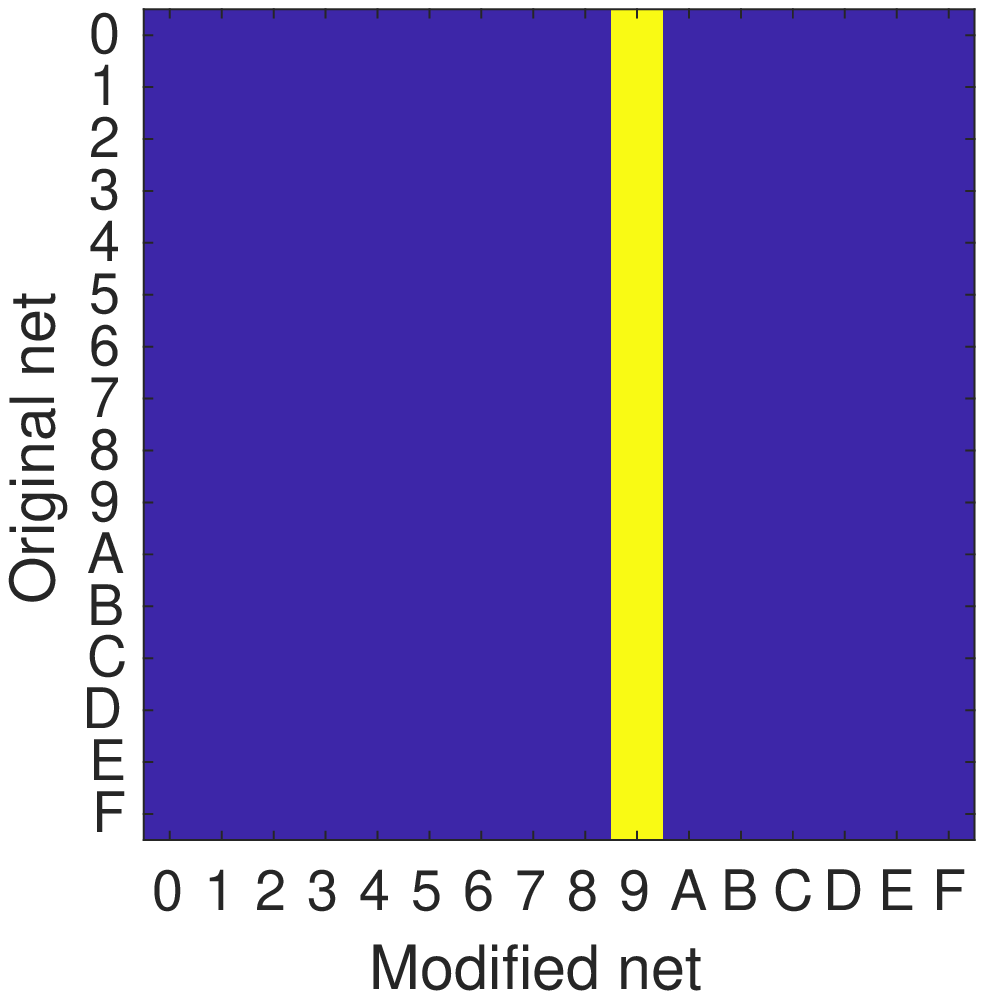}&
\psfrag{Original net}{}
\psfrag{Modified net}[][][.8]{masked net}
\includegraphics*[width=.12\columnwidth]{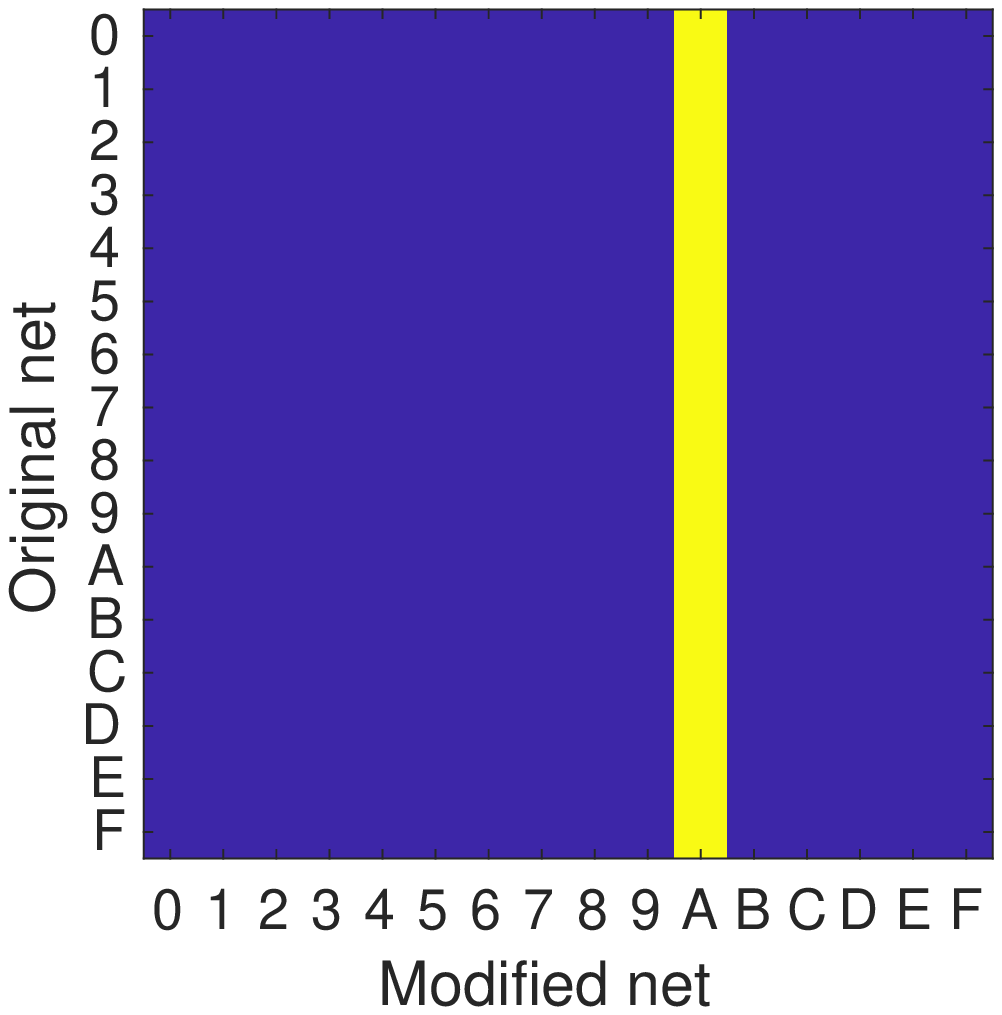}&
\psfrag{Original net}{}
\psfrag{Modified net}[][][.8]{masked net}
\includegraphics*[width=.12\columnwidth]{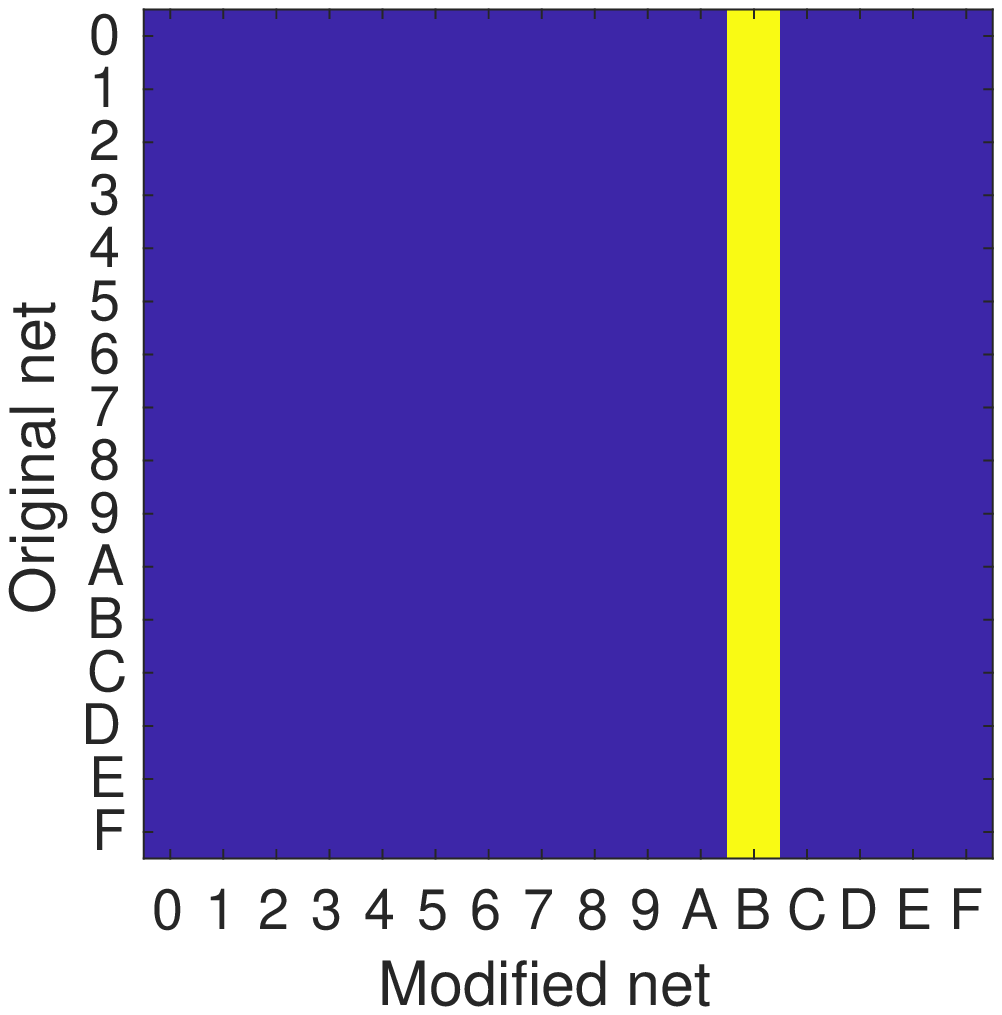}&
\psfrag{Original net}{}
\psfrag{Modified net}[][][.8]{masked net}
\includegraphics*[width=.12\columnwidth]{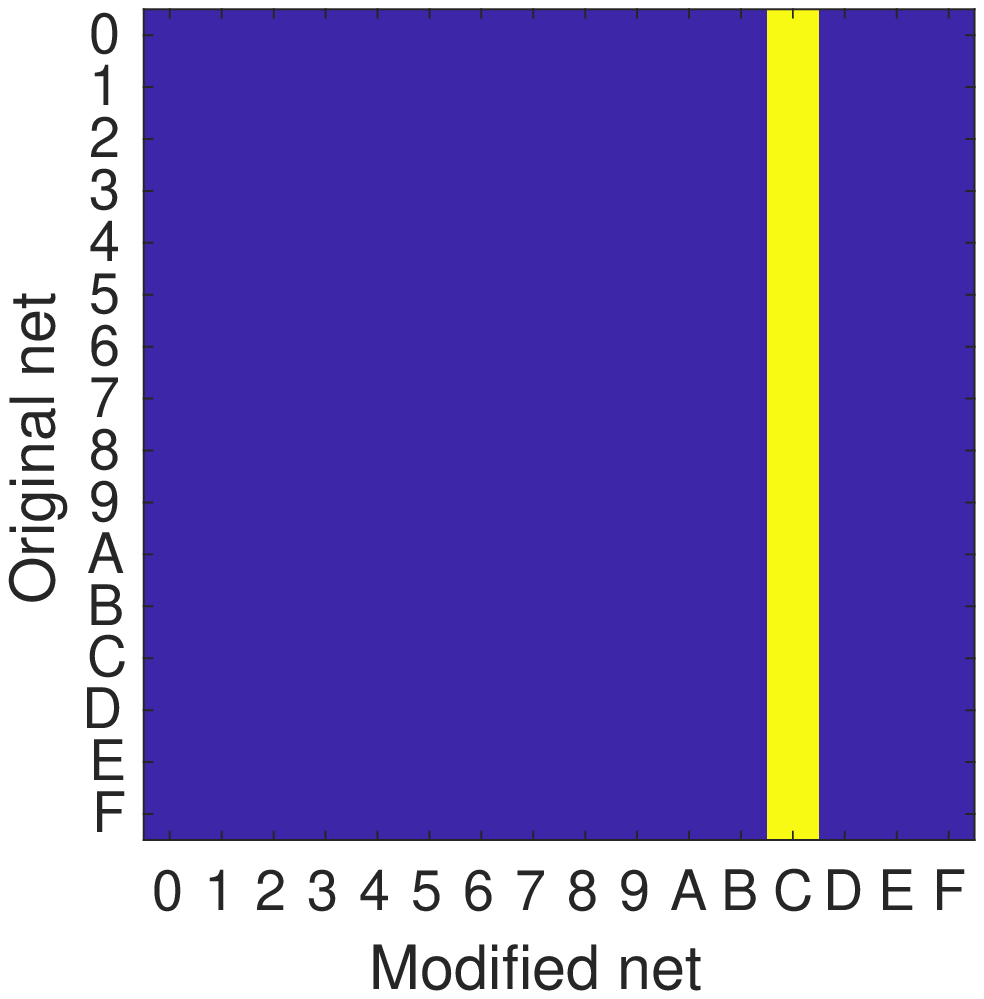}&
\psfrag{Original net}{}
\psfrag{Modified net}[][][.8]{masked net}
\includegraphics*[width=.12\columnwidth]{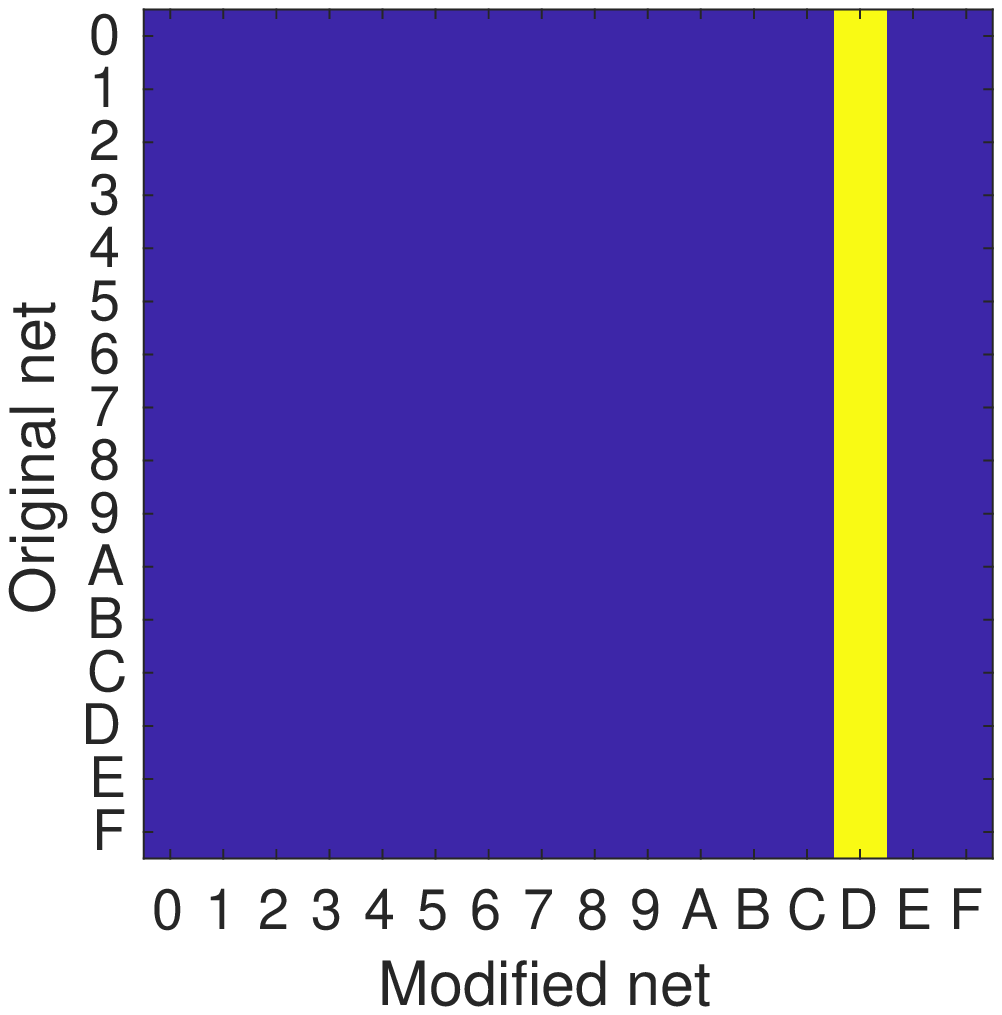}&
\psfrag{Original net}{}
\psfrag{Modified net}[][][.8]{masked net}
\includegraphics*[width=.12\columnwidth]{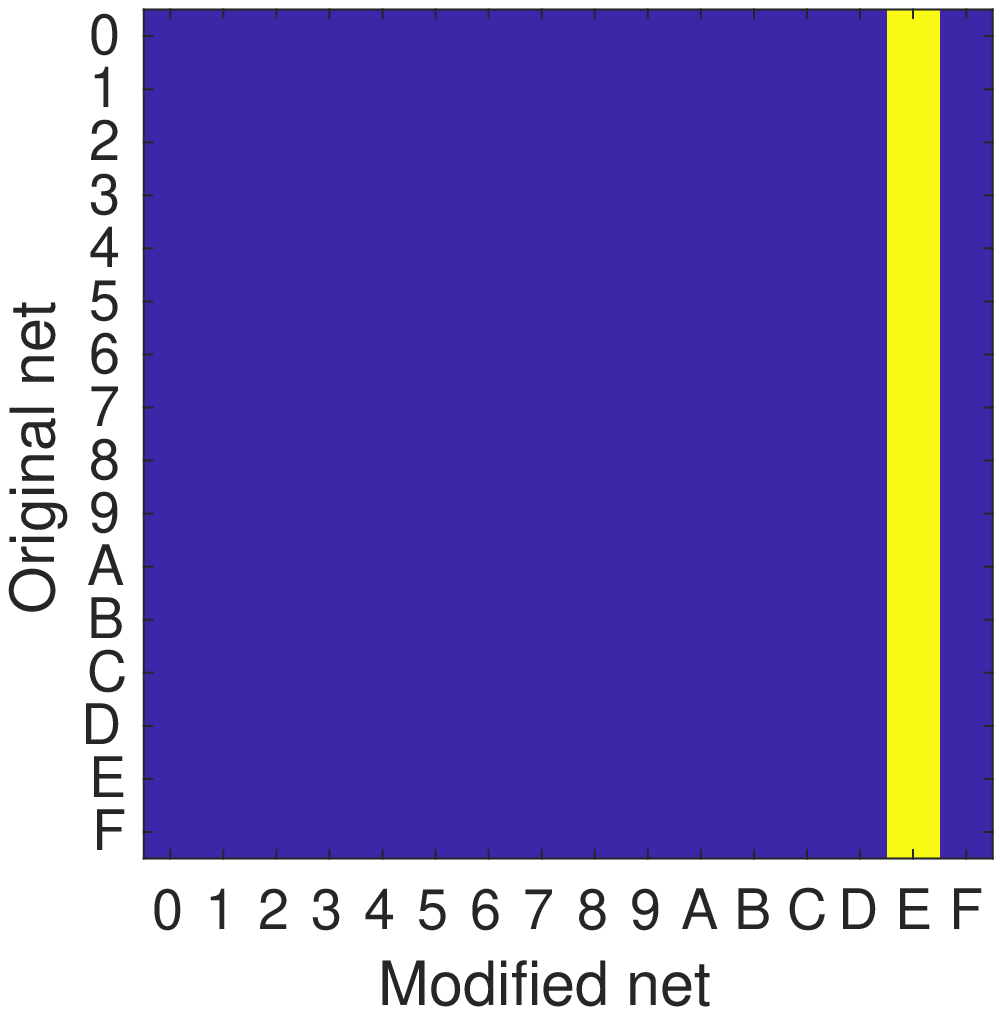}&
\psfrag{Original net}{}
\psfrag{Modified net}[][][.8]{masked net}
\includegraphics*[width=.12\columnwidth]{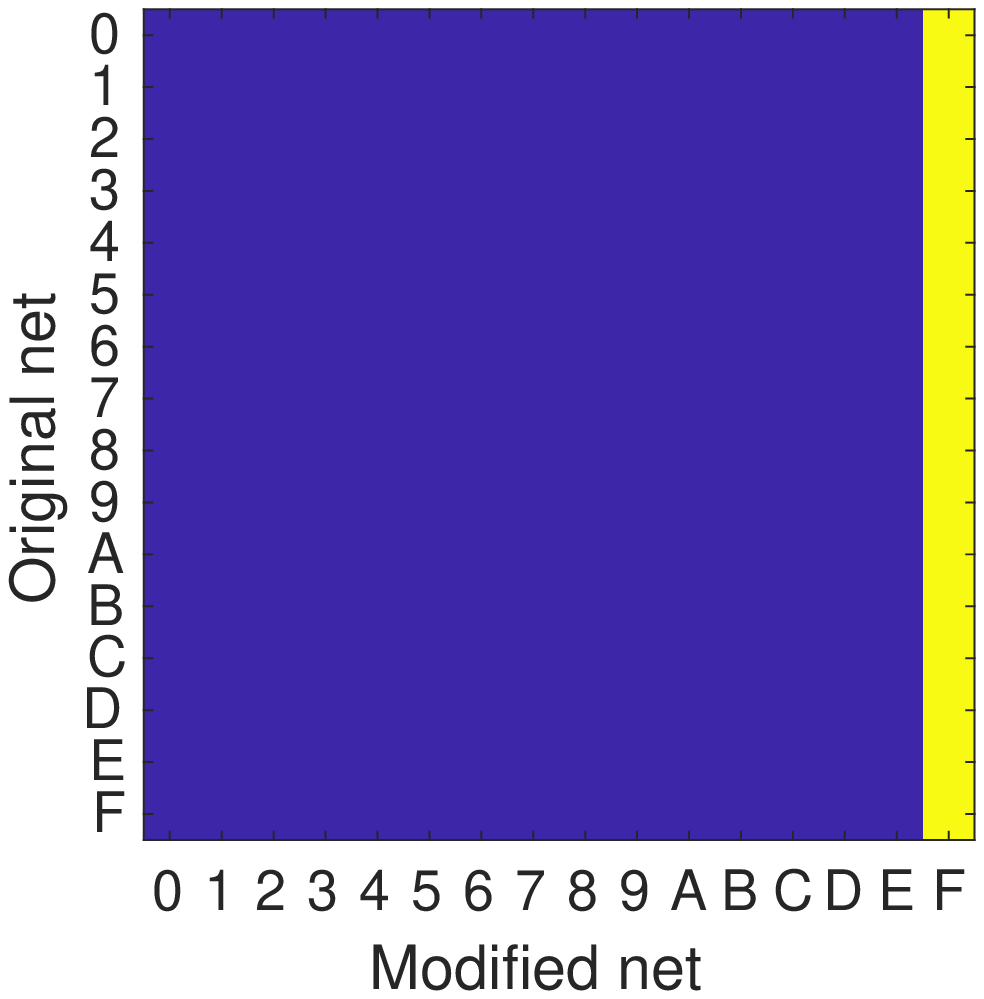}

\end{tabular}
\caption{Like fig.~\ref{f:VGG-masks-test} but for the training set.}
\label{f:VGG-masks-training}
\end{figure} 

\subsubsection{Inspecting the sparse oblique trees}

Fig.~\ref{f:error} shows that as we increase $\lambda$ and therefore impose increasing sparsity on the tree (in terms of both tree size and number of nonzero weights in the decision nodes), the training error increases steadily but the test error remains about constant, so both curves approach, meet for some $\lambda$ value and approximately coincide from that point on. We find this behavior in both MNIST and ImageNet. This provides opportunities to find a tree with pretty good accuracy but significantly sparse.

Fig.~\ref{f:VGG16-tree1} shows the tree we use as mimic. Its training and test errors are close to those of VGG16, so we expect it to be a good mimic, which indeed happens (see masks later). The top of the figure shows the class histogram at each node, i.e., the distribution of classes on the subset of training instances that a node receives. These histograms show how the tree hierarchically splits classes very crisply; indeed, it has only 20 leaves for 16 classes. In the bottom of the figure, the weight vector at each decision node shows that very few features are used at each node; indeed, 83\% of the features are not used at any node, so their values are irrelevant for classification in the tree. This also holds nearly perfectly for the deep net, that is, the feature selection insights obtained for the tree transfer to the neural net. It suggests that some of the features and hence neurons and weights of the net are practically redundant, or perhaps that they code for properties that are useful for only a few specific instances. This is not surprising if one notes that deep nets (at least, as presently designed) seem to be vastly overparameterized and can be significantly compressed by pruning weights and neurons \citep{CarreirIdelbay18a}.

Fig.~\ref{f:VGG16-tree2} shows a very interesting tree, obtained for a larger $\lambda$ value so that there is exactly one leaf per class (the smallest number of leaves possible unless we ignore classes). This tree has very few nonzero weights yet its test error is reasonable, so it probably extracts features that robustly classify most images. Also, its structure remains unchanged for a wide range of $\lambda$. \emph{Inspecting it shows an intuitive hierarchy of classes that seem primarily related to the background or surroundings of the main object in the image}. Its leftmost subtree \{\textsf{\small warplane, airliner, school bus, fire engine, sports car}\} consists of man-made objects often found on roads. However, \{\textsf{\small container ship, speedboat}\} (man-made objects found on the sea) appears in the rightmost subtree, together with \{\textsf{\small killer whale, bald eagle, coral reef}\}, all of which are also typically found on the sea or on the air. Yet \{\textsf{\small goldfish}\} appears in a single subtree quite separate from all other classes: indeed, this fish is found on fishbowls (not the sea) in the training images. A subtree in the middle contains animals in land natural environments (forest, snow, grass, etc.): \{\textsf{\small tiger cat, white wolf, goose, Siberian husky, lion}\}. And so on. This is consistent with previous works that have found that, in some specific cases, the reason why a deep net classifies an object as a certain class is caused by the background or more generally by some confounding variables \citep{Ribeir_16a,Zech_18a}. It points to a possible vulnerability of the net, in that it may badly misclassify an object that happens to appear in an unusual background (say, a bald eagle standing on a road).

\subsubsection{Manipulating the deep net features via masks}

We derive masks using the mimic tree ($\lambda = 1$). Figures~\ref{f:VGG-masks-test}--\ref{f:VGG-masks-training} show confusion matrices, which are self-explanatory, over all instances (test and training, respectively) regarding the deep net, the tree and the masked deep net. Generally, the masks affect the deep net classification in the same way as the tree. This is to be expected since the tree has a very similar error and confusion matrix as the net, but it is still surprising in how well it works in most cases. This also indicates that certain neurons of the deep net (those critically involved in the masks) play a well-defined role in the classification. The number of features that a mask critically needs to perform its job is very small, around 200 (out of 8\,192); for MNIST it is much smaller, around 40.

We emphasize two things. First, these confusion matrices are constructed using \emph{all instances (training or test)}. Hence, our conclusions are robust and global, unlike other works that either work locally on a single instance by design, or work globally but show good results on a handful of instances only. By reporting the results aggregated over all instances, we demonstrate the robustness of our masks. Second, \emph{what we show is the result of applying the original VGG16 deep net to the masked features}, not of applying the tree to the masked features (which would work perfectly by construction). This shows that the masks constructed using the tree mimic transfer almost perfectly to the deep net.

Let us analyze the different panels of the figure in more detail:
\begin{itemize}
\item The two confusion matrices ``ground truth vs deep net vs tree'' (which look like a diagonal yellow line) show that the deep net and the tree predictions are almost identical to each other and to the ground truth, indicating that the tree is a good mimic (blue is 0 and yellow 1).
\item The confusion matrix ``features selected by the tree''  (which looks like a diagonal yellow line) shows that if we mask out all features in VGG16 except those selected by the tree, the classification ability remains the same, indicating that the tree is able to detect a subset of features that are enough for good classification.
\item The 6 confusion matrices ``\textsc{All class $k_1$ to class $k_2$}'' demonstrate (in selected combinations of $k_1$ and $k_2$) that this mask works quite reliably. Ideally, entry $(k_1,k_1)$ should become 0 and entry $(k_2,k_1)$ should become 1, with the rest of the entries remaining unchanged. We often observe a bluish vertical bar at $k_2$, indicating that a small proportion of the $k_1$ instances are classified as a class different from $k_2$.
\item The 16 confusion matrices ``\textsc{None to class $k$}'' (for $k \in \{0,1,\dots,\text{F}\}$) should ideally make the entry at $(k,k)$ equal to zero (this entry is originally equal to (almost) 1) and distribute its mass over any other classes (entries $(k,k')$ with $k \neq k'$), with the rest of the entries remaining unchanged. This succeeds better in some classes than in others, but generally works well.
\item The 16 confusion matrices ``\textsc{All to class $k$}'' (for $k \in \{0,1,\dots,\text{F}\}$) should ideally look like a vertical yellow line at $k$, and indeed they do in all cases.
\end{itemize}
All of the above is true for both training and test instances, although the masks work slightly better for the training instances.

\subsubsection{Illustration of the masks with an actual image}

\begin{figure}[p]
  \centering
  \psfrag{un}[][]{*}
  \begin{tabular}{@{}c@{\hspace{3ex}}c@{\hspace{4ex}}c@{\hspace{4ex}}c@{}}
    \rotatebox{90}{\hspace*{3.5ex}\small Original}&
    \includegraphics*[height= .15 \linewidth,width=.16\linewidth]{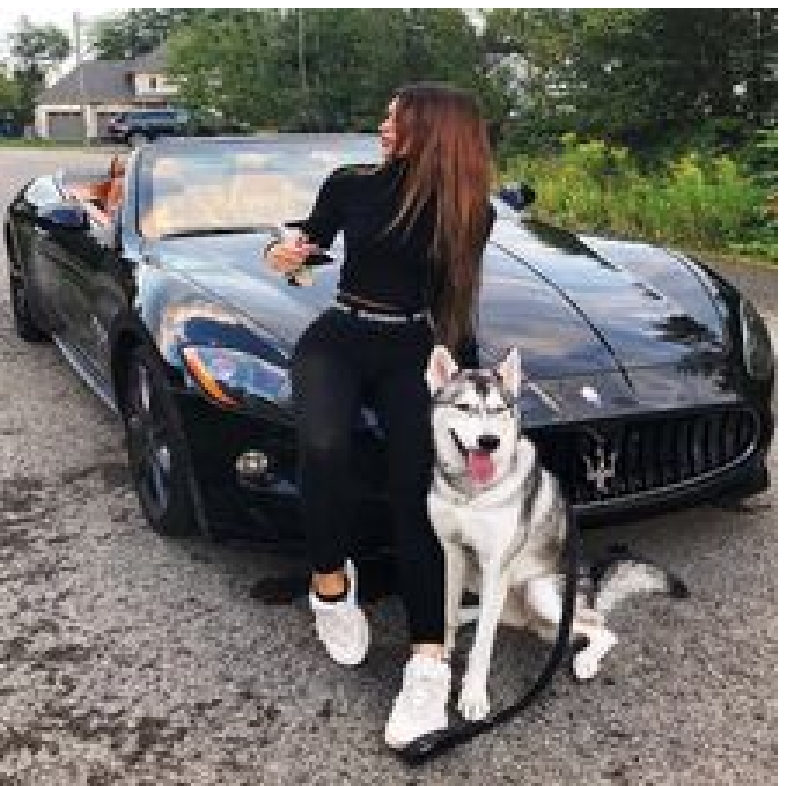}&
    \begin{tabular}[b]{@{}r@{}}
      \psfrag{outside axis limit}[c][][.6]{outside axis limit(6826)$\rightarrow$}
      \psfrag{feature count}[][][.5]{feature count}
      \includegraphics*[height= .075 \linewidth,width=.342\linewidth]{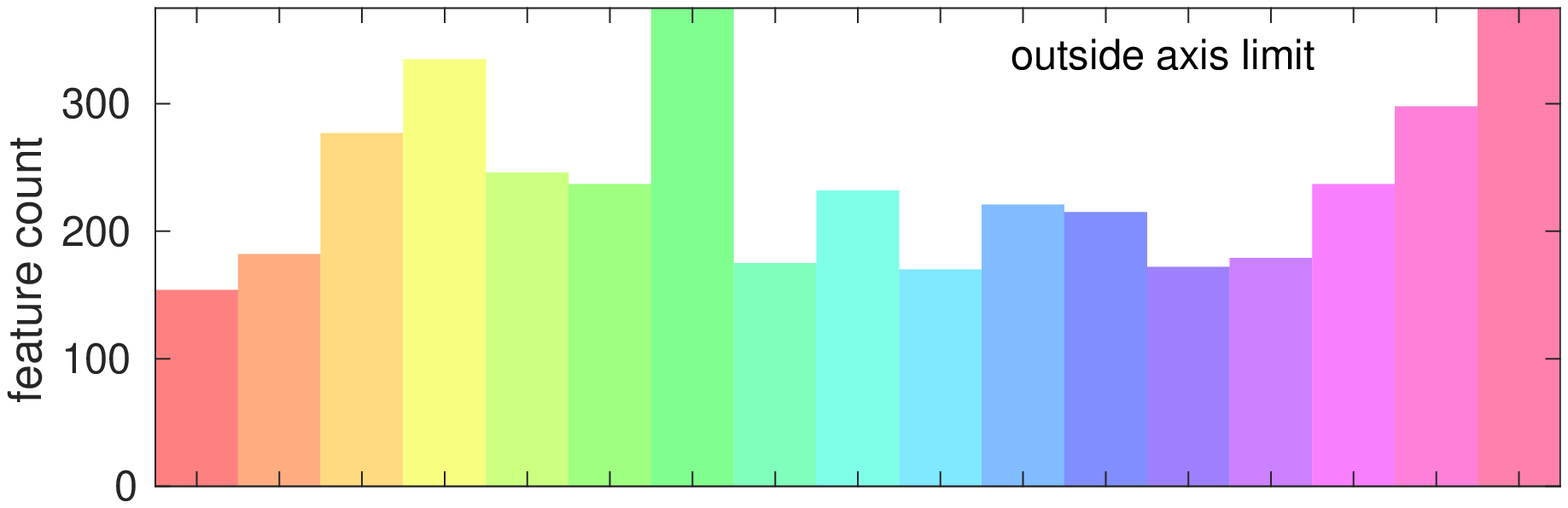}\\
      \psfrag{feature value}[][][.5]{feature value}
      \hspace*{0.5ex}\includegraphics*[height= .075 \linewidth, width=.345\linewidth]{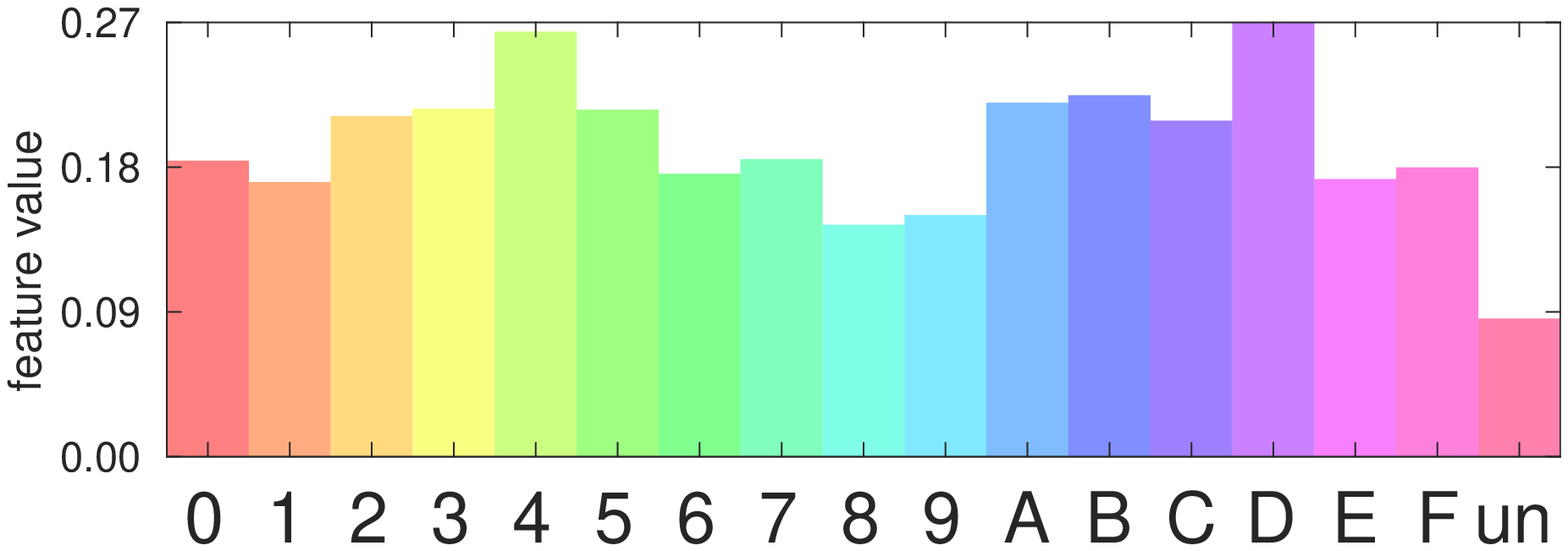}
    \end{tabular} &
    \psfrag{l1}[tl][][.6][0]{}
    \psfrag{l2}[l][][.6][90]{}
    \psfrag{l3}[l][][.6][90]{}
    \psfrag{l4}[cb][b][.6][0]{\textsf{\caja[0.5]{b}{c}{Siberian \\ husky}}}
    \psfrag{l5}[l][][.6][90]{}
    \psfrag{l10}[l][][.6][90]{}
    \psfrag{l11}[cb][b][.6][0]{\textsf{\caja[0.5]{b}{c}{School \\ bus}}}
    \psfrag{l12}[l][][.6][90]{}
    \psfrag{l13}[cb][b][.6][0]{\textsf{\caja[0.5]{b}{c}{Sports \\ car}}}
    \psfrag{softmax value}[b][]{softmax value}
    \includegraphics*[height= .16 \linewidth,width=.3\linewidth]{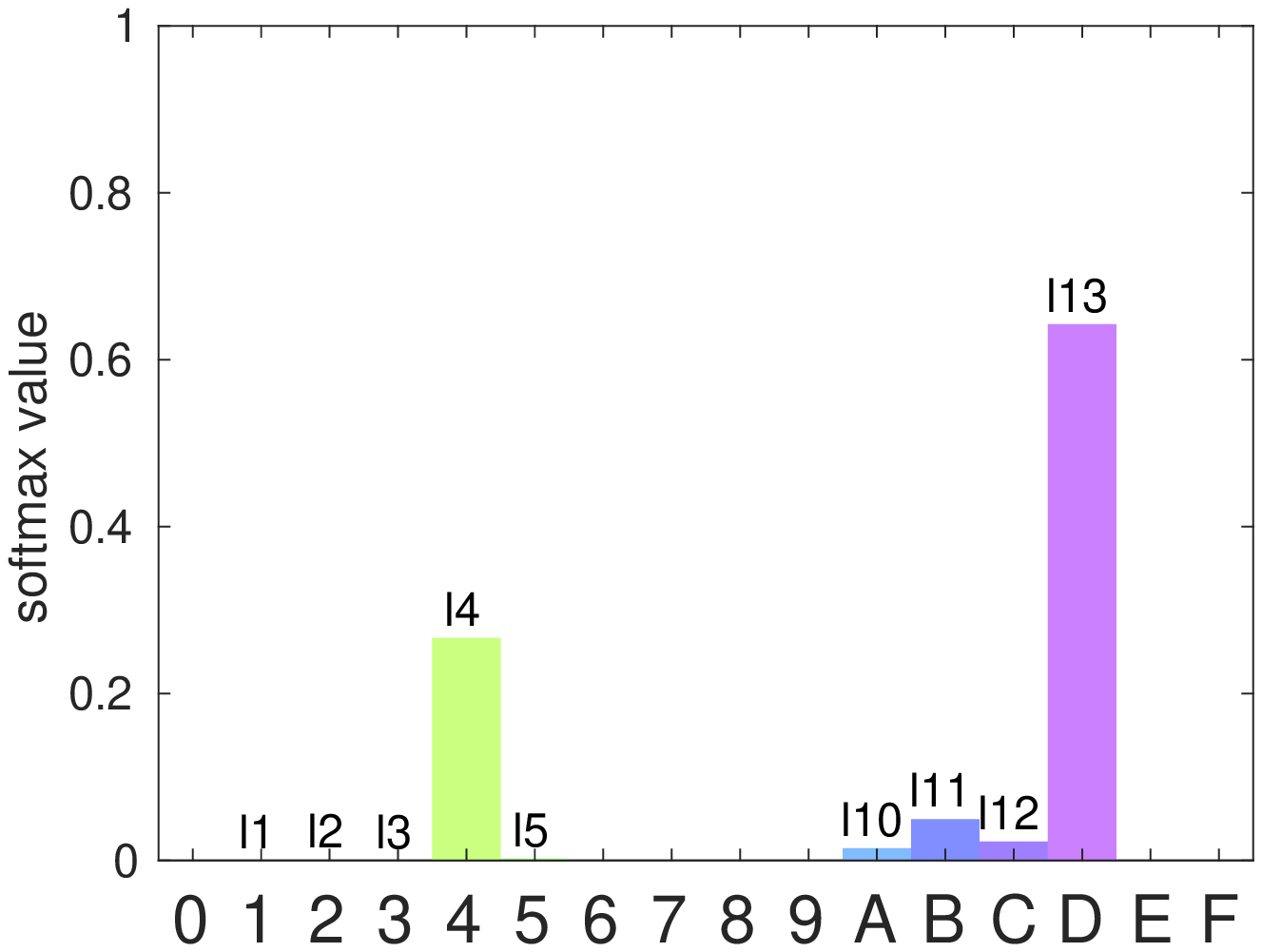}\\[3ex]
    \rotatebox{90}{\hspace*{3ex}\small\caja{c}{c}{Mask in \\ feature space}} &
    \parbox[b][.13\linewidth][c]{.18\linewidth}{\caja{c}{c}{\textsc{All to class} \\ \textsc{``Siberian husky''}\\ mask is applied \\ $\xrightarrow{\hspace*{2cm}}$}}&
    \begin{tabular}[b]{@{}r@{}}
      \psfrag{outside axis limit}[c][][.6]{outside axis limit(6826)$\rightarrow$}
      \psfrag{feature count}[][][.5]{feature count}
      \includegraphics*[height= .075 \linewidth,width=.342\linewidth]{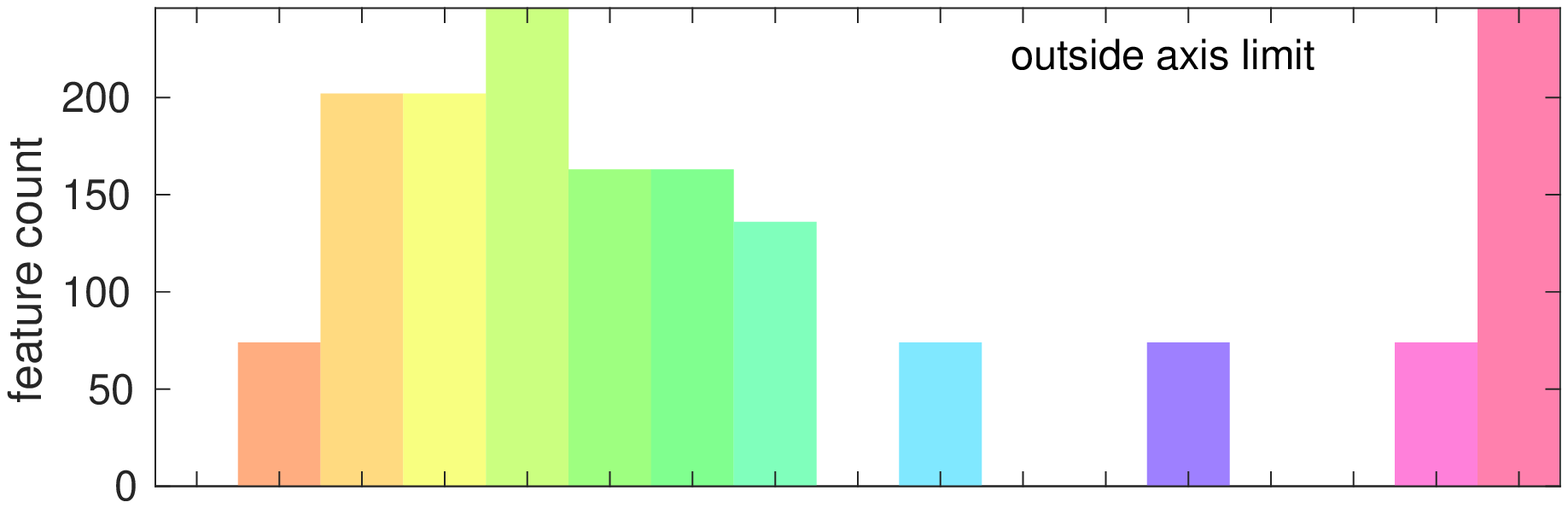}\\
      \psfrag{feature value}[][][.5]{feature value}
      \hspace*{0.5ex}\includegraphics*[ height= .075 \linewidth,width=.345\linewidth]{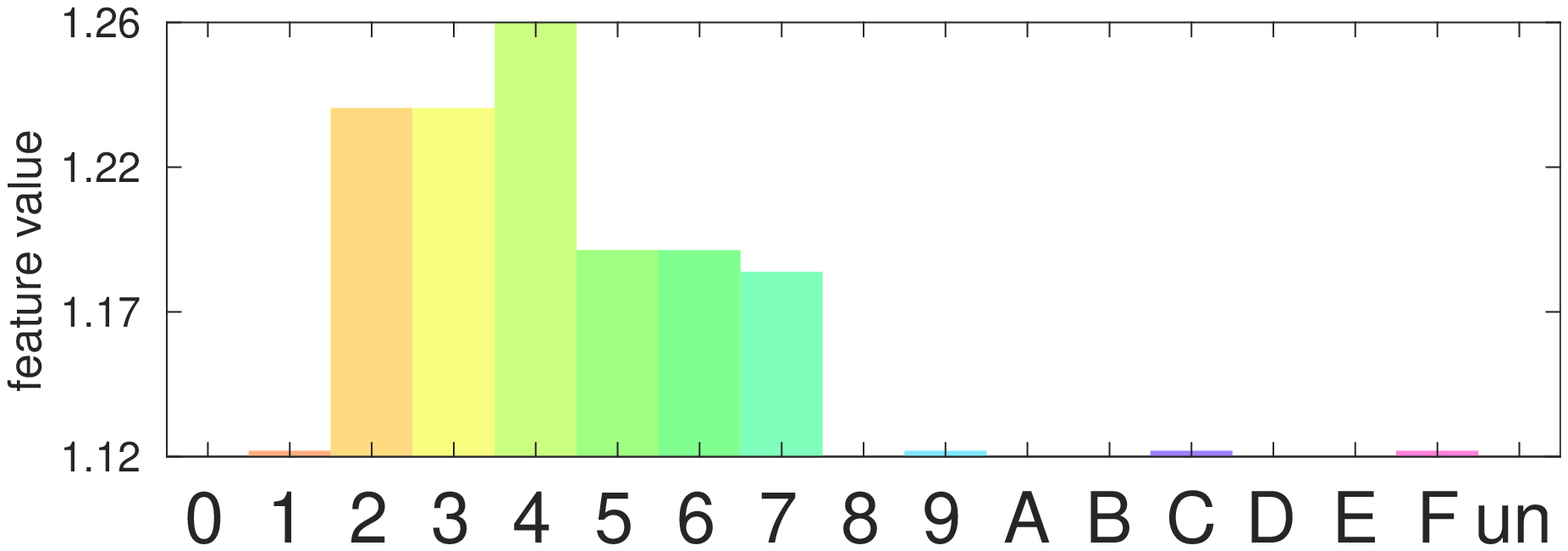}
    \end{tabular} &
    \psfrag{l1}[tl][][.6][0]{}
    \psfrag{l2}[l][][.6][90]{}
    \psfrag{l3}[l][][.6][90]{}
    \psfrag{l4}[l][][.6][90]{}
    \psfrag{l5}[l][][.6][90]{}
    \psfrag{l10}[l][][.6][90]{}
    \psfrag{l11}[l][][.6][90]{}
    \psfrag{l12}[l][][.6][90]{}
    \psfrag{l13}[l][][.6][90]{}
    \psfrag{softmax value}[b][]{softmax value}
    \includegraphics*[height= .16 \linewidth,width=.3\linewidth]{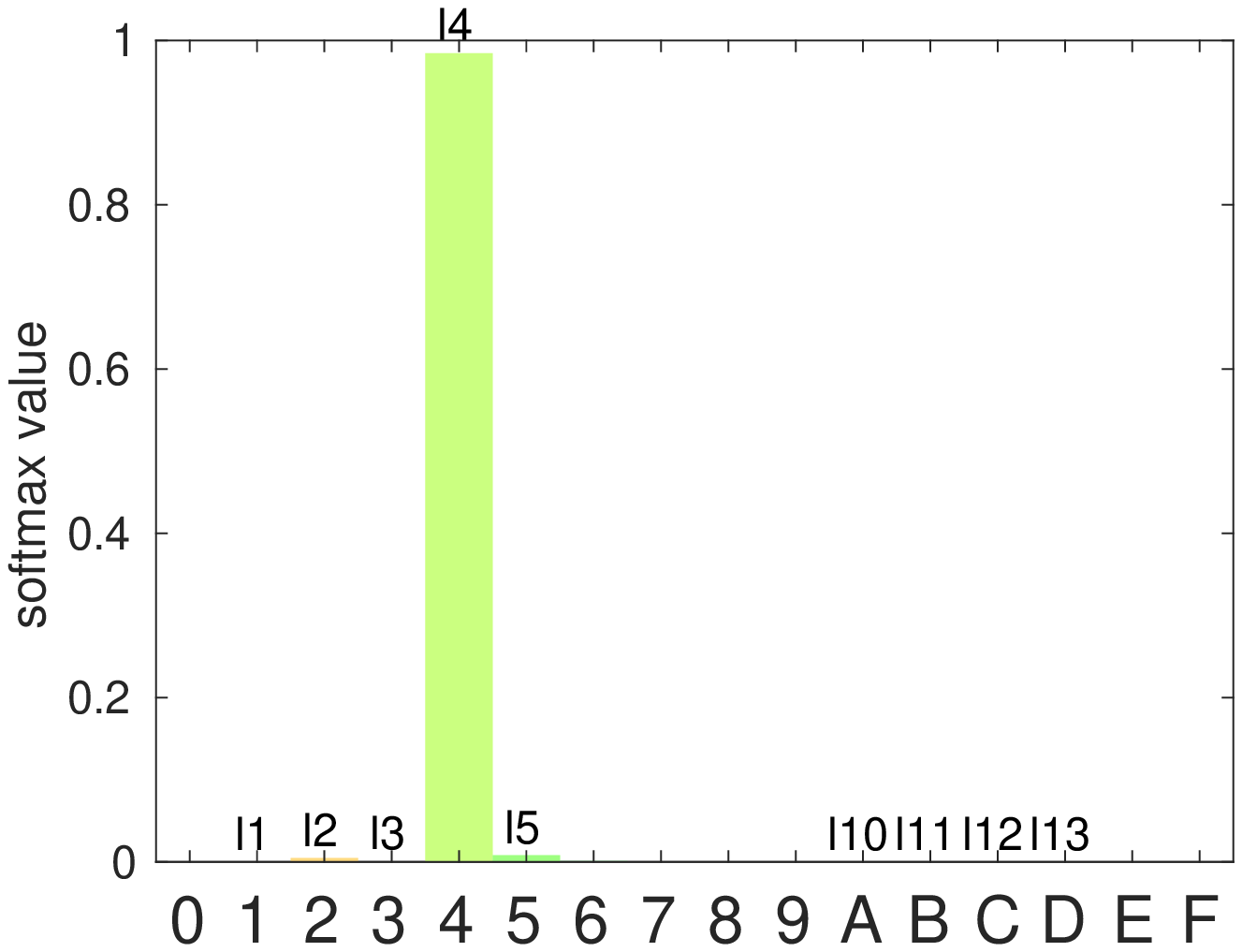}\\[3ex]
    \rotatebox{90}{\hspace*{1ex}\small\caja{c}{c}{Manual mask in \\ image space}} &
    \includegraphics*[height= .15 \linewidth,width=.16\linewidth]{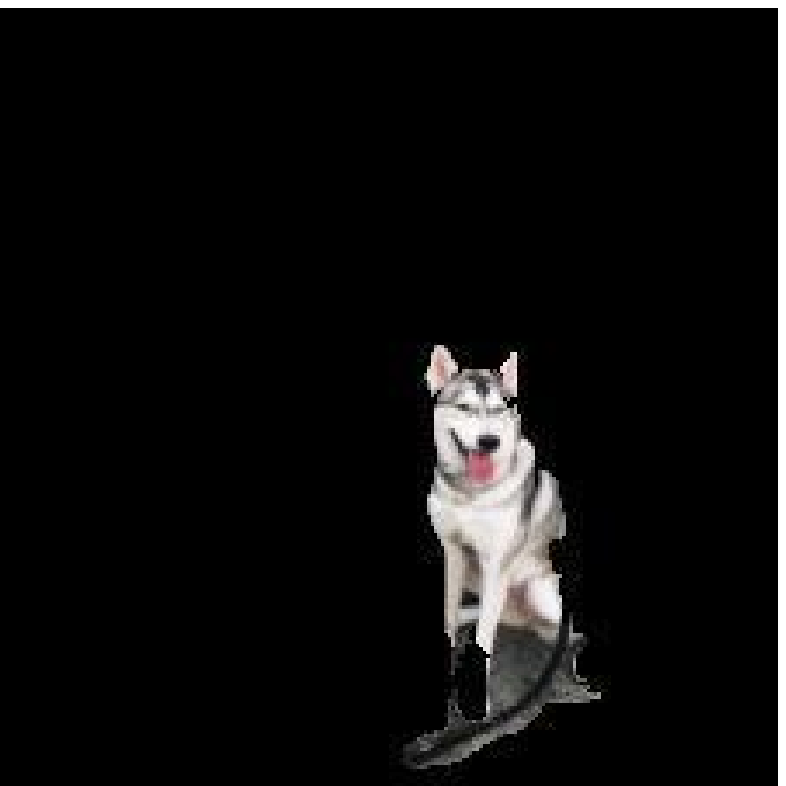}&
    \begin{tabular}[b]{@{}r@{}}
      \psfrag{outside axis limit}[c][][.6]{outside axis limit(6826)$\rightarrow$}
      \psfrag{feature count}[][][.5]{feature count}
      \includegraphics*[height= .075 \linewidth,width=.342\linewidth]{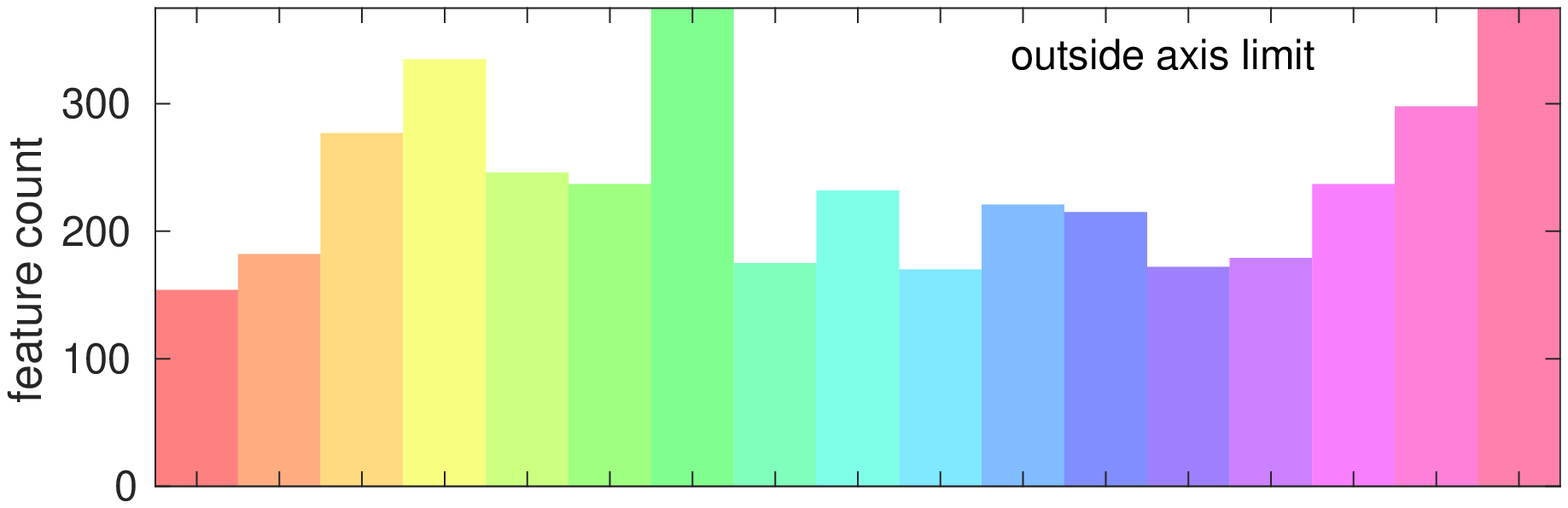}\\
      \psfrag{feature value}[][][.5]{feature value}
      \hspace*{0.5ex}\includegraphics*[ height= .075 \linewidth,width=.345\linewidth]{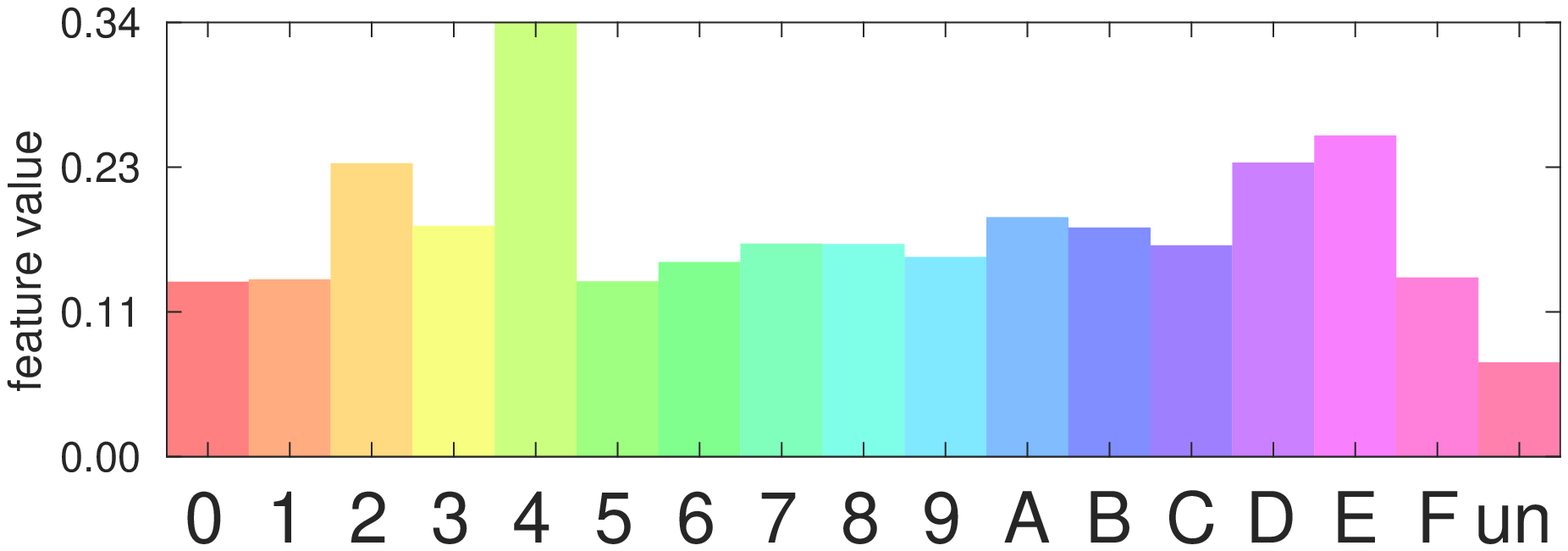}
    \end{tabular} &
    \psfrag{l1}[tl][][.6][0]{}
    \psfrag{l2}[l][][.6][90]{}
    \psfrag{l3}[l][][.6][90]{}
    \psfrag{l4}[l][][.6][90]{}
    \psfrag{l5}[l][][.6][90]{}
    \psfrag{l10}[l][][.6][90]{}
    \psfrag{l11}[l][][.6][90]{}
    \psfrag{l12}[l][][.6][90]{}
    \psfrag{l13}[l][][.6][90]{}
    \psfrag{softmax value}[b][]{softmax value}
    \includegraphics*[height= .16 \linewidth,width=.3\linewidth]{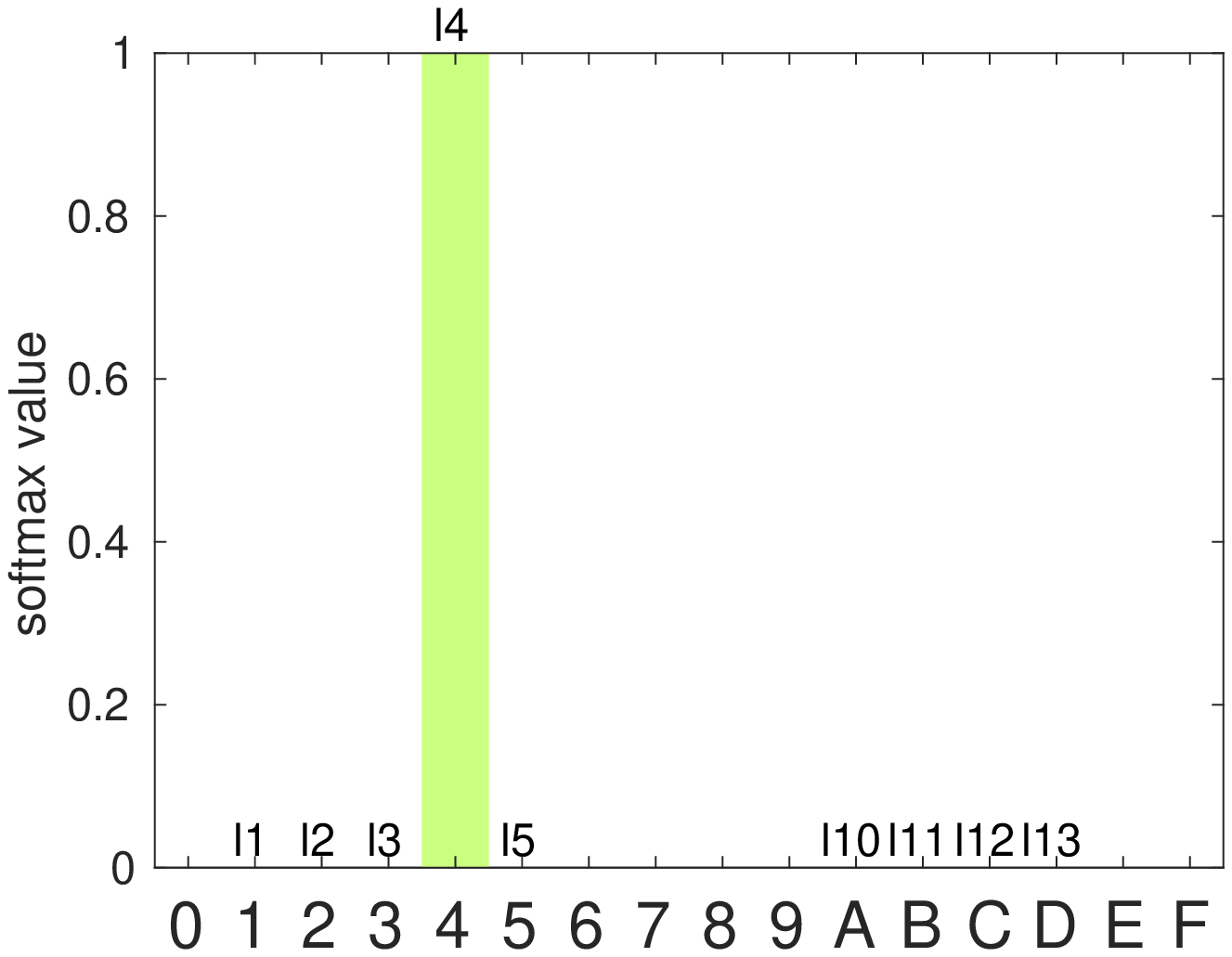}\\[3ex]
    \rotatebox{90}{\hspace*{1ex}\small\caja{c}{c}{Mask in image \\ space obtained \\ by features}} &
    \includegraphics*[height= .15 \linewidth,width=.16\linewidth]{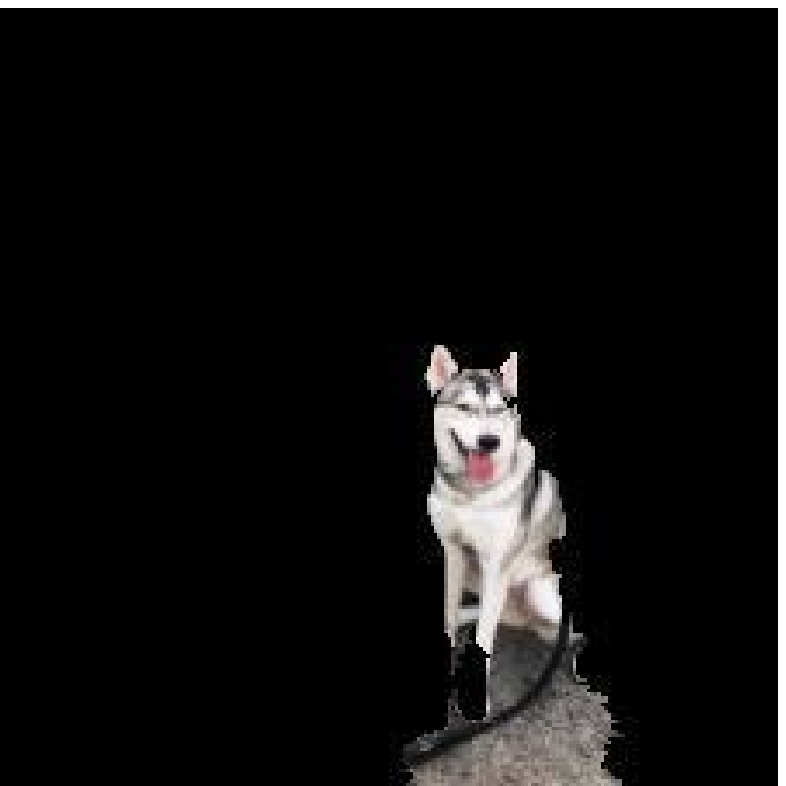}&
    \begin{tabular}[b]{@{}r@{}}
      \psfrag{outside axis limit}[c][][.6]{outside axis limit(6826)$\rightarrow$}
      \psfrag{feature count}[][][.5]{feature count}
      \includegraphics*[height= .075 \linewidth,width=.342\linewidth]{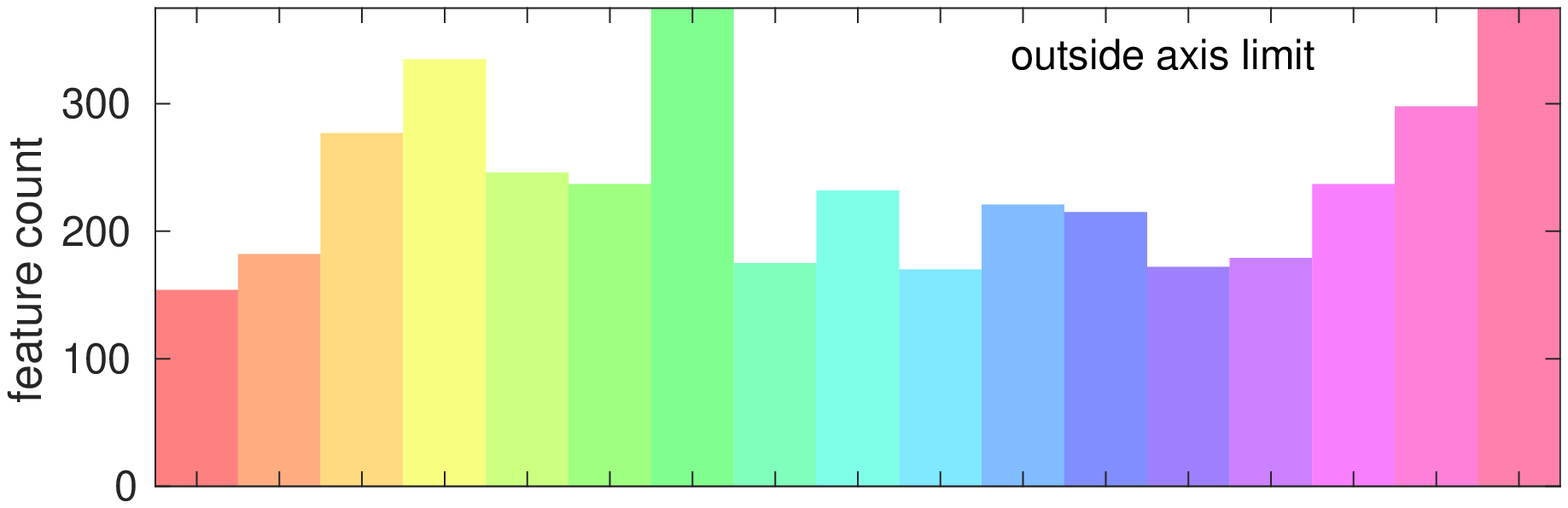}\\
      \psfrag{feature value}[][][.5]{feature value}
      \hspace*{0.5ex}\includegraphics*[ height= .075 \linewidth,width=.345\linewidth]{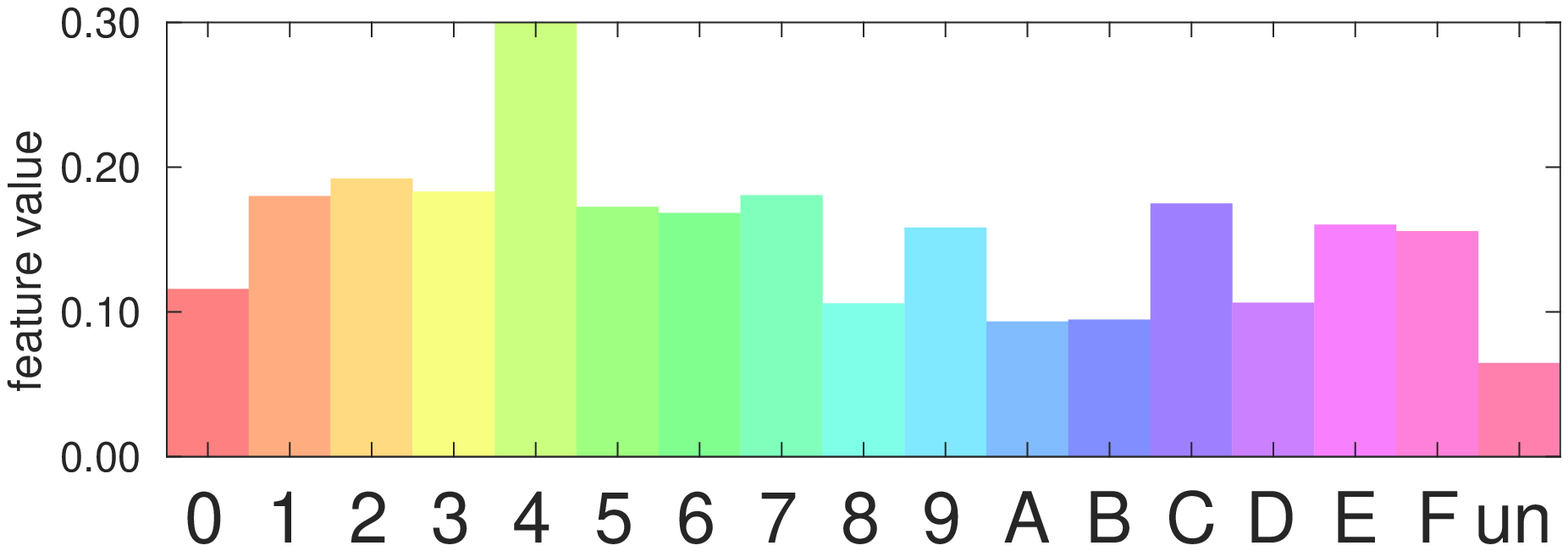}
    \end{tabular} &
    \psfrag{l1}[tl][][.6][0]{}
    \psfrag{l2}[l][][.6][90]{}
    \psfrag{l3}[l][][.6][90]{}
    \psfrag{l4}[l][][.6][90]{}
    \psfrag{l5}[lb][b][.6][0]{\textsf{\caja[0.5]{b}{c}{White \\ wolf}}}
    \psfrag{l10}[l][][.6][90]{}
    \psfrag{l11}[l][][.6][90]{}
    \psfrag{l12}[l][][.6][90]{}
    \psfrag{l13}[l][][.6][90]{}
    \psfrag{softmax value}[b][]{softmax value}
    \includegraphics*[height= .16 \linewidth,width=.3\linewidth]{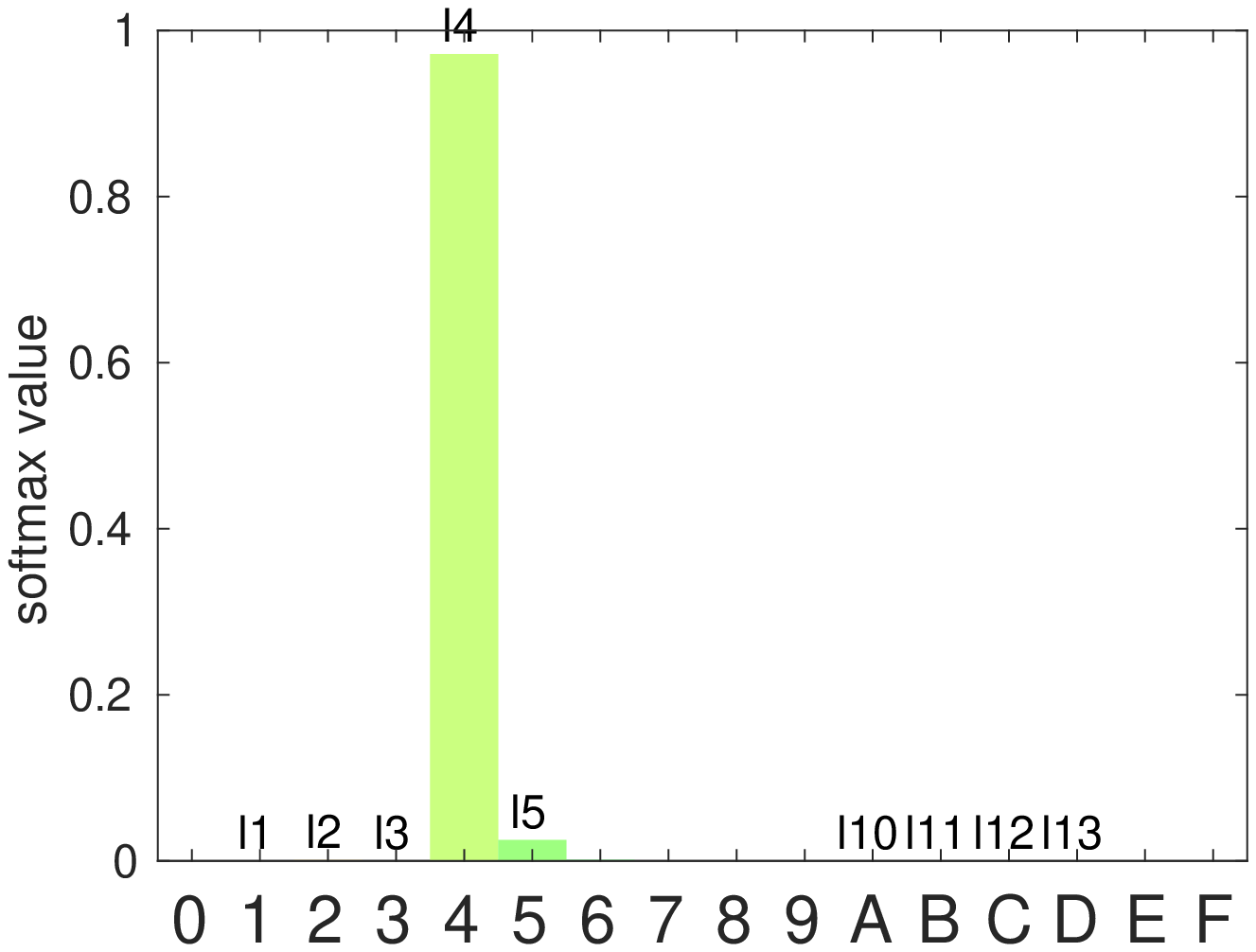}\\[3ex]
    \rotatebox{90}{\hspace*{6ex}\small\caja{c}{c}{Mask in \\ feature space}} &
    \parbox[b][.18\linewidth][c]{.16\linewidth}{\caja{c}{c}{\textsc{All to class} \\ \textsc{``Bald eagle''}\\ mask is applied \\ $\xrightarrow{\hspace*{2cm}}$}}&
    \begin{tabular}[b]{@{}r@{}}
      \psfrag{outside axis limit}[c][][.6]{outside axis limit(6826)$\rightarrow$}
      \psfrag{feature count}[][][.5]{feature count}
      \includegraphics*[height= .075 \linewidth,width=.342\linewidth]{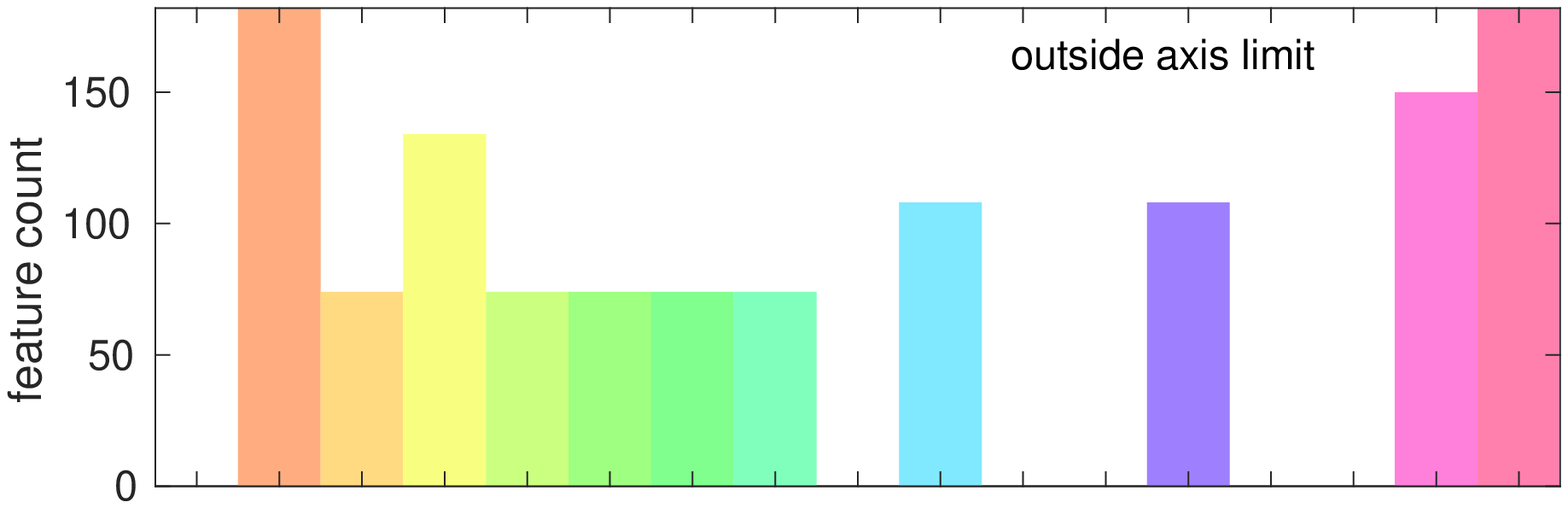}\\
      \psfrag{class-based feature group}[][][1]{class-based feature group}
      \psfrag{feature value}[][][.5]{feature value}
      \hspace*{0.5ex}\includegraphics*[height= 0.11 \linewidth,width=.345\linewidth]{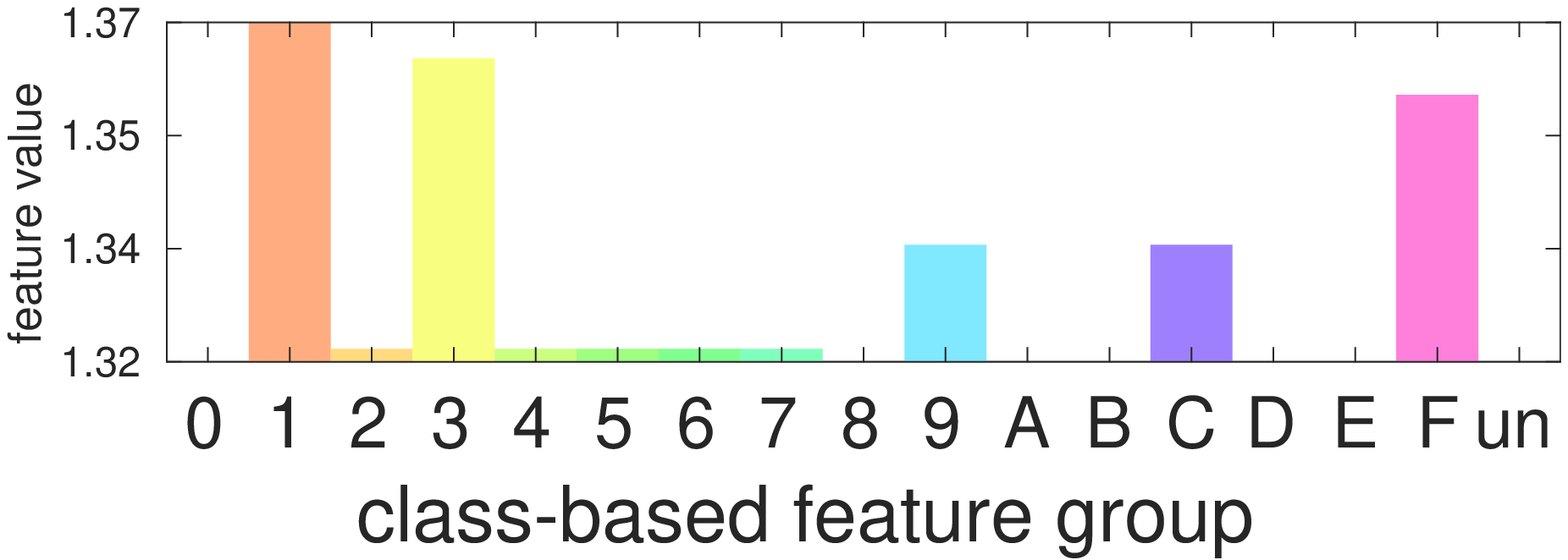}
    \end{tabular} &
    \psfrag{l1}[tl][t][.6][0]{\textsf{\caja[0.5]{t}{c}{Bald \\ eagle}}}
    \psfrag{l2}[c][][.6][0]{}
    \psfrag{l3}[lb][b][.6][0]{\textsf{\caja[0.5]{b}{c}{Killer \\ whale}}}
    \psfrag{l4}[l][][.6][90]{}
    \psfrag{l5}[l][][.6][90]{}
    \psfrag{l10}[l][][.6][90]{}
    \psfrag{l11}[l][][.6][90]{}
    \psfrag{l12}[l][][.6][90]{}
    \psfrag{l13}[l][][.6][90]{}
    \psfrag{softmax value}[b][]{softmax value}
    \psfrag{classes}[][][1]{classes}
    \includegraphics*[height= .185 \linewidth,width=.3\linewidth]{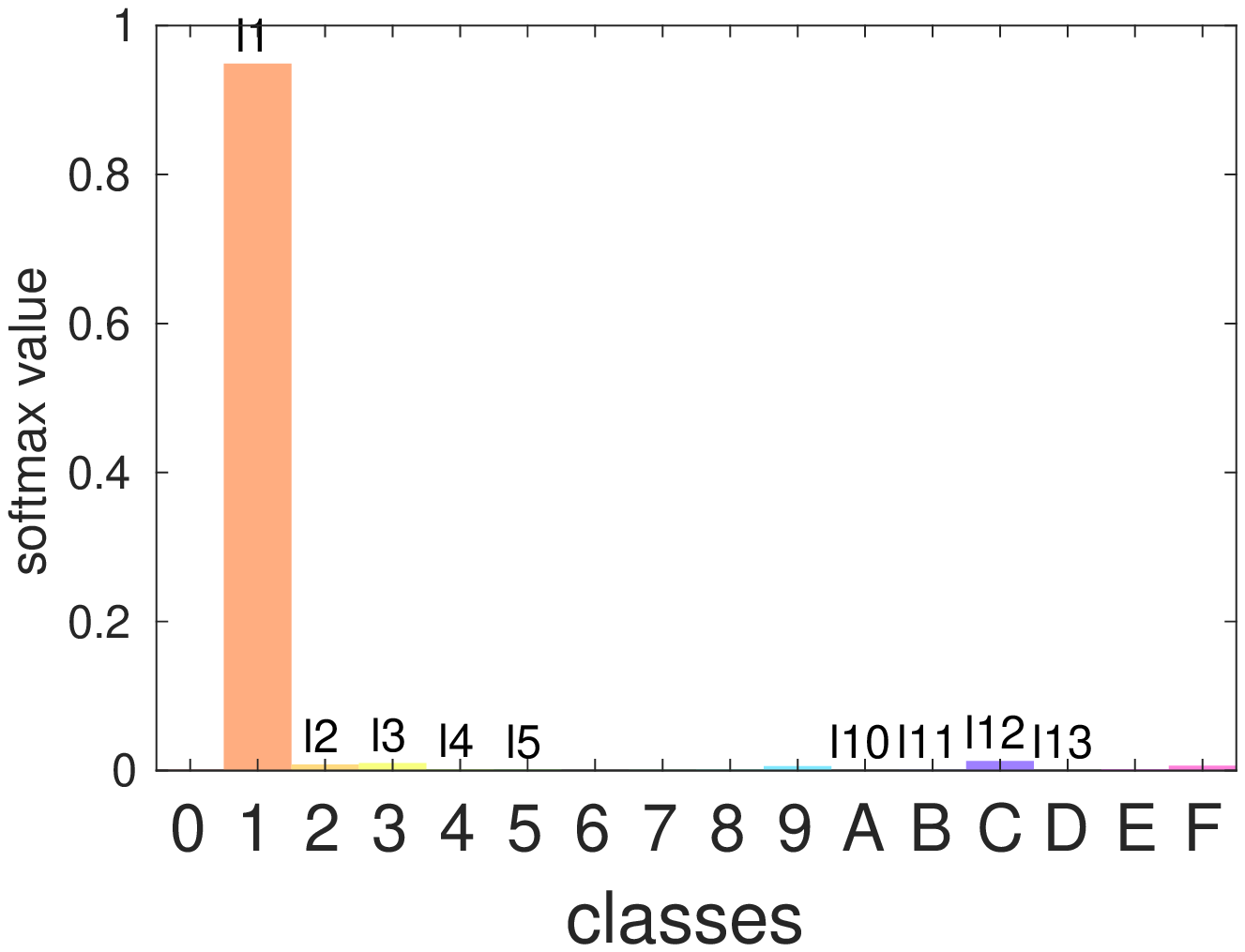}
  \end{tabular}
  \caption{Illustration of masks for a particular image (VGG16 network on ImageNet subset). Column 1 shows the image masks (when available). Column 2 summarizes the 8\,192 feature values as two histograms: on the upper panel, the number of features in each class group (listed in the X axis as 0--F, where ``$\ast$'' means features not used by the tree); on the lower panels, the average feature value (neuron activation) per class group. Column 3 shows the histogram of corresponding softmax values. Row 1 shows the original image. Row 2 shows a mask in feature space to classify it as ``Siberian husky''. Row 3 shows a mask manually cropped in the image, whose features resemble those of row 2. Row 4 shows a mask in feature space obtained by finding the top-3 superpixels whose features most resemble those of the masked features of row 2. Row 5 shows a mask in feature space to classify the image as ``bald eagle''.}
  \label{f:mask-img}
\end{figure}

Fig.~\ref{f:mask-img} illustrates the mask behavior in an image not in the dataset. The middle column histograms show the deep net features (grouped by class). In each row, the top histogram shows the feature values, and the bottom histogram shows the number of features selected for each class. Next, we show how masking the features drastically alters in a controlled way the softmax output. In row 2, when we apply the \textsc{All to class ``Siberian husky''} mask, the network now classifies the image as ``siberian husky''. Similarly, in row 6, when we apply the \textsc{All to class ``bald Eagle''} mask, the network now classifies the original image as ``bald eagle'' with large confidence, compared to row 1, where without the mask the softmax value for ``bald eagle'' is close to zero. We also show how the mask correlates with superpixels (perceptual groups of pixels obtained by oversegmentation) in the image, either manually cropped (row 3) or optimized to invert the desired deep net features (row 4). 

To obtain results like those above, the general procedure is as follows. Firstly (in an offline phase), we train the tree mimic and construct a subset of features $\calS_k$ for each class $k$, using the \textsc{All to class $k$} mask. This defines a score for an input image \x\ as $s_k(\x) = \sum_{i \in \calS_k}{ F_i(\x) }$, where $F_i(\x)$ is the feature $i$ computed by the deep neural net for \x. We can then discard the tree and the classifier part of the deep net. All we need is the feature-extraction part of the deep net and the class sets $\calS_1,\dots,\calS_K$.

Then (in an online phase), given an input image and a target class $k$, we split the image into superpixels (using some oversegmentation algorithm), compute the score for each superpixel, and report the superpixels with lowest score (most salient).

\subsubsection{Control experiments}

We run some experiments to test the robustness of our findings (see details in appendix~\ref{s:control}):
\begin{enumerate}
\item We trained trees with TAO using 5 different random initial trees and verified we can achieve very similar results to those we report (masks, etc.). \\
  Note that, for a given neural network and dataset, it is conceivable that we could learn very different trees (even resulting in different masks) because of local optima in the tree training, correlation or redundancy of features in the neural network, etc. But regardless of that, because we evaluate the masks not just in the tree but in the original neural network, we can claim that the chosen masks work as described earlier.
\item We trained a CART tree \citep{Breiman_84a,Therneau_19a,Pedreg_11a} on the VGG16 features. This created a huge axis-aligned tree with error of 1.97\% (training) and 21.2\% (test), depth 54, 1\,381 nodes and using 619 features. The large test error means the tree is inadequate as a mimic of VGG16, and the tree size would make it useless for explanation purposes anyway. This demonstrates the inability of CART to learn accurate trees for complex, high-dimensional data.
\item We trained a TAO tree on a rotated version of the VGG16 features, i.e., multiplying the feature vector by an orthogonal matrix. This has the effect of mixing all the features in an invertible way. Without sparsity ($\lambda = 0$), the TAO tree achieves identical error to the unrotated features' tree (since the oblique nodes can absorb any linear transformation). However, when increasing $\lambda$ in order to force the tree to use few features, the test error jumps to 11.3\%, much higher than with the unrotated features. This shows that the features learnt by VGG are special in that they seem to operate in small groups associated with classes, rather than all or most features participating in each class.
\end{enumerate}

\subsection{Results on LeNet5 (MNIST dataset)}
\label{s:LeNet5}

\begin{figure}[p]
  \centering
  \begin{tabular}{@{}c@{}}
    \psfrag{error}[][]{Error (LeNet5)}
    \psfrag{c}[][B]{$\log_{10}\lambda$}
    \psfrag{net(train)}[l][][0.9]{\hspace{-2ex}net (train)}
    \psfrag{net(test)}[c][][0.9]{\hspace{-2ex}net (test)}
    \psfrag{tree(train)}[l][][0.9]{\hspace{-3ex}tree (train)}
    \psfrag{tree(test)}[l][][0.9]{\hspace{-3ex}tree (test)}
    \psfrag{Selected tree}[r][][0.9]{\hspace{0ex}Selected tree}
    \includegraphics*[width=.54\linewidth]{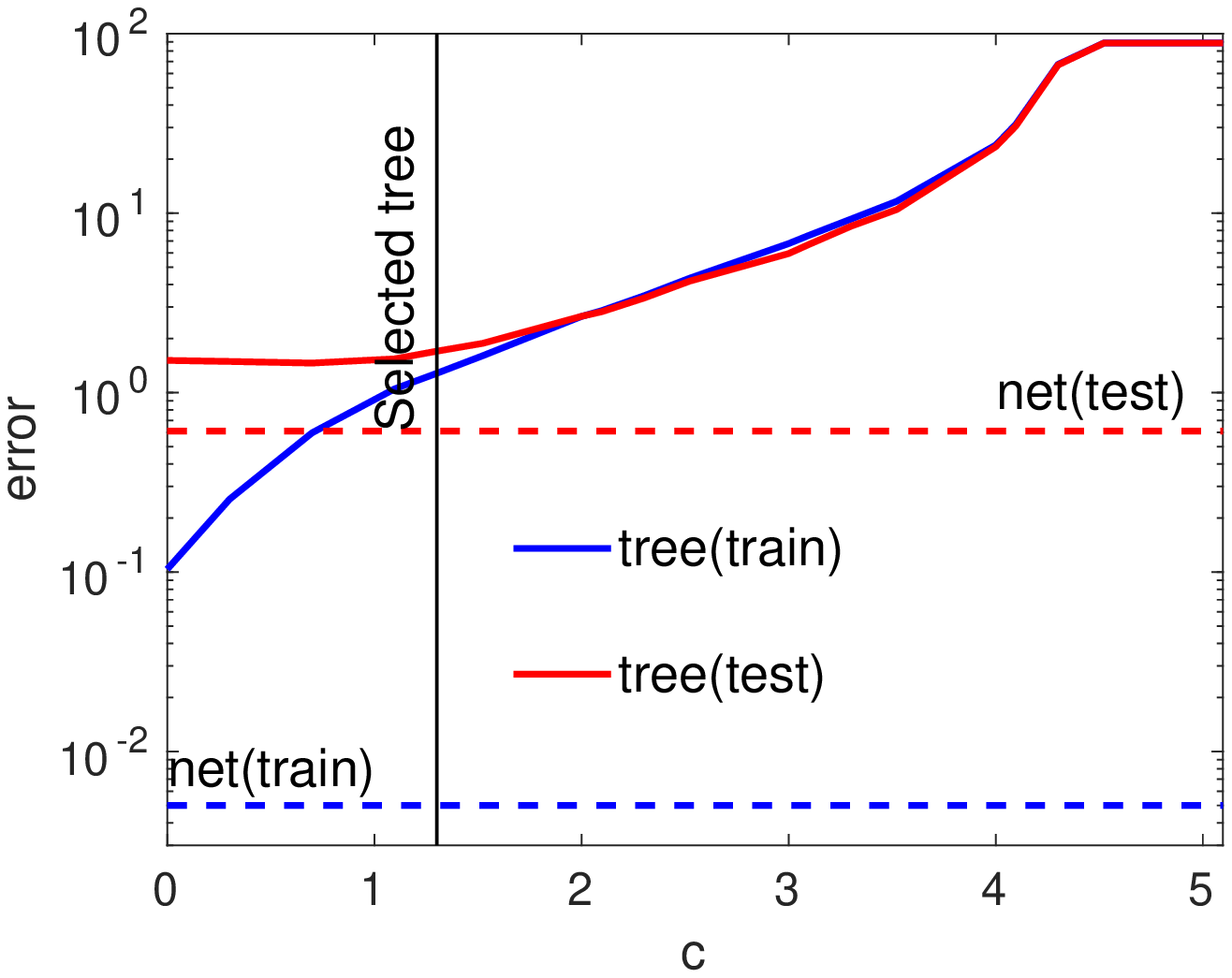} \\[1ex]
    \psfrag{nodes}[][]{}
    \psfrag{depths}[][]{}
    \psfrag{c}[][B]{$\log_{10}\lambda$}
    \psfrag{\# nodes}[c][][0.9]{\hspace{4ex}\# nodes}
    \psfrag{depth}[c][][0.9]{\hspace{4ex}depth}
    \psfrag{Selected tree}[c][][0.9]{\hspace{4ex}Selected tree}
    \hspace{5ex}\includegraphics*[width=.57\linewidth]{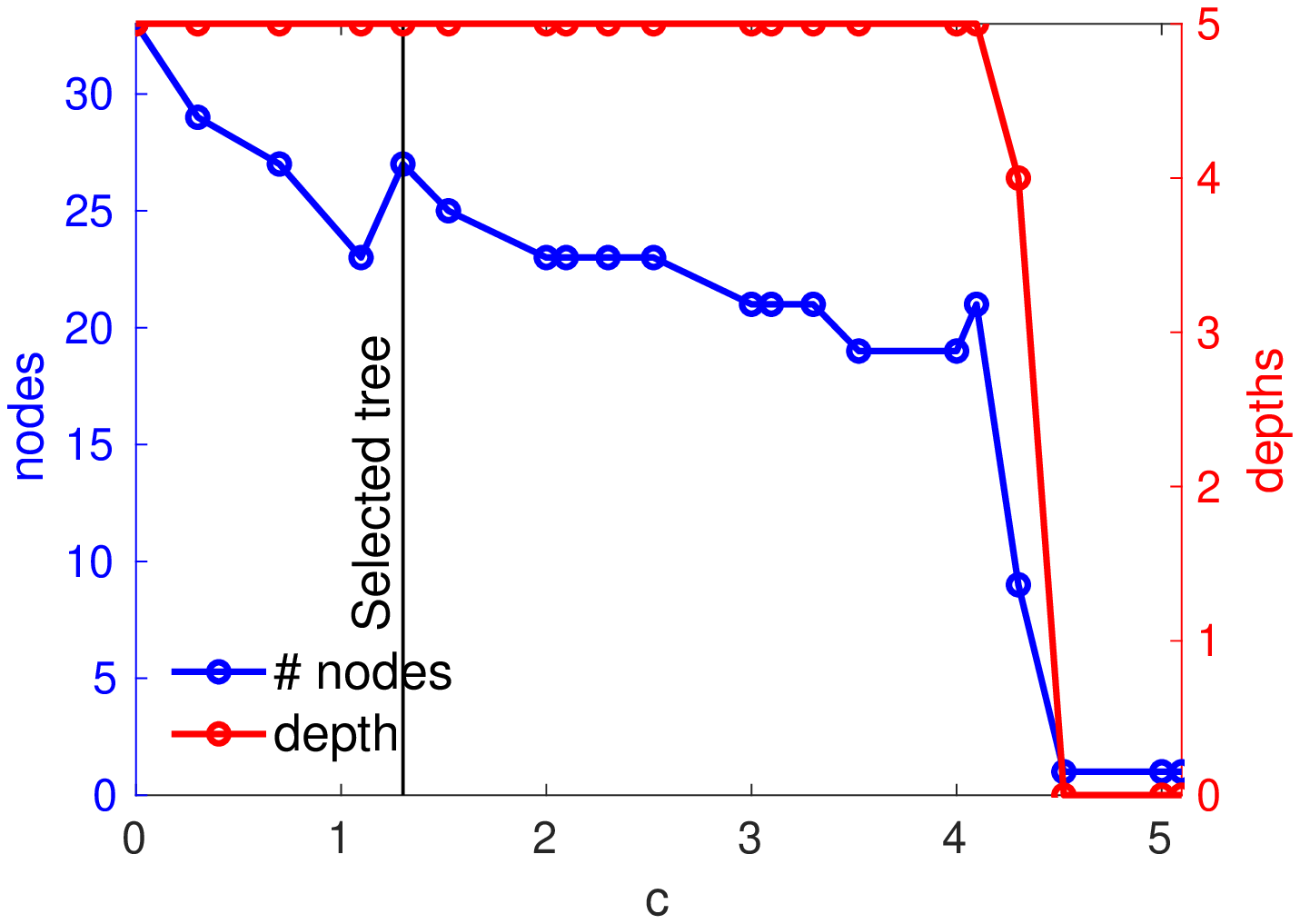} \\[1ex]
    \psfrag{sparse}[][][0.9]{\hspace{-2ex}\# nonzero weights ($\times$1000)}
    \psfrag{c}[][B]{$\log_{10}\lambda$}
    \psfrag{Selected tree}[c][][0.9]{\hspace{2ex}Selected tree}
    \includegraphics*[width=.52\linewidth]{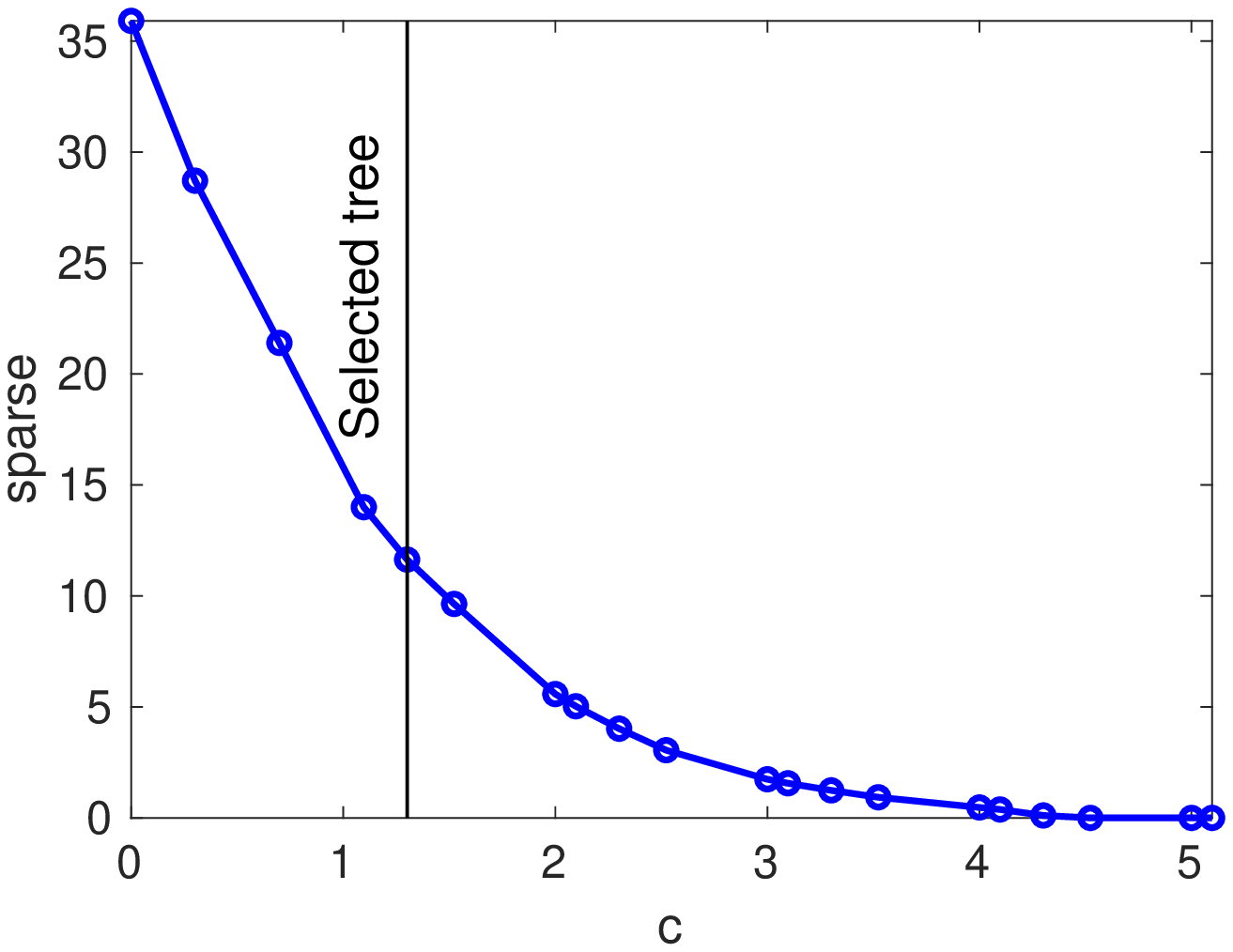}
  \end{tabular}
  \caption{Classification error (training and test) and number of nodes and of nonzero weights of the trees as a function of $\lambda$ for LeNet5. The vertical line indicates the tree we selected as mimic ($\lambda = 20$). Compare this with fig.~\ref{f:error}.}
  \label{f:LeNet5-error}
\end{figure}

\begin{figure}[p]
  \centering
  \begin{tabular}{@{}c@{}}
    \includegraphics*[width=\linewidth]{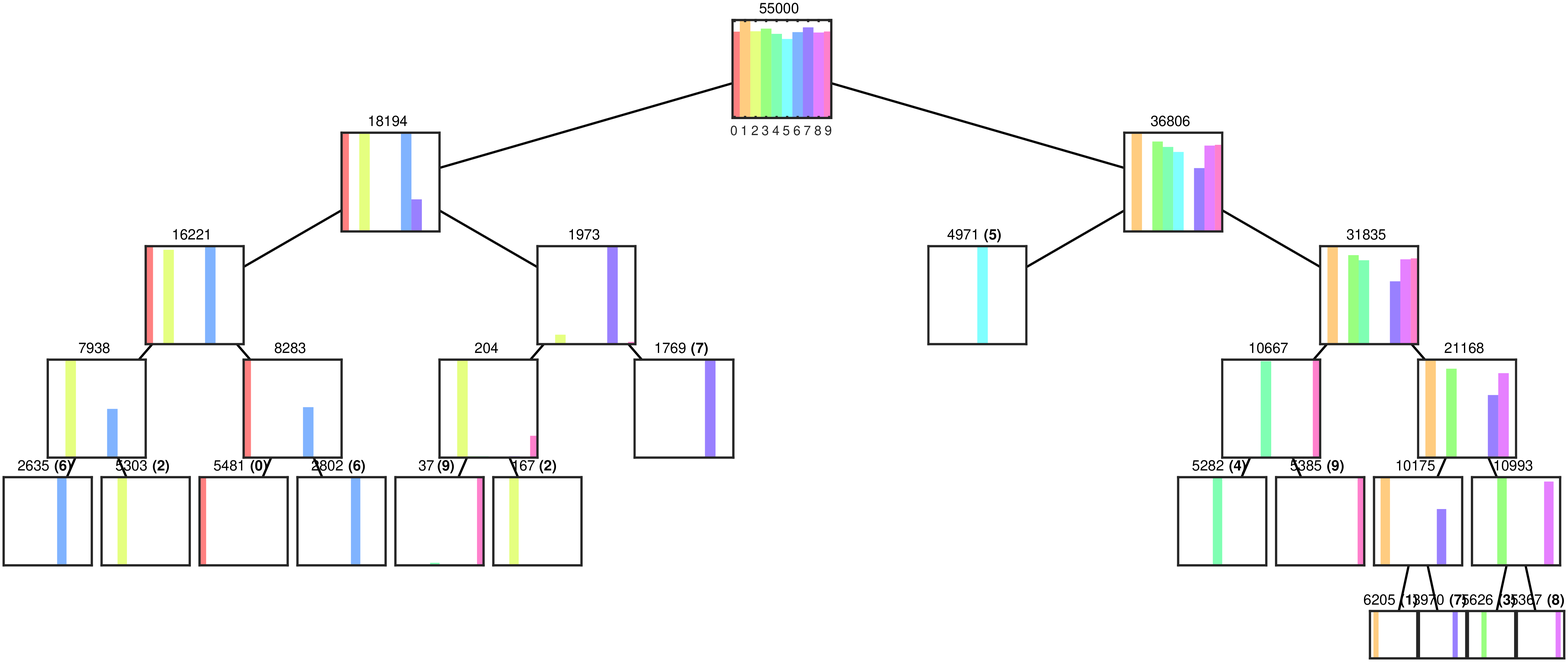} \\[5ex]
    \includegraphics*[width=\linewidth]{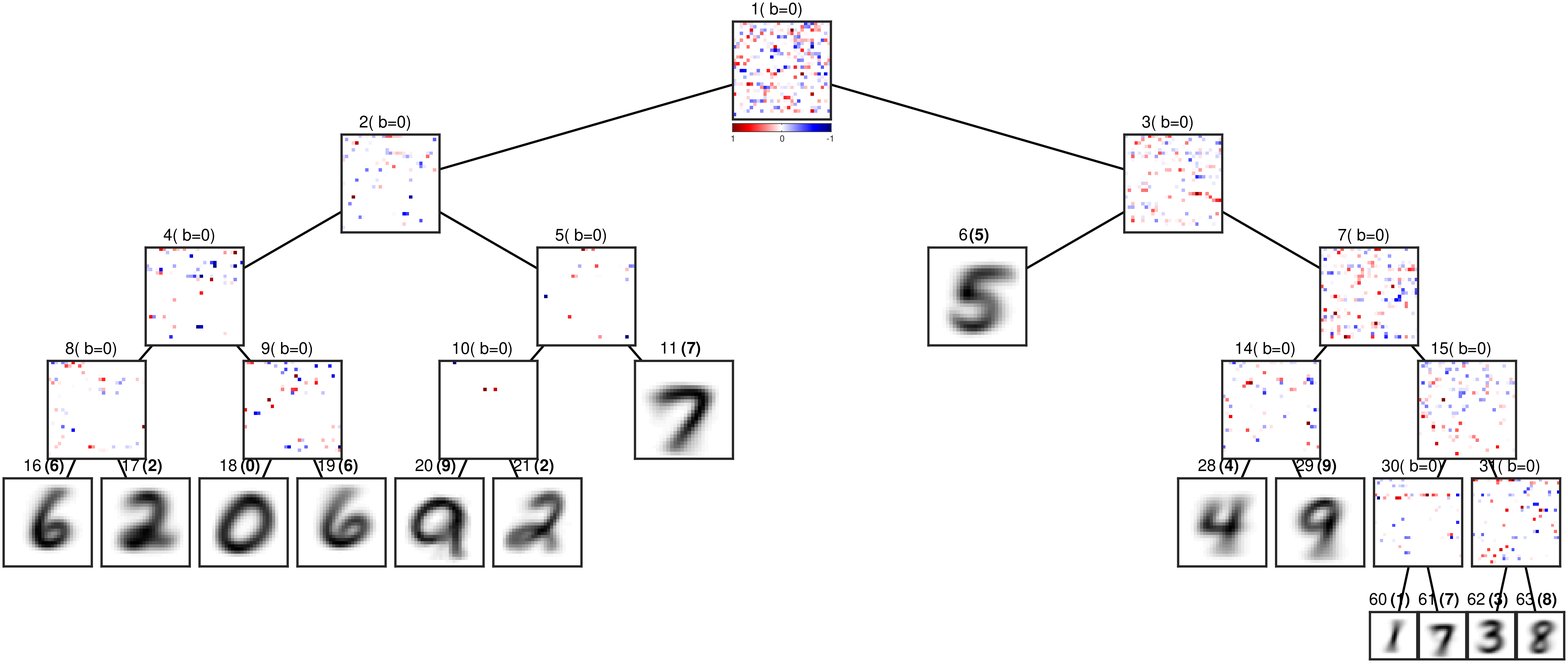}
  \end{tabular}
  \caption{Tree selected as mimic for LeNet5 features ($\lambda = 20$). \emph{Top}: class histograms; we show the number of training instances reaching the node and, for leaves, their label. \emph{Bottom}: weight vector at each decision node and average of training instances at each leaf; we show the node index, bias (always zero) and, for leaves, their label. We plot the weight vector, of dimension 800, as a 29$\times$29 square (the last pixels are unused), with features in the original order in LeNet5 (which is determined during training and arbitrary, hence the random aspect of the images), and colored according to their sign and magnitude (positive, negative and zero values are blue, red and white, respectively). You may need to zoom in the plot.}
  \label{f:LeNet5-tree}
\end{figure}

\begin{figure}[p]
  \centering
  \begin{tabular}{@{}c@{}c@{}c@{}c@{}c@{}c@{}c@{}c@{}c@{}c@{}}
    \multicolumn{2}{c}{ground truth vs}& 
    \multicolumn{3}{c}{features selected}&
    \multicolumn{5}{c}{\dotfill \textsc{All class $k_1$ to class $k_2$} \dotfill}\\
    \multicolumn{2}{c}{deep net vs tree}& 
    \multicolumn{3}{c}{by the tree}&
    {$4\rightarrow9$}&{$9\rightarrow4$}&{$3\rightarrow8$}&{$8\rightarrow3$}&{$1\rightarrow7$}\\ 
    \psfrag{Ground truth}[b][t][.7]{ground truth}
    \psfrag{Original net}[t][b][.7]{original net}
    \includegraphics*[width=.1\linewidth]{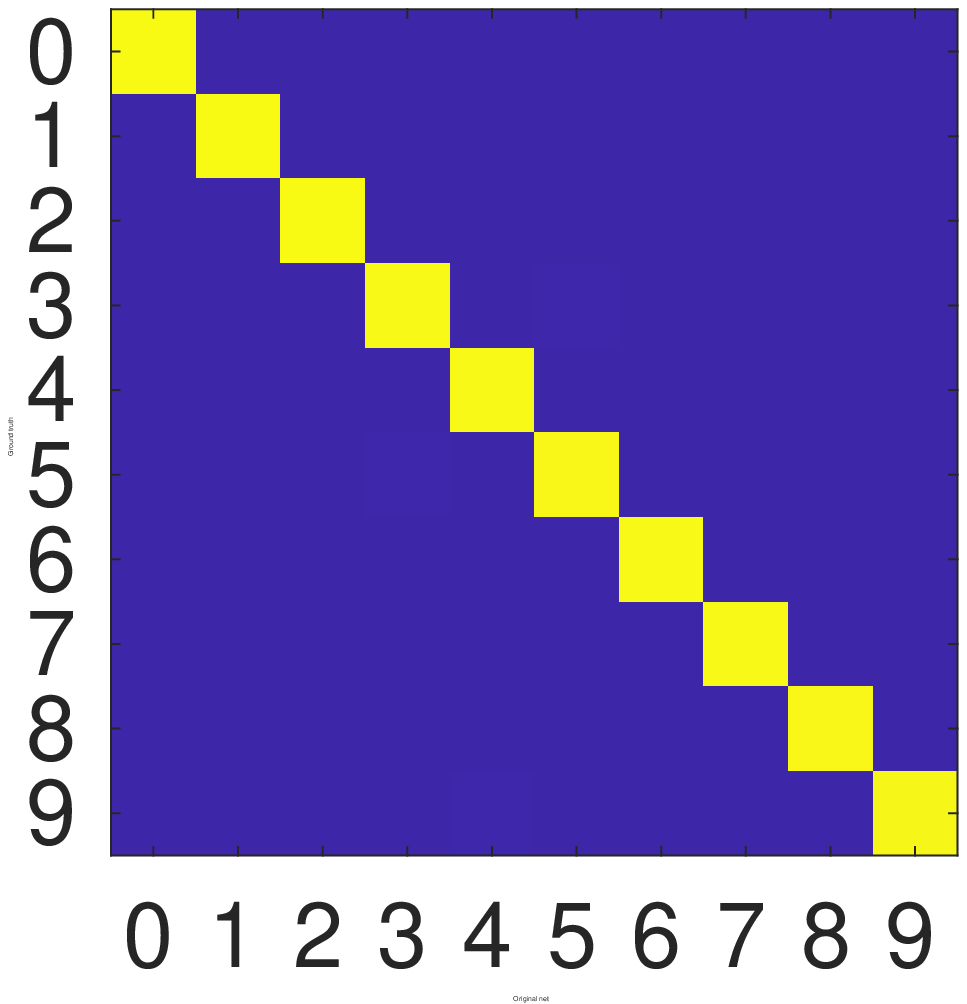}&
    \psfrag{Original net}[b][t][.7]{}
    \psfrag{Tree}[t][b][.7]{tree}
    \includegraphics*[width=.1\linewidth]{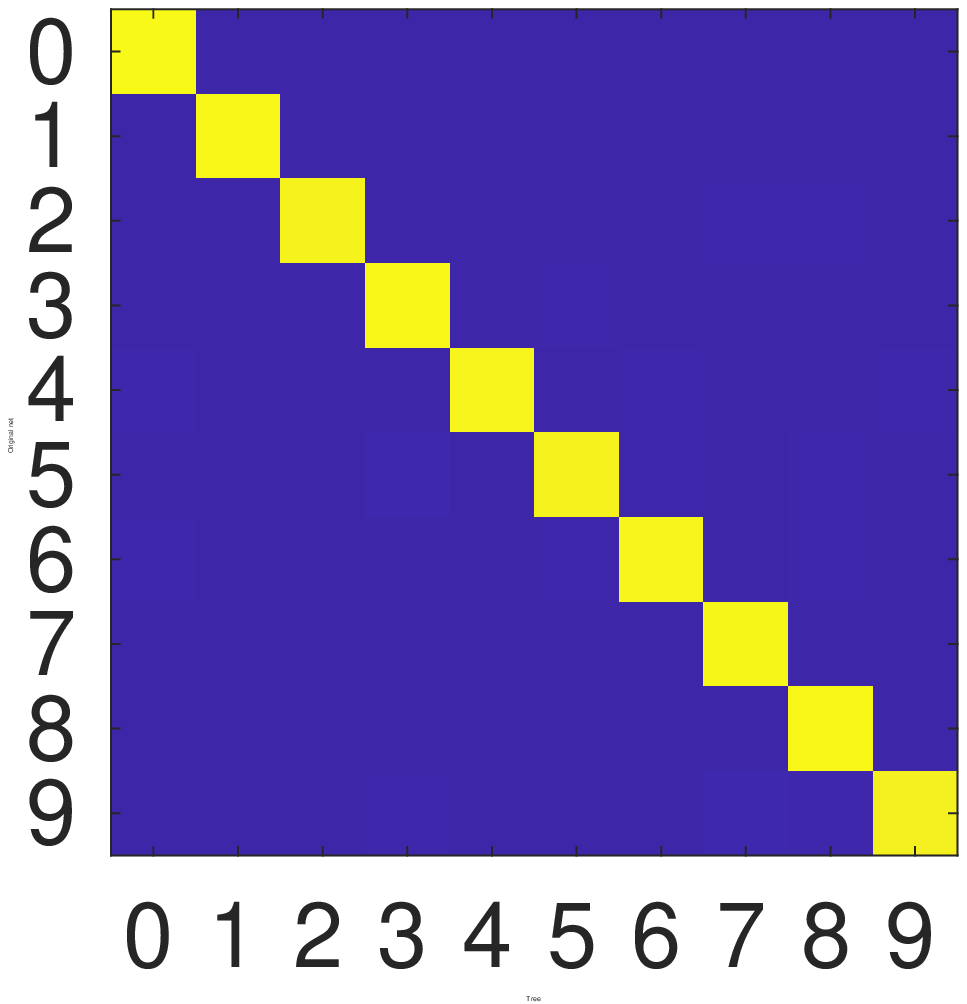}&
    \multicolumn{3}{c}{
      \psfrag{Original net}[b][t][.7]{original net}
      \psfrag{Modified net}[t][b][.7]{masked net}
      \includegraphics*[width=.1\linewidth]{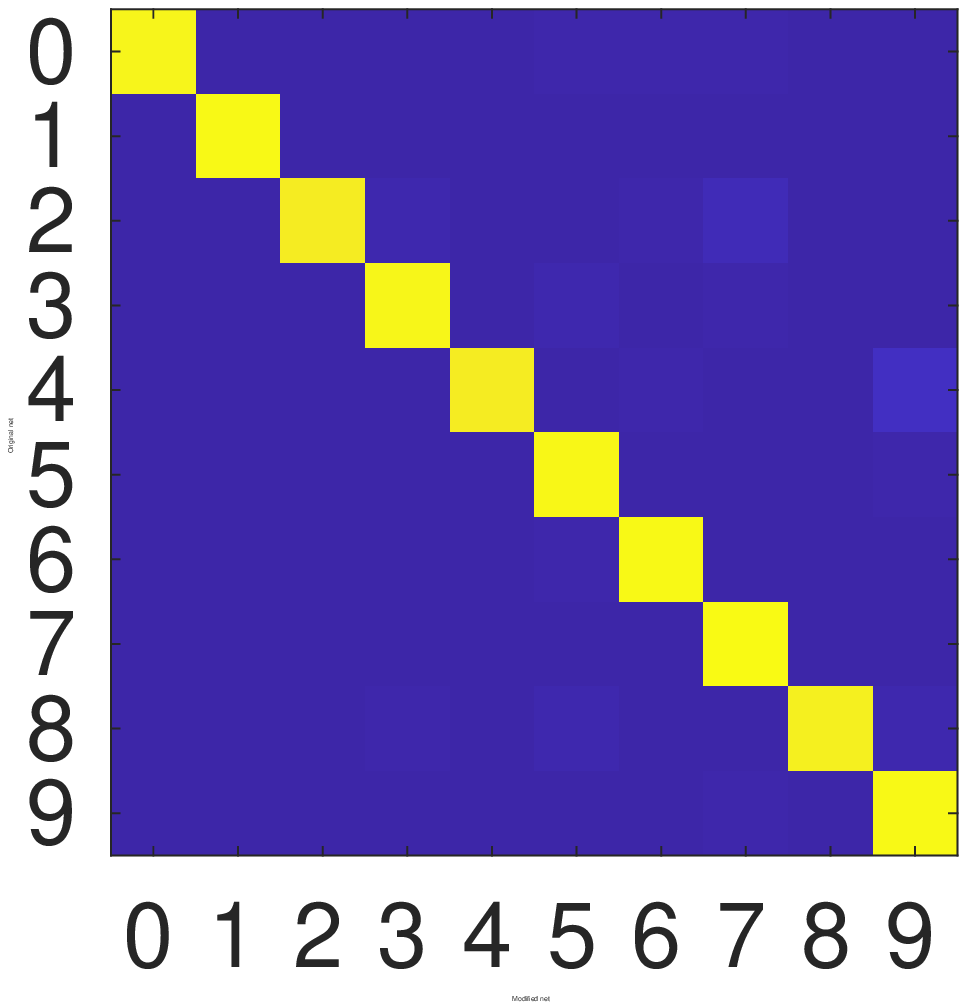}
      \raisebox{1.5ex}[0pt][0pt]{\psfrag{0}[][][0.5]{0}\psfrag{1}[][][0.5]{1}\makebox[0pt][l]{\includegraphics*[height=.09\textwidth]{VGG/colormap.eps}}}
    }&
    \psfrag{Original net}[b][t][.7]{original net}
    \psfrag{Modified net}[t][b][.7]{masked net}
    \includegraphics*[width=.1\linewidth]{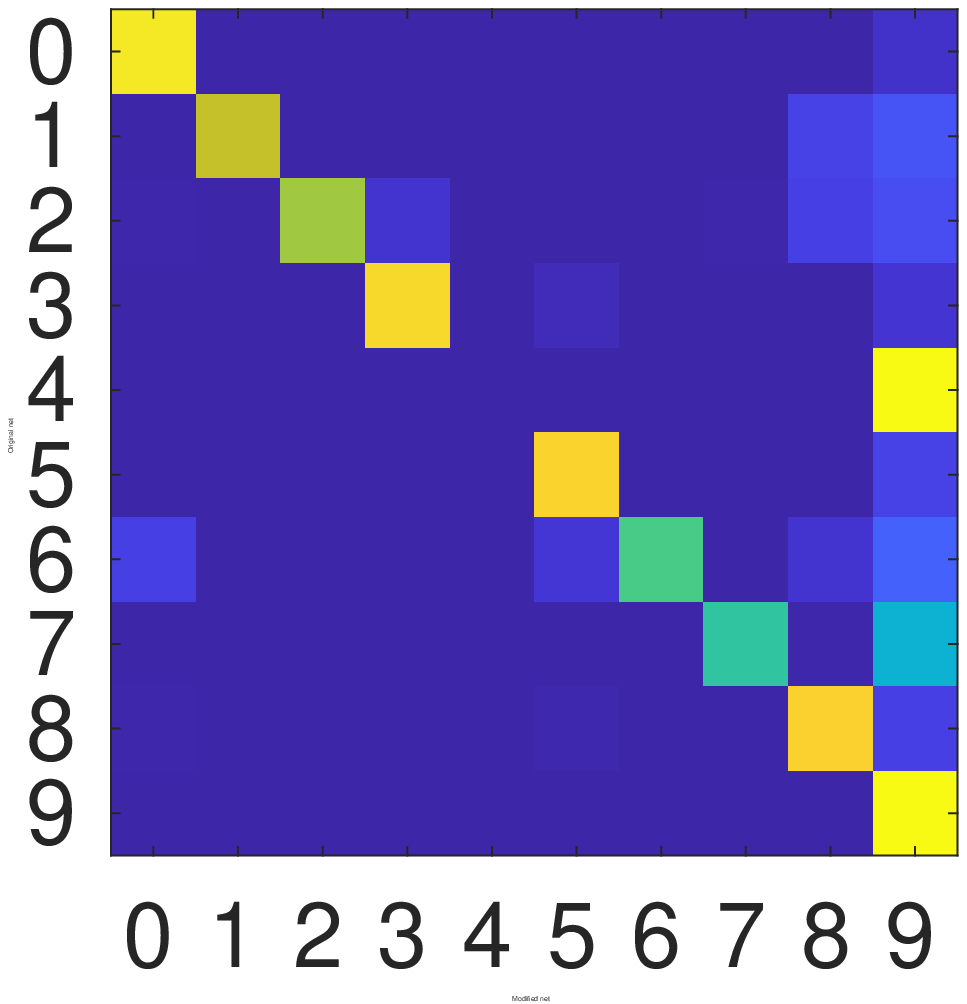}&
    \psfrag{Original net}{}
    \psfrag{Modified net}[t][b][.7]{masked net}
    \includegraphics*[width=.1\linewidth]{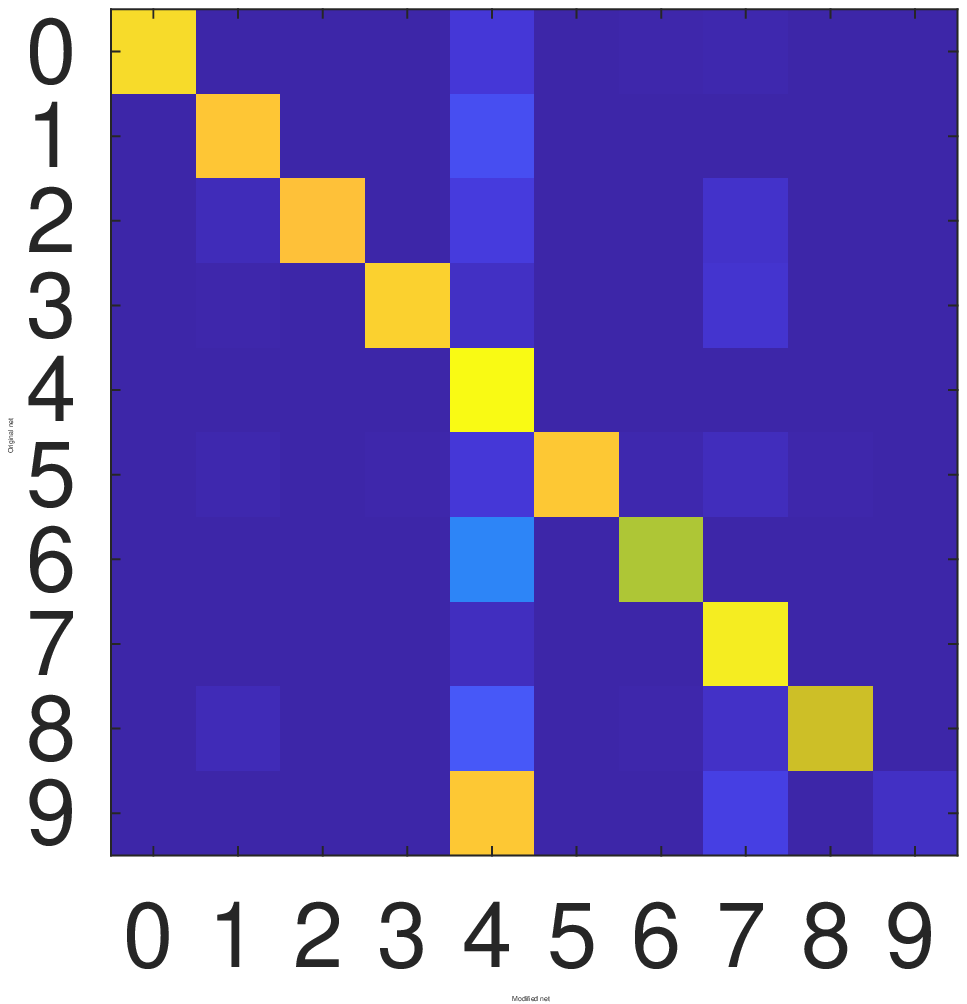}&
    \psfrag{Original net}{}
    \psfrag{Modified net}[t][b][.7]{masked net}
    \includegraphics*[width=.1\linewidth]{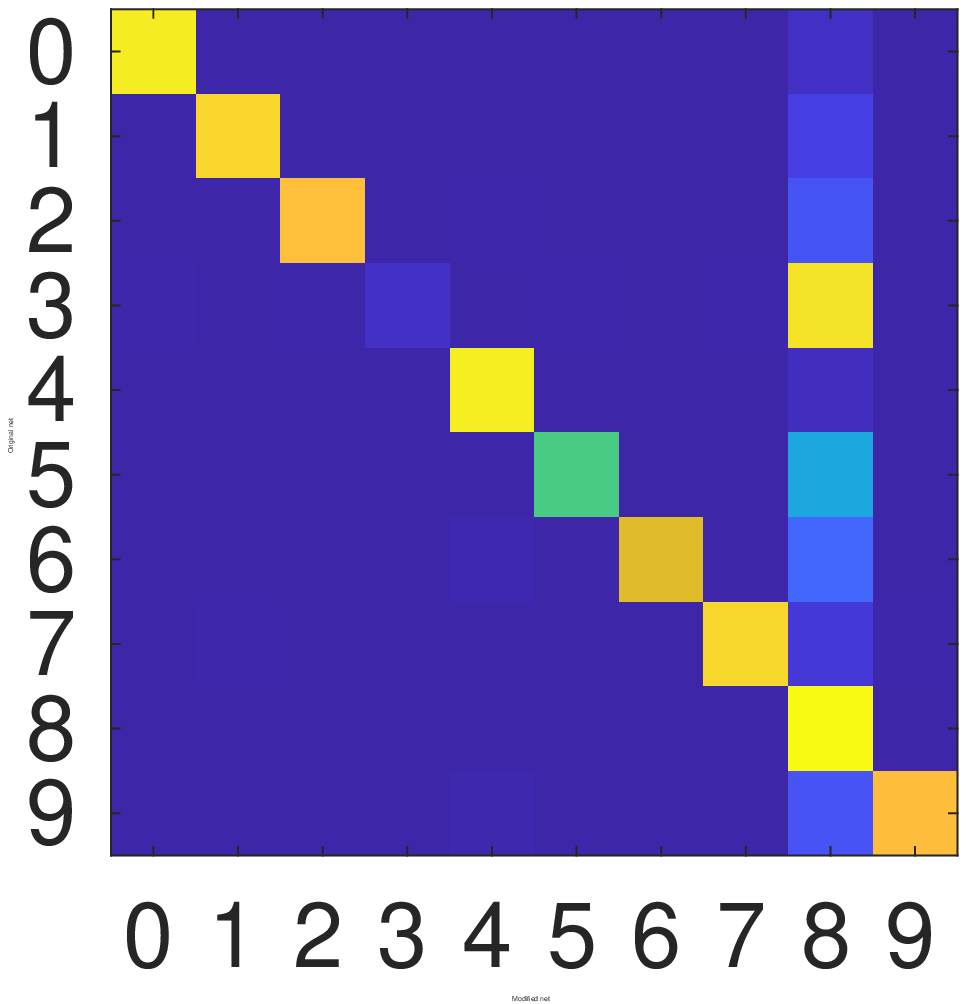}&
    \psfrag{Original net}{}
    \psfrag{Modified net}[t][b][.7]{masked net}
    \includegraphics*[width=.1\linewidth]{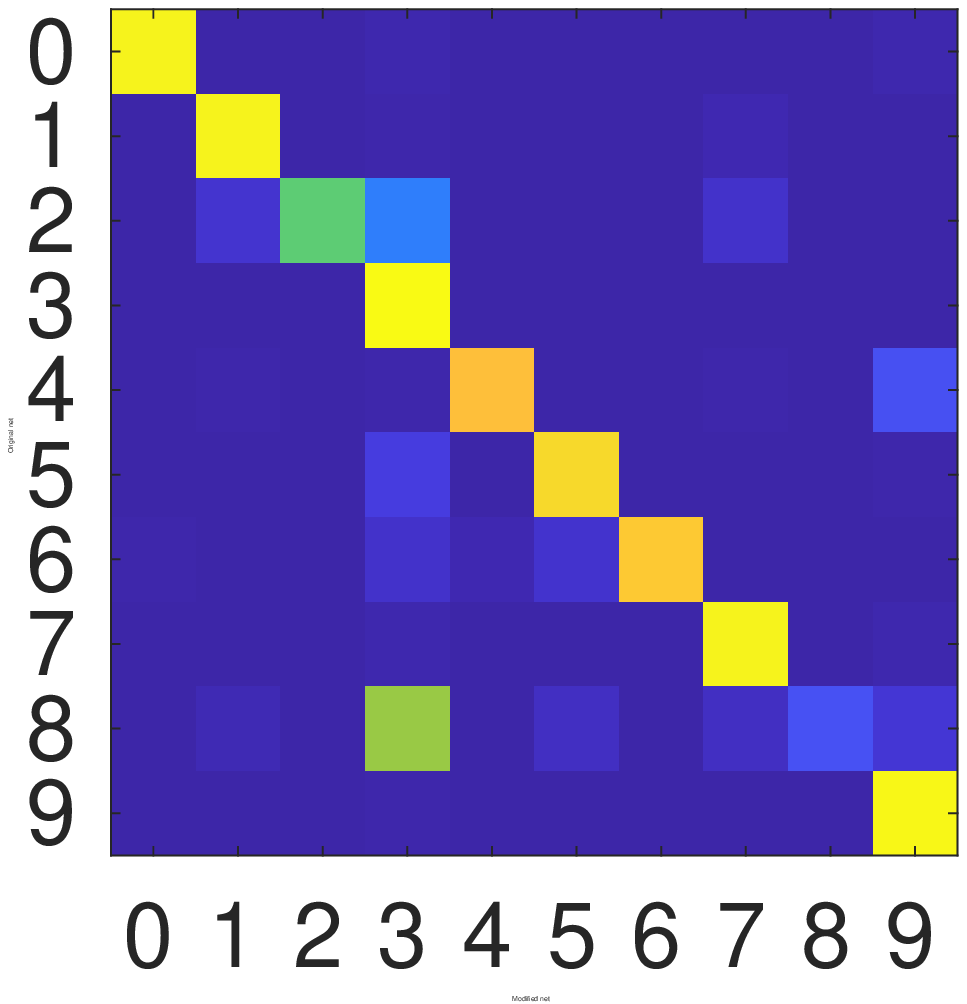}&
    \psfrag{Original net}{}
    \psfrag{Modified net}[t][b][.7]{masked net}
    \includegraphics*[width=.1\linewidth]{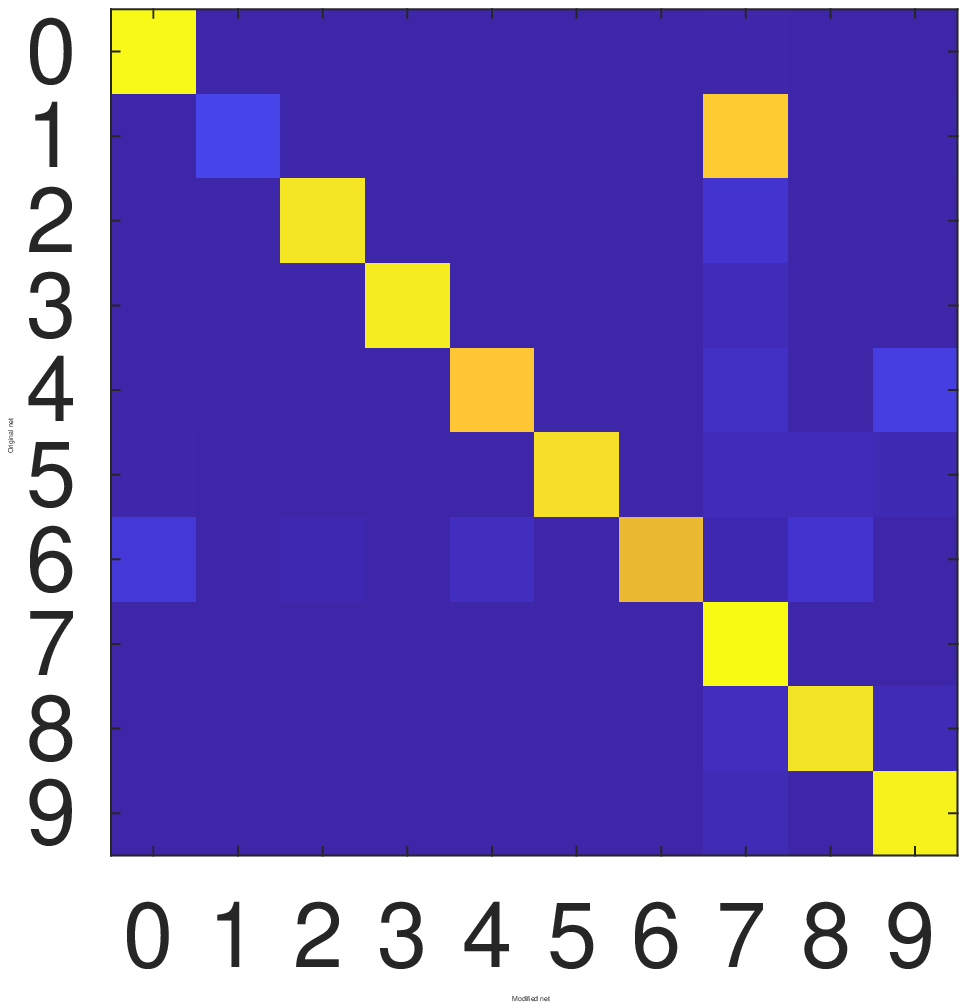} \\[2ex]
    \multicolumn{10}{c}{\dotfill \textsc{None to class $k$} \dotfill}\\
    $k = 0$ & $k = 1$ & $k = 2$ & $k = 3$ & $k = 4$ & $k = 5$ & $k = 6$ & $k = 7$ & $k = 8$ & $k = 9$ \\
    \psfrag{Original net}[b][t][.7]{original net}
    \psfrag{Modified net}[t][b][.7]{masked net}
    \includegraphics*[width=.1\linewidth]{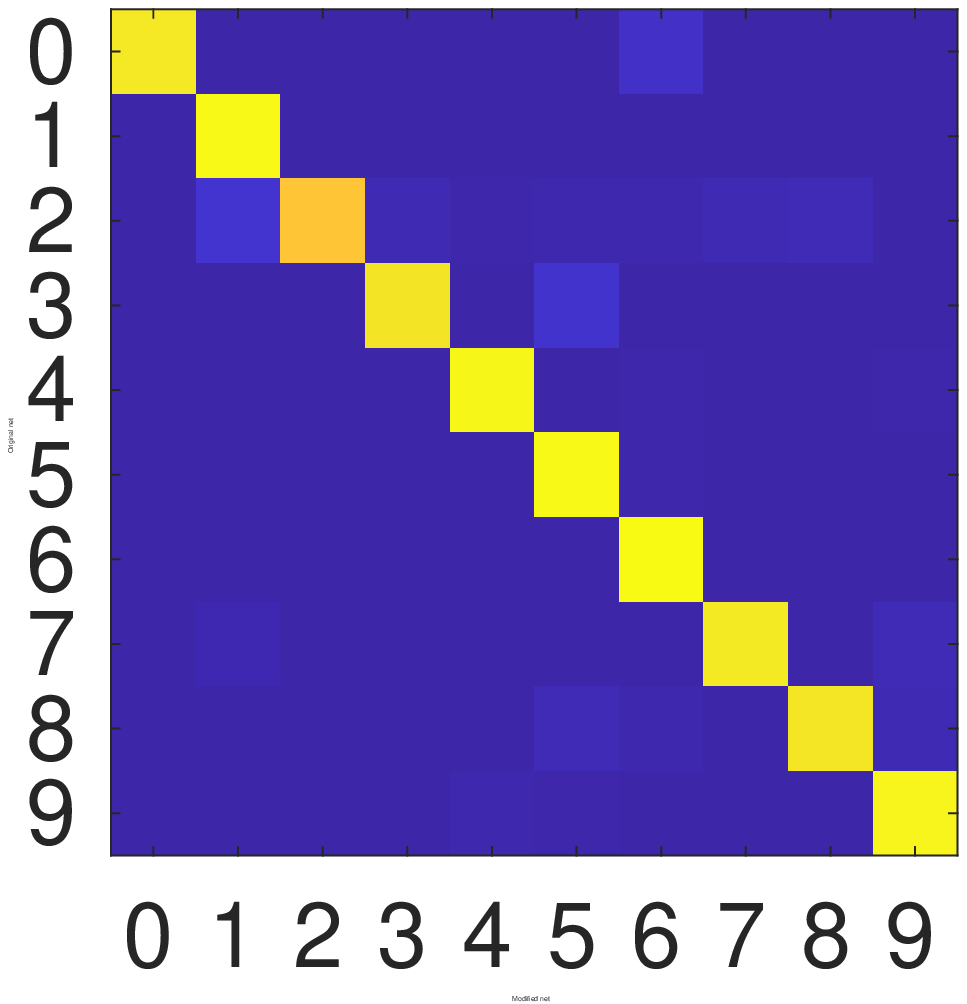}&
    \psfrag{Original net}{}
    \psfrag{Modified net}[t][b][.7]{masked net}
    \includegraphics*[width=.1\linewidth]{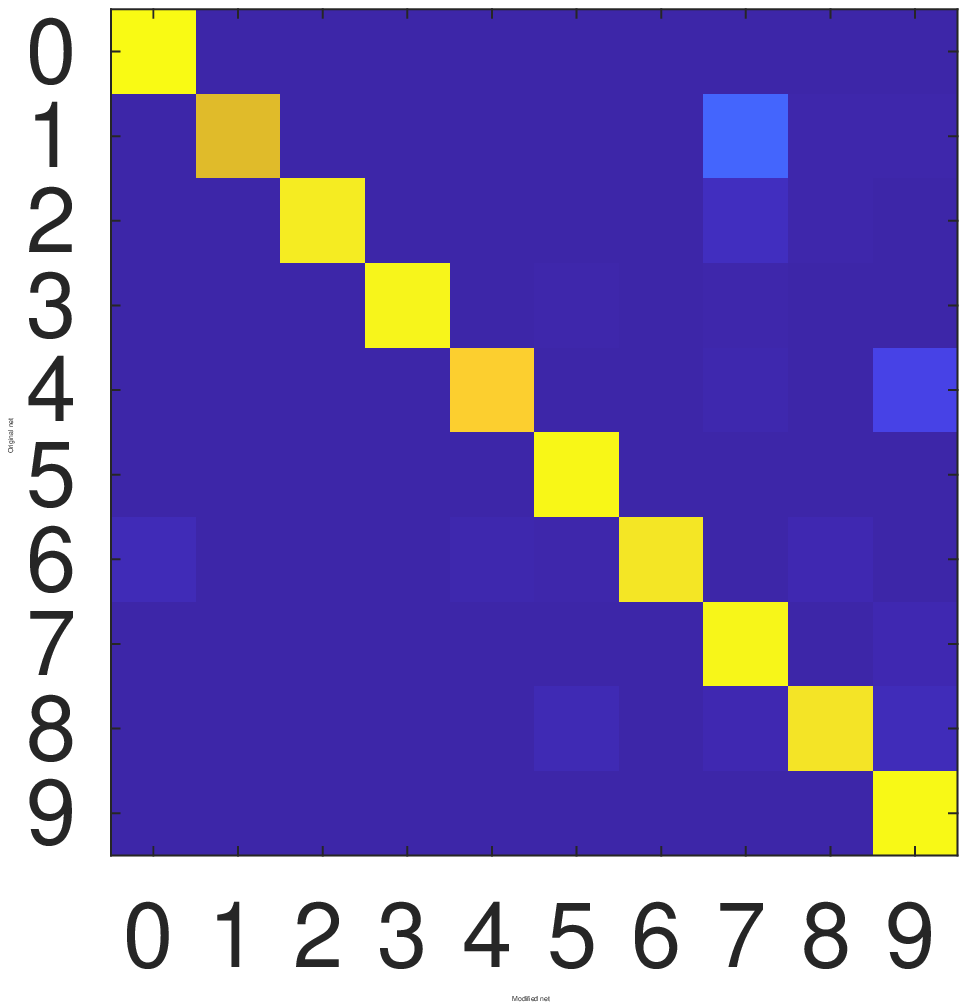}&
    \psfrag{Original net}{}
    \psfrag{Modified net}[t][b][.7]{masked net}
    \includegraphics*[width=.1\linewidth]{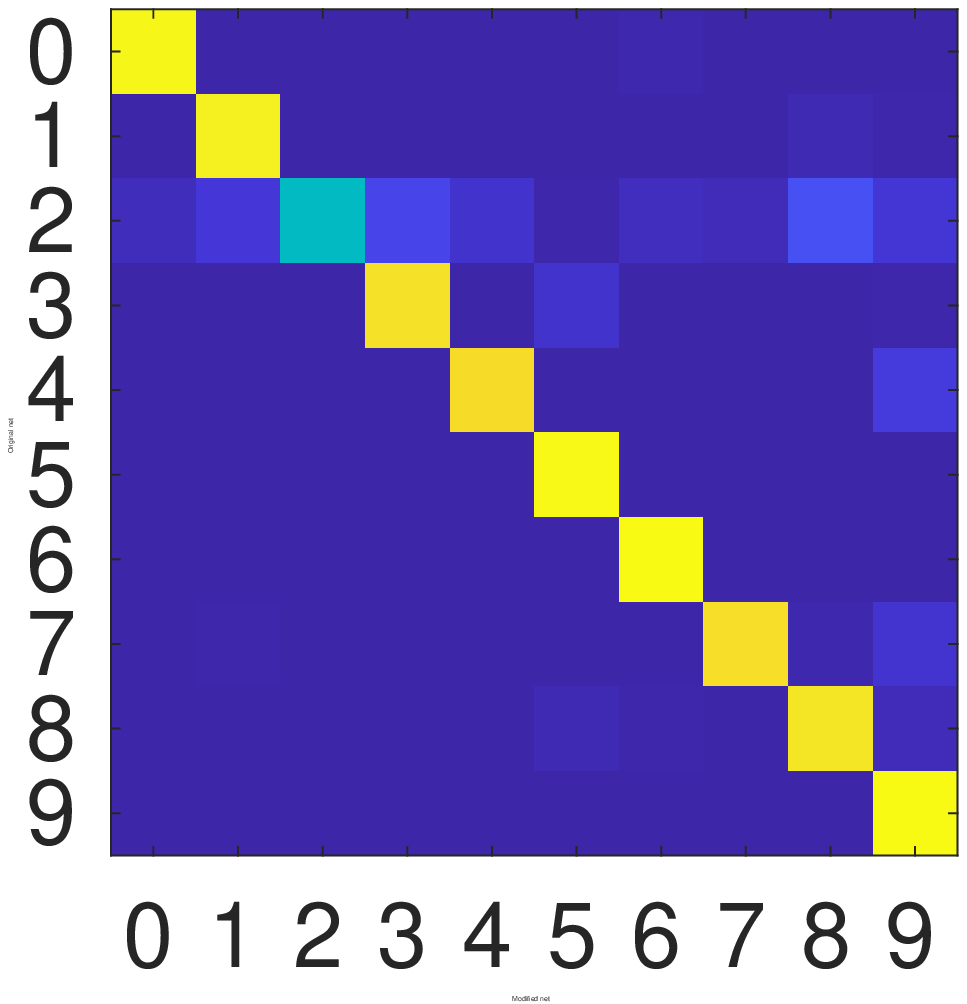}&
    \psfrag{Original net}{}
    \psfrag{Modified net}[t][b][.7]{masked net}
    \includegraphics*[width=.1\linewidth]{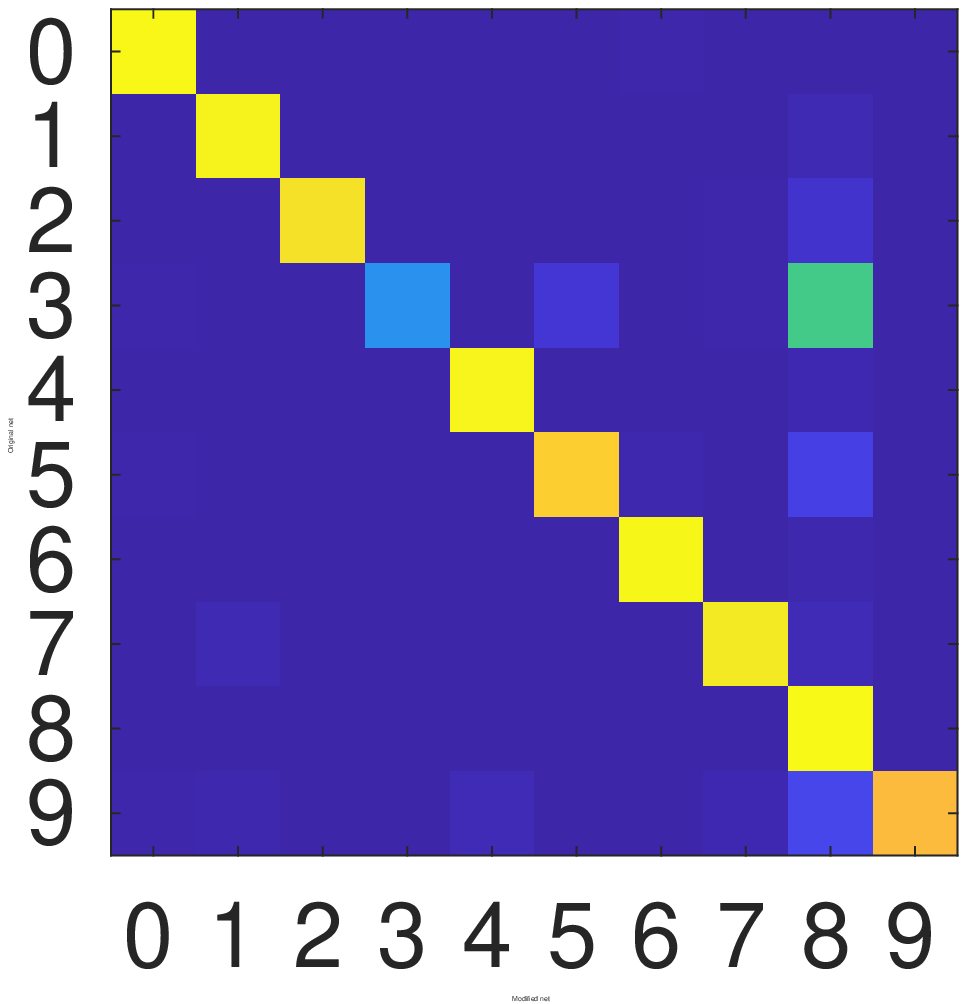}&
    \psfrag{Original net}{}
    \psfrag{Modified net}[t][b][.7]{masked net}
    \includegraphics*[width=.1\linewidth]{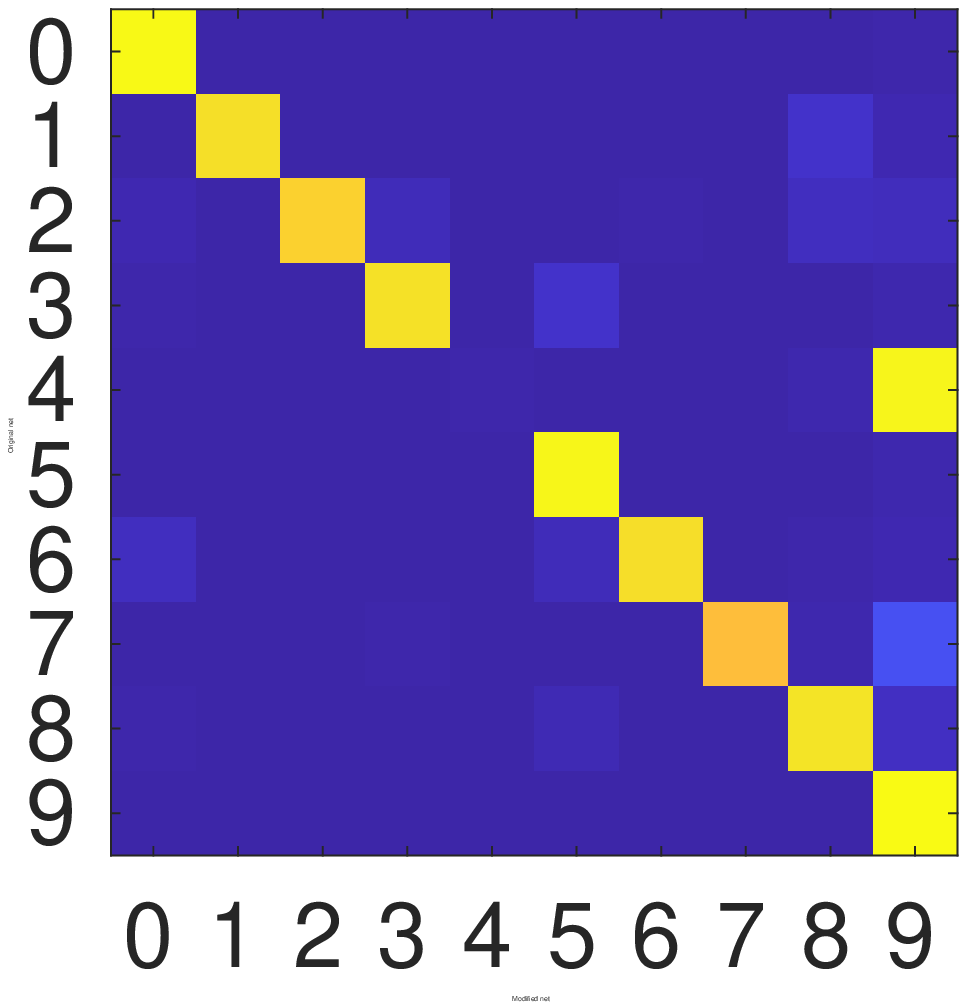}&
    \psfrag{Original net}{}
    \psfrag{Modified net}[t][b][.7]{masked net}
    \includegraphics*[width=.1\linewidth]{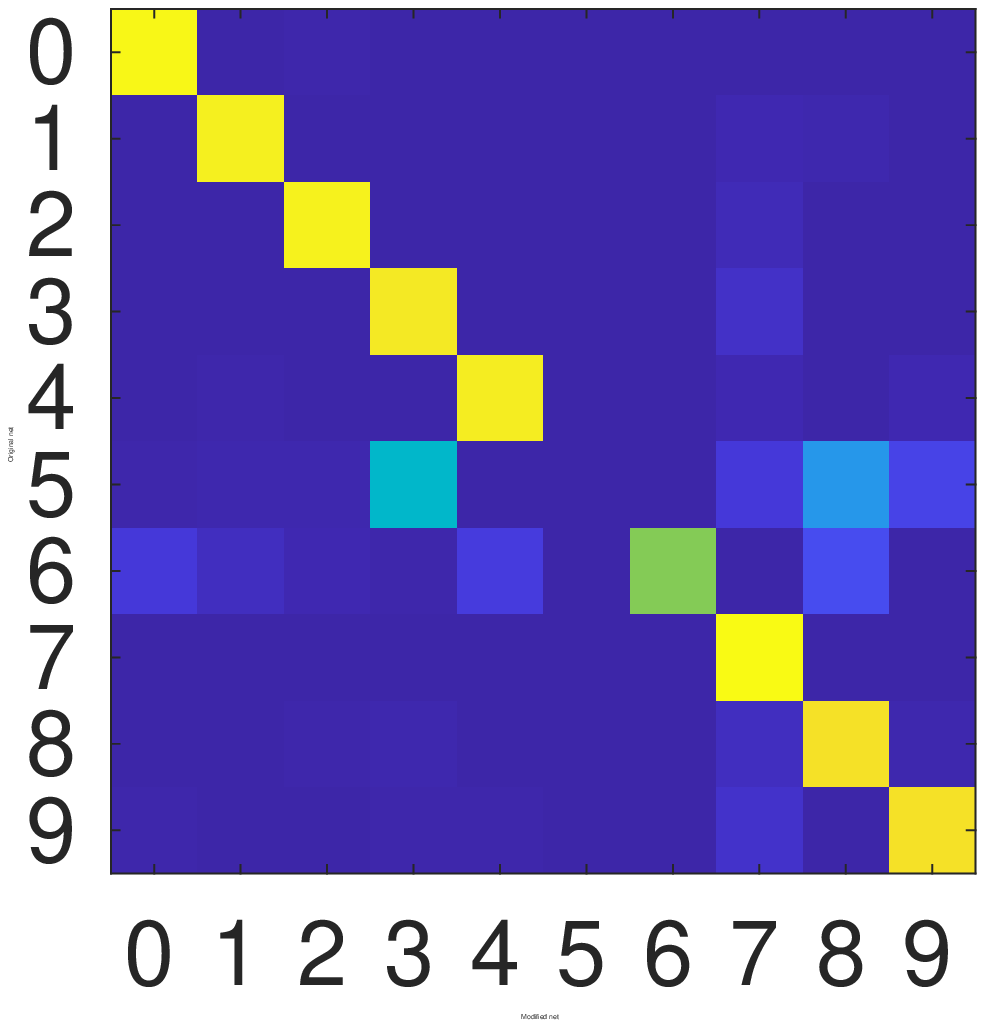}&
    \psfrag{Original net}{}
    \psfrag{Modified net}[t][b][.7]{masked net}
    \includegraphics*[width=.1\linewidth]{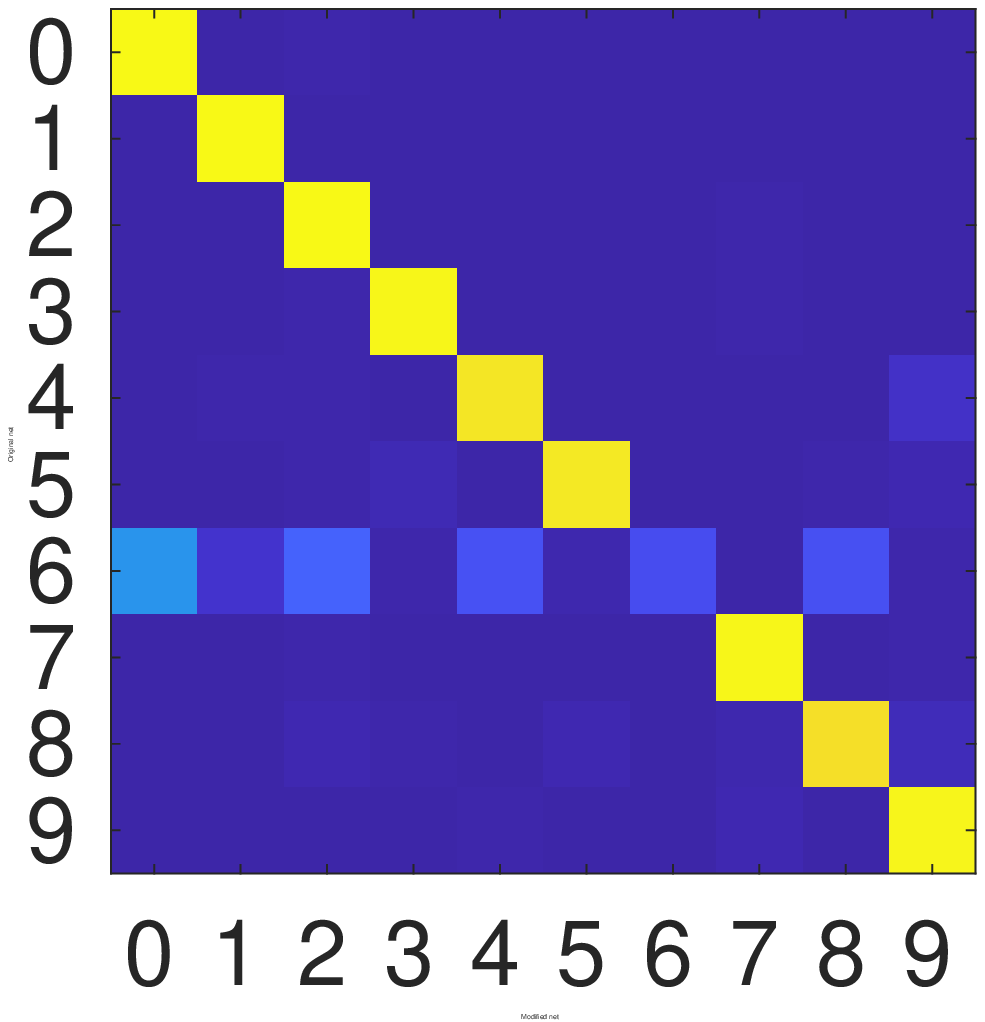}&
    \psfrag{Original net}{}
    \psfrag{Modified net}[t][b][.7]{masked net}
    \includegraphics*[width=.1\linewidth]{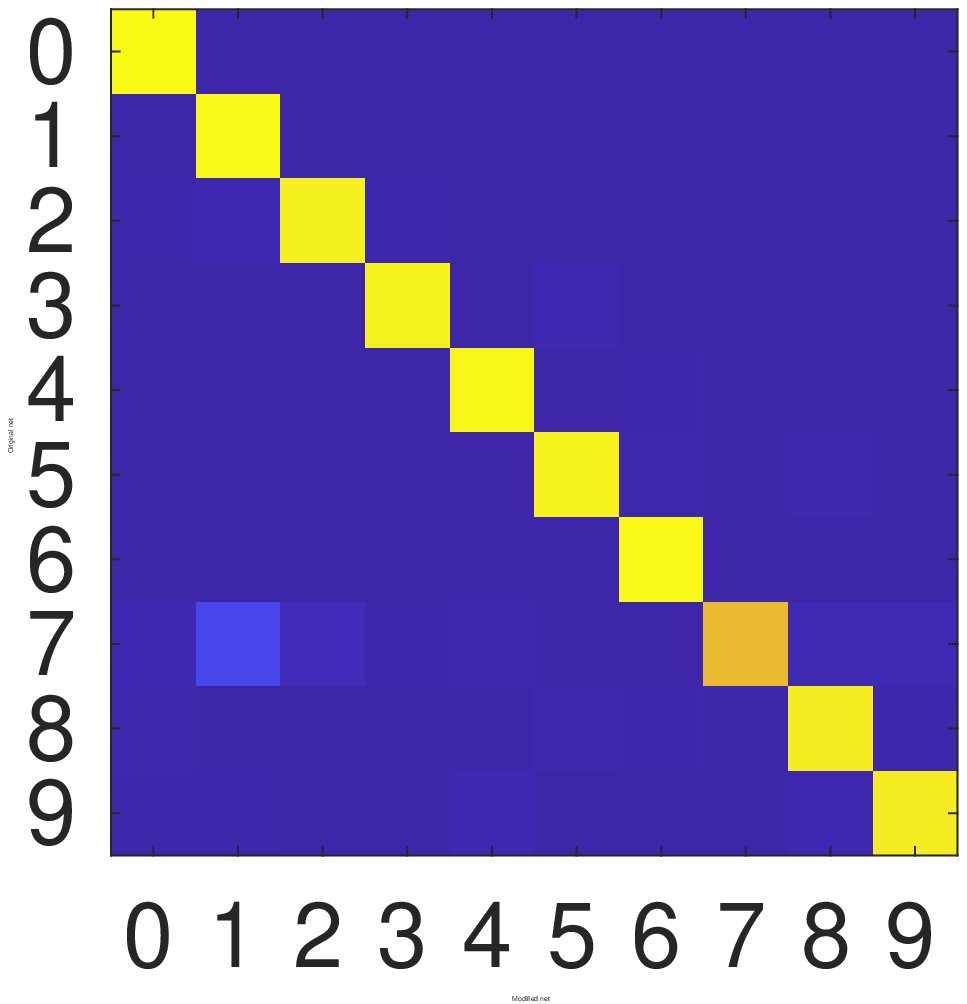}&
    \psfrag{Original net}{}
    \psfrag{Modified net}[t][b][.7]{masked net}
    \includegraphics*[width=.1\linewidth]{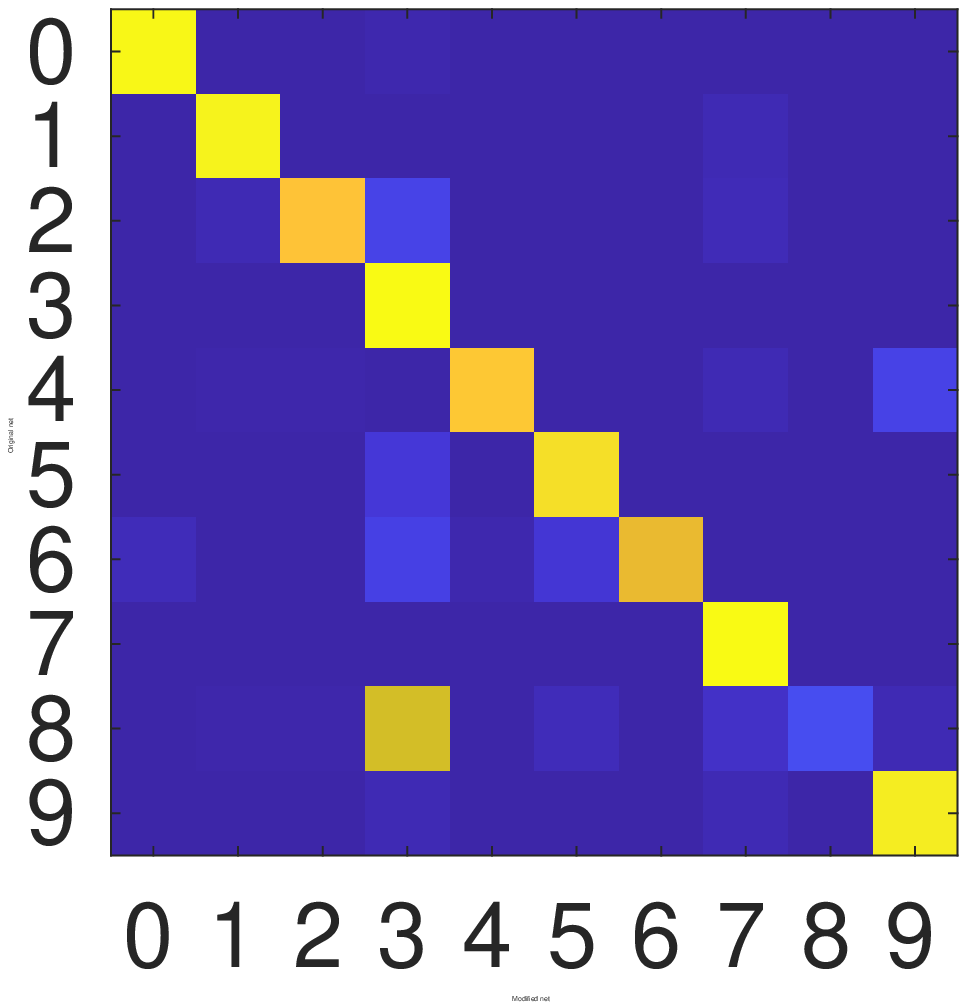}&
    \psfrag{Original net}{}
    \psfrag{Modified net}[t][b][.7]{masked net}
    \includegraphics*[width=.1\linewidth]{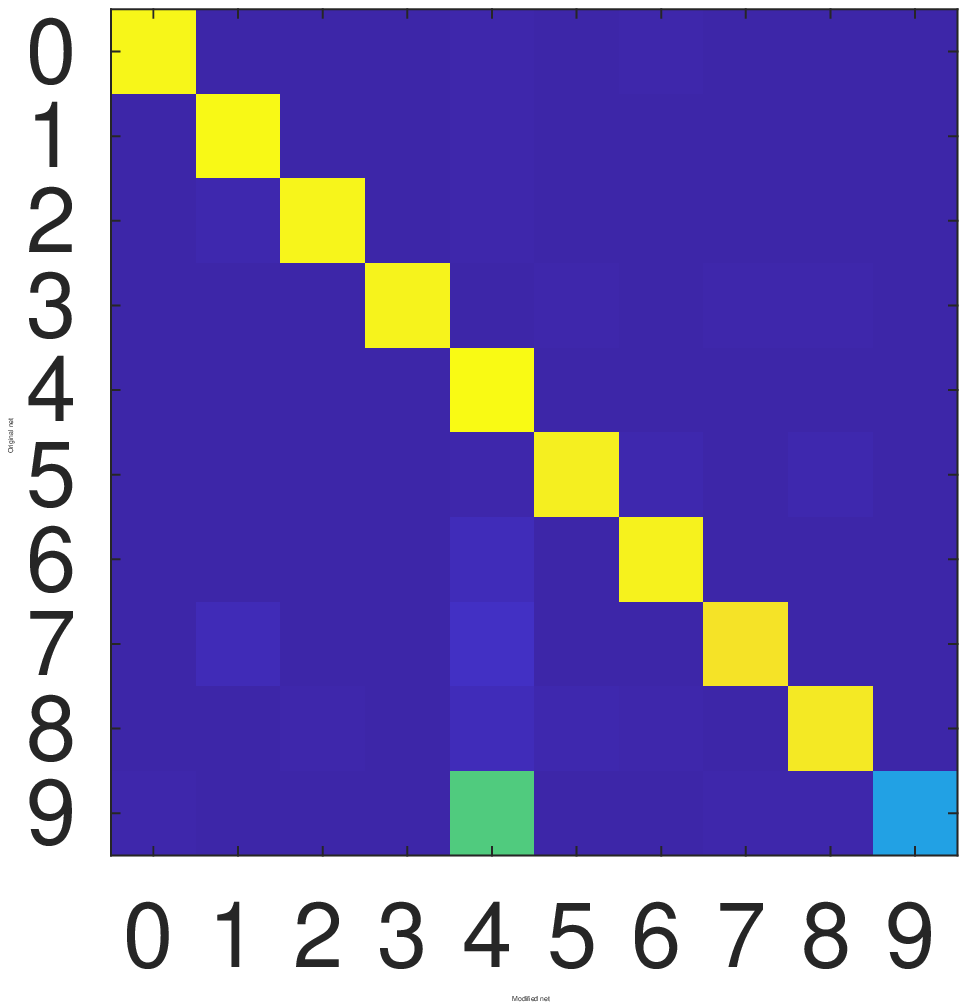} \\[2ex]
    \multicolumn{10}{c}{\dotfill \textsc{All to class $k$} \dotfill}\\
    $k = 0$ & $k = 1$ & $k = 2$ & $k = 3$ & $k = 4$ & $k = 5$ & $k = 6$ & $k = 7$ & $k = 8$ & $k = 9$ \\
    \psfrag{Original net}[b][t][.7]{original net}
    \psfrag{Modified net}[t][b][.7]{masked net}
    \includegraphics*[width=.1\linewidth]{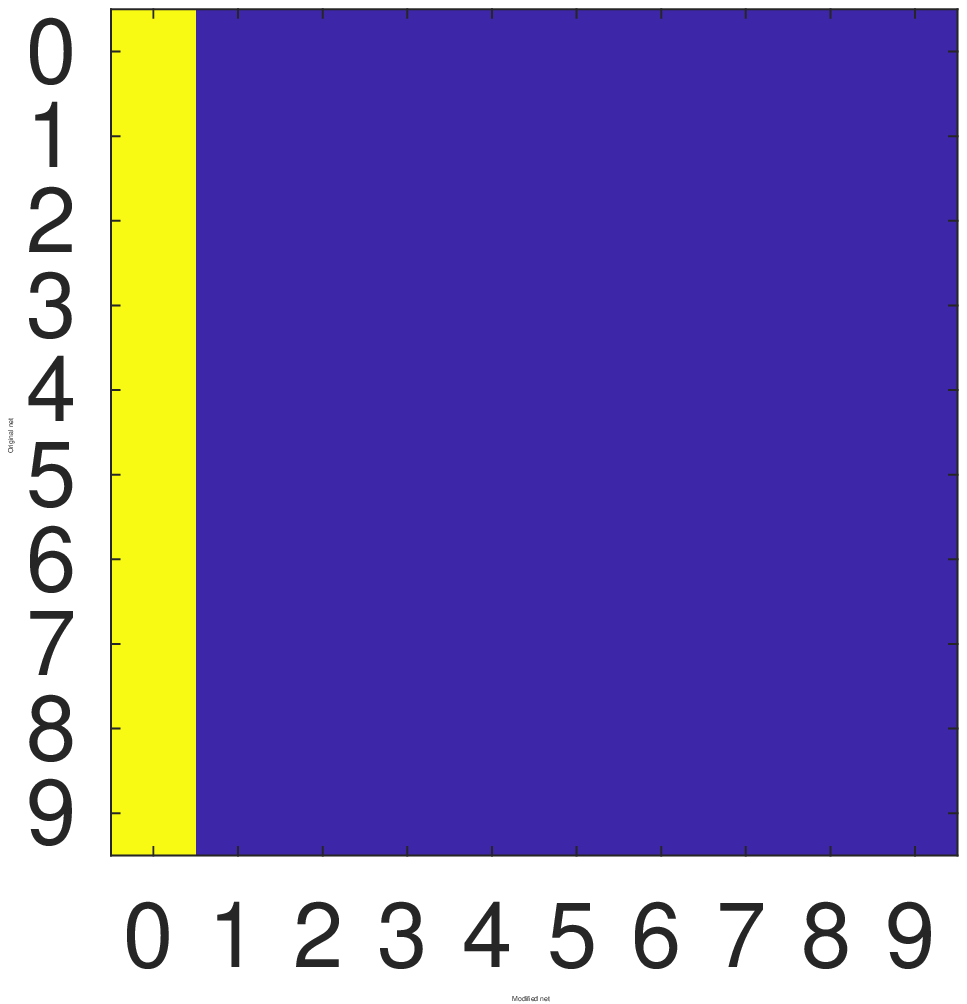}&
    \psfrag{Original net}{}
    \psfrag{Modified net}[t][b][.7]{masked net}
    \includegraphics*[width=.1\linewidth]{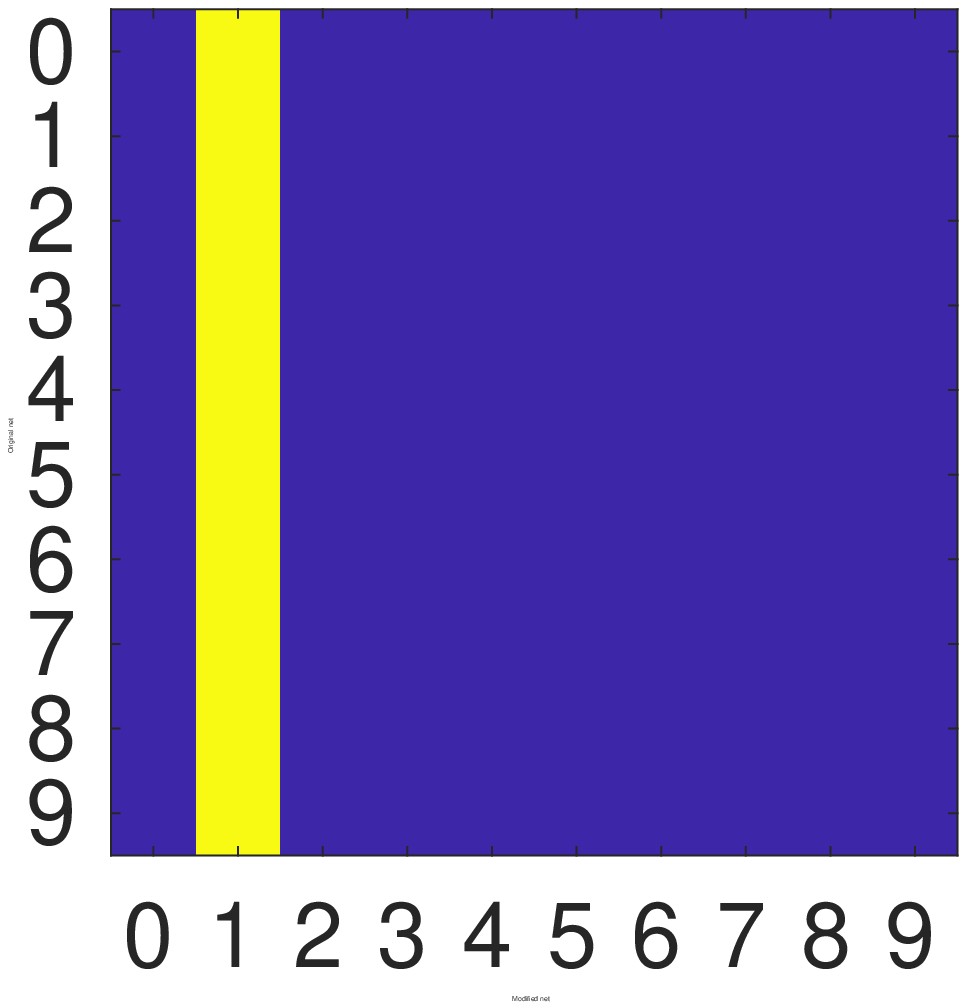}&
    \psfrag{Original net}{}
    \psfrag{Modified net}[t][b][.7]{masked net}
    \includegraphics*[width=.1\linewidth]{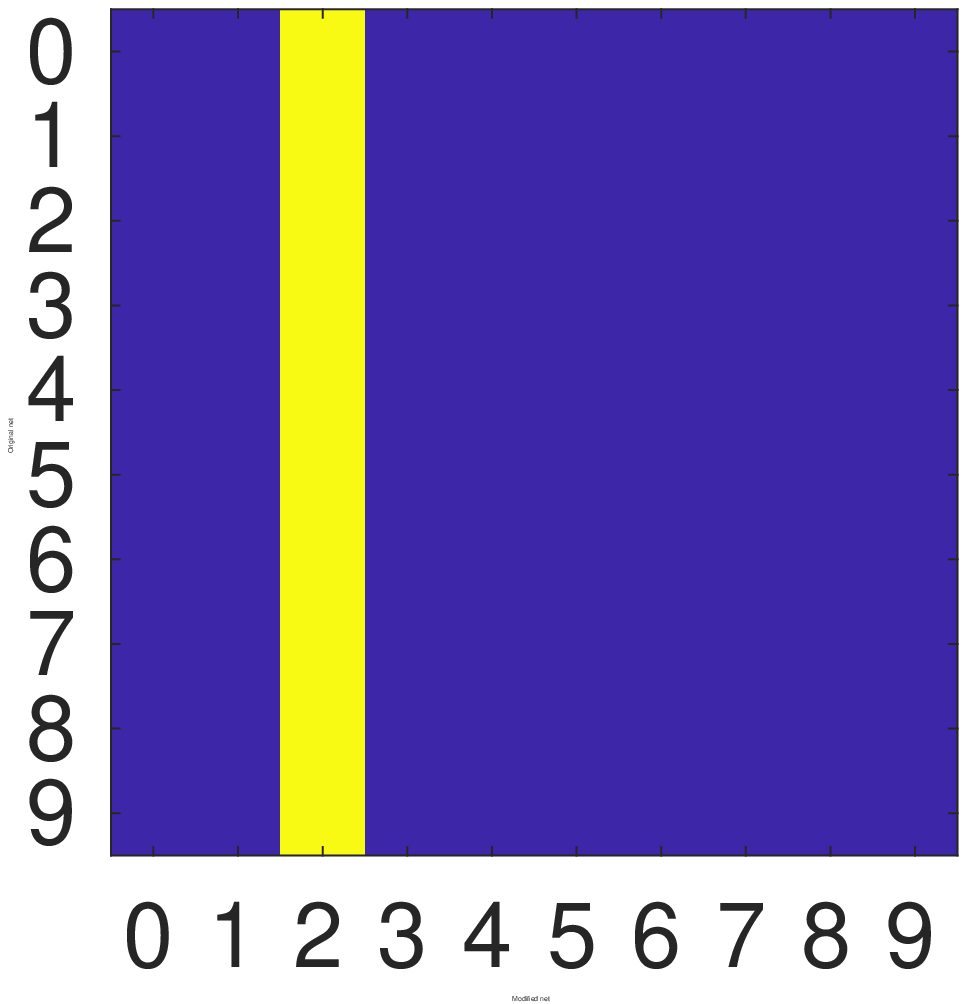}&
    \psfrag{Original net}{}
    \psfrag{Modified net}[t][b][.7]{masked net}
    \includegraphics*[width=.1\linewidth]{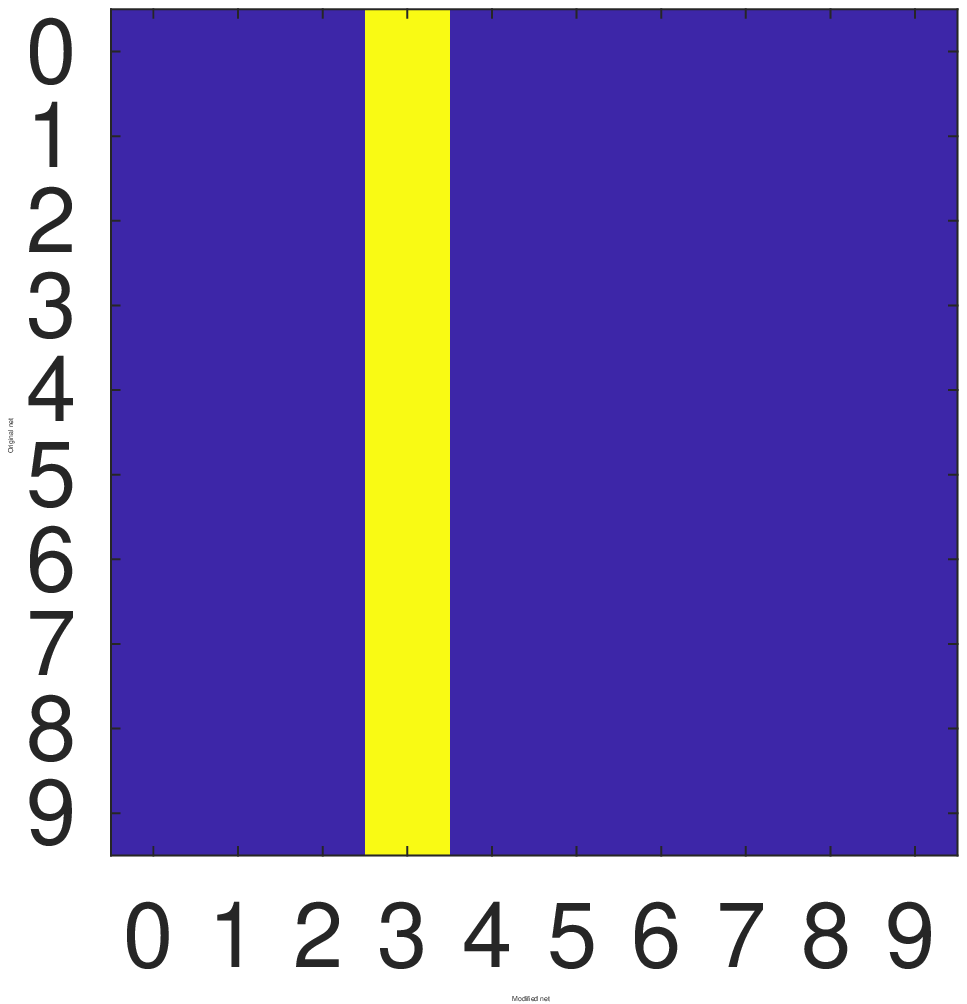}&
    \psfrag{Original net}{}
    \psfrag{Modified net}[t][b][.7]{masked net}
    \includegraphics*[width=.1\linewidth]{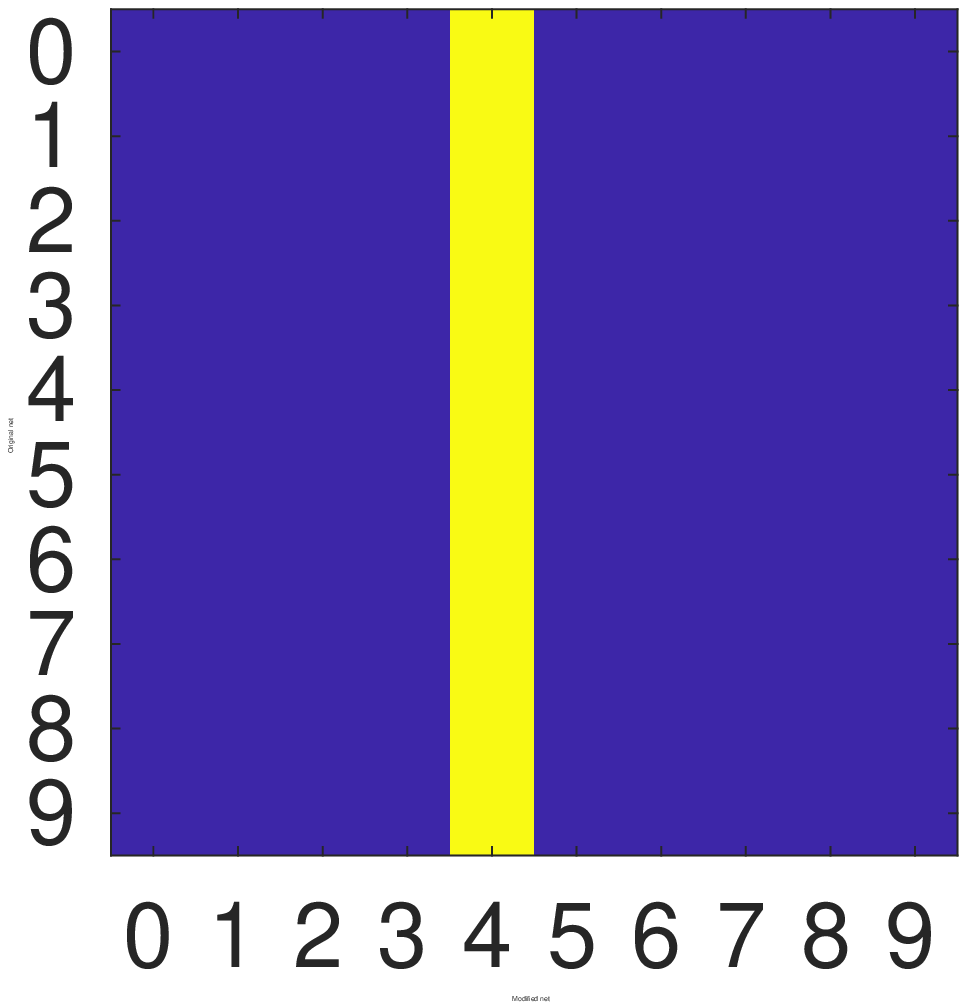}&
    \psfrag{Original net}{}
    \psfrag{Modified net}[t][b][.7]{masked net}
    \includegraphics*[width=.1\linewidth]{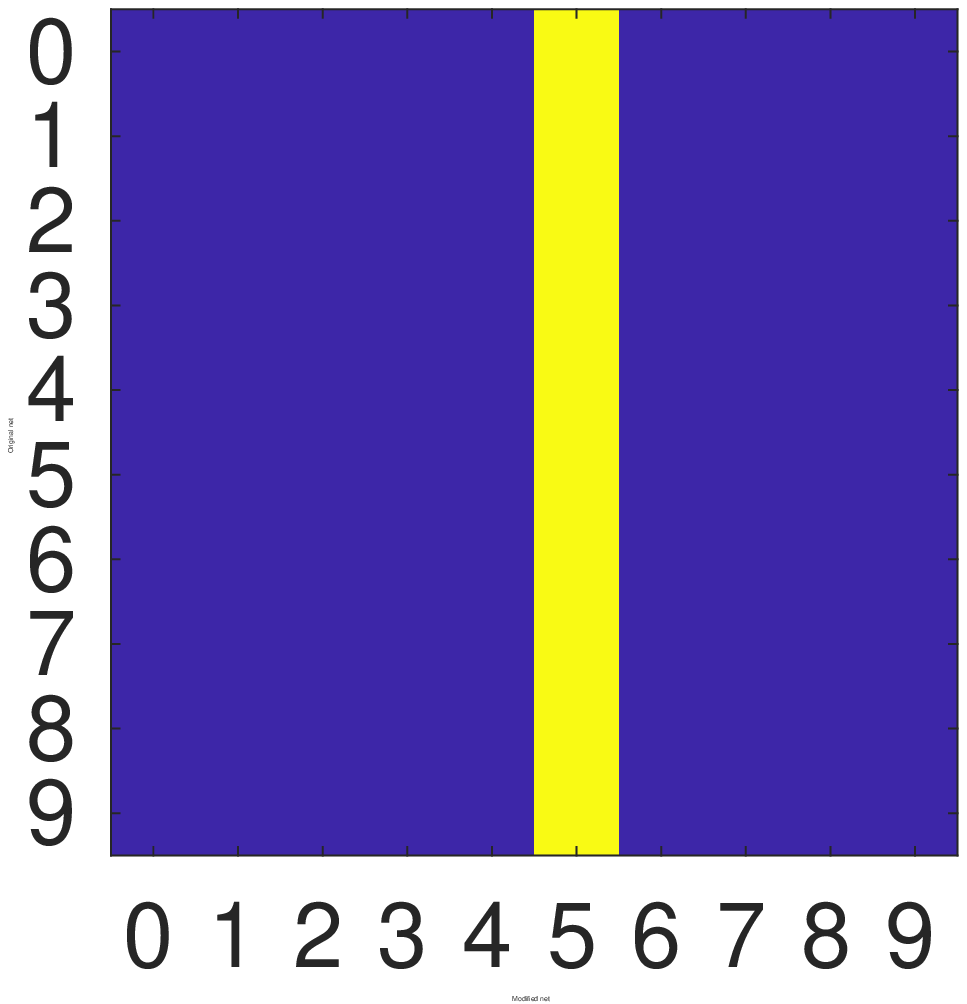}&
    \psfrag{Original net}{}
    \psfrag{Modified net}[t][b][.7]{masked net}
    \includegraphics*[width=.1\linewidth]{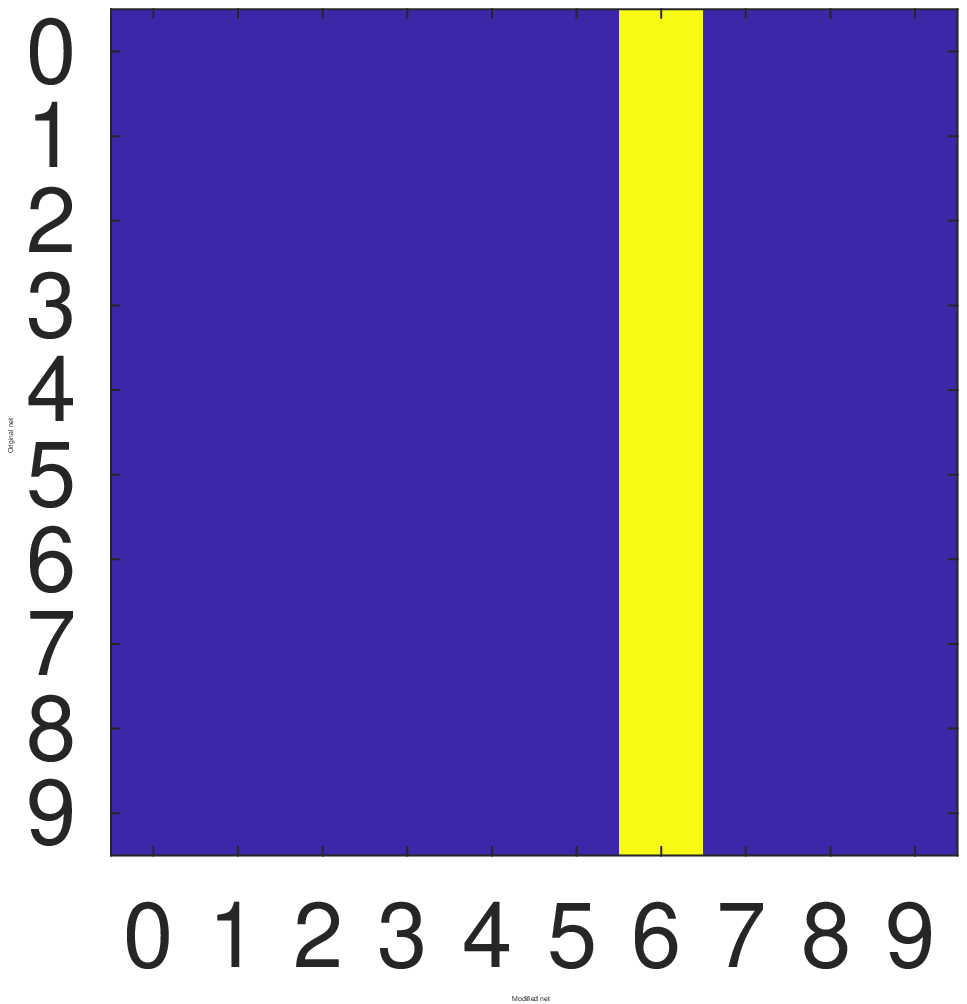}&
    \psfrag{Original net}{}
    \psfrag{Modified net}[t][b][.7]{masked net}
    \includegraphics*[width=.1\linewidth]{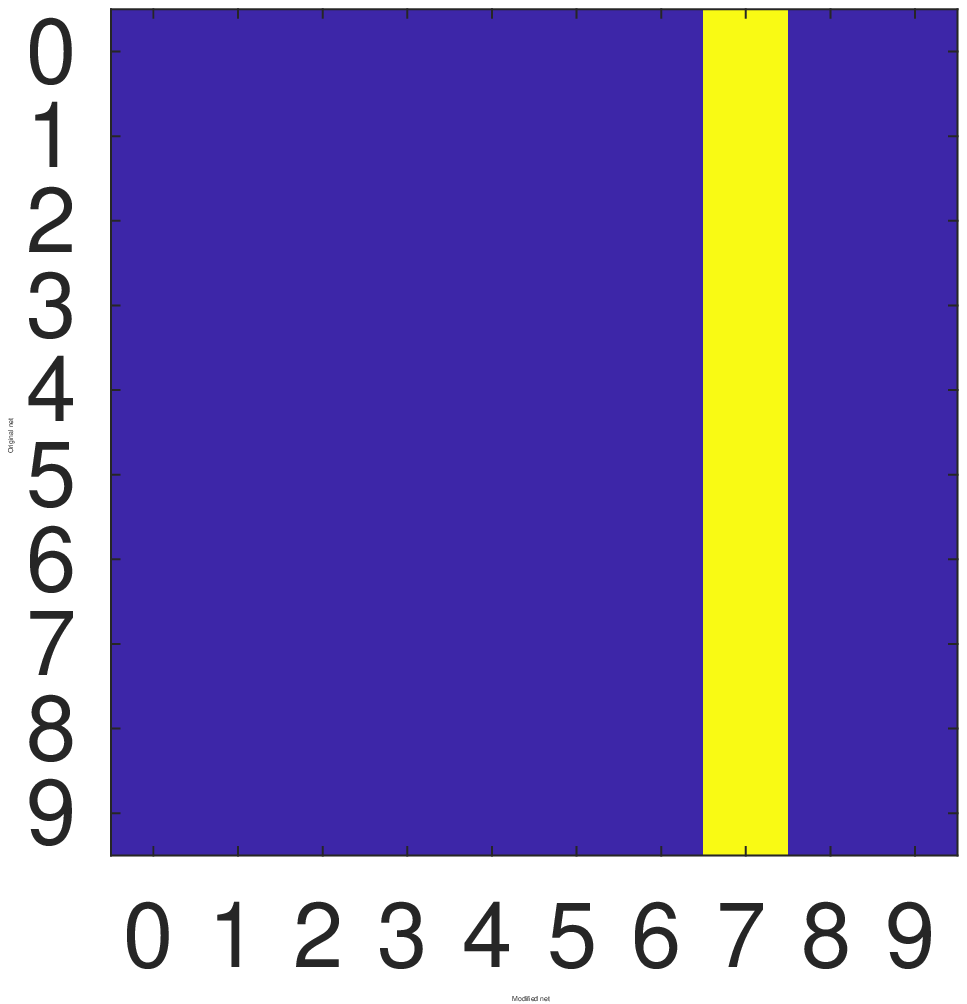}&
    \psfrag{Original net}{}
    \psfrag{Modified net}[t][b][.7]{masked net}
    \includegraphics*[width=.1\linewidth]{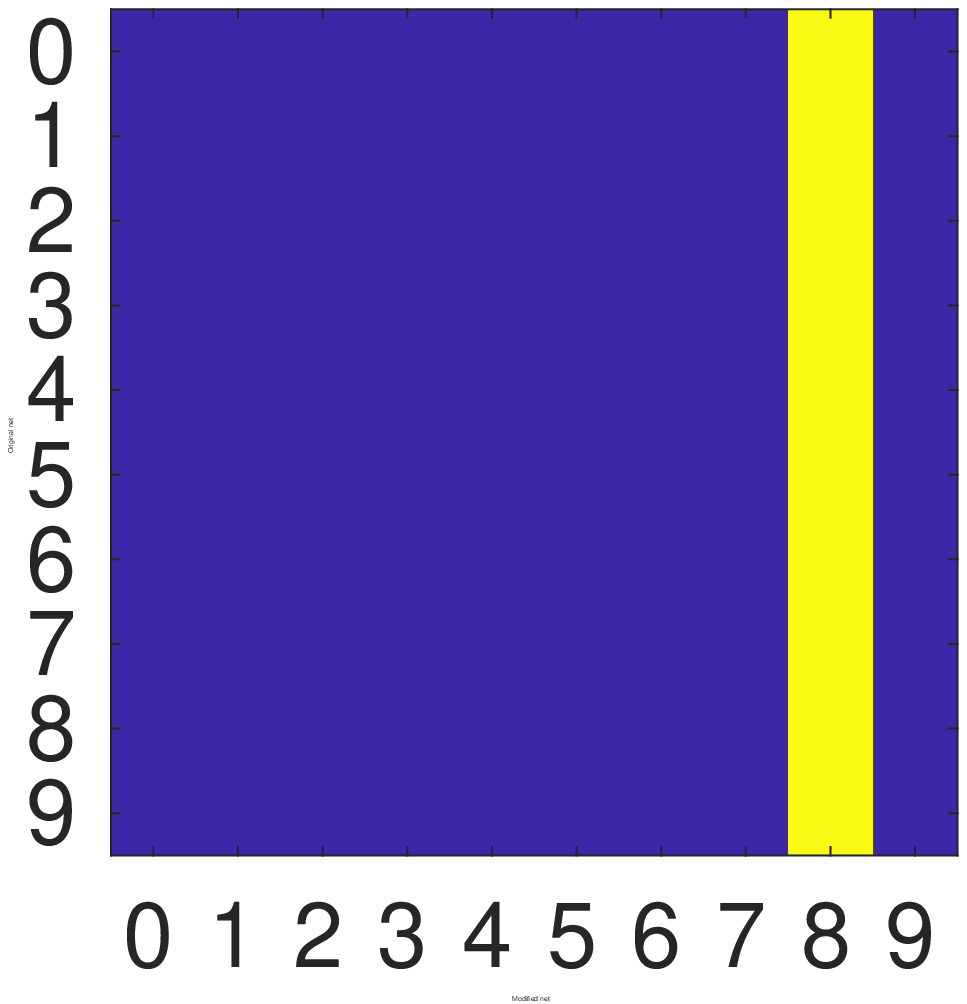}&
    \psfrag{Original net}{}
    \psfrag{Modified net}[t][b][.7]{masked net}
    \includegraphics*[width=.1\linewidth]{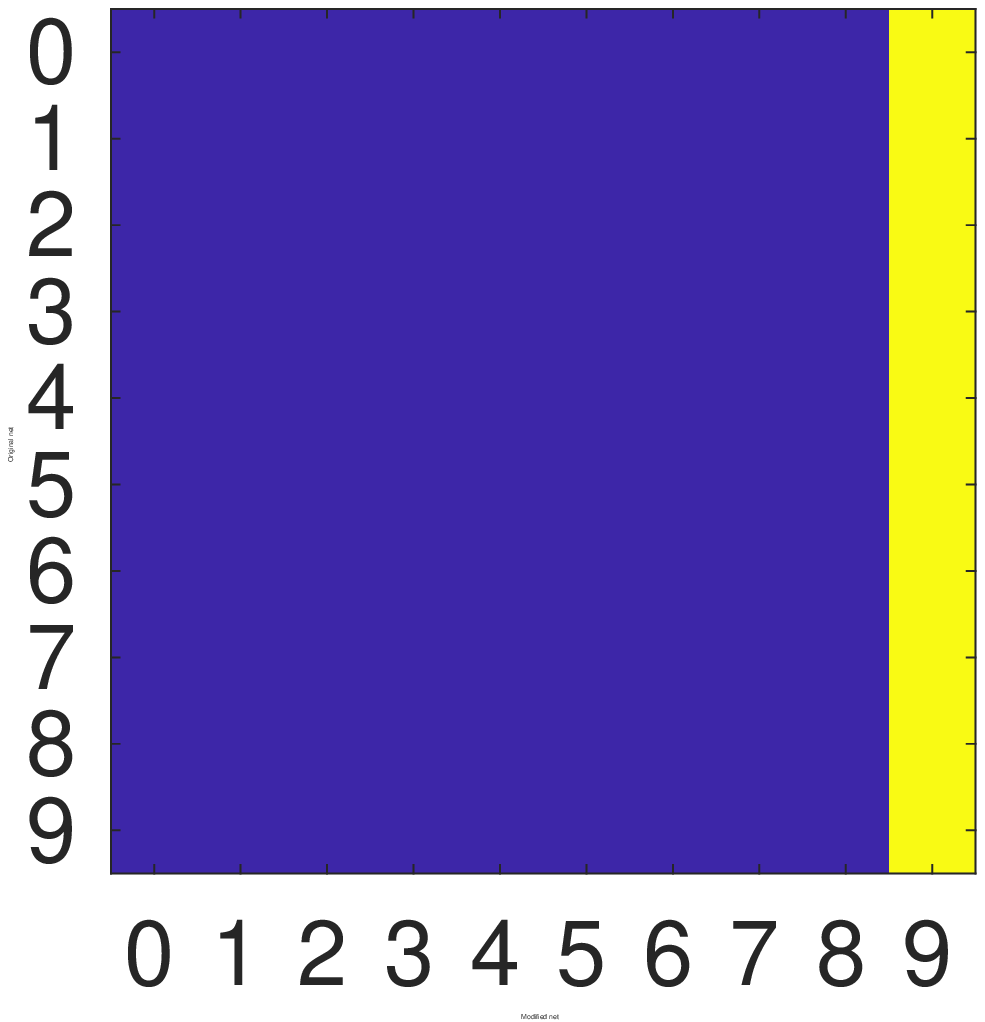}\\
  \end{tabular}
  \caption{Confusion matrices for LeNet5 (test set). \emph{Top left}: ground-truth vs deep net, and deep net vs tree. \emph{Top middle}: deep net vs deep net with only the features selected by the tree. \emph{Top right}: \textsc{All class $k_1$ to class $k_2$} (selected examples). \emph{Middle}: \textsc{None to class $k$}. \emph{Bottom}: \textsc{All to class $k$}.}
  \label{f:LeNet5-masks-test}
\end{figure}

\begin{figure}[p]
  \centering
  \begin{tabular}{@{}c@{}c@{}c@{}c@{}c@{}c@{}c@{}c@{}c@{}c@{}}
    \multicolumn{2}{c}{ground truth vs}& 
    \multicolumn{3}{c}{features selected}&
    \multicolumn{5}{c}{\dotfill \textsc{All class $k_1$ to class $k_2$} \dotfill}\\
    \multicolumn{2}{c}{deep net vs tree}& 
    \multicolumn{3}{c}{by the tree}&
    {$4\rightarrow9$}&{$9\rightarrow4$}&{$3\rightarrow8$}&{$8\rightarrow3$}&{$1\rightarrow7$}\\ 
    \psfrag{Ground truth}[b][t][.7]{ground truth}
    \psfrag{Original net}[t][b][.7]{original net}
    \includegraphics*[width=.1\linewidth]{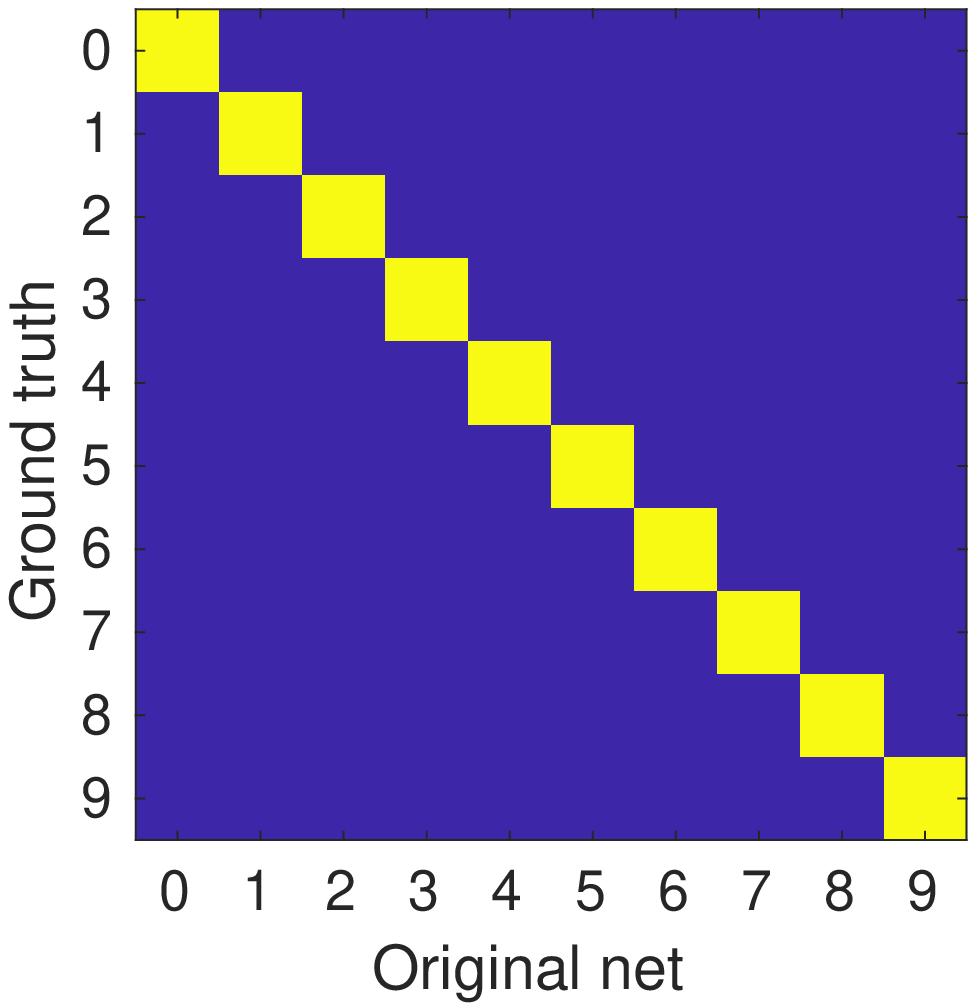}&
    \psfrag{Original net}[b][t][.7]{}
    \psfrag{Tree}[t][b][.7]{tree}
    \includegraphics*[width=.1\linewidth]{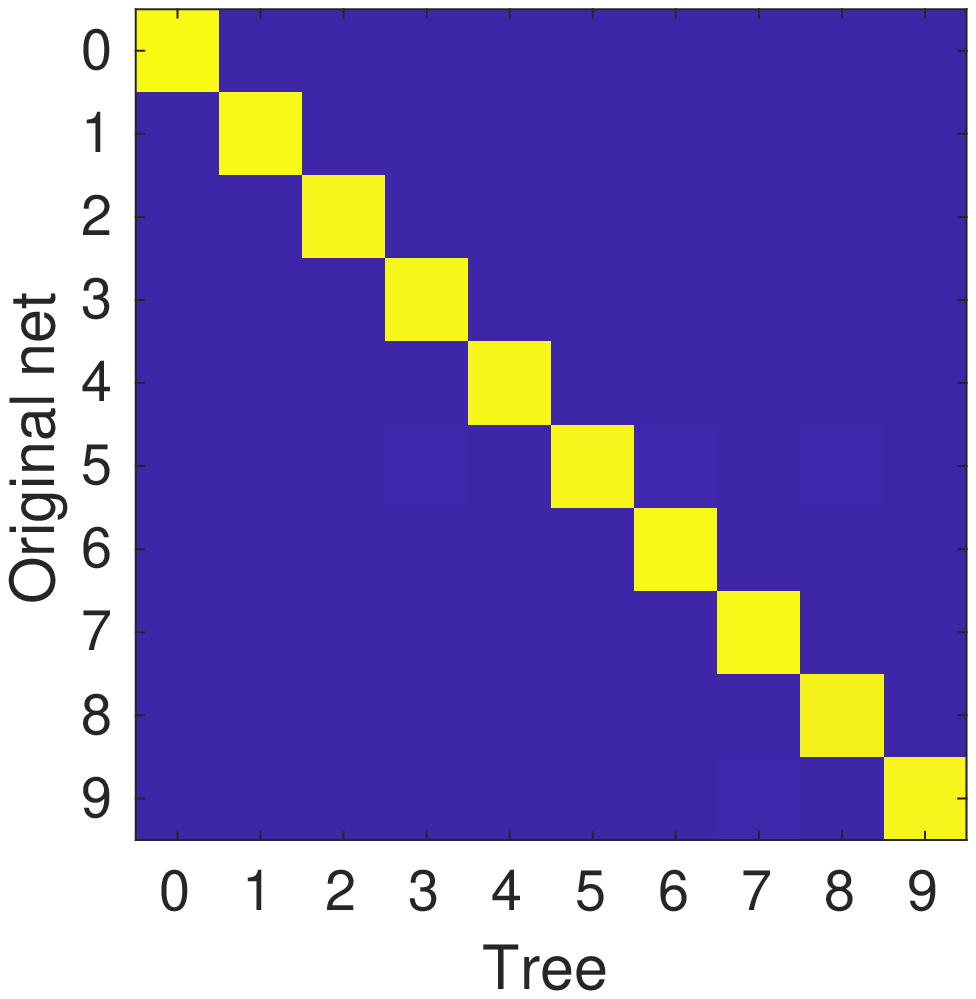}&
    \multicolumn{3}{c}{
      \psfrag{Original net}[b][t][.7]{original net}
      \psfrag{Modified net}[t][b][.7]{masked net}
      \includegraphics*[width=.1\linewidth]{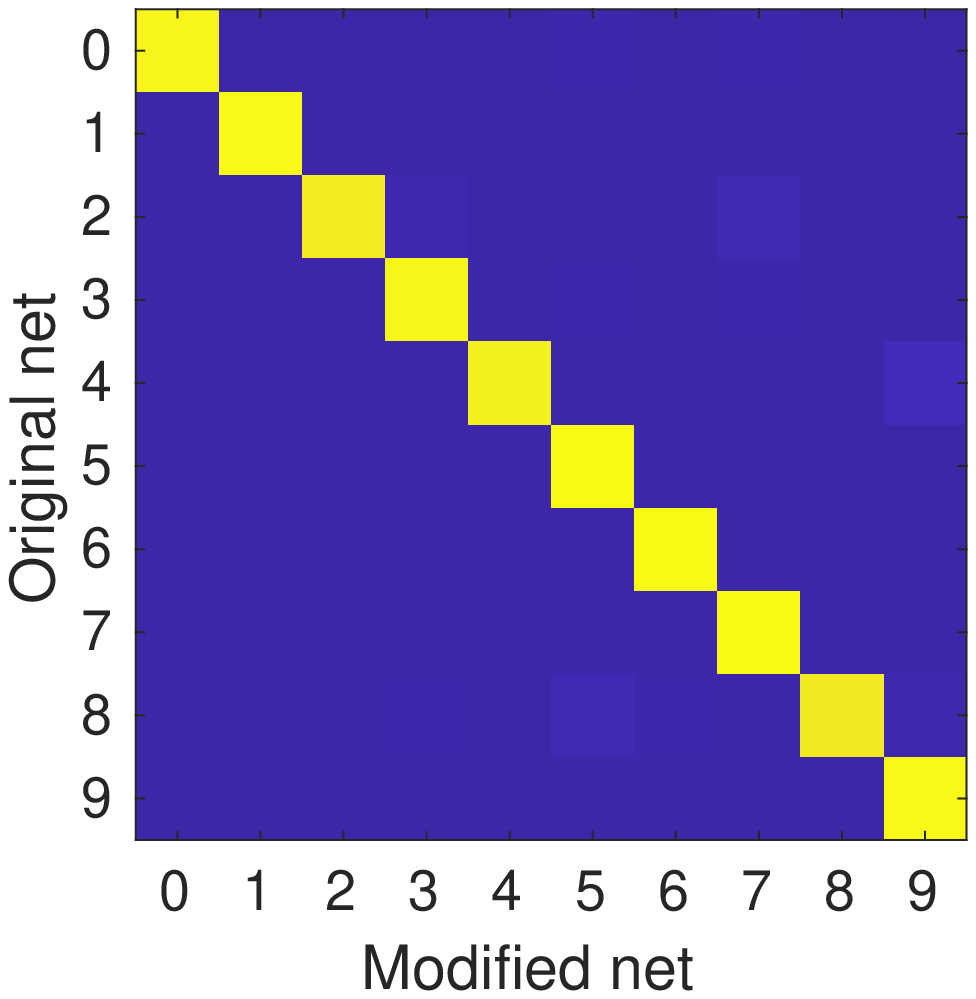}
      \raisebox{1.5ex}[0pt][0pt]{\psfrag{0}[][][0.5]{0}\psfrag{1}[][][0.5]{1}\makebox[0pt][l]{\includegraphics*[height=.09\textwidth]{VGG/colormap.eps}}}
    }&
    \psfrag{Original net}[b][t][.7]{original net}
    \psfrag{Modified net}[t][b][.7]{masked net}
    \includegraphics*[width=.1\linewidth]{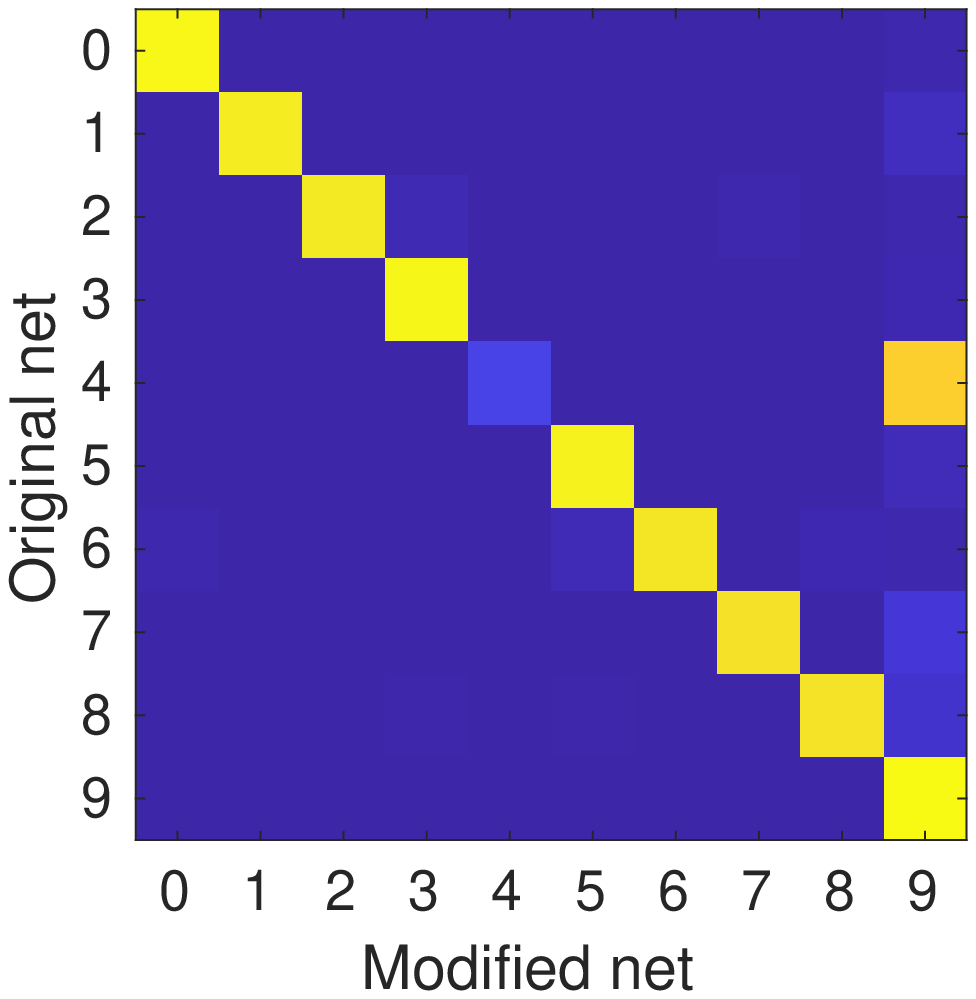}&
    \psfrag{Original net}{}
    \psfrag{Modified net}[t][b][.7]{masked net}
    \includegraphics*[width=.1\linewidth]{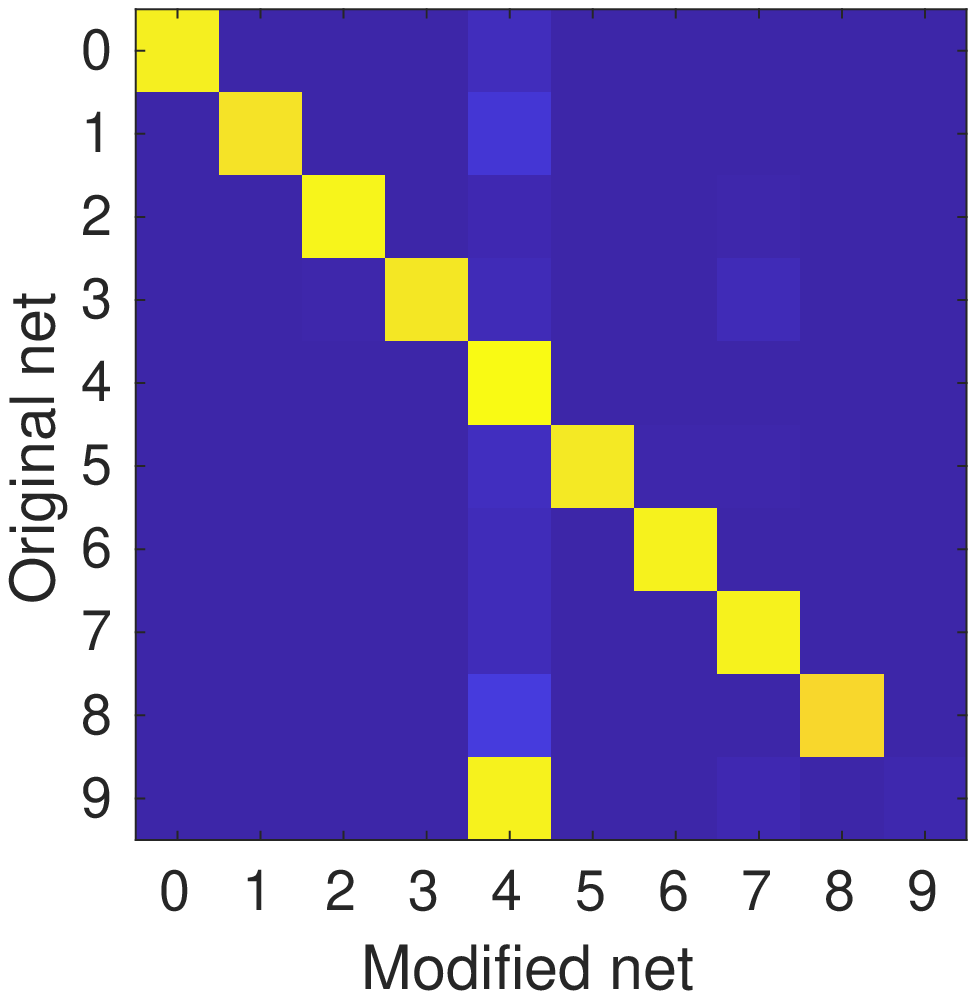}&
    \psfrag{Original net}{}
    \psfrag{Modified net}[t][b][.7]{masked net}
    \includegraphics*[width=.1\linewidth]{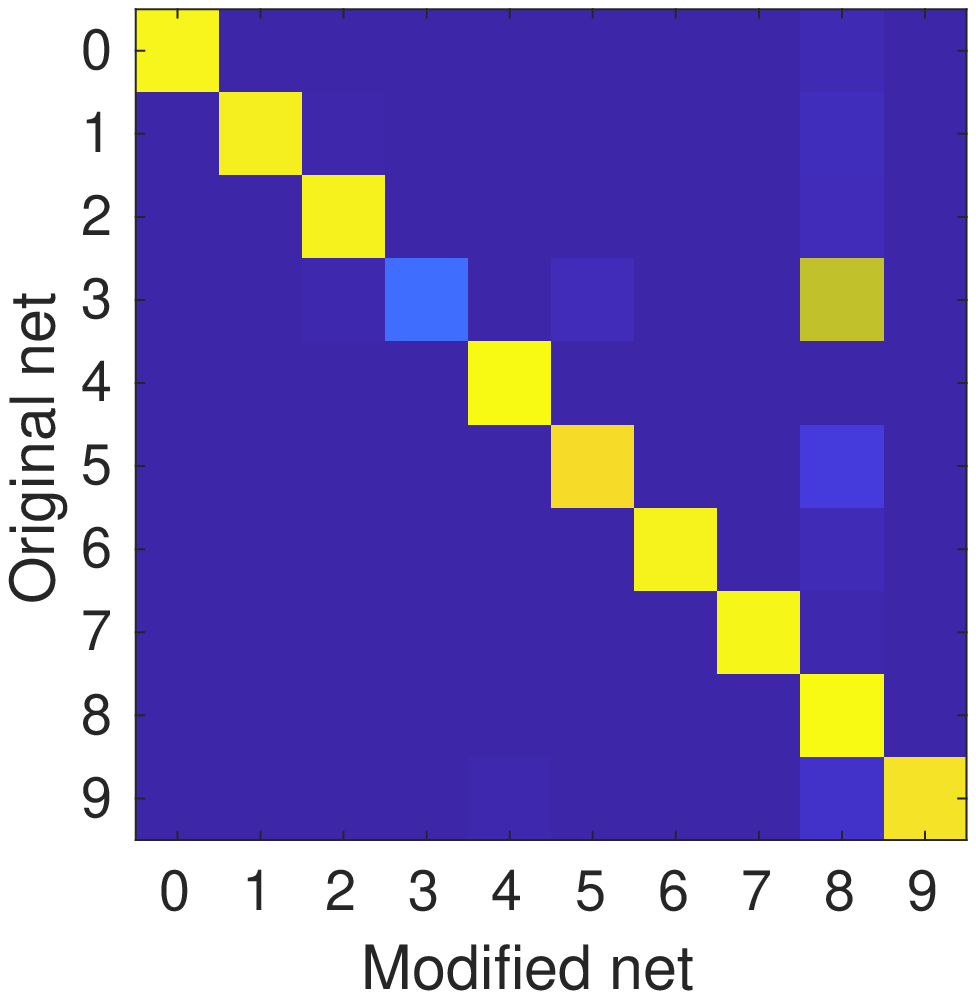}&
    \psfrag{Original net}{}
    \psfrag{Modified net}[t][b][.7]{masked net}
    \includegraphics*[width=.1\linewidth]{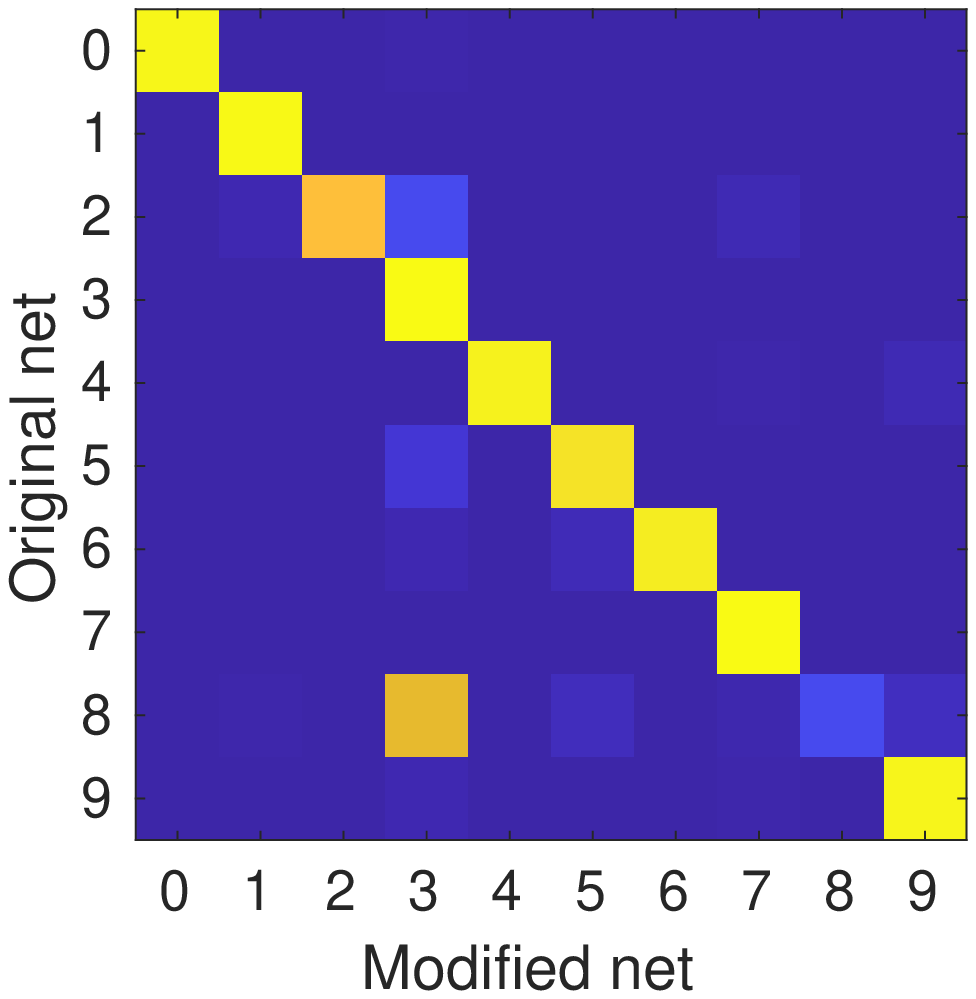}&
    \psfrag{Original net}{}
    \psfrag{Modified net}[t][b][.7]{masked net}
    \includegraphics*[width=.1\linewidth]{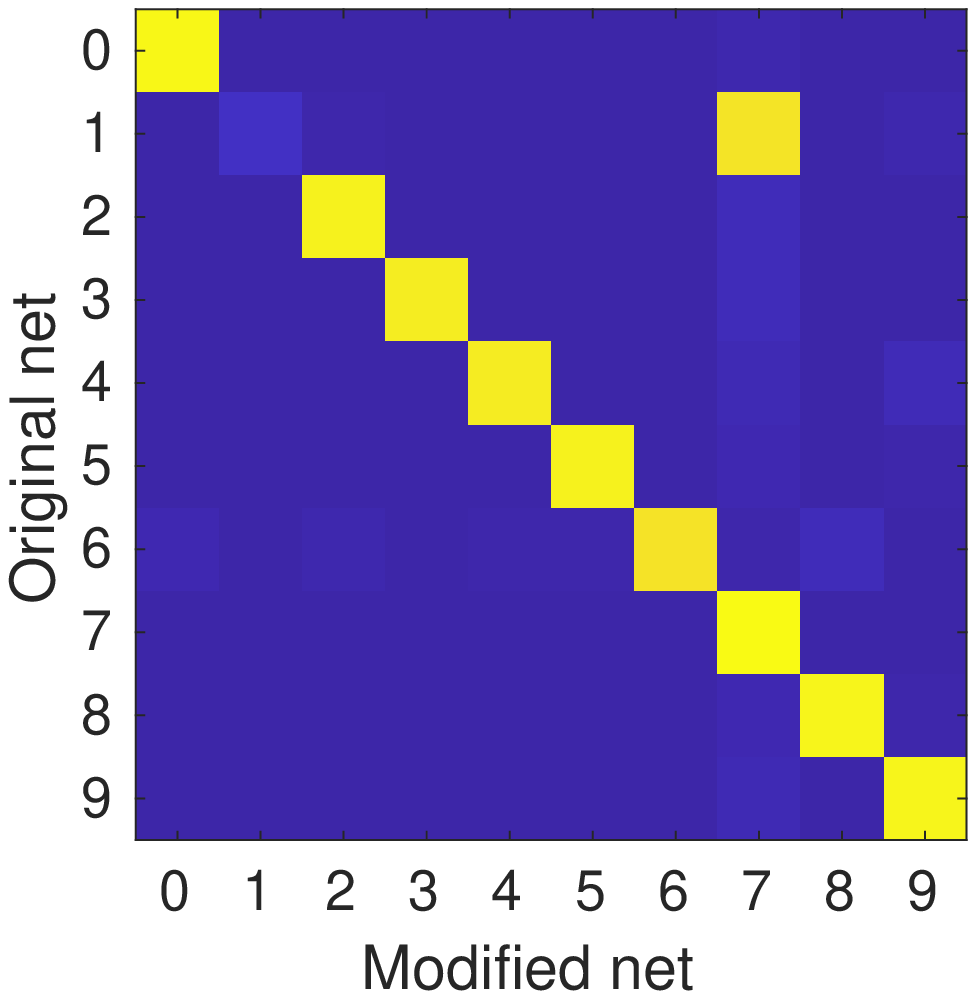} \\[2ex]
    \multicolumn{10}{c}{\dotfill \textsc{None to class $k$} \dotfill}\\
    $k = 0$ & $k = 1$ & $k = 2$ & $k = 3$ & $k = 4$ & $k = 5$ & $k = 6$ & $k = 7$ & $k = 8$ & $k = 9$ \\
    \psfrag{Original net}[b][t][.7]{original net}
    \psfrag{Modified net}[t][b][.7]{masked net}
    \includegraphics*[width=.1\linewidth]{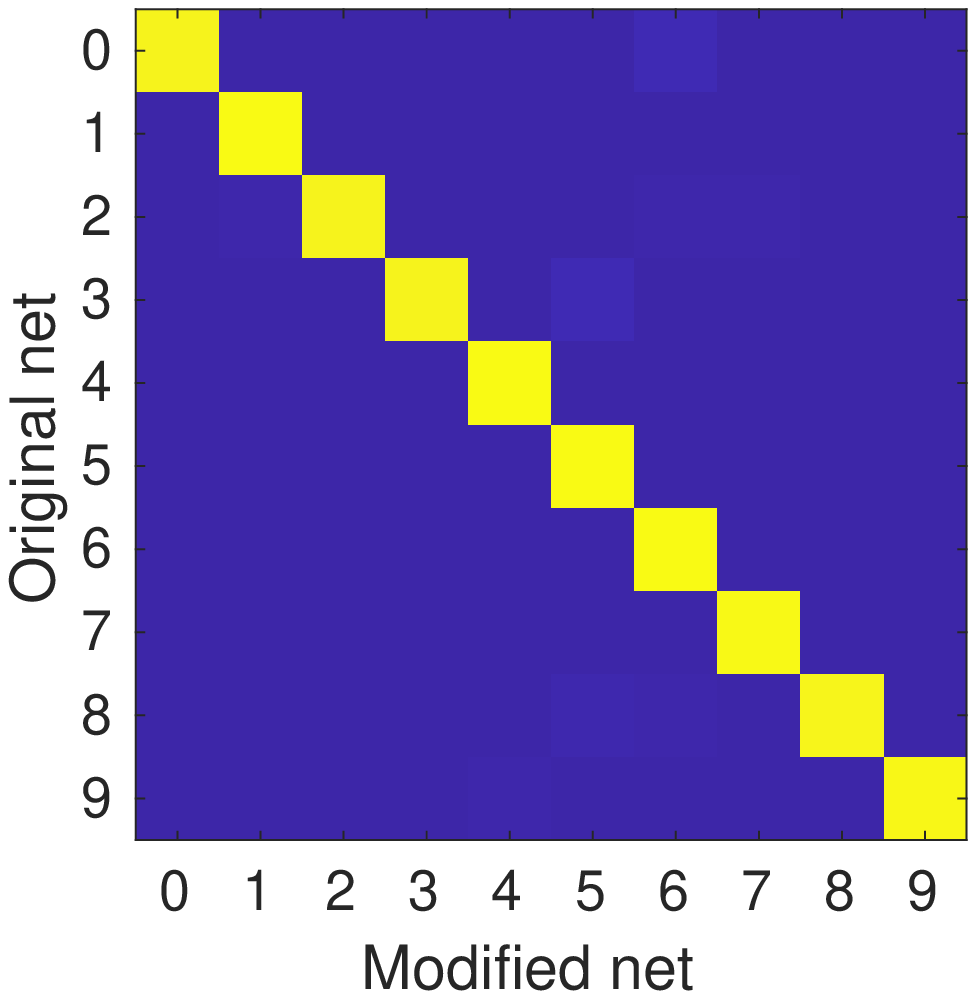}&
    \psfrag{Original net}{}
    \psfrag{Modified net}[t][b][.7]{masked net}
    \includegraphics*[width=.1\linewidth]{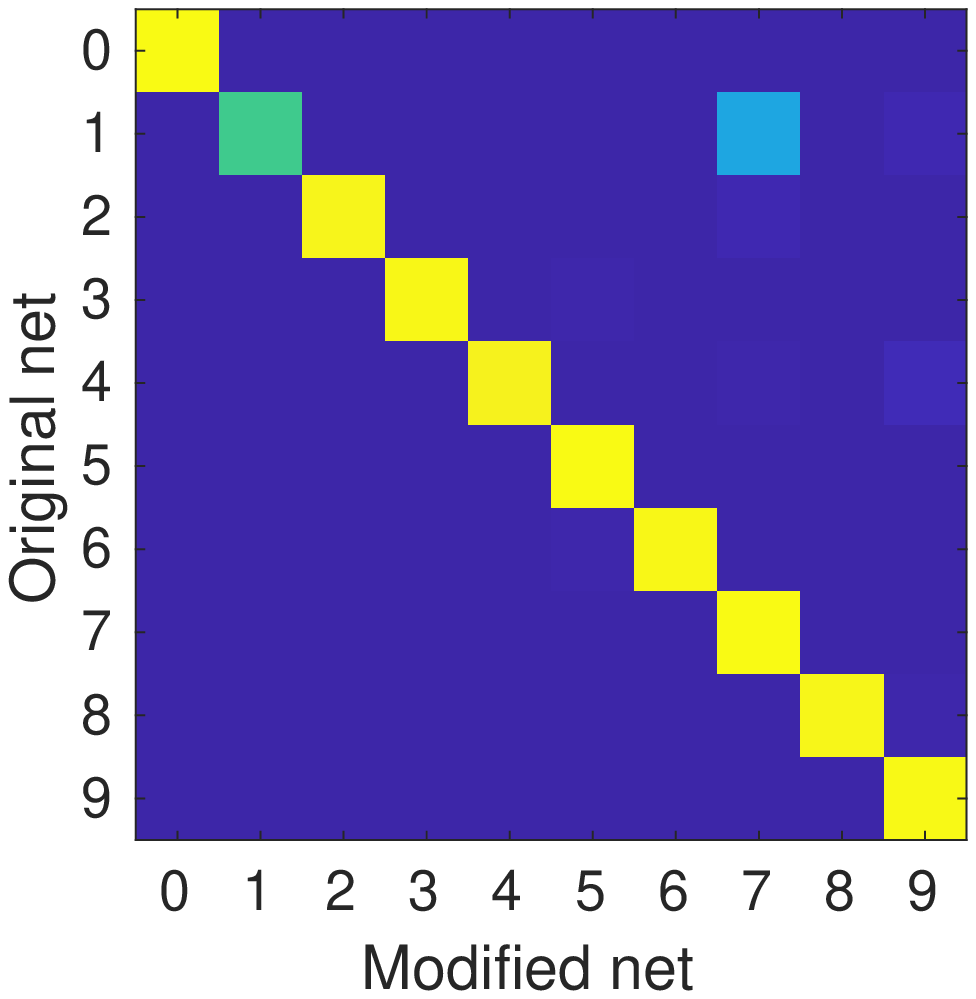}&
    \psfrag{Original net}{}
    \psfrag{Modified net}[t][b][.7]{masked net}
    \includegraphics*[width=.1\linewidth]{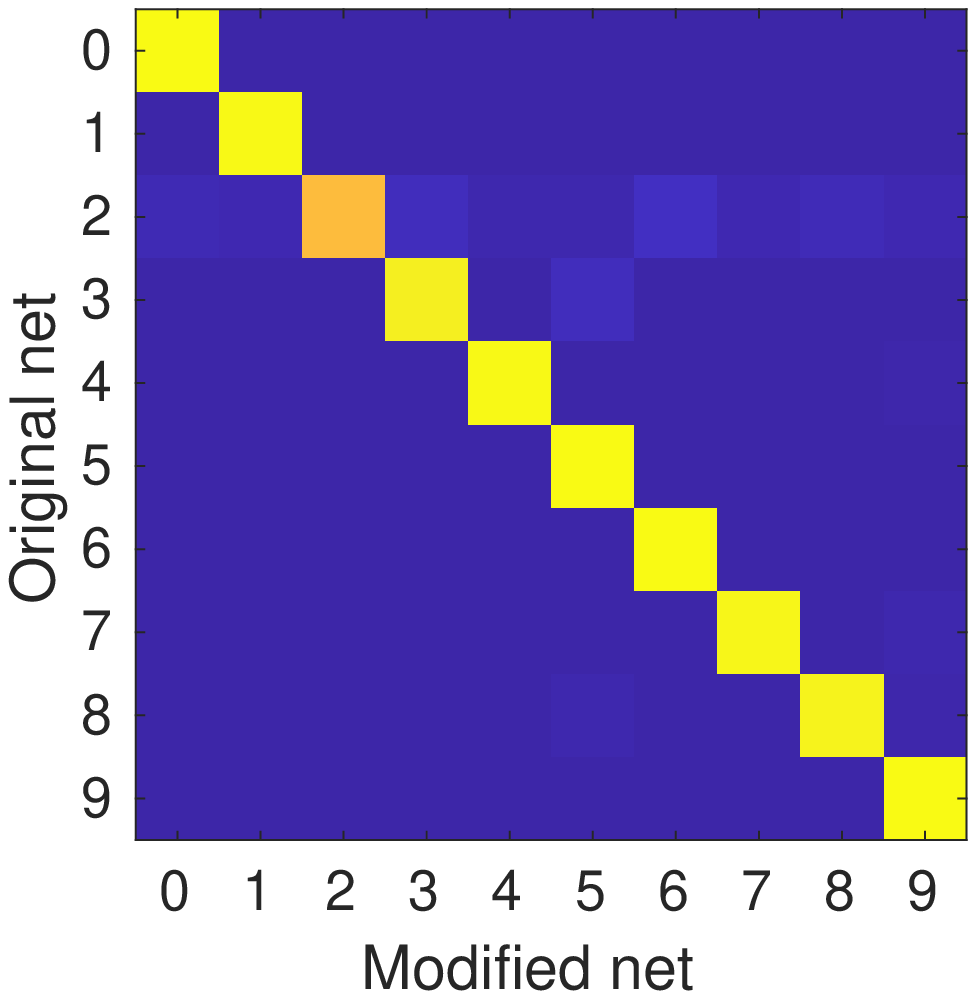}&
    \psfrag{Original net}{}
    \psfrag{Modified net}[t][b][.7]{masked net}
    \includegraphics*[width=.1\linewidth]{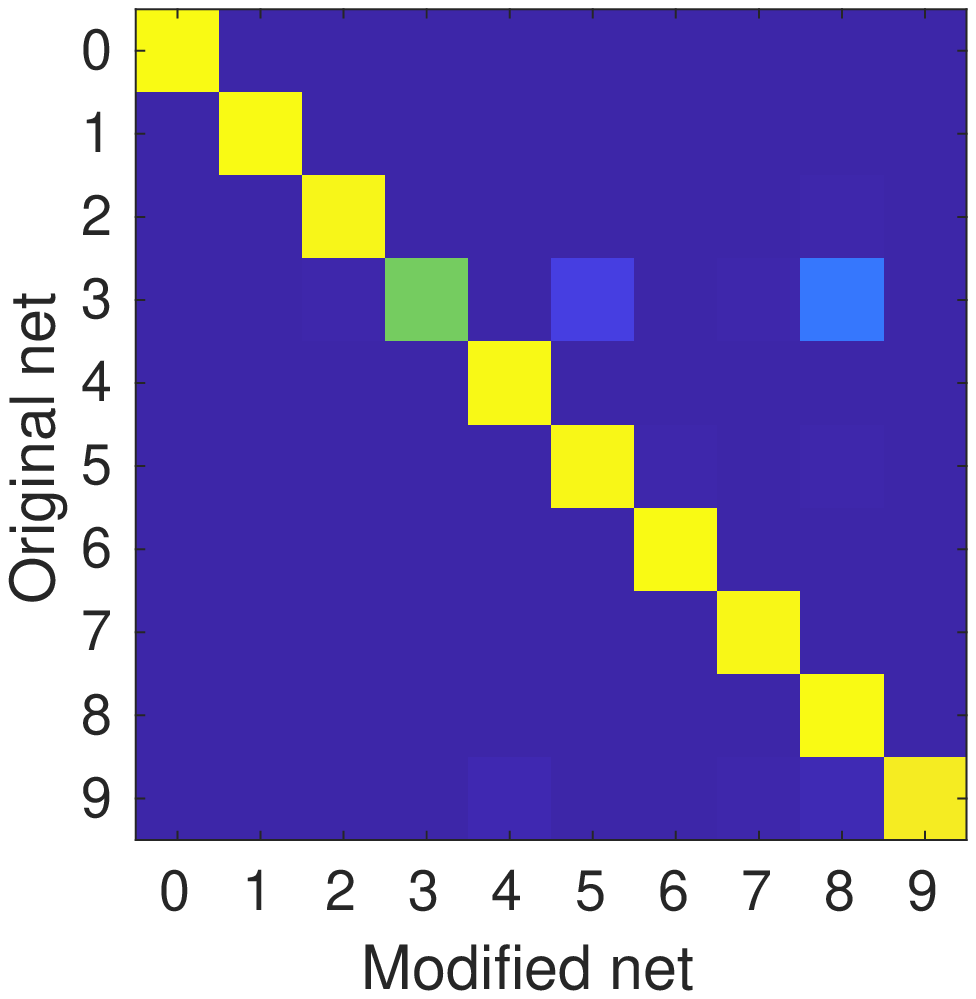}&
    \psfrag{Original net}{}
    \psfrag{Modified net}[t][b][.7]{masked net}
    \includegraphics*[width=.1\linewidth]{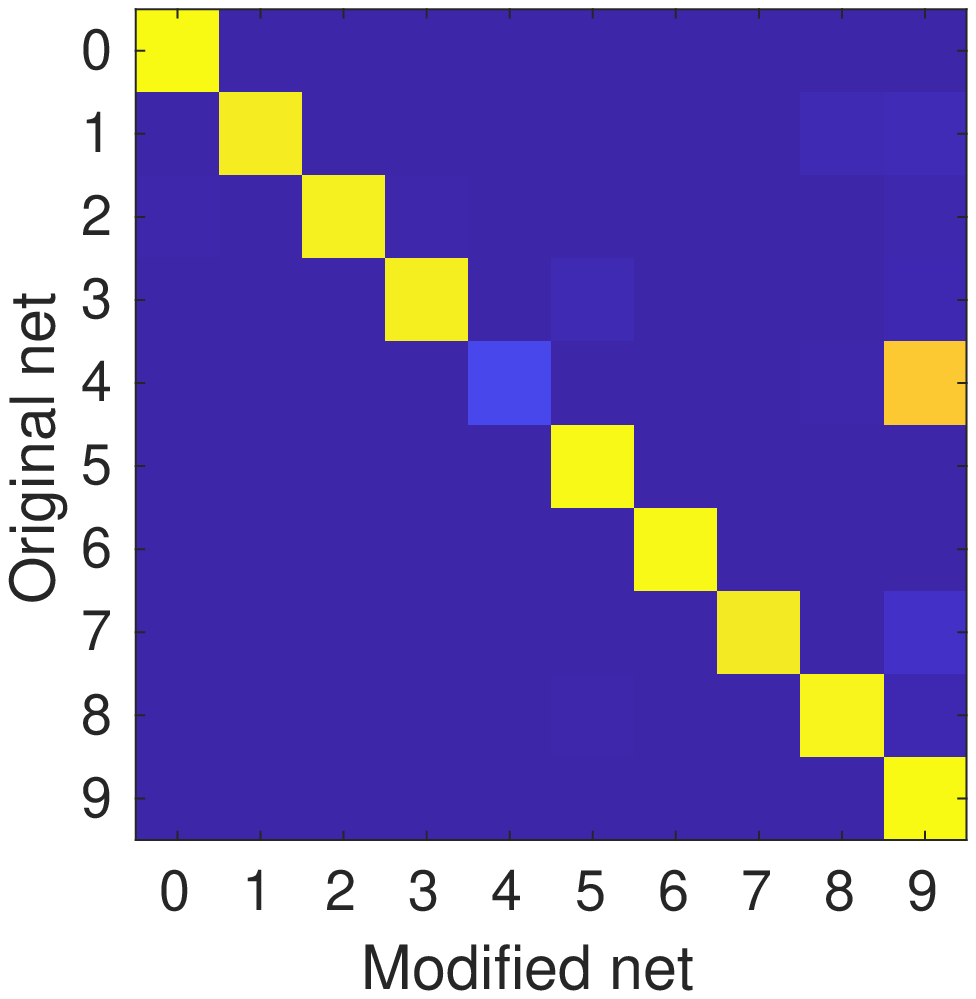}&
    \psfrag{Original net}{}
    \psfrag{Modified net}[t][b][.7]{masked net}
    \includegraphics*[width=.1\linewidth]{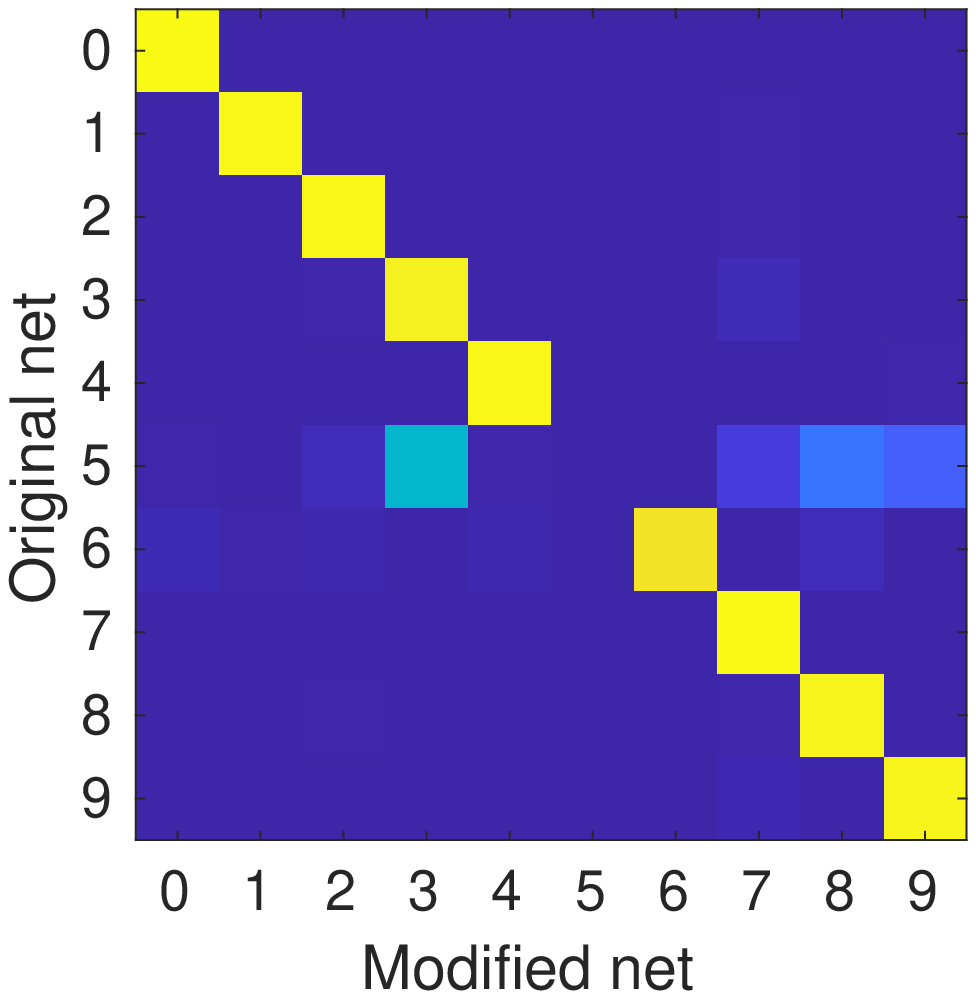}&
    \psfrag{Original net}{}
    \psfrag{Modified net}[t][b][.7]{masked net}
    \includegraphics*[width=.1\linewidth]{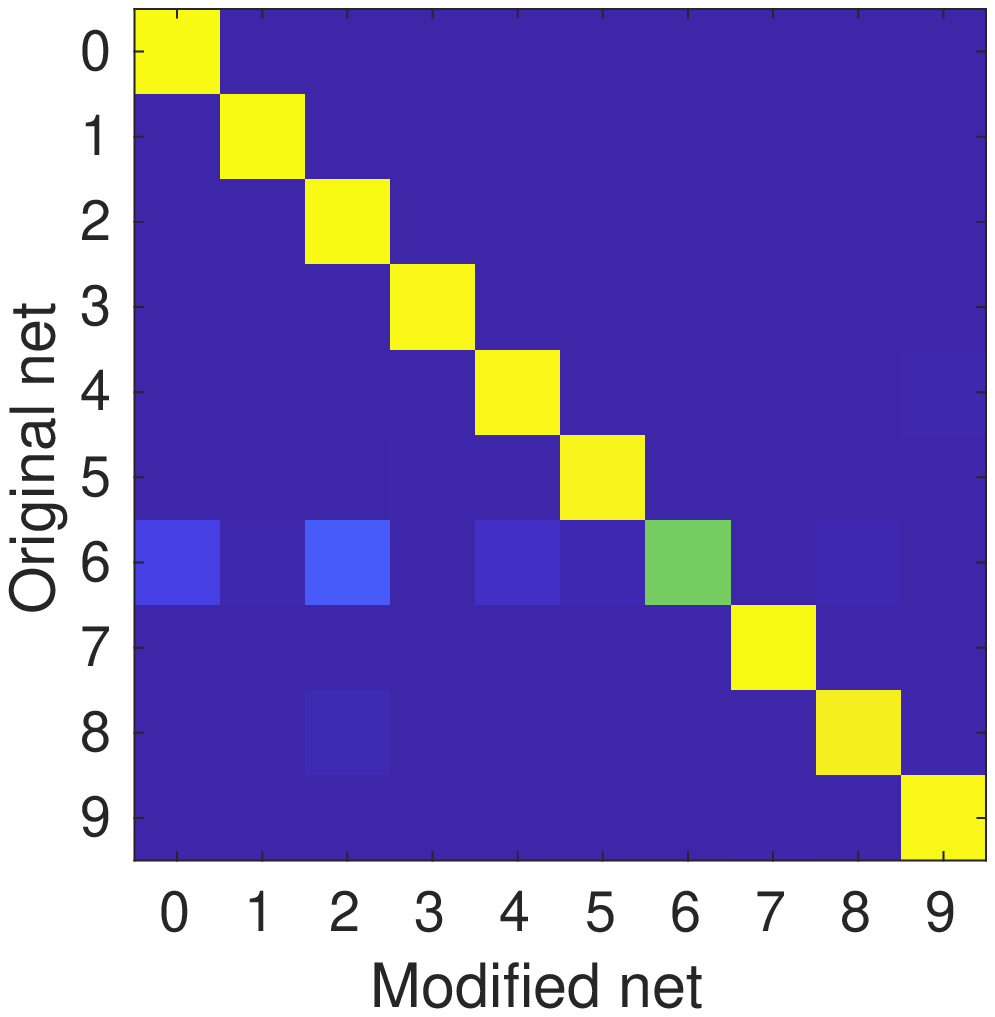}&
    \psfrag{Original net}{}
    \psfrag{Modified net}[t][b][.7]{masked net}
    \includegraphics*[width=.1\linewidth]{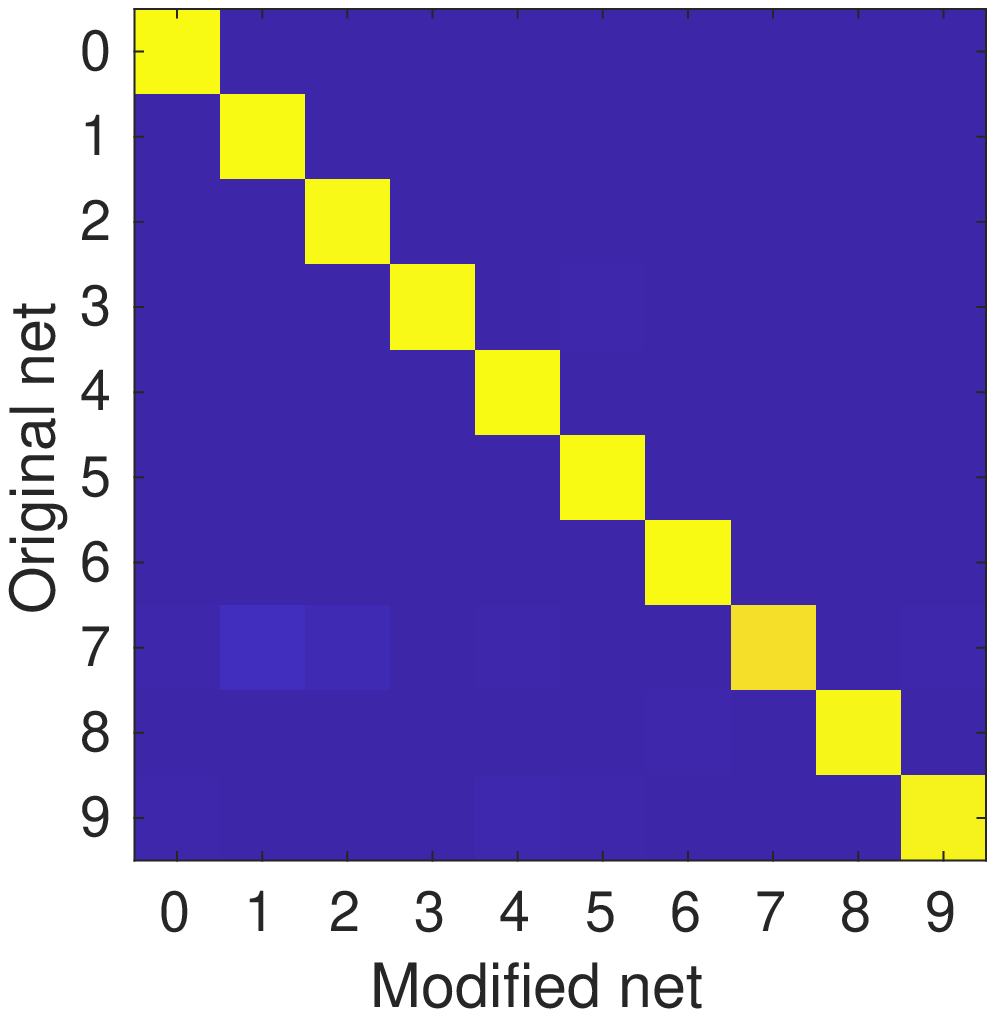}&
    \psfrag{Original net}{}
    \psfrag{Modified net}[t][b][.7]{masked net}
    \includegraphics*[width=.1\linewidth]{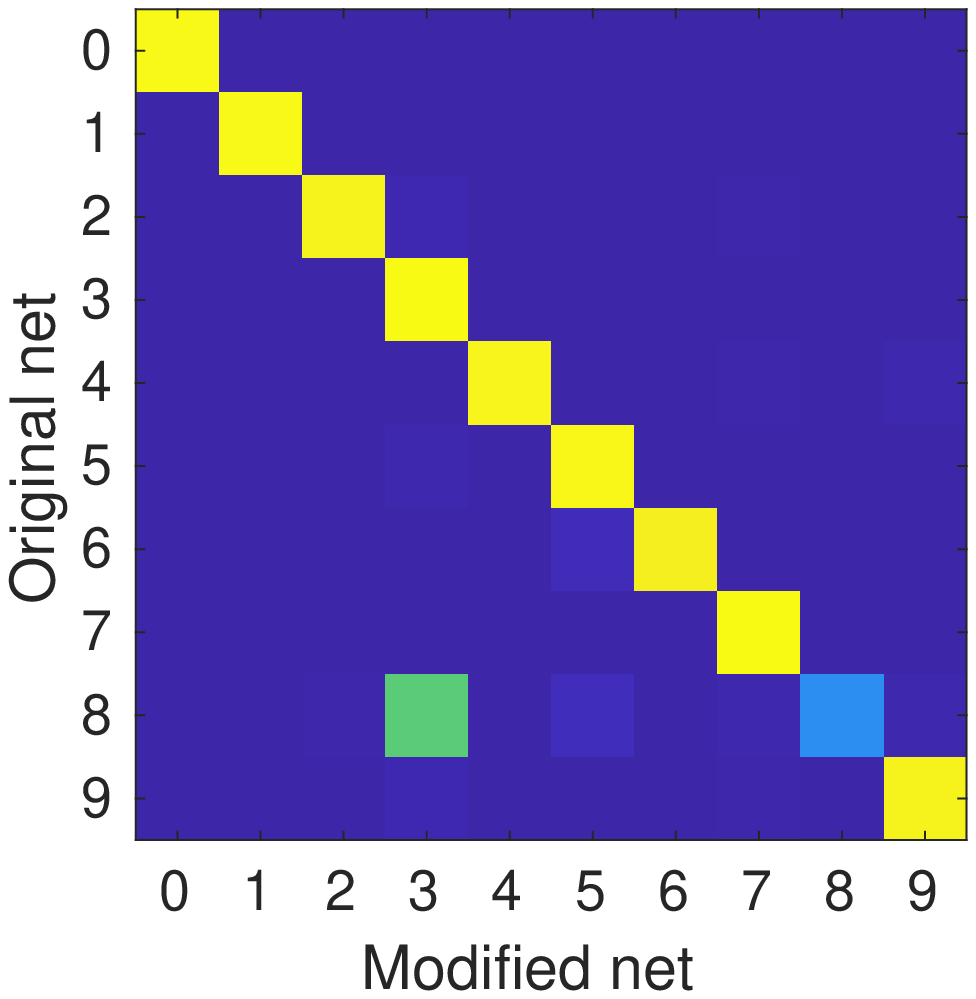}&
    \psfrag{Original net}{}
    \psfrag{Modified net}[t][b][.7]{masked net}
    \includegraphics*[width=.1\linewidth]{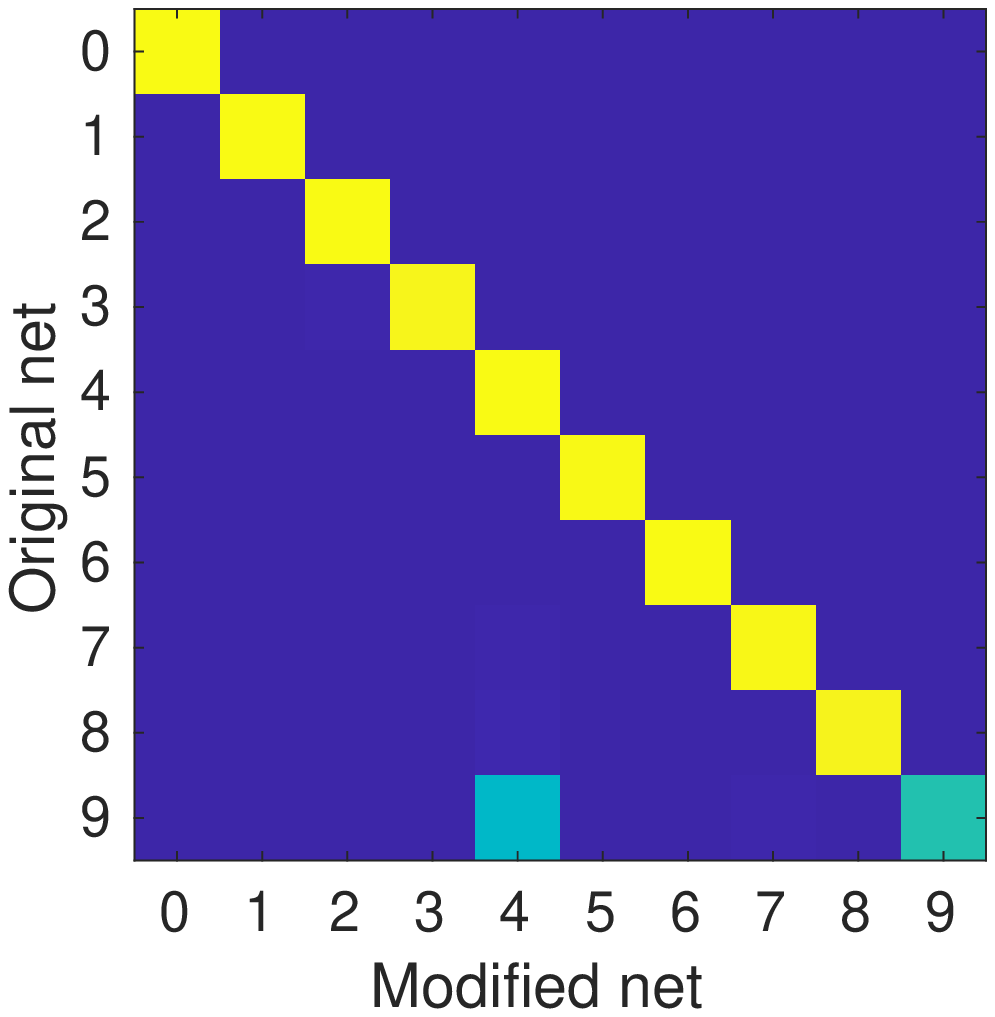} \\[2ex]
    \multicolumn{10}{c}{\dotfill \textsc{All to class $k$} \dotfill}\\
    $k = 0$ & $k = 1$ & $k = 2$ & $k = 3$ & $k = 4$ & $k = 5$ & $k = 6$ & $k = 7$ & $k = 8$ & $k = 9$ \\
    \psfrag{Original net}[b][t][.7]{original net}
    \psfrag{Modified net}[t][b][.7]{masked net}
    \includegraphics*[width=.1\linewidth]{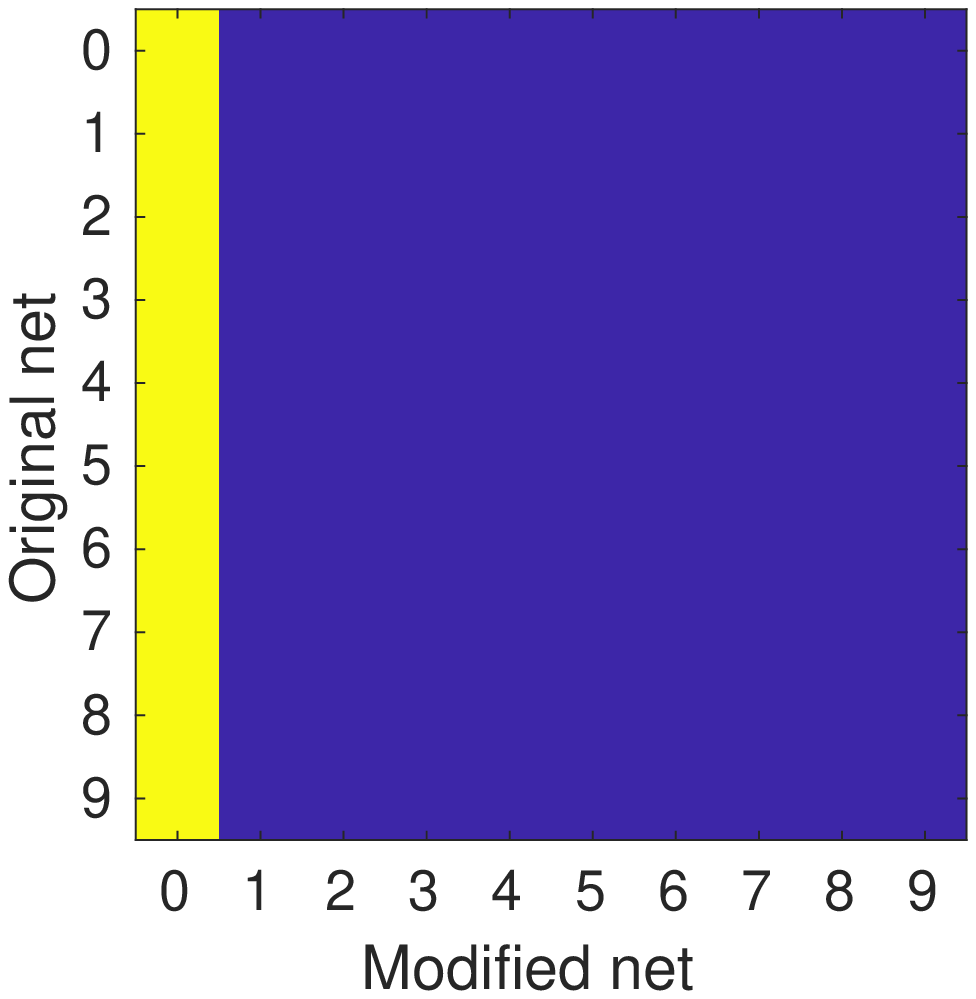}&
    \psfrag{Original net}{}
    \psfrag{Modified net}[t][b][.7]{masked net}
    \includegraphics*[width=.1\linewidth]{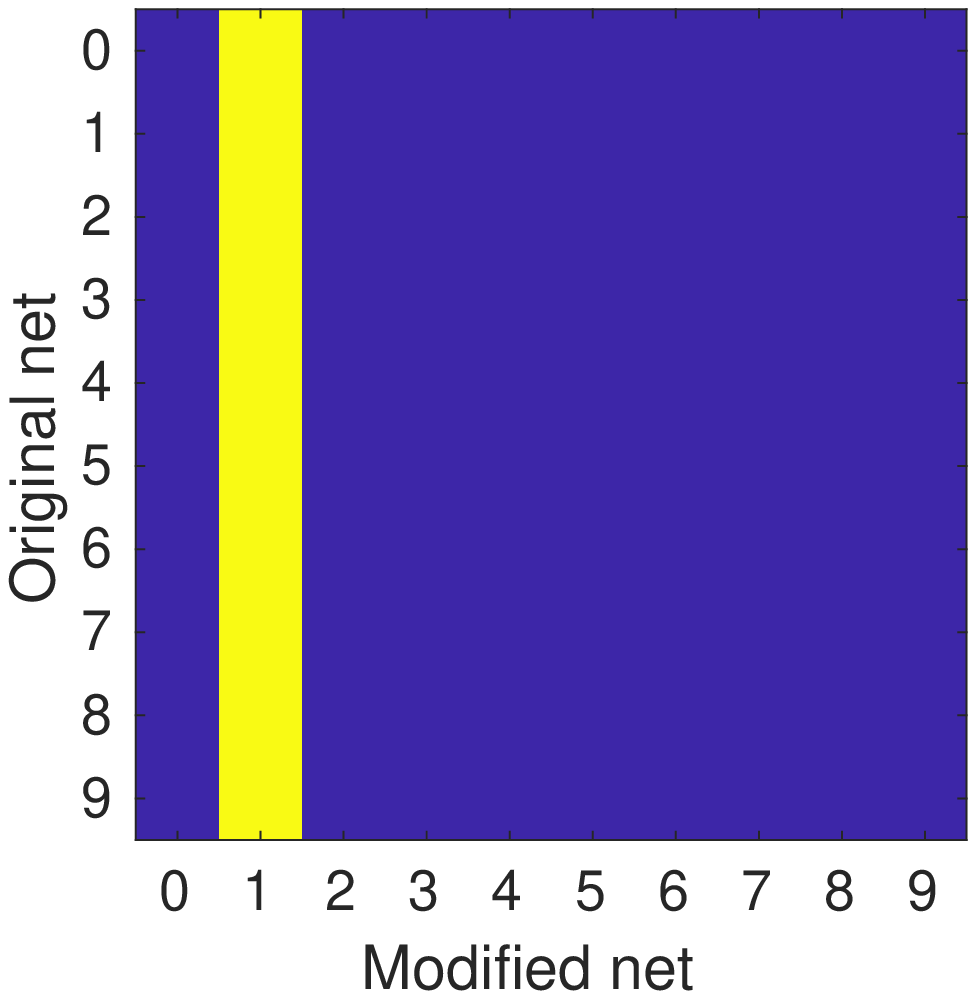}&
    \psfrag{Original net}{}
    \psfrag{Modified net}[t][b][.7]{masked net}
    \includegraphics*[width=.1\linewidth]{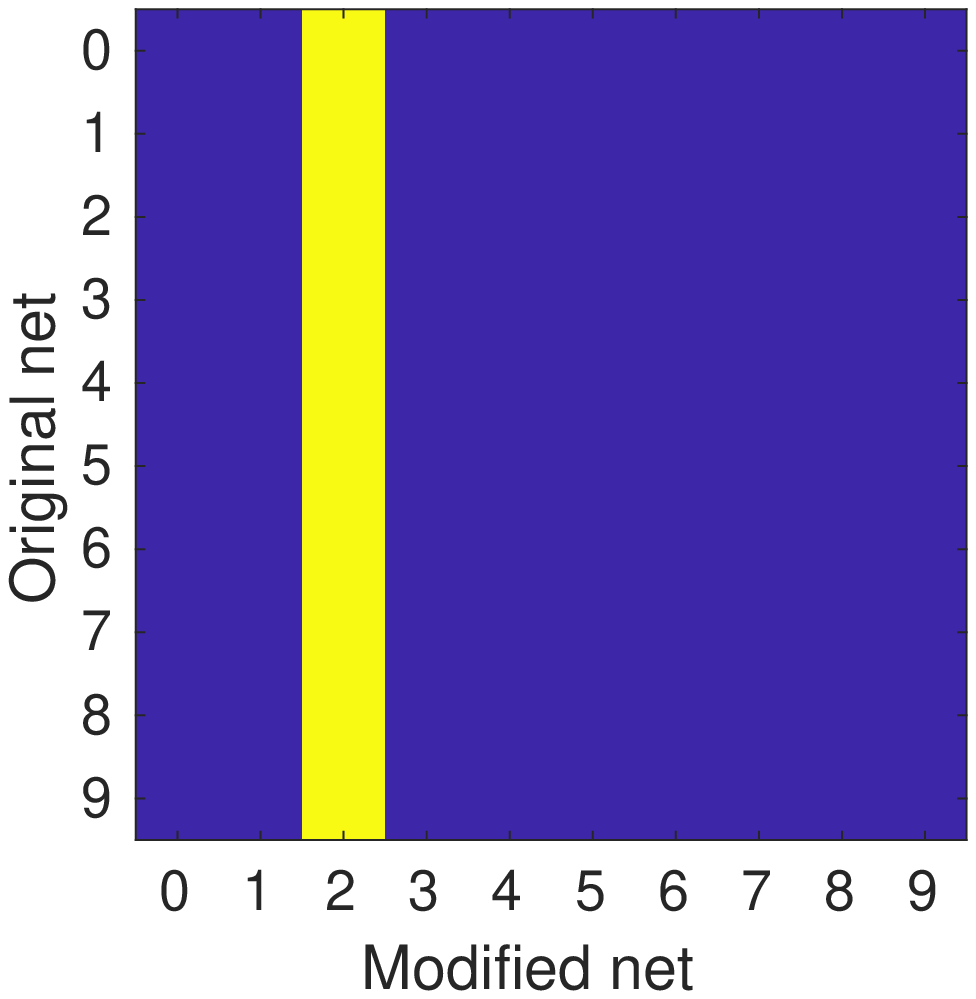}&
    \psfrag{Original net}{}
    \psfrag{Modified net}[t][b][.7]{masked net}
    \includegraphics*[width=.1\linewidth]{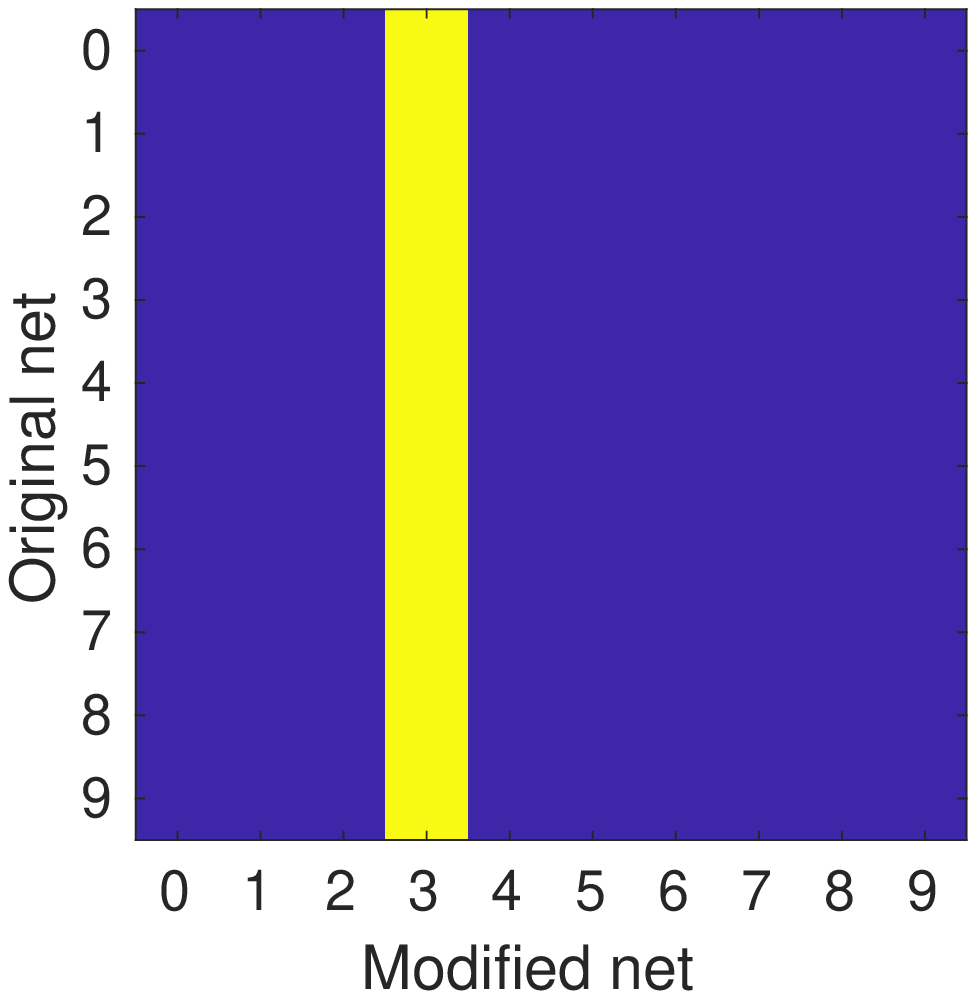}&
    \psfrag{Original net}{}
    \psfrag{Modified net}[t][b][.7]{masked net}
    \includegraphics*[width=.1\linewidth]{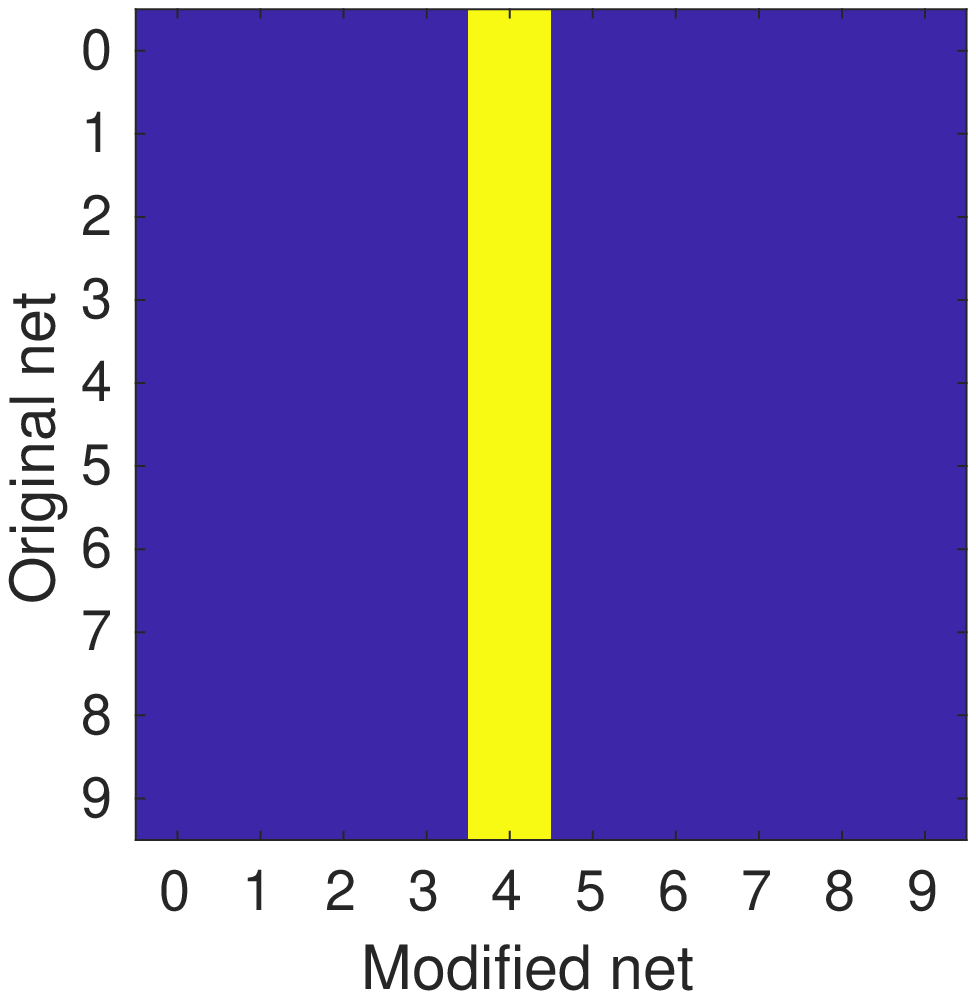}&
    \psfrag{Original net}{}
    \psfrag{Modified net}[t][b][.7]{masked net}
    \includegraphics*[width=.1\linewidth]{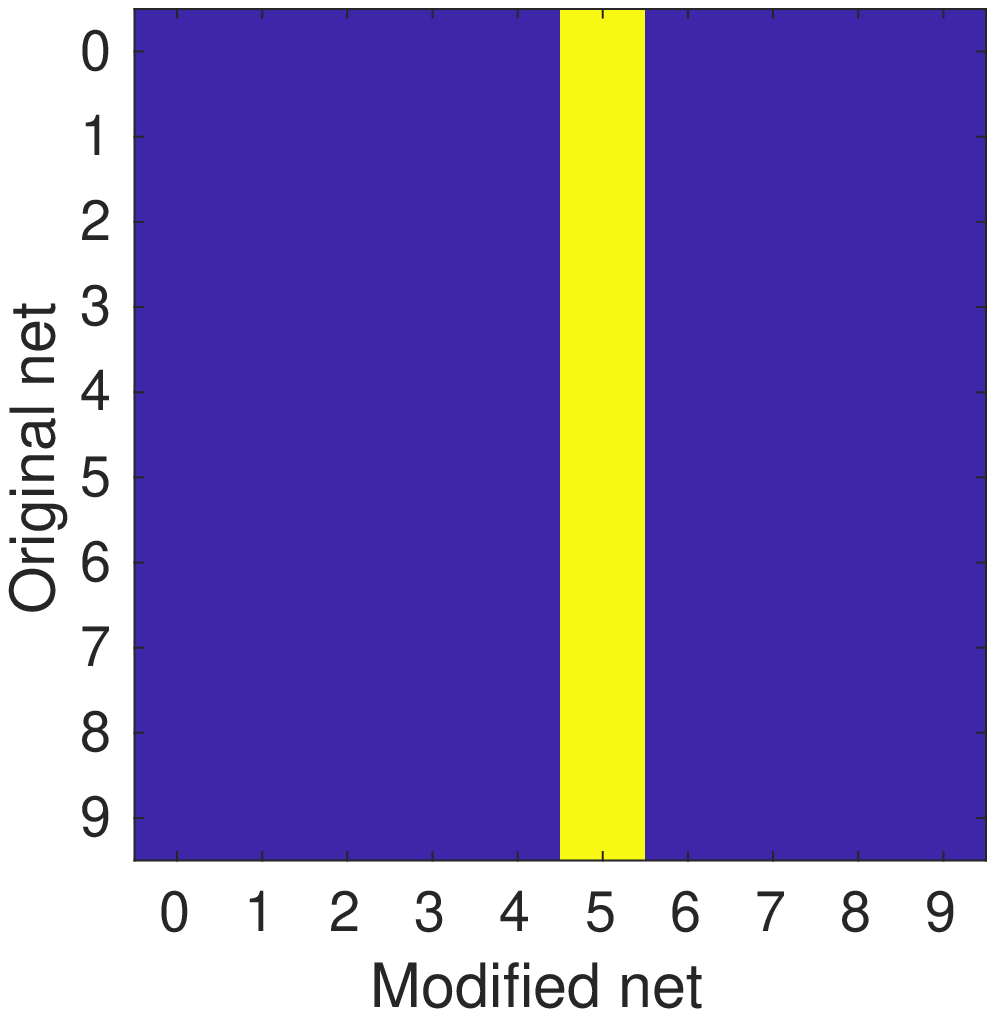}&
    \psfrag{Original net}{}
    \psfrag{Modified net}[t][b][.7]{masked net}
    \includegraphics*[width=.1\linewidth]{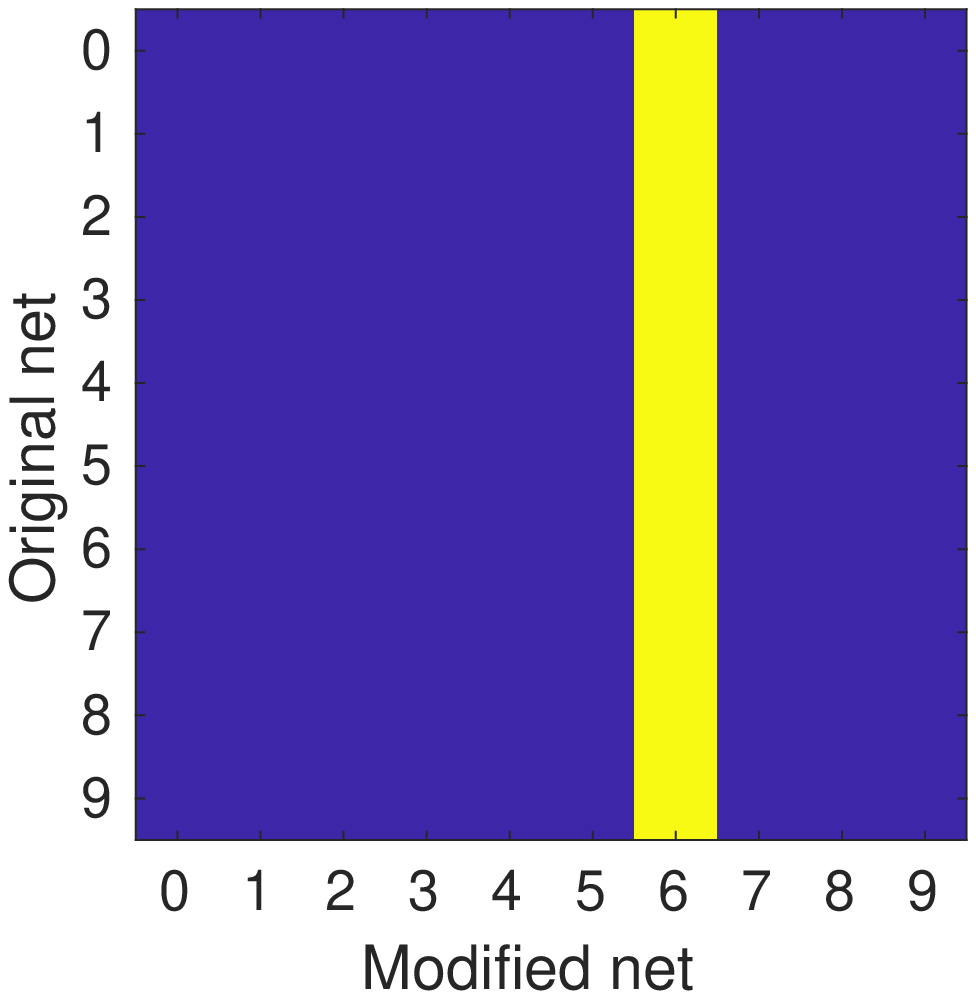}&
    \psfrag{Original net}{}
    \psfrag{Modified net}[t][b][.7]{masked net}
    \includegraphics*[width=.1\linewidth]{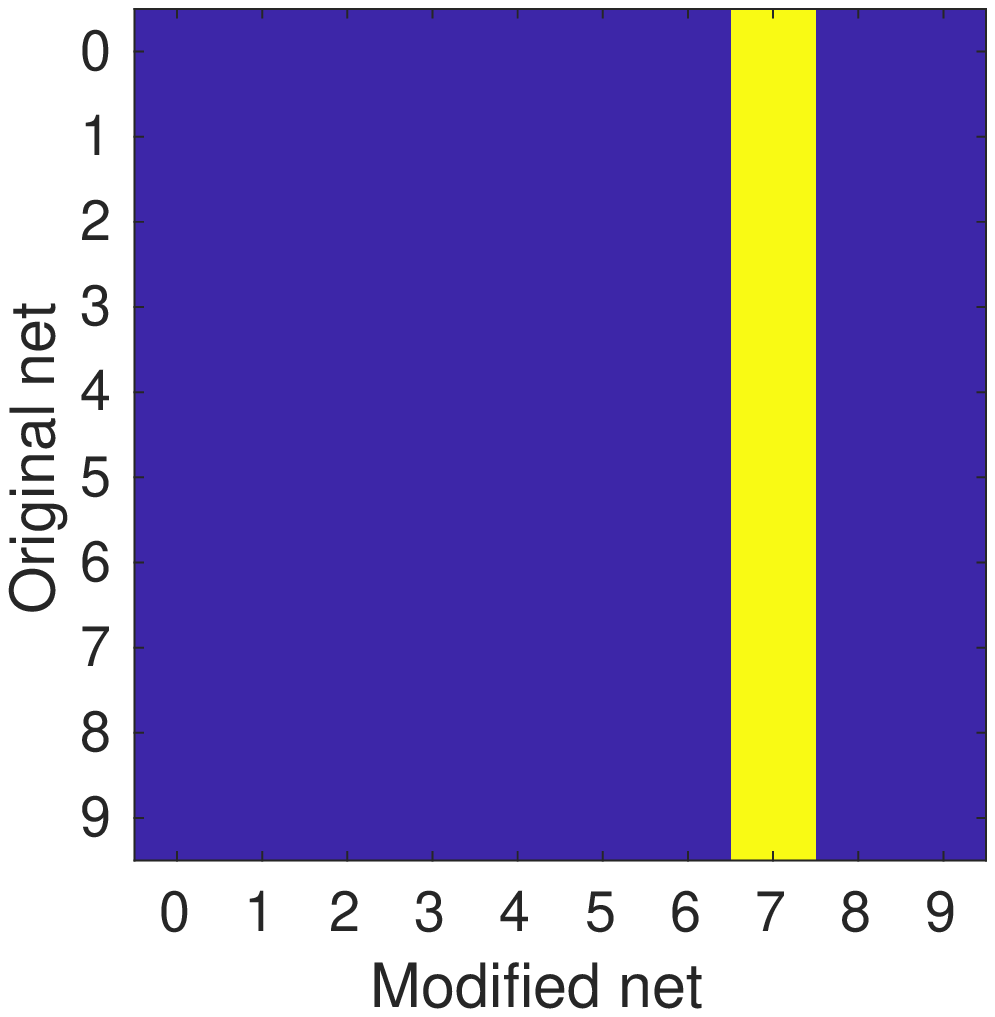}&
    \psfrag{Original net}{}
    \psfrag{Modified net}[t][b][.7]{masked net}
    \includegraphics*[width=.1\linewidth]{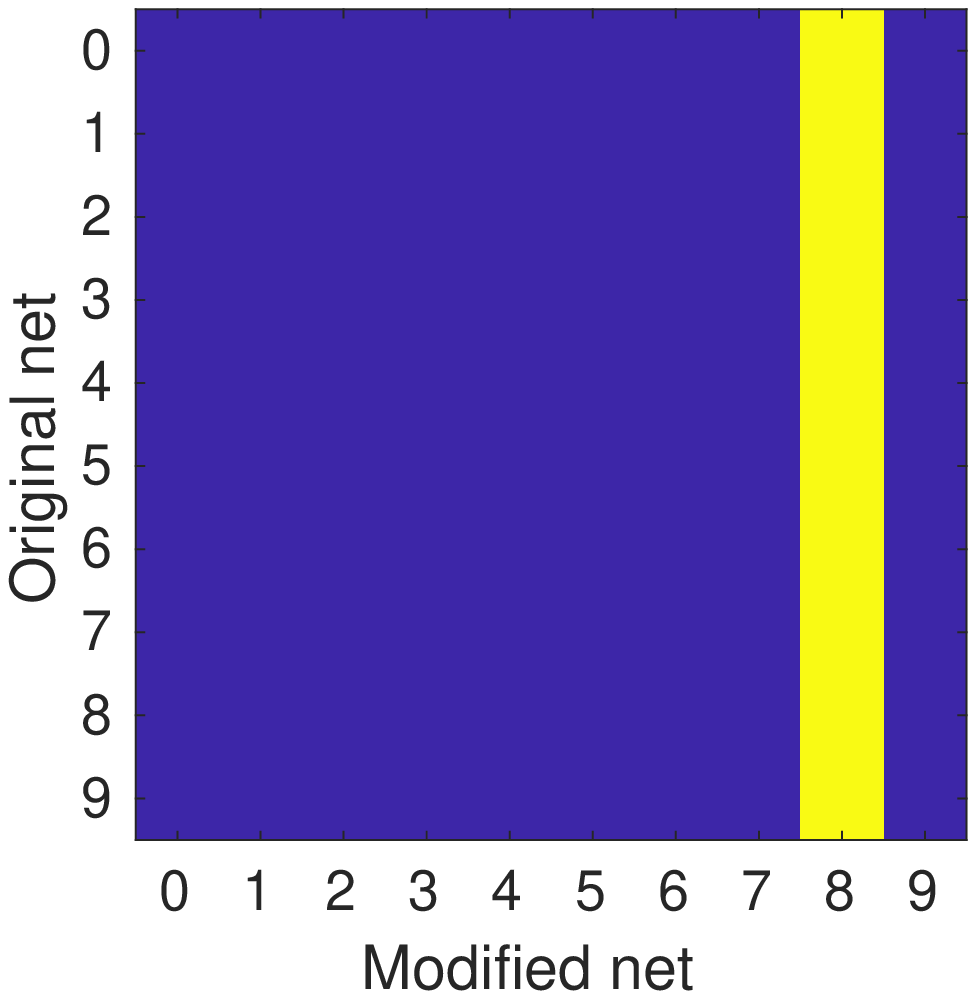}&
    \psfrag{Original net}{}
    \psfrag{Modified net}[t][b][.7]{masked net}
    \includegraphics*[width=.1\linewidth]{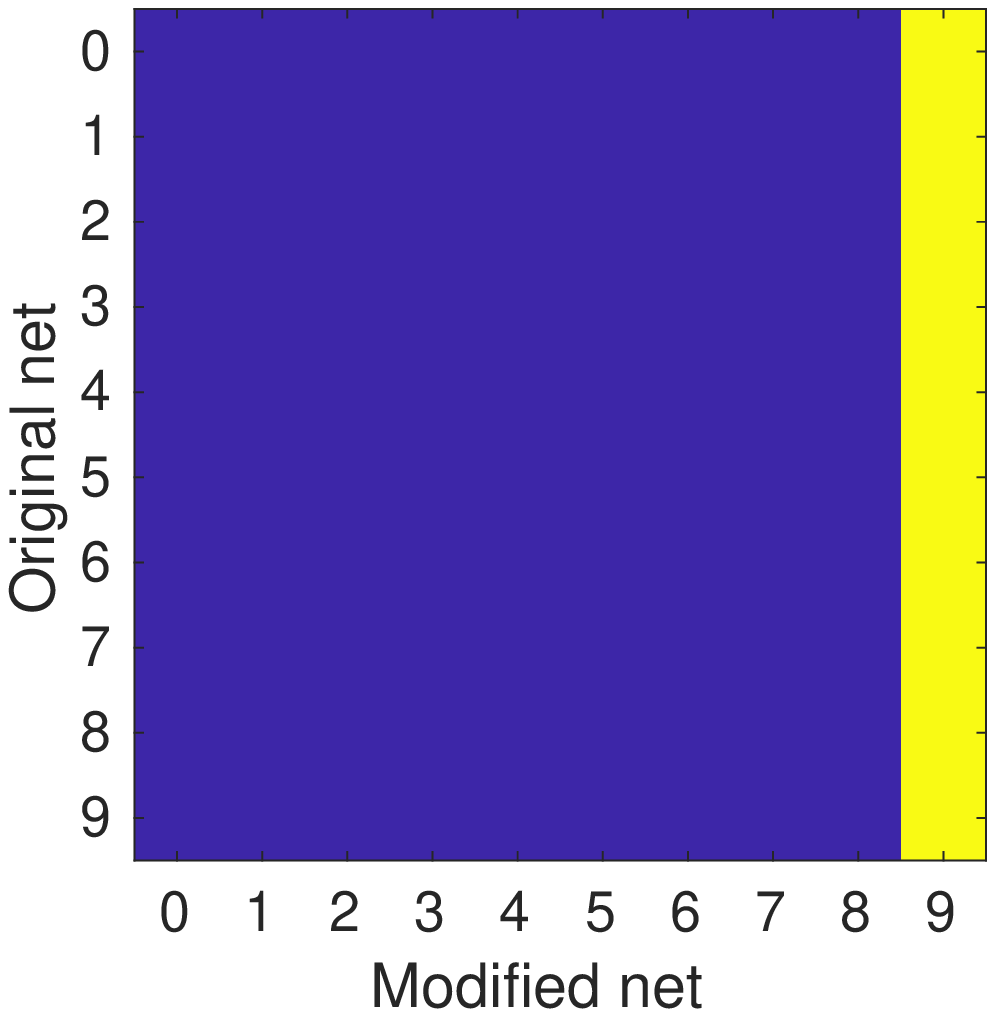}\\
  \end{tabular}
  \caption{Like fig.~\ref{f:LeNet5-masks-test} but for the training set.}
  \label{f:LeNet5-masks-training}
\end{figure}

We trained sparse oblique trees on features obtained by the LeNet5 neural net architecture for MNIST. The results agree qualitatively with those for VGG16: we are able to obtain trees with nearly the same error as the original net, hence good mimics; the masks we construct work as desired in nearly all the training and test instances; and the tree is highly interpretable.

Specifically, we train a LeNet5 net on the 60\,000 training images for MNIST, of $28 \times 28$ pixels (10 handwritten digit classes), and report test errors on the 10\,000 test instances \citep{Lecun_98a}. The LeNet5 architecture is in table~\ref{t:LeNet5}. We train the network using Nesterov's accelerated gradient method for 100 epochs using minibatches of size 512, learning rate 0.02 (updated every 20 epochs with a factor of $0.99^2$) and momentum rate 0.9. The training error is $0.00545$\% and the test error is $0.61$\%. We select the $F =$ 800 neurons at layer conv2 as features on which to train the tree.

We used TAO with an initial tree structure of depth 5 (total 63 nodes) and random initial values for the weights at the nodes. We constructed a collection of trees over a range of the sparsity hyperparameter $\lambda \in [0,\infty)$ (the regularization path), running TAO for 40 iterations (when it approximately converged) to learn each tree. Fig.~\ref{f:LeNet5-error} shows, as a function of $\lambda$, the error (training and test), number of nodes and depth of each tree, and number of nonzero weights in each tree (total over all its nodes). As with the VGG16 trees, as we increase $\lambda$ and therefore impose increasing sparsity on the tree (in terms of both tree size and number of nonzero weights in the decision nodes), the training error increases steadily but the test error remains about constant, so both curves approach, meet for some $\lambda$ value and approximately coincide from that point on. This provides opportunities to find a tree with pretty good accuracy but significantly sparse. As we increase $\lambda$, the tree size decreases (number of nodes and depth) and the number of nonzero weights decreases.

We selected as mimic the tree for $\lambda =$ 20, with depth 5 and only 27 nodes. It has an error of 1.28\% (training) and 1.67\% (test), which is very close to that of LeNet5, so we expect the tree to be a good mimic of the net. The best classifier tree (for $\lambda =$ 5) had a depth of 5 and 27 nodes, and an error or 0.59\% (training) and 1.46\% (test). It is possible to reduce this error even more by using a larger tree, but the one we obtained is good enough as a mimic and to obtain masks.

Fig.~\ref{f:LeNet5-tree} shows the tree selected as mimic. The class histograms at each node show how the tree hierarchically splits classes very crisply; indeed, it has only 14 leaves for 10 classes (only digit classes `2', `6', `7', `9' appear with 2 leaves each). The blurry average image at each leaf shows significant shape variability, indicating the features have successfully learned to ignore such within-class variability. The weight vector at each decision node shows that very few features are used at each node; 295 features (37\% of the total 800) are not used at any node, so their values are irrelevant for classification in the tree. This also holds nearly perfectly for the deep net.

Figures~\ref{f:LeNet5-masks-test}--\ref{f:LeNet5-masks-training} show the confusion matrices for our different masks, which work nearly perfectly in the training instances and only slightly less so in the test ones. The number of features we need to mask out in each case is very small, around 40 (out of 800 features). Some masks work more reliably than others. Classifying all instances as class $k$ works surprisingly well no matter the choice of $k$. Misclassifying class $k_1$ as $k_2$ (where $k_1$ must have a single leaf which is a sibling of $k_2$) works also well, although a few instances from other classes are sometimes classified as $k_2$. Not classifying any instance as class $k$ works also well but fails with some instances, which remain as class $k$.

In is interesting to compare our tree with a TAO tree trained directly on the pixel values (hence not associated with any deep net), such as that in \citep[fig.~1]{CarreirTavall18a}. The tree trained on LeNet5 features is much smaller, sparser and accurate. The best test error for a tree on pixels (panel 2 of \citep[fig.~1]{CarreirTavall18a}) is 5.69\% for a tree of depth 8 and 75 nodes (in our own experiments we can get this down to around 5\%). This error is remarkably low for a tree but, compared to our tree, the error is much bigger, and the tree is quite larger and messier. A smaller, more interpretable tree on pixel values shown in panel 3 of \citep[fig.~1]{CarreirTavall18a} achieves an error of around 10\% (training or test) with depth 7 and 33 nodes. Since the tree operates on the raw pixels, inspection of the weight vectors is very informative in identifying ``strokes'' that characterize the difference between a digit 4 and a digit 9, for example. As to applying our masks at the pixel level, this is far less successful if we want guarantees for (nearly) all instances; however, it may be possible to make this work for a specific image.

\section{Discussion and limitations of our work}

Decision trees are a good choice of interpretable classifier because they handle multiple classes naturally, do feature selection automatically, have a hierarchical structure that promotes an increasing specialization from the root towards the leaves, and can be inspected. Sparse oblique trees improve this by using few features at each node, so they explicitly show the influence of groups of features on classes. This makes it possible to find important subsets of features efficiently among thousands of possible features. For such trees to be useful, it is critical to be able to train them to high accuracy so they can mimic (part of) a deep net, which the TAO algorithm does.

As shown by our masks, deep net features participate in a coordinated way in predicting a class, where small groups of specific features encode information specific for some decisions (rather than, say, all features participating in all classes for all instances). That we find features specialized for classes is not that surprising---some features must provide information for some classes, after all. What is surprising is the small size of these groups of features. This is probably partly due to the neural network being heavily overparameterized. What do these groups of neural net features represent anyway? Unfortunately, in spite of the high activity in this area (as summarized in section~\ref{s:related}), at present the research community does not have a systematic understanding of what ``concepts'' these neural net features may be encoding; elucidating this remains an open research problem. 

Our findings are remarkably similar to recent findings in visual neuroscience \citep{Marshel_19a,Carril_19a} that show that very small groups of neurons (around 20) in mouse primary visual cortex seem to code for specific properties or behaviors. In fact, removing all visual input to the mouse and directly stimulating those neurons triggers the same behaviors---analogously to what our masks do. 

Recent work \citep{KuhnKacker19a} tries to address the problem of identifying a group of $n$ features (out of $m$ total features) that are related with a specific class. They consider explainability of a classifier as a problem of fault location in combinatorial testing. Specifically, they seek to identify combinations of features that are present in members of a given class but absent or rare in non-members. However, finding a group of $n$ features out of $m$ existing features involves $m$-choose-$n$ combinations. This does not scale to neural networks such as LeNet5 or VGG16, which use a large number of features.

Our results apply to features extracted by a deep net with specific weights. Since deep nets are typically overparameterized and have local optima, it is possible to obtain numerically very different weights depending on the initialization and optimization protocol. We have not explored how this may affect the resulting features, tree and masks. We did observe that our results seem robust to the initialization of the tree itself.

One fair criticism of mimic-based interpretations is that, in learning a mimic, what we are interpreting is the mimic, not the original model (here, the neural net). For example, if the mimic strongly uses feature $i$ to classify into class $k$ (say), then it is tempting to conclude that the original model also does that. This is risky, especially if the mimic is not very accurate with respect to the neural net. However, in our experiments we transferred the insights obtained through the tree mimic to the original neural net (e.g.\ in masking features) and confirmed that they still hold there for most of the training and test instances.

\section{Conclusion}

Our paper demonstrates the use of sparse oblique decision trees as a powerful ``microscope'' to investigate the behavior of deep nets, by learning interpretable yet accurate trees that mimic the classifier part of a deep net. The tree takes as input the neural net ``features'' produced by the neural net for a given input instance (that is, the neural net activations at an internal layer). The tree then predicts the corresponding class for those features, emulating the classification behavior of the neural net with very high accuracy, at least in the neural nets we considered. Using oblique trees trained with the TAO algorithm is critical for this to succeed.

The resulting tree gives insights about the relation between neurons and classes, such as what groups of neurons are involved in predicting what classes. It also enables the design of simple manipulations of the neuron activations that can, for any training or test instance, change the class predicted in various, controllable ways (thus making adversarial attacks possible at the level of the deep net features).

This approach to interpreting or manipulating features applies to other types of deep nets and data, such as audio or language. It may also prove helpful in other areas where deep nets are being applied, such as in biology, where we may be able to relate neurons to genes or diseases, and observe the effect of ``knocking out'' such genes, which is essentially what our proposed masks do.

\section{Acknowledgments}

Work partially supported by NSF award IIS--2007147.

\appendix

\section{Neural network architectures}
\label{s:nn-arch}

Tables~\ref{t:VGG16} and~\ref{t:LeNet5} describe the architectures for our VGG16 network and the LeNet5 network.

\begin{table}[p]
  \begin{minipage}{.52\textwidth}	
    \begin{tabular}{l l}
      \hline
      Layer& Connectivity\\
      \hline
      Input&{64$\times$64$\times$3 Image }\\
      &\\		
      1&{ convolutional, 64 3$\times$3 filters}\\
      &{ (stride=1,padding = 1) }\\
      &{ followed by BatchNormalization  }\\
      &{   $\rightarrow $ ReLU  }\\
      &\\		
      2&{ convolutional, 64 3$\times$3 filters}\\
      &{ (stride=1,padding = 1) }\\
      &{ followed by BatchNormalization  }\\
      &{   $\rightarrow $ ReLU  }\\
      &\\		
      3&{max pool, 2$\times$ 2 window (stride=2)}\\
      &\\
      4&{ convolutional, 128 3$\times$3 filters}\\
      &{ (stride=1,padding = 1) }\\
      &{ followed by BatchNormalization  }\\
      &{   $\rightarrow $ ReLU  }\\
      &\\
      5&{ convolutional, 128 3$\times$3 filters}\\
      &{ (stride=1,padding = 1) }\\
      &{ followed by BatchNormalization  }\\
      &{   $\rightarrow $ ReLU  }\\
      &\\
      6&{max pool, 2$\times$ 2 window (stride=2)}\\
      &\\
      7&{ convolutional, 256 3$\times$3 filters}\\
      &{ (stride=1,padding = 1) }\\
      &{ followed by BatchNormalization  }\\
      &{   $\rightarrow $ ReLU  }\\
      &\\
      8&{ convolutional, 256 3$\times$3 filters}\\
      &{ (stride=1,padding = 1) }\\
      &{ followed by BatchNormalization  }\\
      &{   $\rightarrow $ ReLU  }\\
      &\\
      9&{ convolutional, 256 3$\times$3 filters}\\
      &{ (stride=1,padding = 1) }\\
      &{ followed by BatchNormalization  }\\
      &{   $\rightarrow $ ReLU  }\\
      &\\
      10&{max pool, 2$\times$ 2 window (stride=2)}\\
    \end{tabular}
  \end{minipage}
  \begin{minipage}{.2\textwidth}	
    \begin{tabular}{l l}
      \hline
      Layer& Connectivity\\
      \hline
      &\\
      11&{ convolutional, 512 3$\times$3 filters}\\
      &{ (stride=1,padding = 1) }\\
      &{ followed by BatchNormalization  }\\
      &{   $\rightarrow $ ReLU  }\\
      &\\
      12&{ convolutional, 512 3$\times$3 filters}\\
      &{ (stride=1,padding = 1) }\\
      &{ followed by BatchNormalization  }\\
      &{   $\rightarrow $ ReLU  }\\
      &\\
      13&{ convolutional, 512 3$\times$3 filters}\\
      &{ (stride=1,padding = 1) }\\
      &{ followed by BatchNormalization  }\\
      &{   $\rightarrow $ ReLU  }\\
      &\\
      14&{max pool, 2$\times$ 2 window (stride=2)}\\
      &\\
      15&{ convolutional, 512 3$\times$3 filters}\\
      &{ (stride=1,padding = 1) }\\
      &{ followed by BatchNormalization  }\\
      &{   $\rightarrow $ ReLU  }\\
      &\\
      16&{ convolutional, 512 3$\times$3 filters}\\
      &{ (stride=1,padding = 1) }\\
      &{ followed by BatchNormalization  }\\
      &{   $\rightarrow $ ReLU  }\\
      &\\
      17&{ convolutional, 512 3$\times$3 filters}\\
      &{ (stride=1,padding = 1) }\\
      &{ followed by BatchNormalization  }\\
      &{   $\rightarrow $ ReLU  }\\
      &\\
      18&{Dense Layer 4096 neuron }\\
      &{followed by ReLU $\rightarrow $ Dropout (p=0.6) }\\
      &\\
      19&{Dense Layer 4096 neuron }\\
      &{followed by ReLU $\rightarrow $ Dropout (p=0.6) }\\
      &\\
      20&{Dense Layer 16 neuron }\\
      \\
      21&{softmax}
    \end{tabular}
  \end{minipage}
  \caption{Architecture of our modified VGG16 neural net for our ImageNet subset (64$\times$ 64 image size).}
  \label{t:VGG16}
\end{table}

\begin{table}[p]
  \centering
  \begin{tabular}[c]{@{}lc@{}}
    \hline 
    Block name & Block description \\
    \hline 
    \\
    conv1 &	convolution with kernel size $5\times5$ and $20$ channels \\
    & ReLU \\
    & maxpooling with kernel size $2\times2$\\
    \\
    \hline 
    \\
    conv2 &	convolution with kernel size $5\times5$ and $50$ channels \\
    & ReLU \\
    & maxpooling with kernel size $2\times2$\\
    \\
    \hline 
    \\
    fc1 & Fully connected layer with 500 hidden units\\
    & ReLU\\
    \\
    \hline 
    \\
    dropout & p=0.5 \\
    \\
    \hline 
    \\
    fc2 & Fully connected layer with 10 hidden units \\
    \\
    \hline 
  \end{tabular}
  \caption{LeNet5 architecture.}
  \label{t:LeNet5}
\end{table}

\clearpage

\section{Control experiments}
\label{s:control}

\subsection{CART trees are unsuitable as mimics for the deep net}

\begin{figure}[p]
  \centering
  \includegraphics*[width=\linewidth]{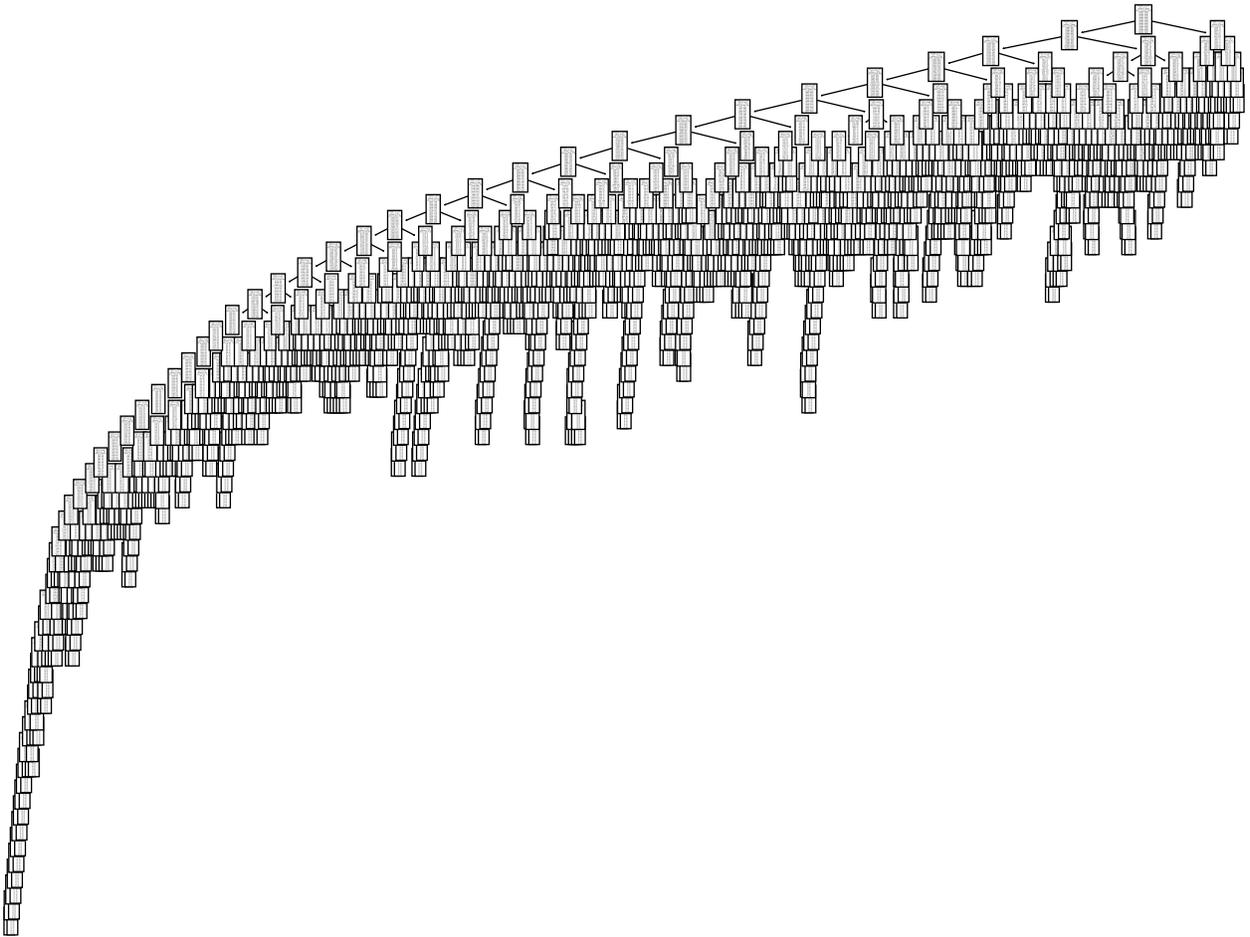}
  \caption{CART axis-aligned tree trained on VGG16 features.}
  \label{f:VGG16-CART}
\end{figure}

Sparse oblique trees trained with TAO have been shown \citep{CarreirTavall18a} to outperform traditional tree learning algorithms by a large margin, in particular axis-aligned trees trained with CART \citep{Breiman_84a,Therneau_19a,Pedreg_11a}. Still, we tried to construct a mimic by using an axis-aligned tree trained with CART. In an axis-aligned tree, each decision node tests a single feature (rather than a linear combination) in order to send an instance down its left or right child. We used the CART implementation of scikit-learn \citep{Pedreg_11a}. As is customary with CART, we first allow the tree to grow in full and then apply cost-complexity pruning, choosing the best pruning hyperparameter by cross-validation. We learned the CART tree on the same dataset of VGG16 features as the TAO tree. Table~\ref{t:VGG16-CART} shows statistics of the resulting tree and figure~\ref{f:VGG16-CART} the tree itself. It is obvious that the tree is both grossly inaccurate in test error (21.2\%) and huge in size (depth 54 and 1\,381 nodes). This makes it unsuitable as a mimic for VGG16 and practically impossible to interpret or to construct masks.

\begin{table}[p]
  \centering 
  \begin{tabular}{@{}l@{\hspace{2ex}}rr@{\hspace{5ex}}rrr@{}}
    \toprule
    & \multicolumn{2}{c}{CART} & \multicolumn{3}{c}{TAO} \\
    \cline{2-3}\cline{4-6}
    & after pruning & before pruning & $\lambda=0.01$ & $\lambda=1$ & $\lambda= 33$ \\
    \midrule
    Training error (\%) & 1.97 & 0.00 & 0.00 & 0.00 &1.79\\
    Test error (\%) & 21.24 & 21.47 & 7.63 & 7.91 &9.56\\
    Features used (out of 8\,192) & 619 & 795 & 4423 & 1366 & 408\\
    Depth & 54  & 57 & 6&6 & 5\\
    Number of nodes & 1381  & 1785 & 51 & 39 & 31\\
    \bottomrule
  \end{tabular}
  \caption{Training a tree on VGG16 features: axis-aligned tree with CART, sparse oblique tree with TAO.}
  \label{t:VGG16-CART}
\end{table}

\subsection{Rotated deep net features do not admit sparsity well}

Our finding that a very small subset of deep net features are sufficient to control classification for a given class is surprising, because it is not strictly necessary for this to happen. That is, each class could be dependent on the collaborative values of all (or nearly all) features. To verify this, we manufactured a version of the features that contains all the information in the original features, but mixes them in a dense way as a linear combination. Specifically, we multiplied the original features \f\ by a dense, invertible matrix \Q, to obtain transformed features $\overline{\f} = \Q \f$. Clearly, both the fully-connected layers of VGG16 or an oblique tree can absorb this transformation, namely by using a new matrix $\overline{\W} = \W \Q^{-1}$ in the first fully-connected layer or by using a new weight vector $\overline{\w}_i = \smash{\Q^{-T}} \w_i$ in each decision node $i$ of the tree, respectively (since then we have $\overline{\W} \overline{\f} = \W \f$ and $\overline{\w}^T_i \overline{\f} = \w^T_i \f$, where \W\ and $\w_i$ were the original matrix or weight vector). However, this need not be true anymore if we force the weight vector $\overline{\w}_i$ to be sparse by using a large enough $\lambda$ value. (The same would be true for the deep net if pruning weights in the fully-connected layers.) Hence, we expected that training a sparse oblique tree on the rotated VGG16 features would fail to produce a tree as sparse as before but with low test error. Indeed this is what happened, as described next.

To generate a rotated version of the VGG16 features, we used a dense orthogonal matrix as \Q\ matrix%
\footnote{This satisfies $\Q^{-1} = \Q^T$ and is easier to handle. We use the default matrix in Matlab's \texttt{gallery('orthog',n,k)} function, which is an $n \times n$ matrix with entries defined as $q_{ij} = \sqrt{\frac{2}{n+1}} \, \sin\left( \frac{\pi i j}{n+1} \right)$.}.
This has the desired effect of mixing all the features in an invertible way. Without sparsity ($\lambda = 0$), the TAO tree achieves identical error to the unrotated features' tree (as predicted theoretically). As we increase $\lambda$ in order to achieve sparsity, the behavior of the tree is very different to that of figure~\ref{f:error}, where the tree size, number of nonzeros and training/test error change continuously with $\lambda$. Instead, increasing the value of $\lambda$ over a wide range (including the values used for our original trees) results in no sparsity at all. But once we reach $\lambda = 45$, the tree changes drastically: it becomes quite sparse but its test error jumps to 11.3\%, much higher than with the unrotated features.

This shows that the features learnt by VGG are special in that they seem to operate in small groups associated with classes, rather than all or most features participating in each class. Since VGG16 could learn mixed rather than sparse features, the reason must be not in the architecture of VGG16 but in the training algorithm and/or objective function.

\begin{table}[p]
  \centering 
  \begin{tabular}{@{}cccc@{}}
    \toprule
    \multicolumn{4}{c}{Original features}\\
    \midrule
    $\lambda$ & \caja{c}{c}{\# features selected \\ by the tree out of 8192} & \caja{c}{c}{Training \\ error(\%)} & \caja{c}{c}{Test \\ error(\%)} \\
    \midrule
    0.01 & 4423 & 0.00 & 7.63 \\
    1 & 1366 & 0.00 & 7.91 \\
    33 & 408 & 1.79 & 9.56 \\
    \bottomrule \\[3ex]
    \toprule
    \multicolumn{4}{c}{Rotated features}\\
    \midrule
    $\lambda$ & \caja{c}{c}{\# features selected \\ by the tree out of 8192} & \caja{c}{c}{Training \\ error(\%)} & \caja{c}{c}{Test \\ error(\%)} \\
    \midrule
    1 & 8192 & 0.00 & 7.91 \\
    45 & 335 & 1.91 & 11.34 \\
    \bottomrule
  \end{tabular}
  \caption{TAO sparse oblique trees trained with the original VGG features and the rotated VGG features.}
  \label{t:rotated}
\end{table}

\clearpage
\subsection{\textcolor{black}{Linear classifier and CART trees are not suitable for explanation}}

\textcolor{black}{Instead of sparse oblique trees, one could use other interpretable models to try to understand the relationship between the classes and the neurons. We show results (on the VGG16 features) using two other models that are widely considered as interpretable: a linear classifier and an axis-aligned tree (trained using CART). As shown next, they do not work nearly as well.}
\begin{description}
  \item[Softmax linear classifier] \textcolor{black}{First, we train a model without $\ell_1$ regularization (so the weights are not sparse). This gives a reasonably good mimic, with a training error of 0.52\% and a test error of 8.04\%. However, all features participate in all classes, which makes the model difficult to interpret. Next, we train a model with $\ell_1$ regularization and tune the latter to achieve a similar sparsity (around 83\%) as our tree (fig.~\ref{f:VGG16-tree1}). The linear classifier achieves a training error of 1.12\% and a test error of 9.84\%, which is worse but not too far from our tree (0\% train and 7.63\% test error). However, we cannot find class-specific neurons in this linear classifier. This is evident from fig.~\ref{f:VGG-masks-test-linear}, where we show the results of the \textsc{All to class $k$} mask created by class-specific neurons from the linear model. We can see that the mask fails for every class, meaning the linear model cannot identify the class-specific neurons.}
  \item[CART axis-aligned tree] \textcolor{black}{We use the tree from fig.~\ref{f:VGG16-CART}, which has a training error of 1.97\% and a test error of 21.34\%. Obviously, this is a bad mimic of the network's classifier, and, as shown in fig.~\ref{f:VGG16-CART}, it is impossible to interpret manually. As fig.~\ref{f:VGG-masks-test-cart} shows, the \textsc{All to class $k$} mask also fails.}
\end{description}

\begin{figure}[p]
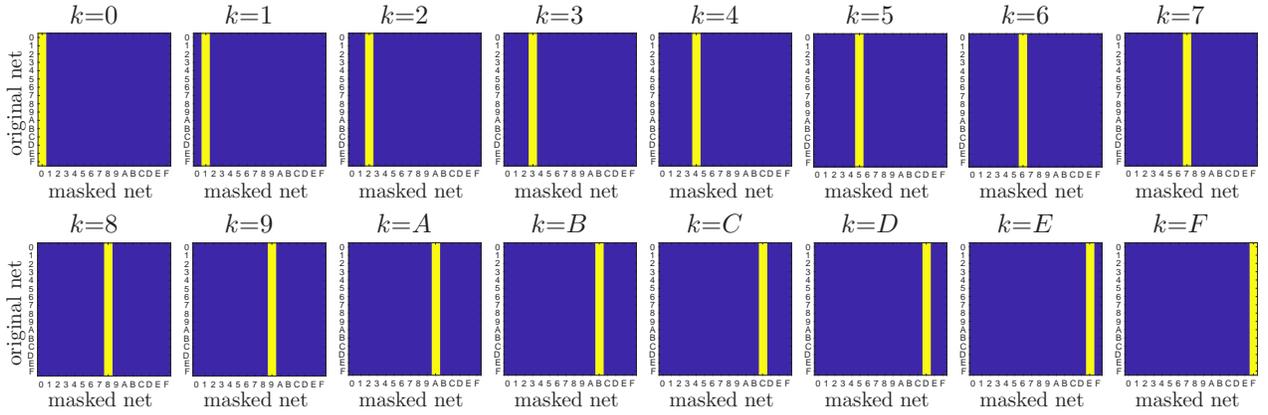

  \centering
  \begin{tabular}{@{}c@{}c@{}c@{}c@{}c@{}c@{}c@{}c@{}}
    $k$$=$$0$ & $k$$=$$1$ & $k$$=$$2$ & $k$$=$$3$ & $k$$=$$4$ & $k$$=$$5$ & $k$$=$$6$ & $k$$=$$7$\\
    \psfrag{Original net}[c][t][.8]{original net}
    \psfrag{Modified net}[][b][.8]{masked net}%{\caja[0.5]{c}{c}{masked \\ net}}
    \includegraphics*[width=.125\textwidth]{VGG/test/modi_train_0_classhot.eps}&
    \psfrag{Original net}{}
    \psfrag{Modified net}[][b][.8]{masked net}%{\caja[0.5]{c}{c}{masked \\ net}}
    \includegraphics*[width=.125\textwidth]{VGG/test/modi_train_1_classhot.eps}&
    \psfrag{Original net}{}
    \psfrag{Modified net}[][b][.8]{masked net}%{\caja[0.5]{c}{c}{masked \\ net}}
    \includegraphics*[width=.125\textwidth]{VGG/test/modi_train_2_classhot.eps}&
    \psfrag{Original net}{}
    \psfrag{Modified net}[][b][.8]{masked net}%{\caja[0.5]{c}{c}{masked \\ net}}
    \includegraphics*[width=.125\textwidth]{VGG/test/modi_train_3_classhot.eps}&
    \psfrag{Original net}{}
    \psfrag{Modified net}[][b][.8]{masked net}%{\caja[0.5]{c}{c}{masked \\ net}}
    \includegraphics*[width=.125\textwidth]{VGG/test/modi_train_4_classhot.eps}&
    \psfrag{Original net}{}
    \psfrag{Modified net}[][b][.8]{masked net}%{\caja[0.5]{c}{c}{masked \\ net}}
    \includegraphics*[width=.125\textwidth]{VGG/test/modi_train_5_classhot.eps}&
    \psfrag{Original net}{}
    \psfrag{Modified net}[][b][.8]{masked net}%{\caja[0.5]{c}{c}{masked \\ net}}
    \includegraphics*[width=.125\textwidth]{VGG/test/modi_train_6_classhot.eps}&
    \psfrag{Original net}{}
    \psfrag{Modified net}[][b][.8]{masked net}%{\caja[0.5]{c}{c}{masked \\ net}}
    \includegraphics*[width=.125\textwidth]{VGG/test/modi_train_7_classhot.eps}\\ [1ex]
    $k$$=$$8$ & $k$$=$$9$ & $k$$=$$A$ & 	$k$$=$$B$ & $k$$=$$C$ & $k$$=$$D$ & $k$$=$$E$ & $k$$=$$F$ \\
    \psfrag{Original net}[c][t][.8]{original net}
    \psfrag{Modified net}[][b][.8]{masked net}%{\caja[0.5]{c}{c}{masked \\ net}}
    \includegraphics*[width=.125\textwidth]{VGG/test/modi_train_8_classhot.eps}&
    \psfrag{Original net}{}
    \psfrag{Modified net}[][b][.8]{masked net}%{\caja[0.5]{c}{c}{masked \\ net}}
    \includegraphics*[width=.125\textwidth]{VGG/test/modi_train_9_classhot.eps}&
    \psfrag{Original net}{}
    \psfrag{Modified net}[][b][.8]{masked net}%{\caja[0.5]{c}{c}{masked \\ net}}
    \includegraphics*[width=.125\textwidth]{VGG/test/modi_train_10_classhot.eps}&
    \psfrag{Original net}{}
    \psfrag{Modified net}[][b][.8]{masked net}%{\caja[0.5]{c}{c}{masked \\ net}}
    \includegraphics*[width=.125\textwidth]{VGG/test/modi_train_11_classhot.eps}&
    \psfrag{Original net}{}
    \psfrag{Modified net}[][b][.8]{masked net}%{\caja[0.5]{c}{c}{masked \\ net}}
    \includegraphics*[width=.125\textwidth]{VGG/test/modi_train_12_classhot.eps}&
    \psfrag{Original net}{}
    \psfrag{Modified net}[][b][.8]{masked net}%{\caja[0.5]{c}{c}{masked \\ net}}
    \includegraphics*[width=.125\textwidth]{VGG/test/modi_train_13_classhot.eps}&
    \psfrag{Original net}{}
    \psfrag{Modified net}[][b][.8]{masked net}%{\caja[0.5]{c}{c}{masked \\ net}}
    \includegraphics*[width=.125\textwidth]{VGG/test/modi_train_14_classhot.eps}&
    \psfrag{Original net}{}
    \psfrag{Modified net}[][b][.8]{masked net}%{\caja[0.5]{c}{c}{masked \\ net}}
    \includegraphics*[width=.125\textwidth]{VGG/test/modi_train_15_classhot.eps}
  \end{tabular}
  \caption{Confusion matrices for VGG (test set) using the \textsc{All to class} $k$ mask created with our sparse oblique trees.}
  \label{f:VGG-masks-test-comp}
\end{figure}

\begin{figure}[p]
  \centering
  \begin{tabular}{@{}c@{}c@{}c@{}c@{}c@{}c@{}c@{}c@{}}
    $k$$=$$0$ & $k$$=$$1$ & $k$$=$$2$ & $k$$=$$3$ & $k$$=$$4$ & $k$$=$$5$ & $k$$=$$6$ & $k$$=$$7$\\
    \psfrag{Original net}[c][t][.8]{original net}
    \psfrag{Modified net}[][b][.8]{masked net}%{\caja[0.5]{c}{c}{masked \\ net}}
    \includegraphics*[width=.125\textwidth]{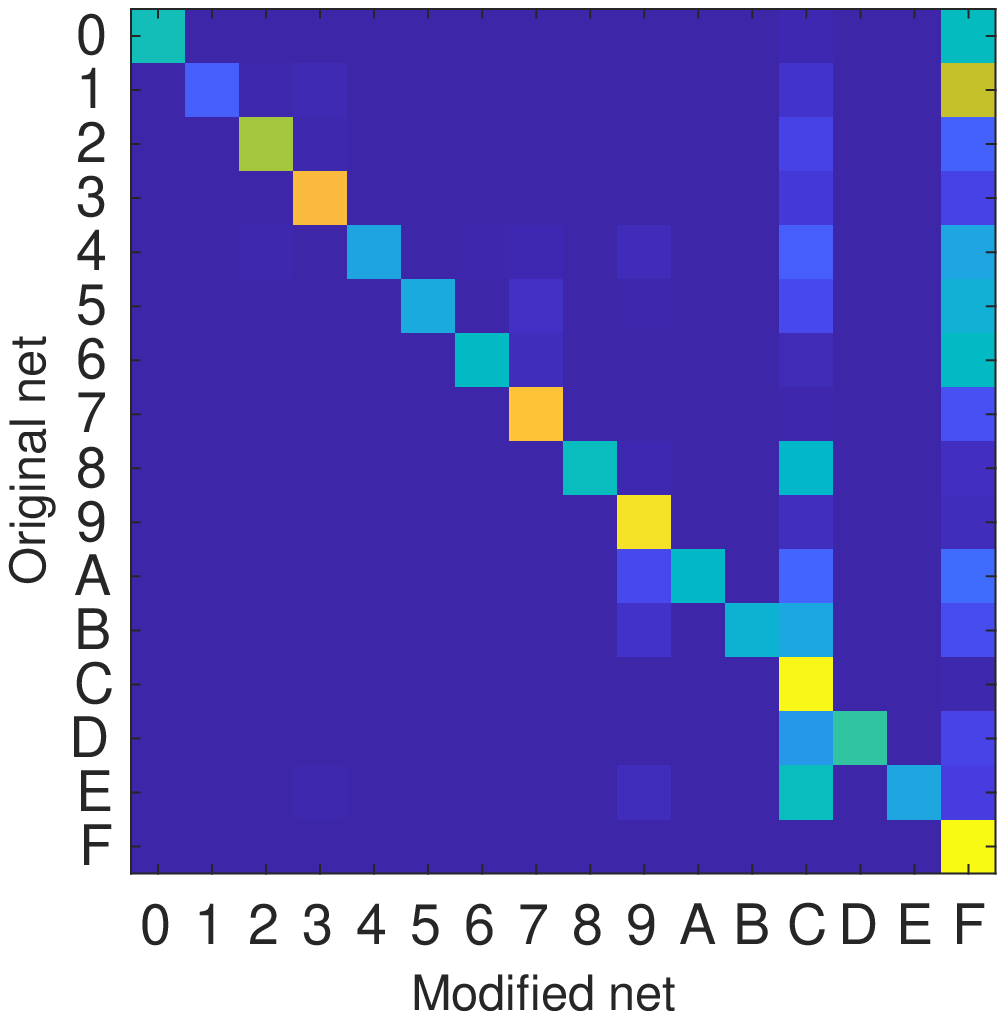}&
    \psfrag{Original net}{}
    \psfrag{Modified net}[][b][.8]{masked net}%{\caja[0.5]{c}{c}{masked \\ net}}
    \includegraphics*[width=.125\textwidth]{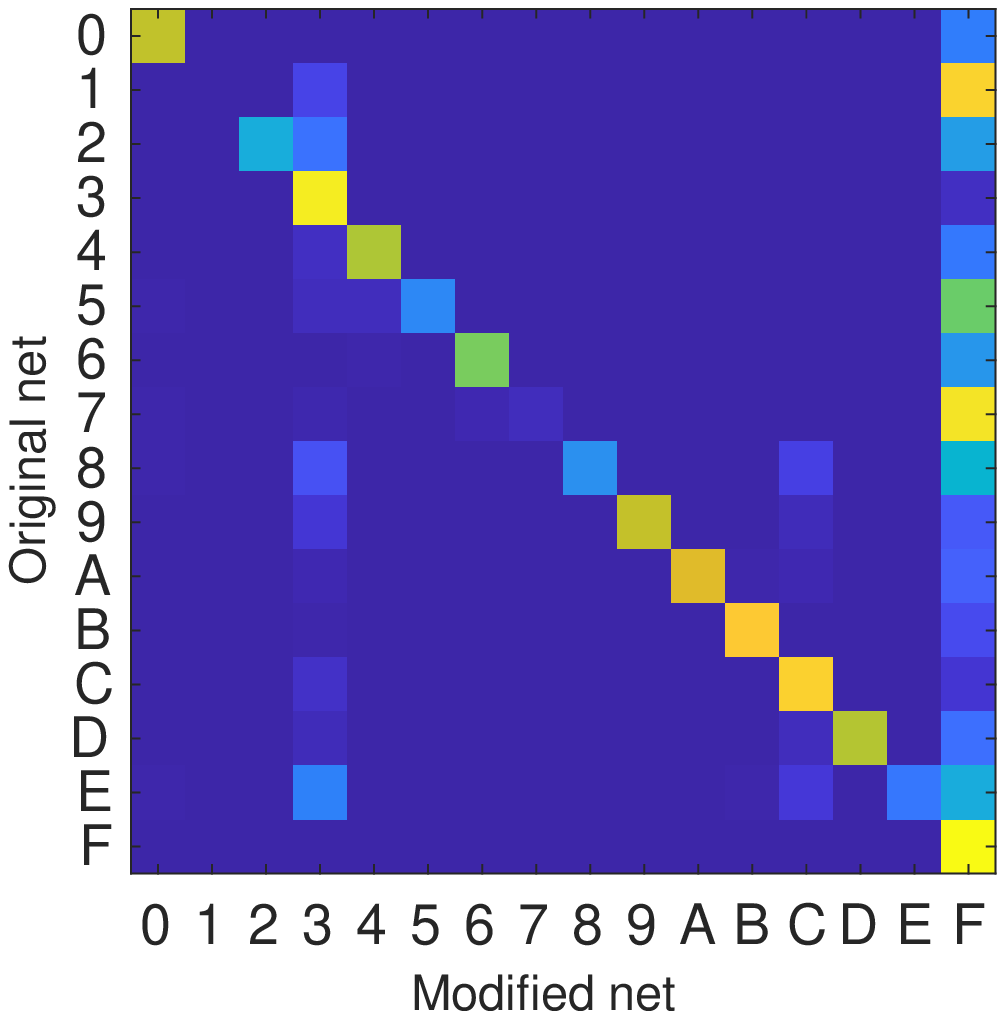}&
    \psfrag{Original net}{}
    \psfrag{Modified net}[][b][.8]{masked net}%{\caja[0.5]{c}{c}{masked \\ net}}
    \includegraphics*[width=.125\textwidth]{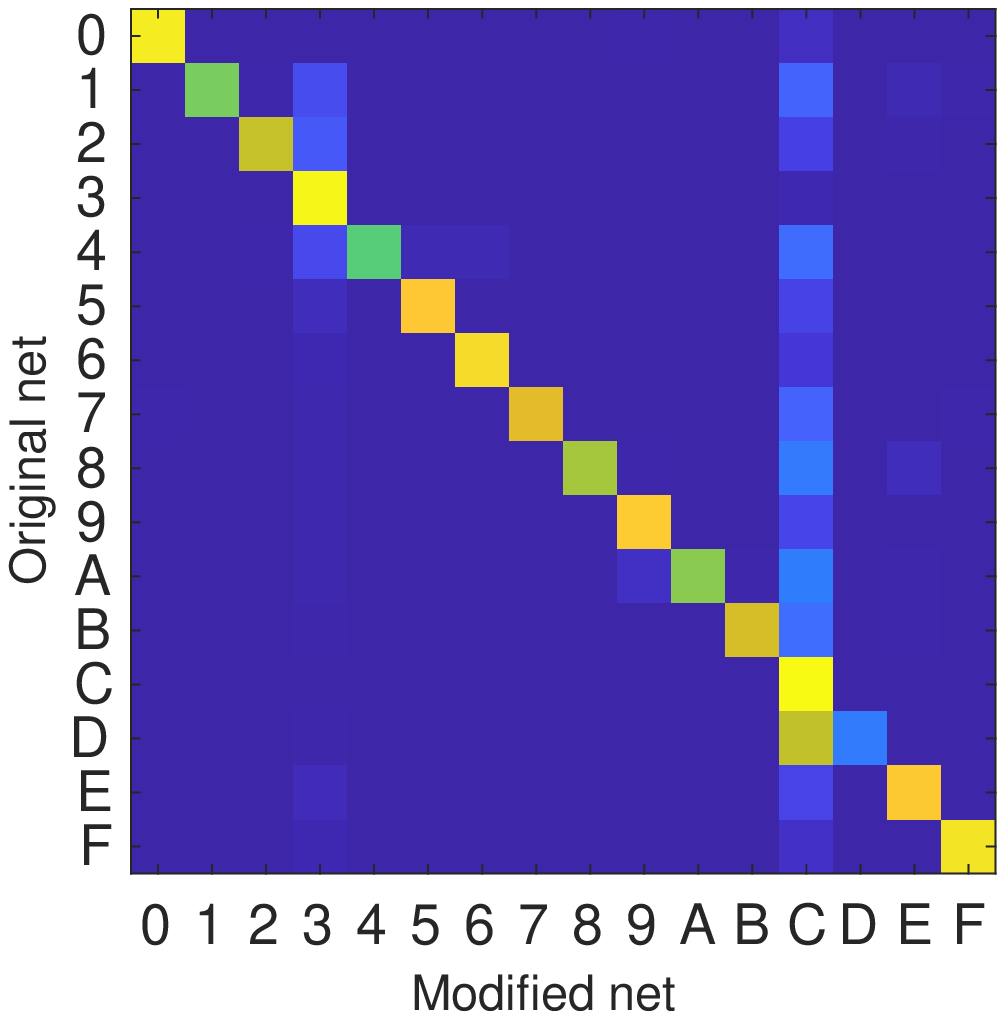}&
    \psfrag{Original net}{}
    \psfrag{Modified net}[][b][.8]{masked net}%{\caja[0.5]{c}{c}{masked \\ net}}
    \includegraphics*[width=.125\textwidth]{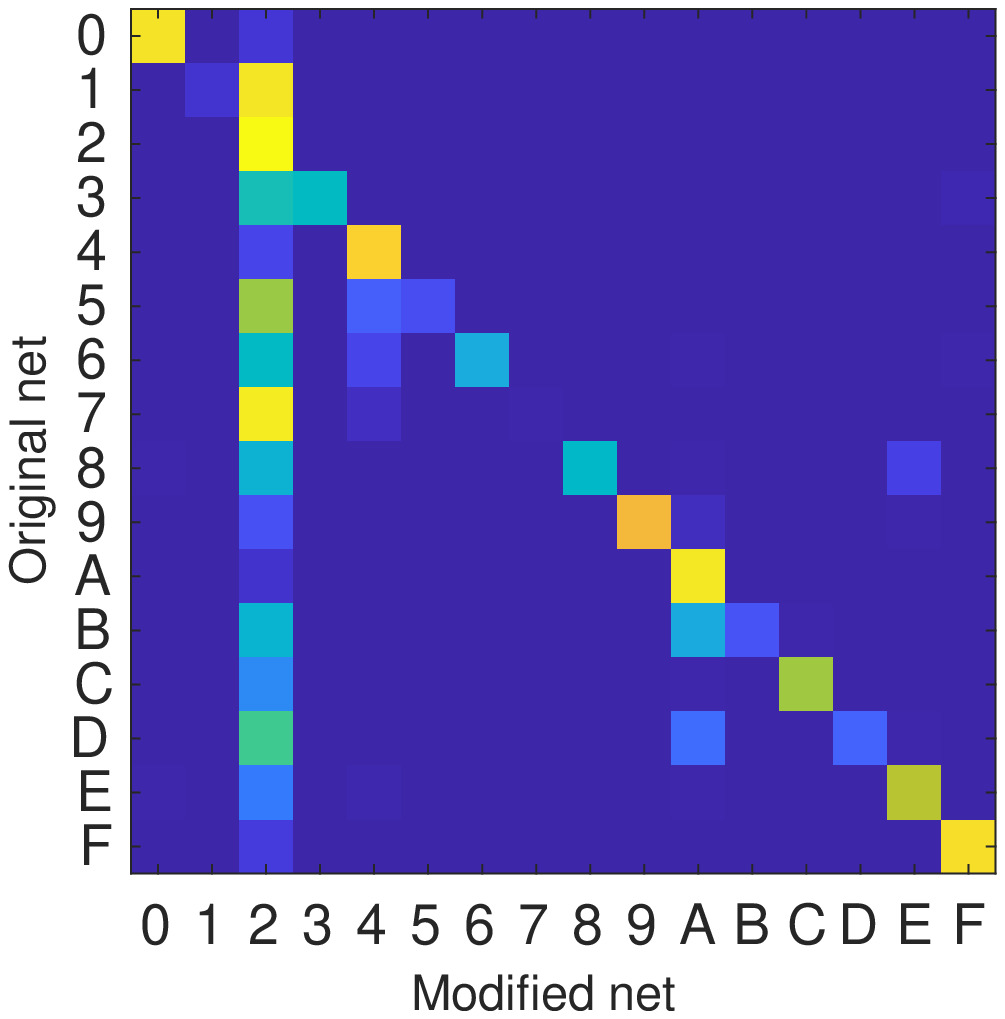}&
    \psfrag{Original net}{}
    \psfrag{Modified net}[][b][.8]{masked net}%{\caja[0.5]{c}{c}{masked \\ net}}
    \includegraphics*[width=.125\textwidth]{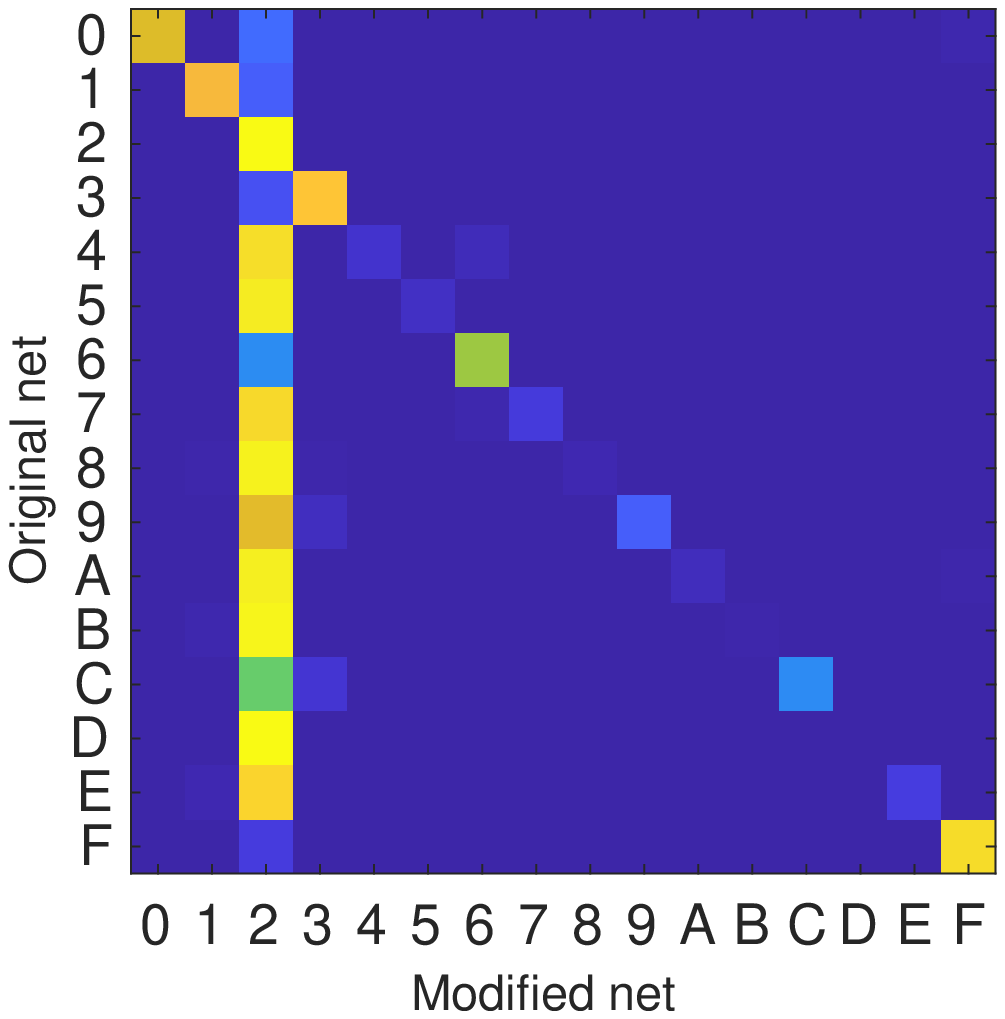}&
    \psfrag{Original net}{}
    \psfrag{Modified net}[][b][.8]{masked net}%{\caja[0.5]{c}{c}{masked \\ net}}
    \includegraphics*[width=.125\textwidth]{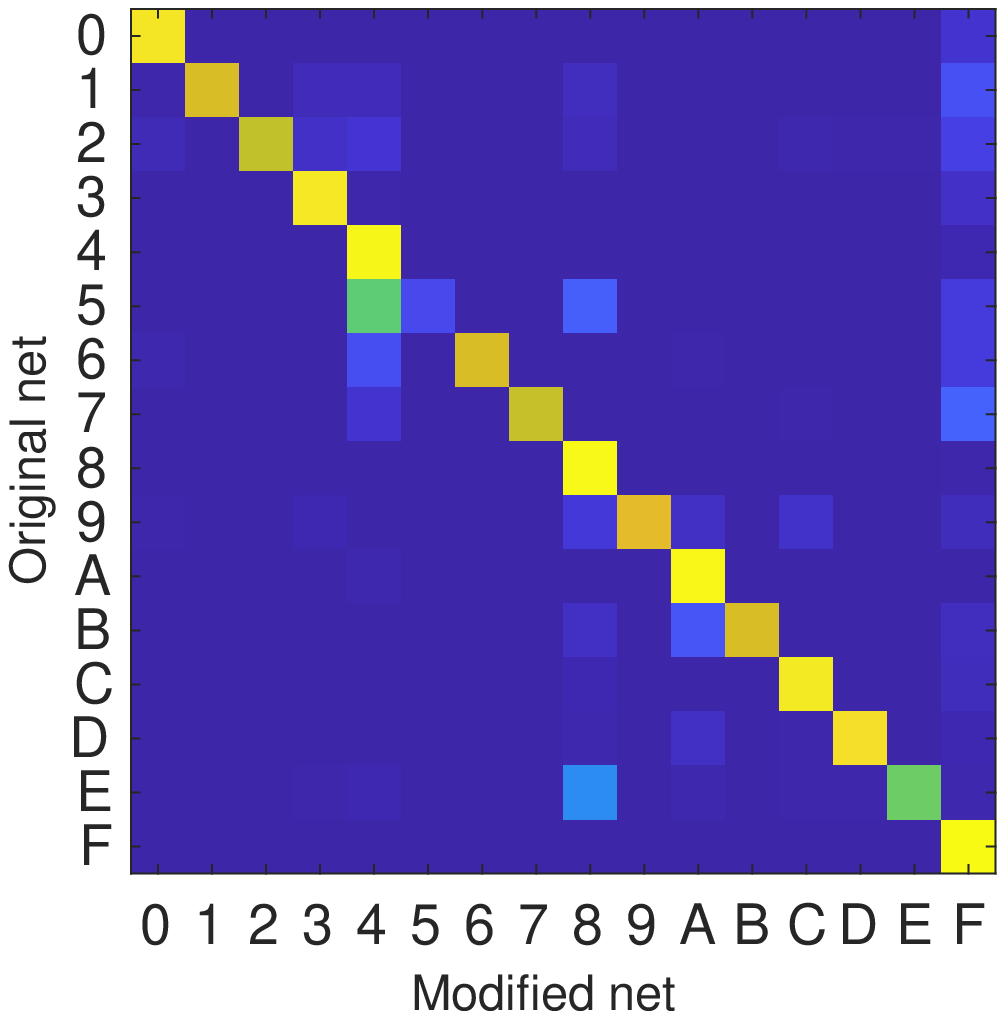}&
    \psfrag{Original net}{}
    \psfrag{Modified net}[][b][.8]{masked net}%{\caja[0.5]{c}{c}{masked \\ net}}
    \includegraphics*[width=.125\textwidth]{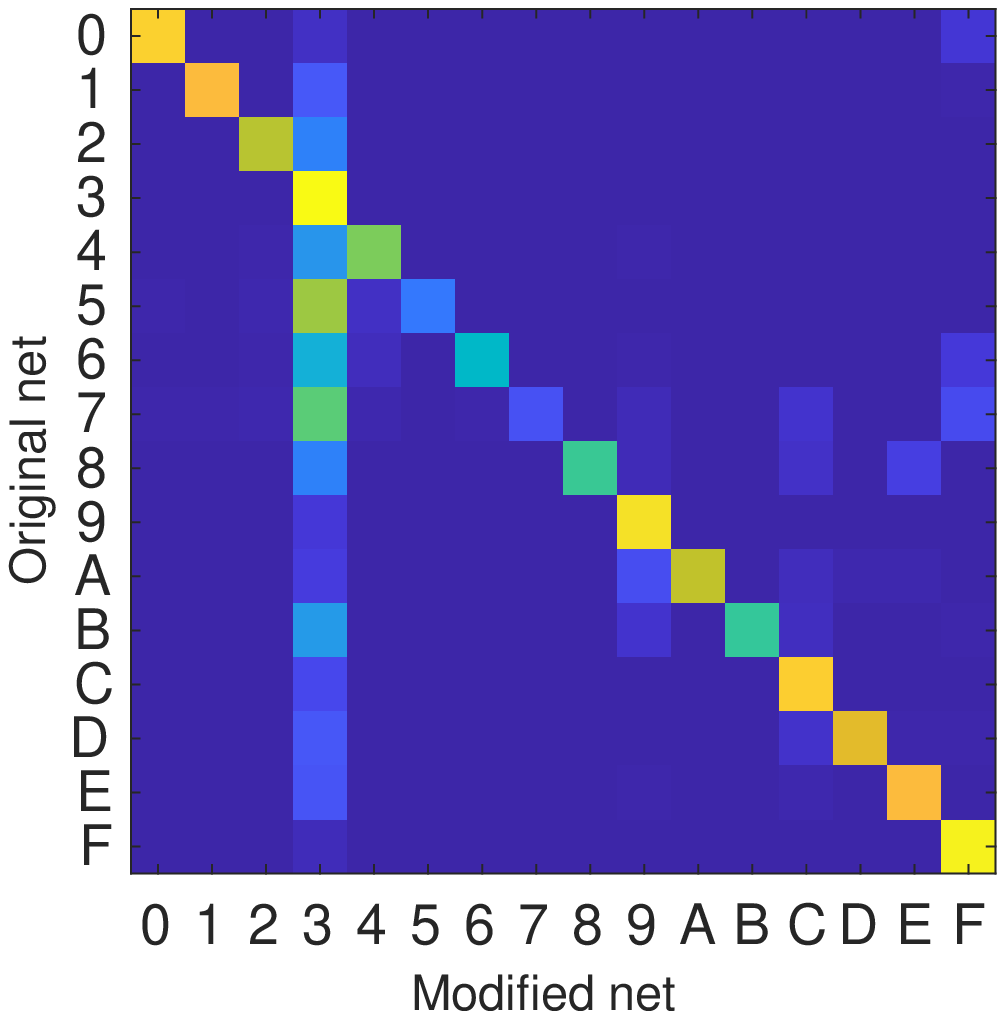}&
    \psfrag{Original net}{}
    \psfrag{Modified net}[][b][.8]{masked net}%{\caja[0.5]{c}{c}{masked \\ net}}
    \includegraphics*[width=.125\textwidth]{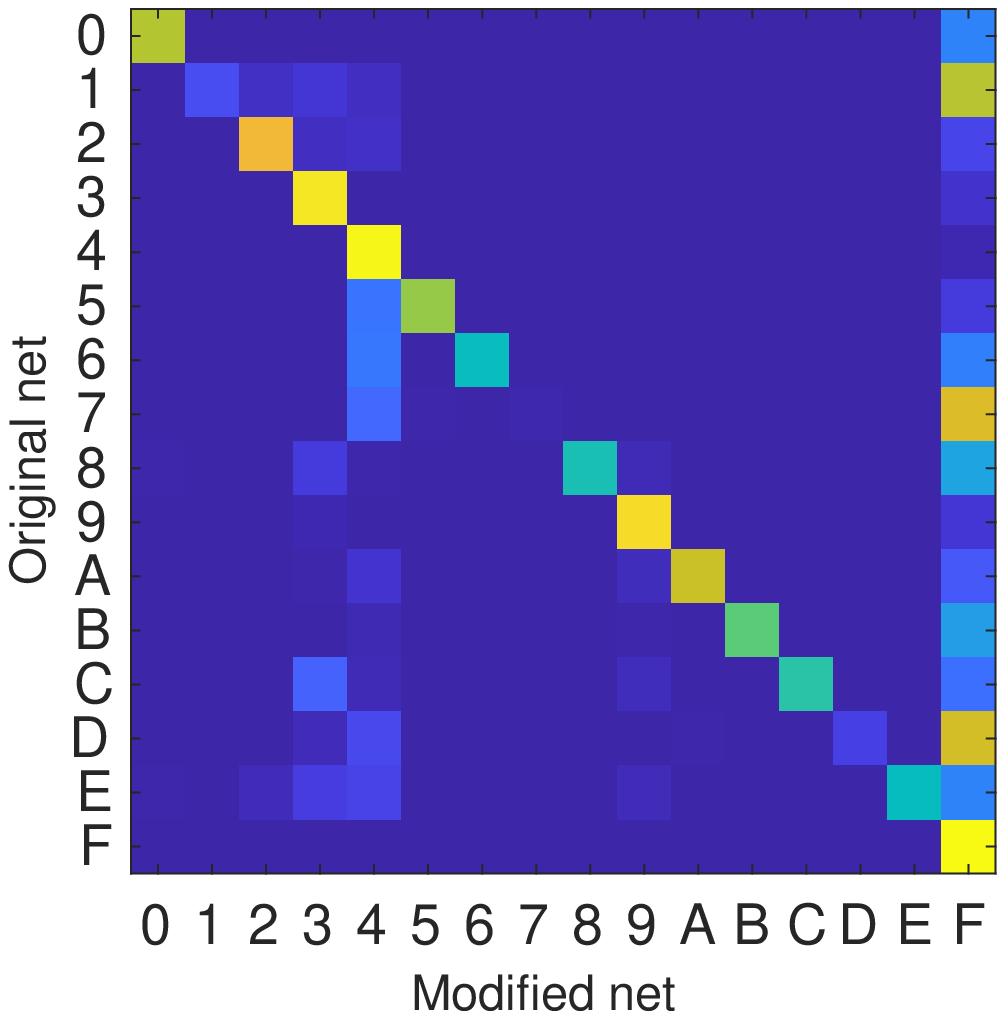}\\ [1ex]
    $k$$=$$8$ & $k$$=$$9$ & $k$$=$$A$ & 	$k$$=$$B$ & $k$$=$$C$ & $k$$=$$D$ & $k$$=$$E$ & $k$$=$$F$ \\
    \psfrag{Original net}[c][t][.8]{original net}
    \psfrag{Modified net}[][b][.8]{masked net}%{\caja[0.5]{c}{c}{masked \\ net}}
    \includegraphics*[width=.125\textwidth]{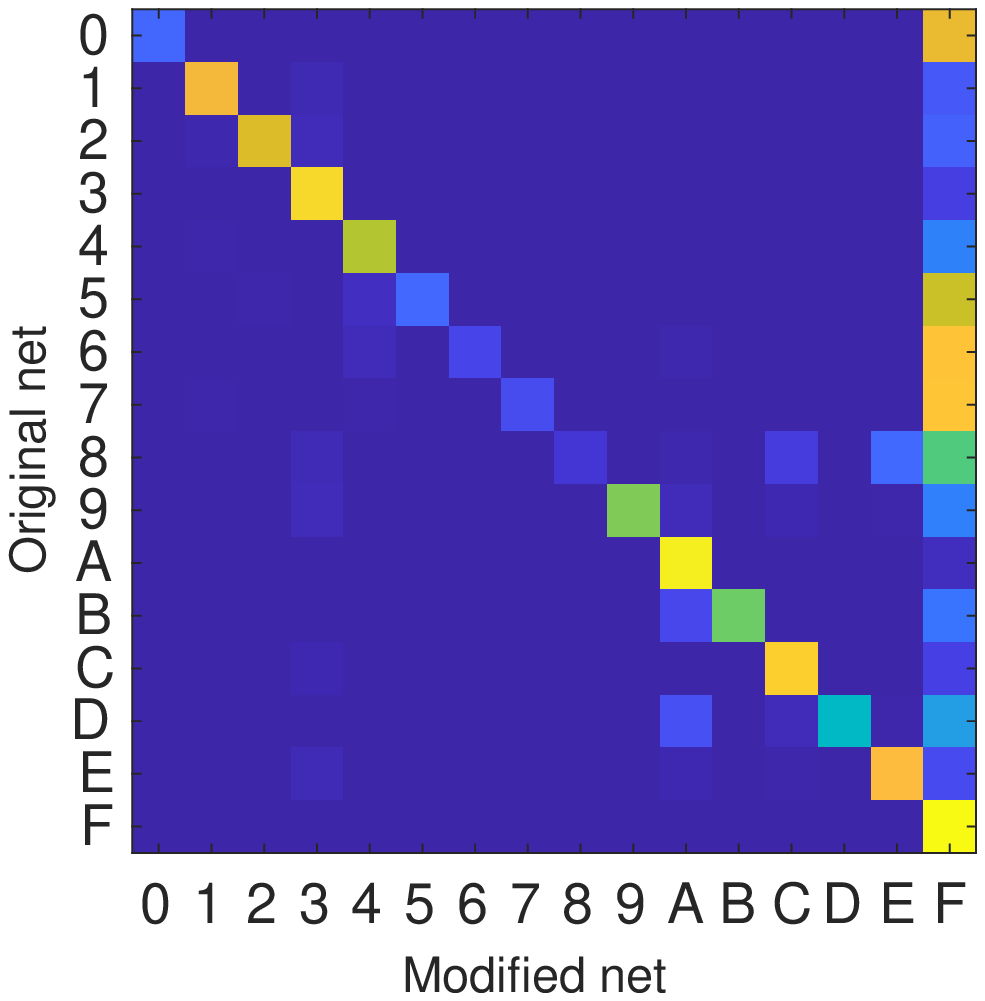}&
    \psfrag{Original net}{}
    \psfrag{Modified net}[][b][.8]{masked net}%{\caja[0.5]{c}{c}{masked \\ net}}
    \includegraphics*[width=.125\textwidth]{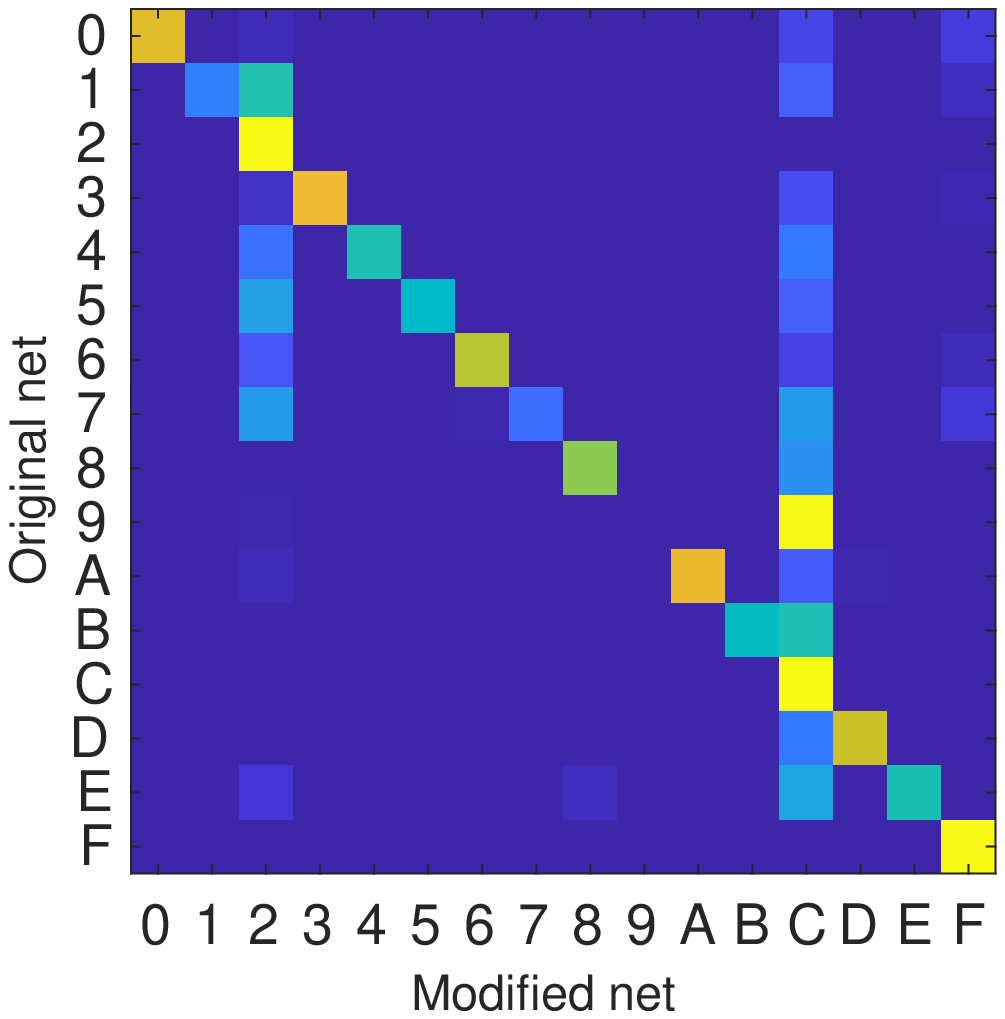}&
    \psfrag{Original net}{}
    \psfrag{Modified net}[][b][.8]{masked net}%{\caja[0.5]{c}{c}{masked \\ net}}
    \includegraphics*[width=.125\textwidth]{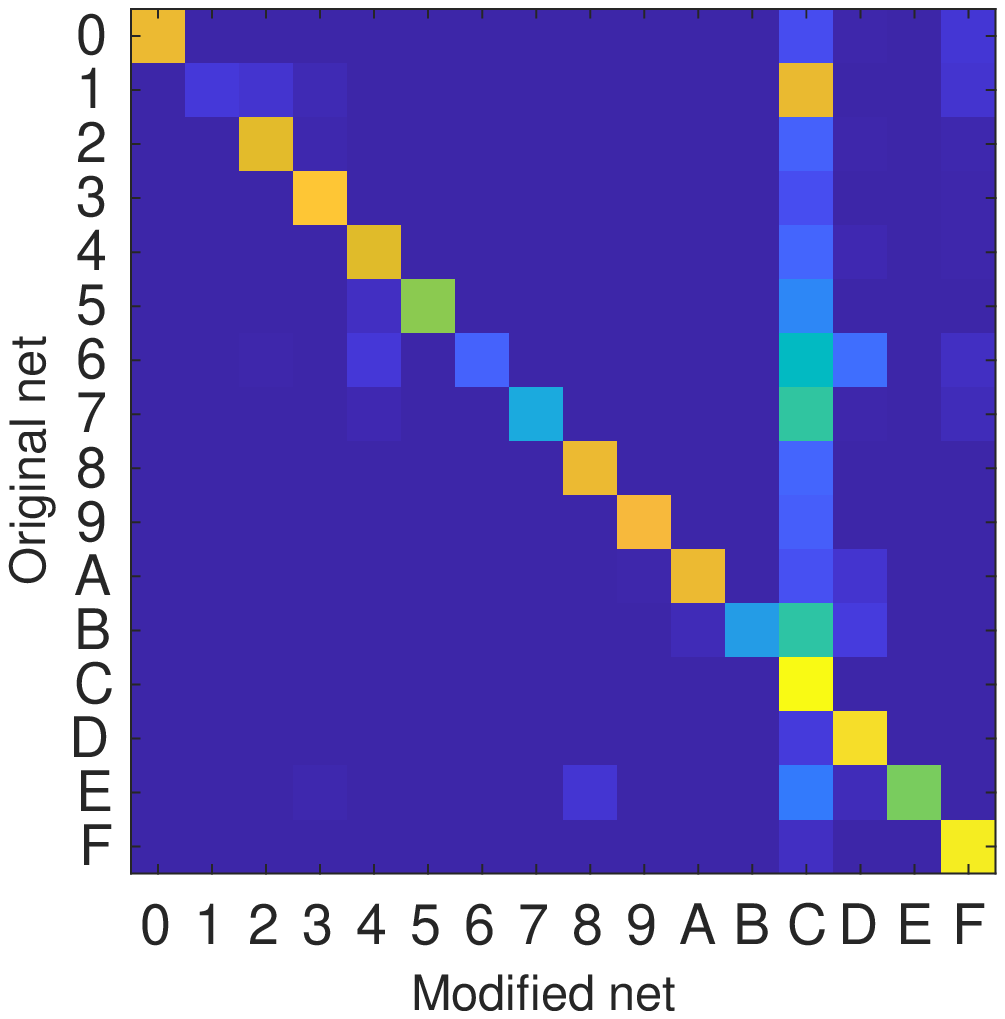}&
    \psfrag{Original net}{}
    \psfrag{Modified net}[][b][.8]{masked net}%{\caja[0.5]{c}{c}{masked \\ net}}
    \includegraphics*[width=.125\textwidth]{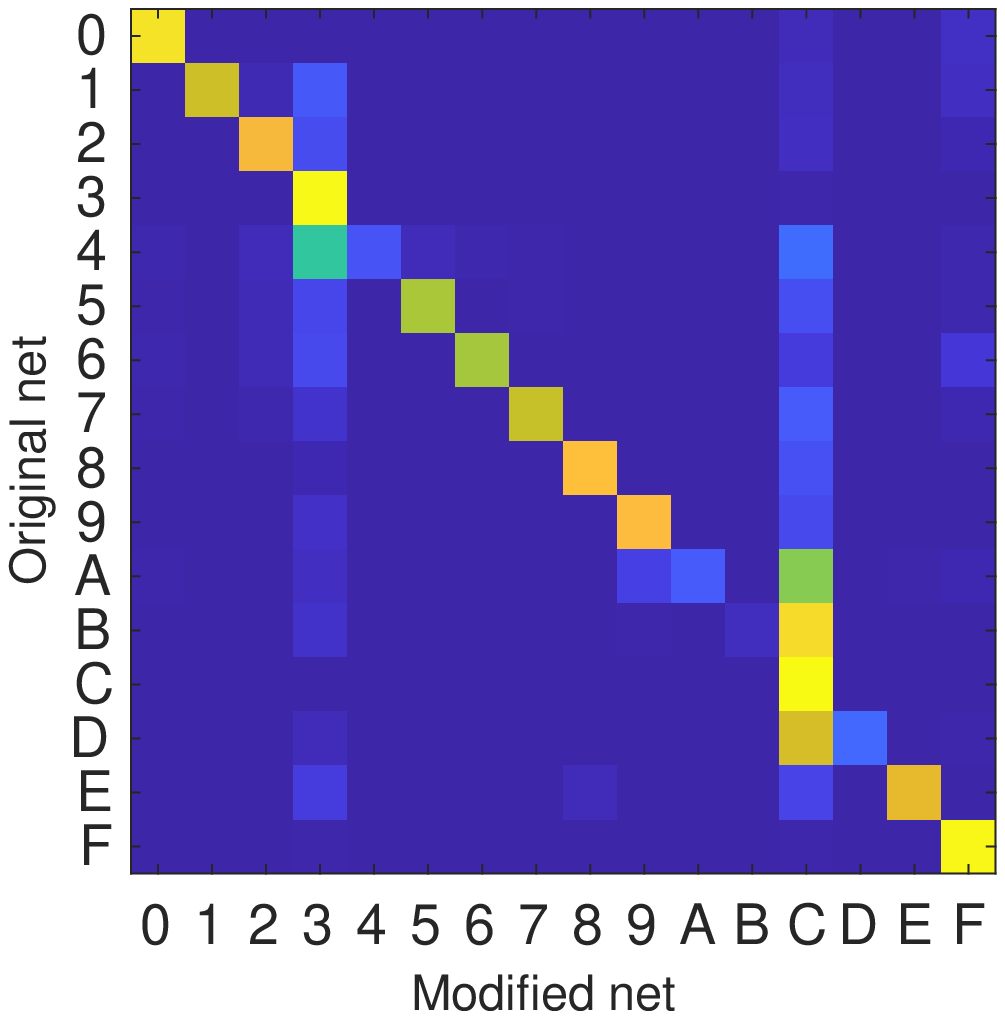}&
    \psfrag{Original net}{}
    \psfrag{Modified net}[][b][.8]{masked net}%{\caja[0.5]{c}{c}{masked \\ net}}
    \includegraphics*[width=.125\textwidth]{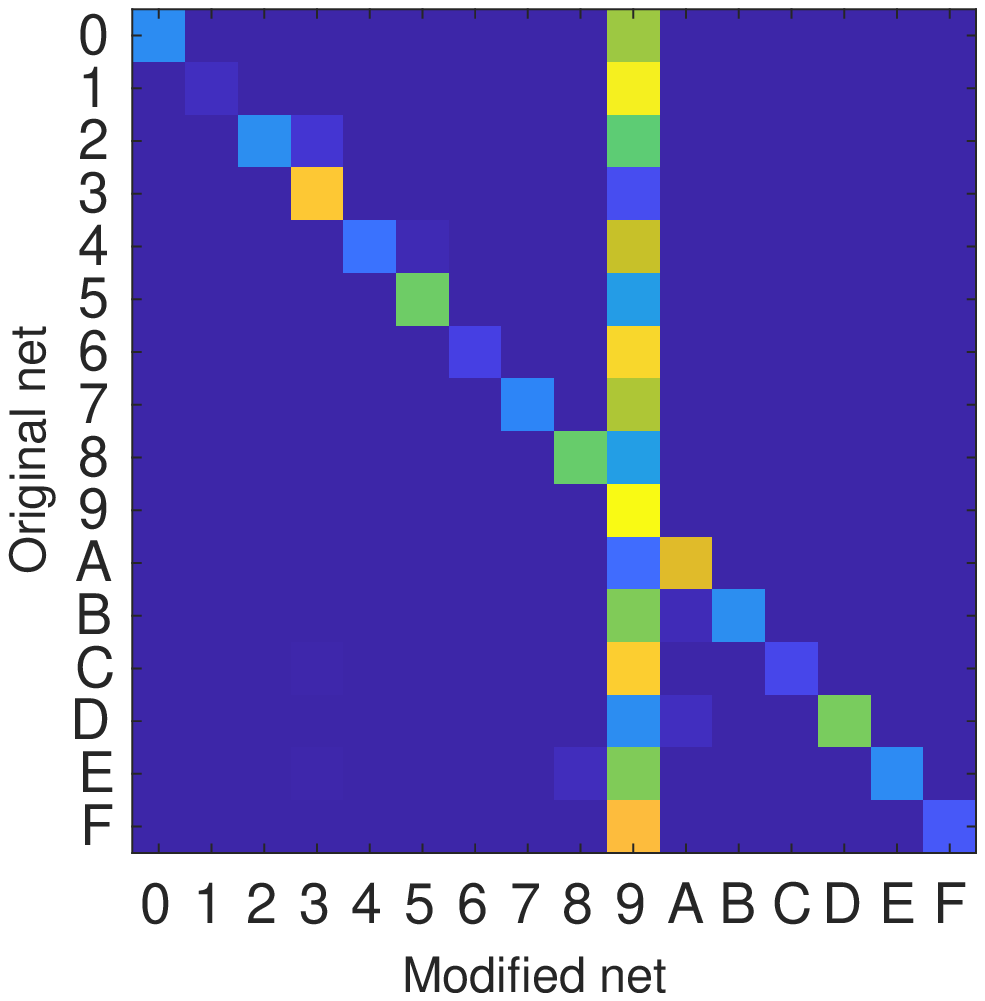}&
    \psfrag{Original net}{}
    \psfrag{Modified net}[][b][.8]{masked net}%{\caja[0.5]{c}{c}{masked \\ net}}
    \includegraphics*[width=.125\textwidth]{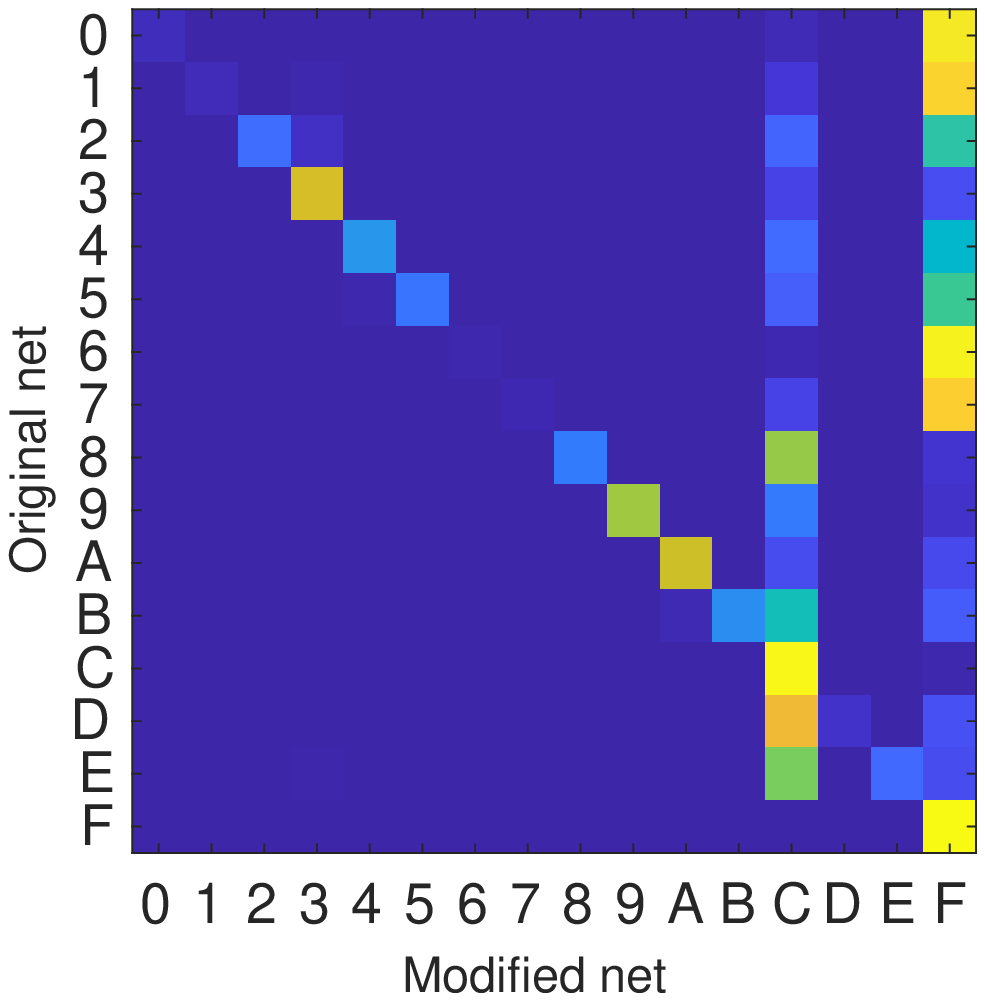}&
    \psfrag{Original net}{}
    \psfrag{Modified net}[][b][.8]{masked net}%{\caja[0.5]{c}{c}{masked \\ net}}
    \includegraphics*[width=.125\textwidth]{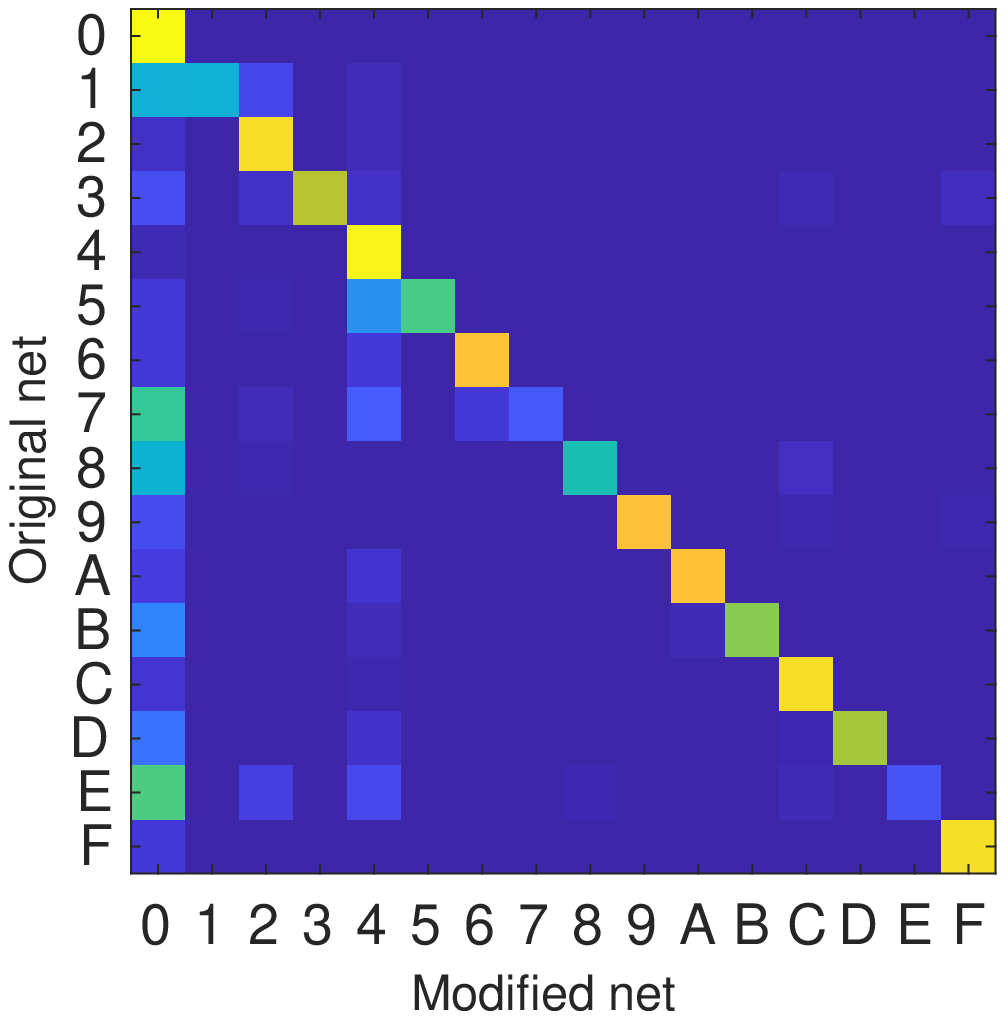}&
    \psfrag{Original net}{}
    \psfrag{Modified net}[][b][.8]{masked net}%{\caja[0.5]{c}{c}{masked \\ net}}
    \includegraphics*[width=.125\textwidth]{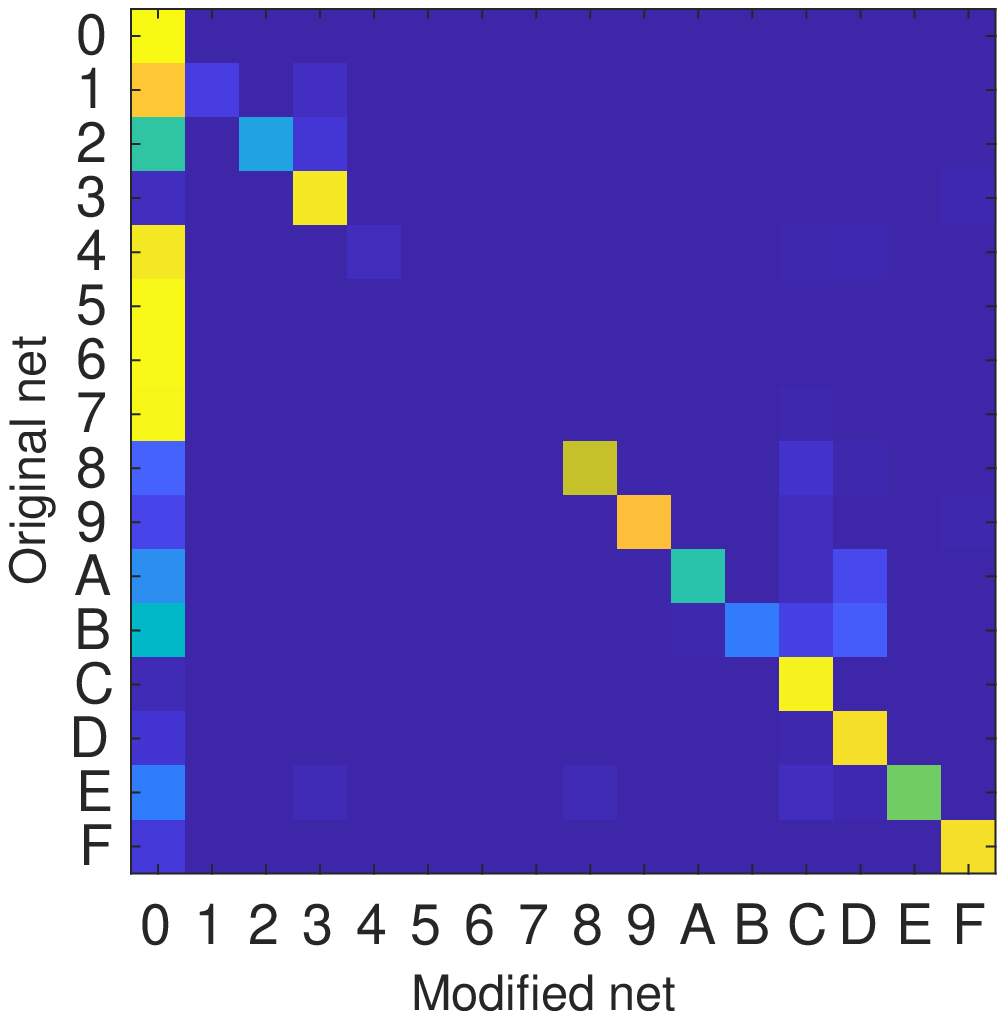}
  \end{tabular}
   \caption{Like fig.~\ref{f:VGG-masks-test-comp} but using a sparse linear classifier.}
  \label{f:VGG-masks-test-linear}
\end{figure}

\begin{figure}[p]
  \centering
  \begin{tabular}{@{}c@{}c@{}c@{}c@{}c@{}c@{}c@{}c@{}}
    $k$$=$$0$ & $k$$=$$1$ & $k$$=$$2$ & $k$$=$$3$ & $k$$=$$4$ & $k$$=$$5$ & $k$$=$$6$ & $k$$=$$7$\\
    \psfrag{Original net}[c][t][.8]{original net}
    \psfrag{Modified net}[][b][.8]{masked net}%{\caja[0.5]{c}{c}{masked \\ net}}
    \includegraphics*[width=.125\textwidth]{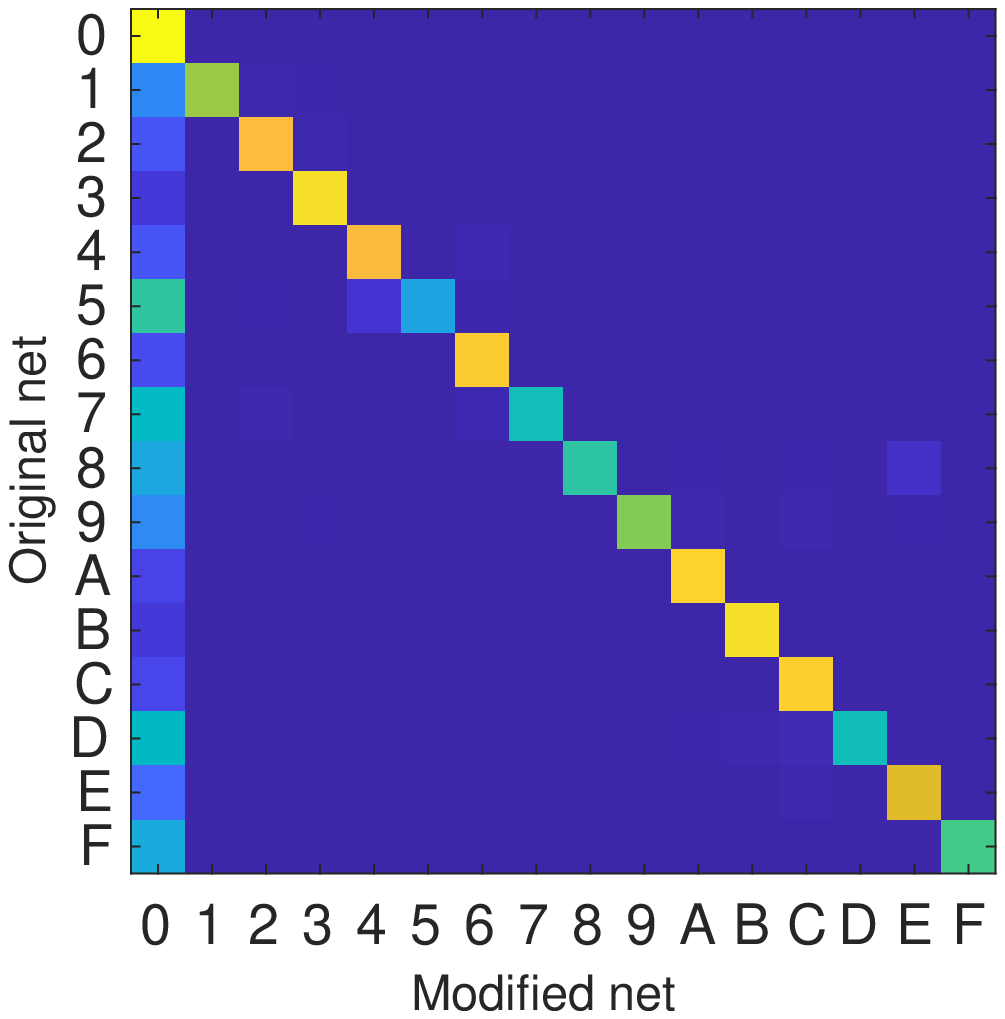}&
    \psfrag{Original net}{}
    \psfrag{Modified net}[][b][.8]{masked net}%{\caja[0.5]{c}{c}{masked \\ net}}
    \includegraphics*[width=.125\textwidth]{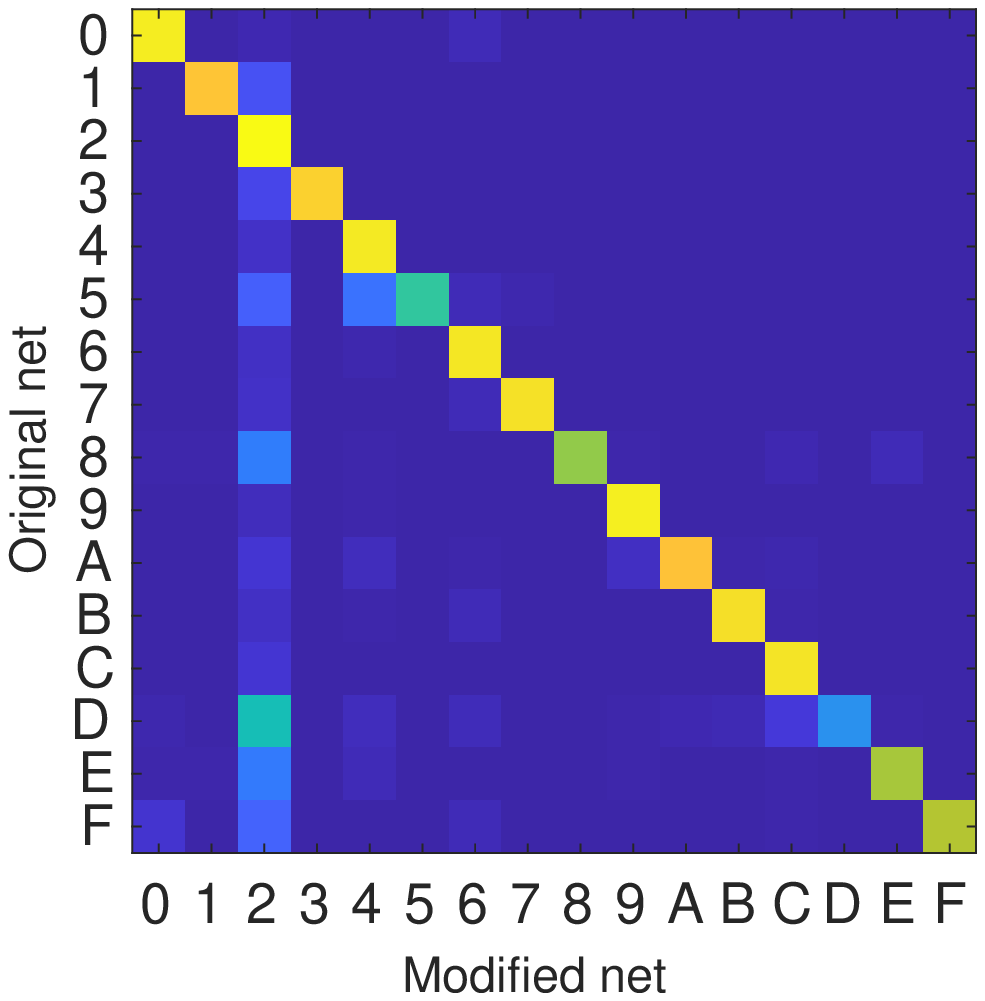}&
    \psfrag{Original net}{}
    \psfrag{Modified net}[][b][.8]{masked net}%{\caja[0.5]{c}{c}{masked \\ net}}
    \includegraphics*[width=.125\textwidth]{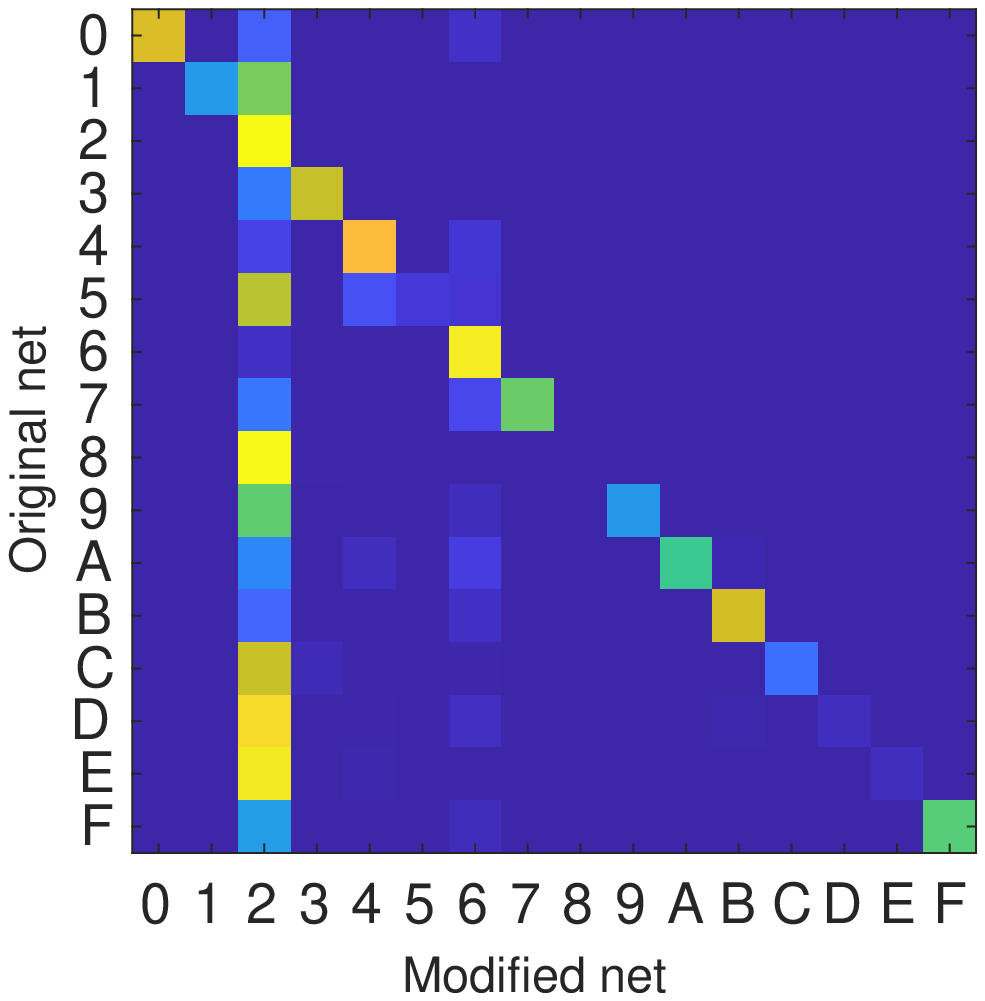}&
    \psfrag{Original net}{}
    \psfrag{Modified net}[][b][.8]{masked net}%{\caja[0.5]{c}{c}{masked \\ net}}
    \includegraphics*[width=.125\textwidth]{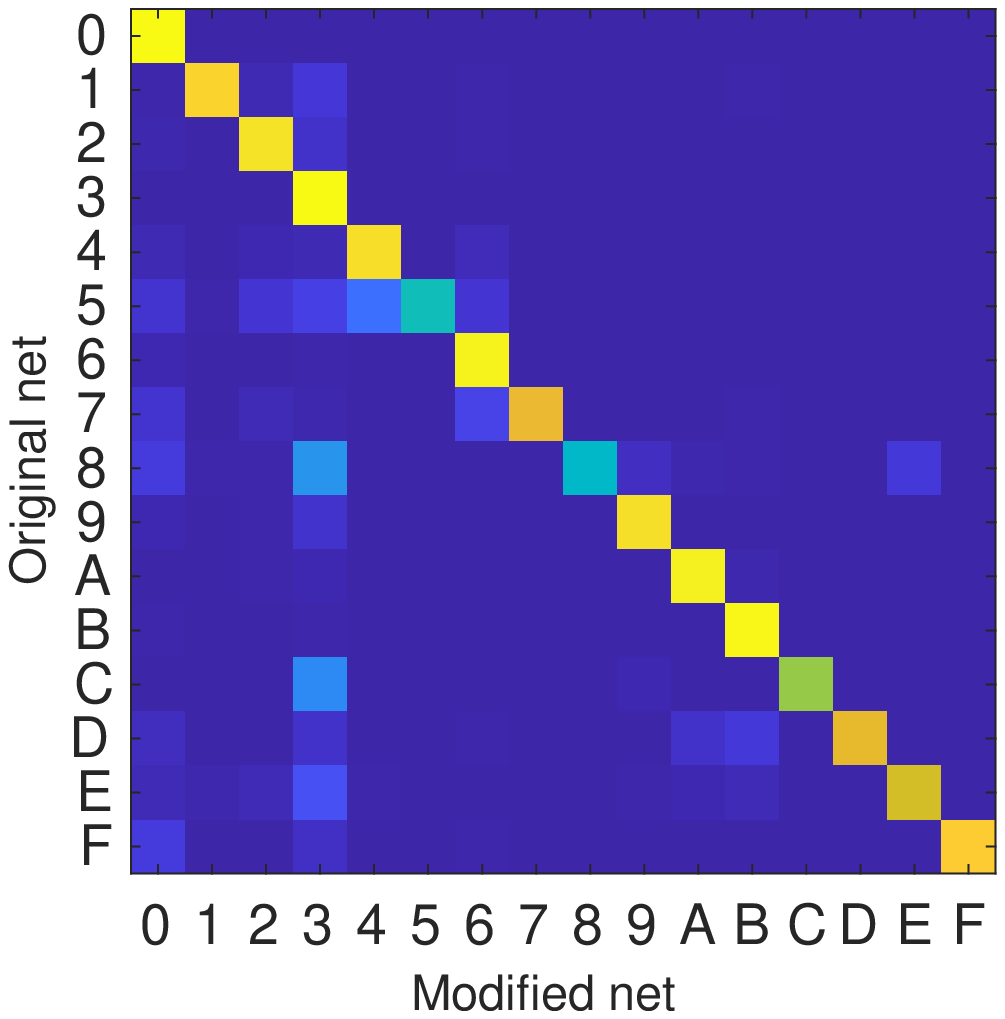}&
    \psfrag{Original net}{}
    \psfrag{Modified net}[][b][.8]{masked net}%{\caja[0.5]{c}{c}{masked \\ net}}
    \includegraphics*[width=.125\textwidth]{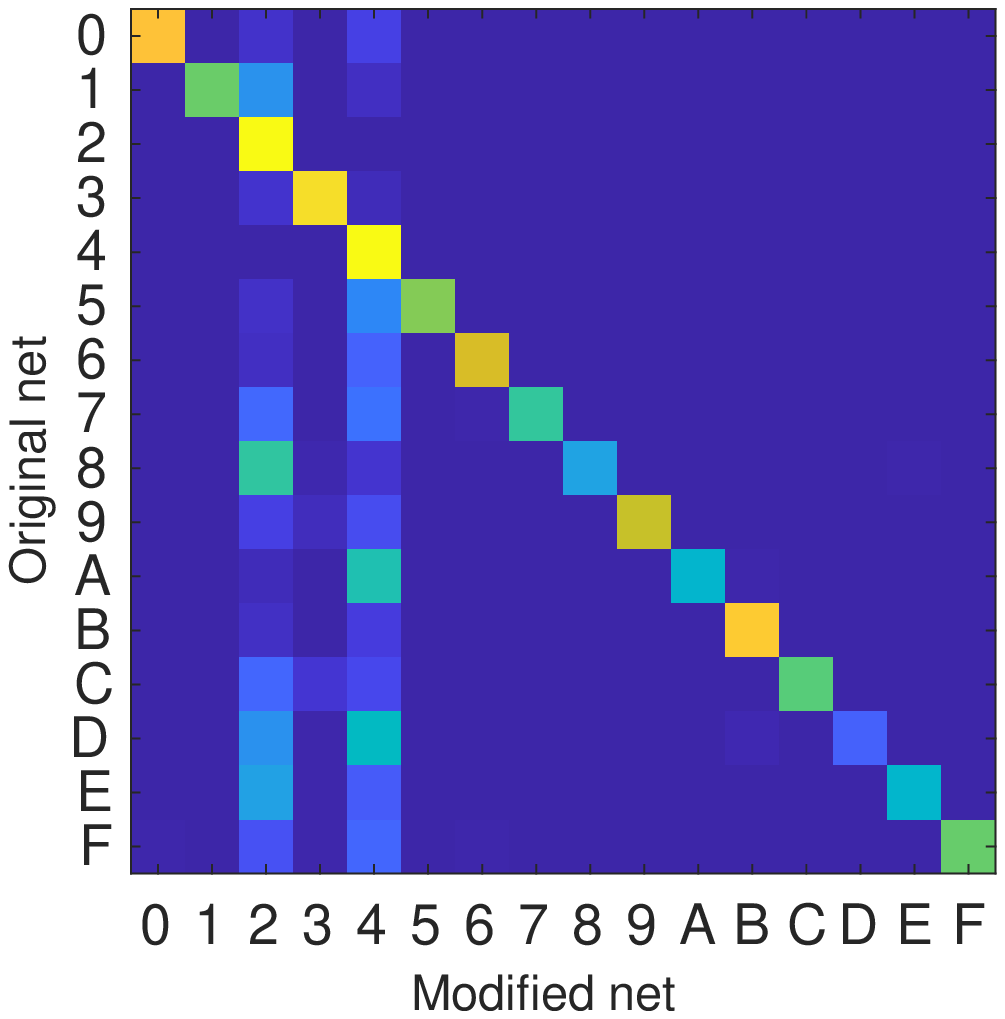}&
    \psfrag{Original net}{}
    \psfrag{Modified net}[][b][.8]{masked net}%{\caja[0.5]{c}{c}{masked \\ net}}
    \includegraphics*[width=.125\textwidth]{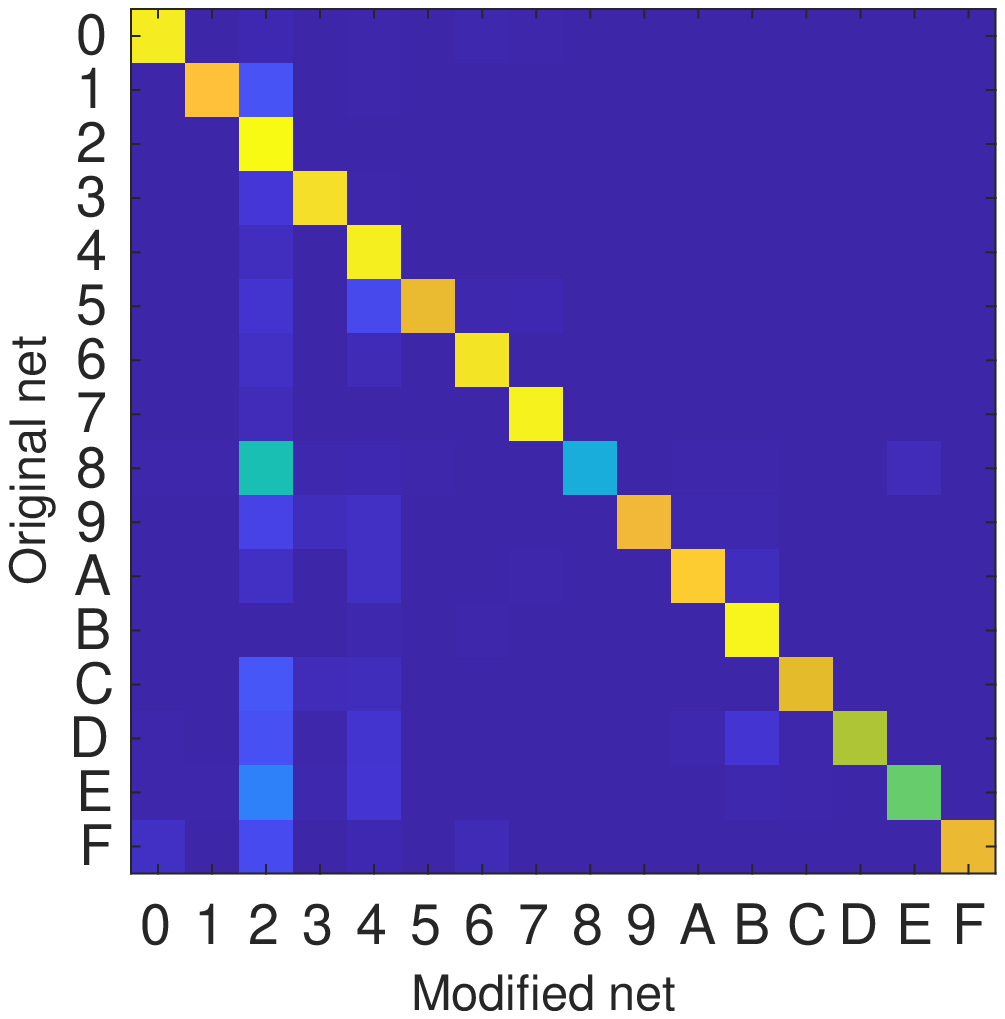}&
    \psfrag{Original net}{}
    \psfrag{Modified net}[][b][.8]{masked net}%{\caja[0.5]{c}{c}{masked \\ net}}
    \includegraphics*[width=.125\textwidth]{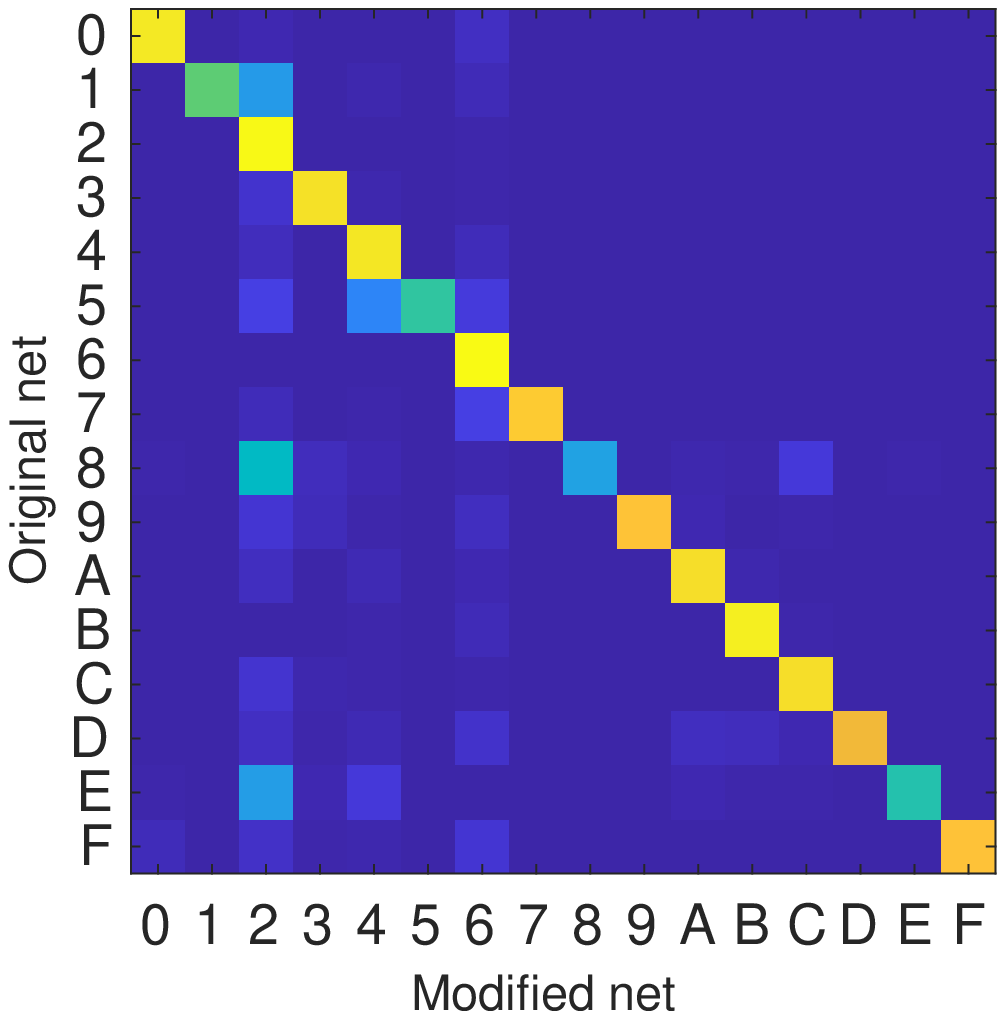}&
    \psfrag{Original net}{}
    \psfrag{Modified net}[][b][.8]{masked net}%{\caja[0.5]{c}{c}{masked \\ net}}
    \includegraphics*[width=.125\textwidth]{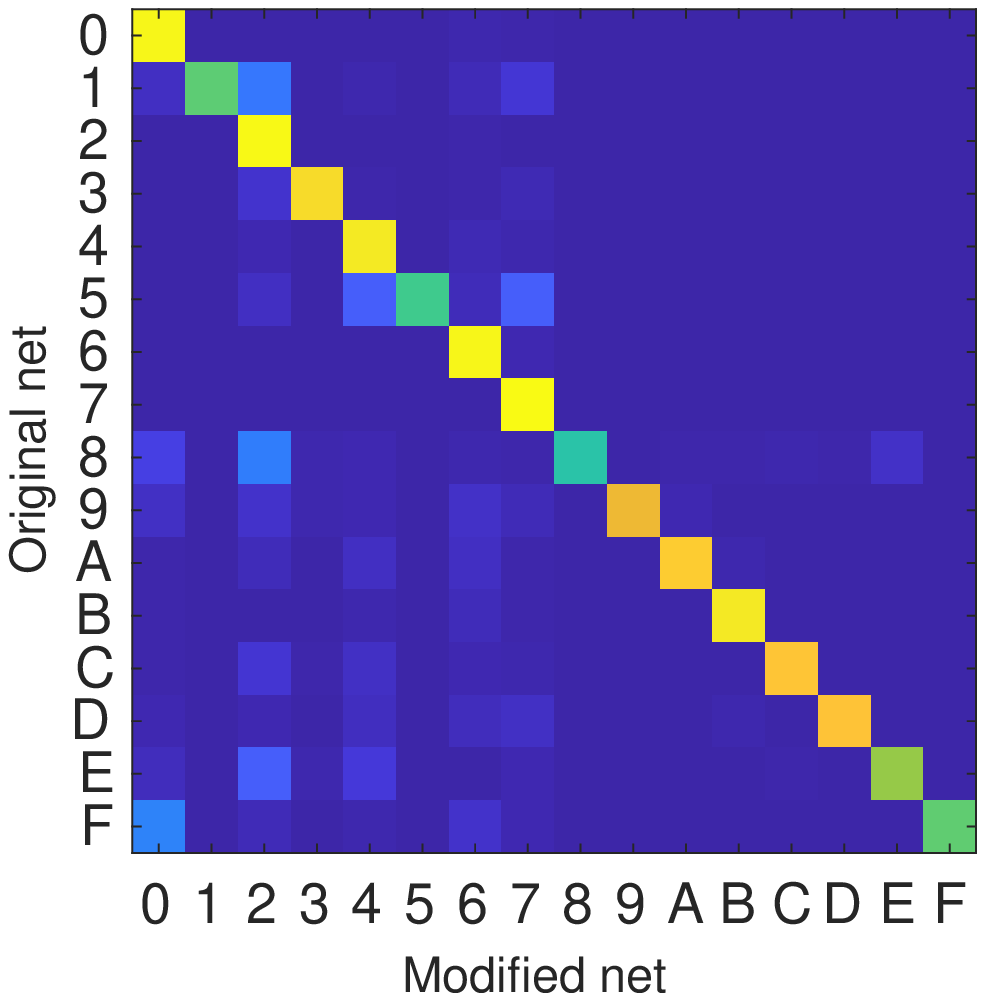}\\ [1ex]
    $k$$=$$8$ & $k$$=$$9$ & $k$$=$$A$ & 	$k$$=$$B$ & $k$$=$$C$ & $k$$=$$D$ & $k$$=$$E$ & $k$$=$$F$ \\
    \psfrag{Original net}[c][t][.8]{original net}
    \psfrag{Modified net}[][b][.8]{masked net}%{\caja[0.5]{c}{c}{masked \\ net}}
    \includegraphics*[width=.125\textwidth]{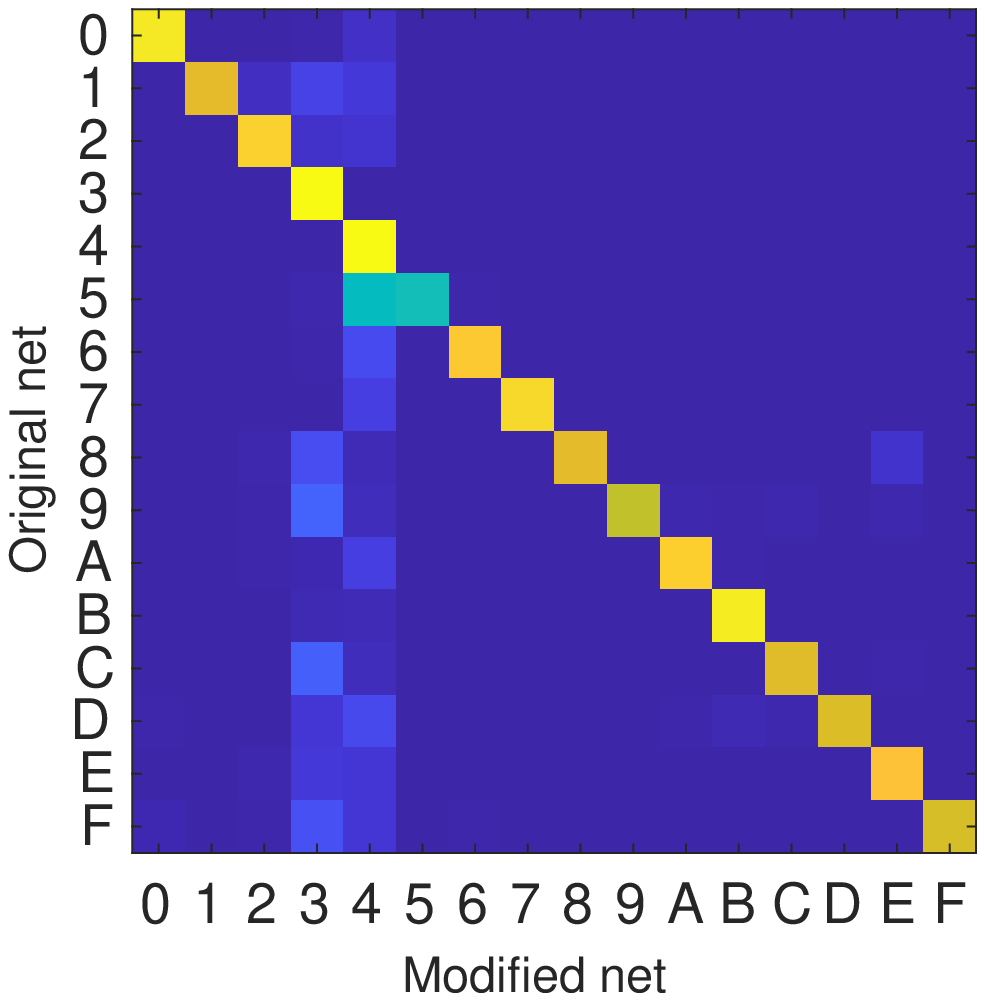}&
    \psfrag{Original net}{}
    \psfrag{Modified net}[][b][.8]{masked net}%{\caja[0.5]{c}{c}{masked \\ net}}
    \includegraphics*[width=.125\textwidth]{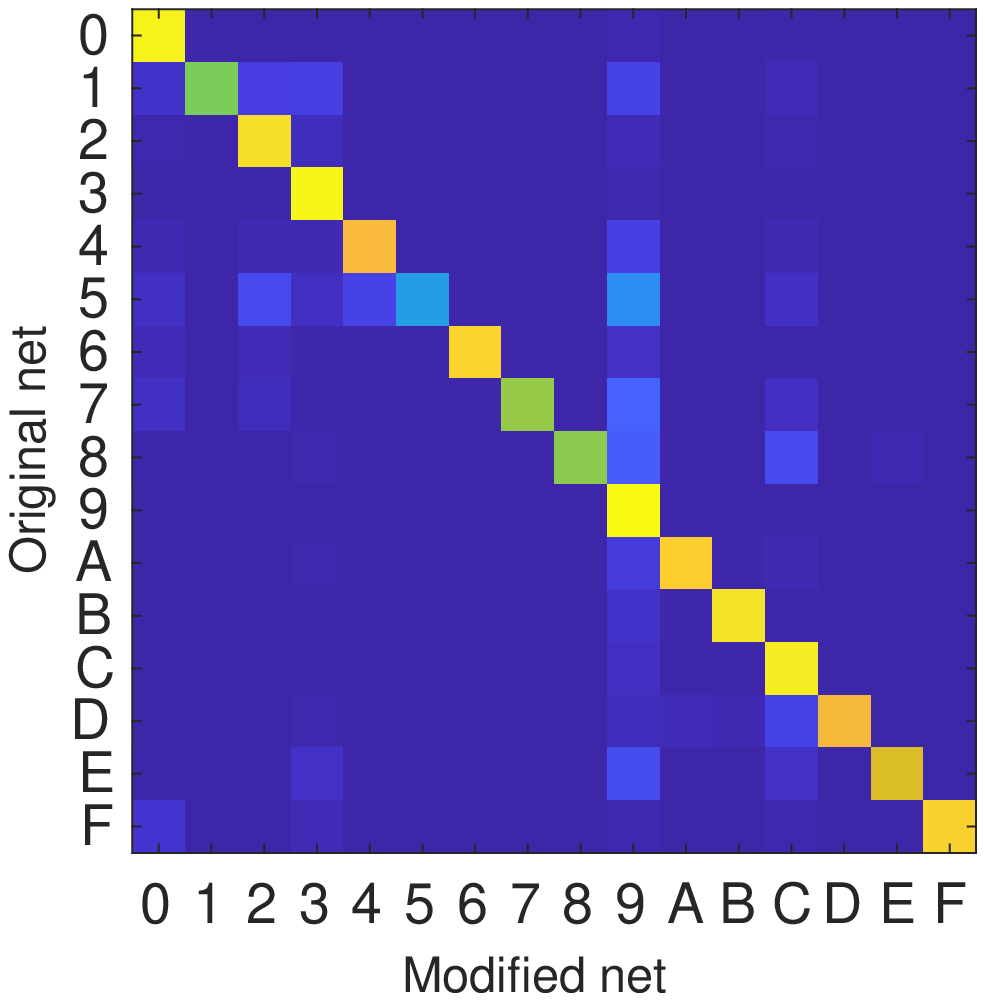}&
    \psfrag{Original net}{}
    \psfrag{Modified net}[][b][.8]{masked net}%{\caja[0.5]{c}{c}{masked \\ net}}
    \includegraphics*[width=.125\textwidth]{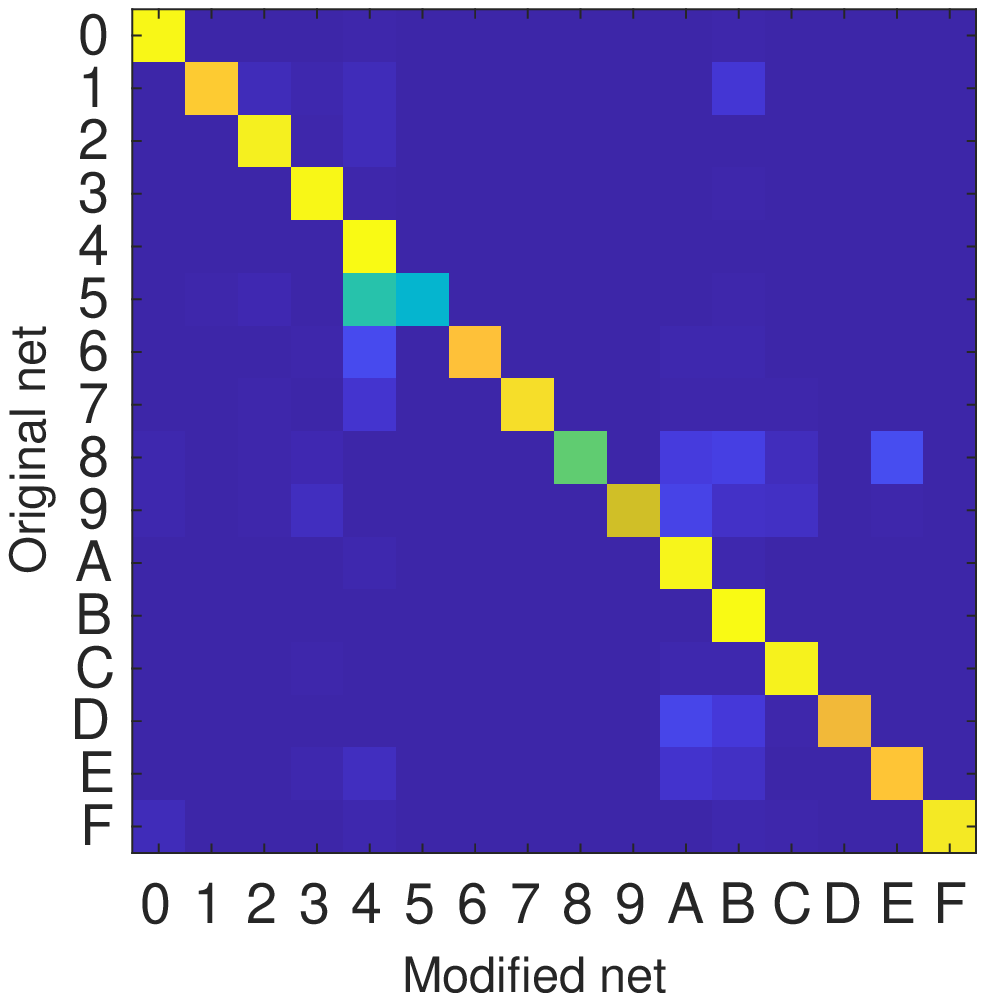}&
    \psfrag{Original net}{}
    \psfrag{Modified net}[][b][.8]{masked net}%{\caja[0.5]{c}{c}{masked \\ net}}
    \includegraphics*[width=.125\textwidth]{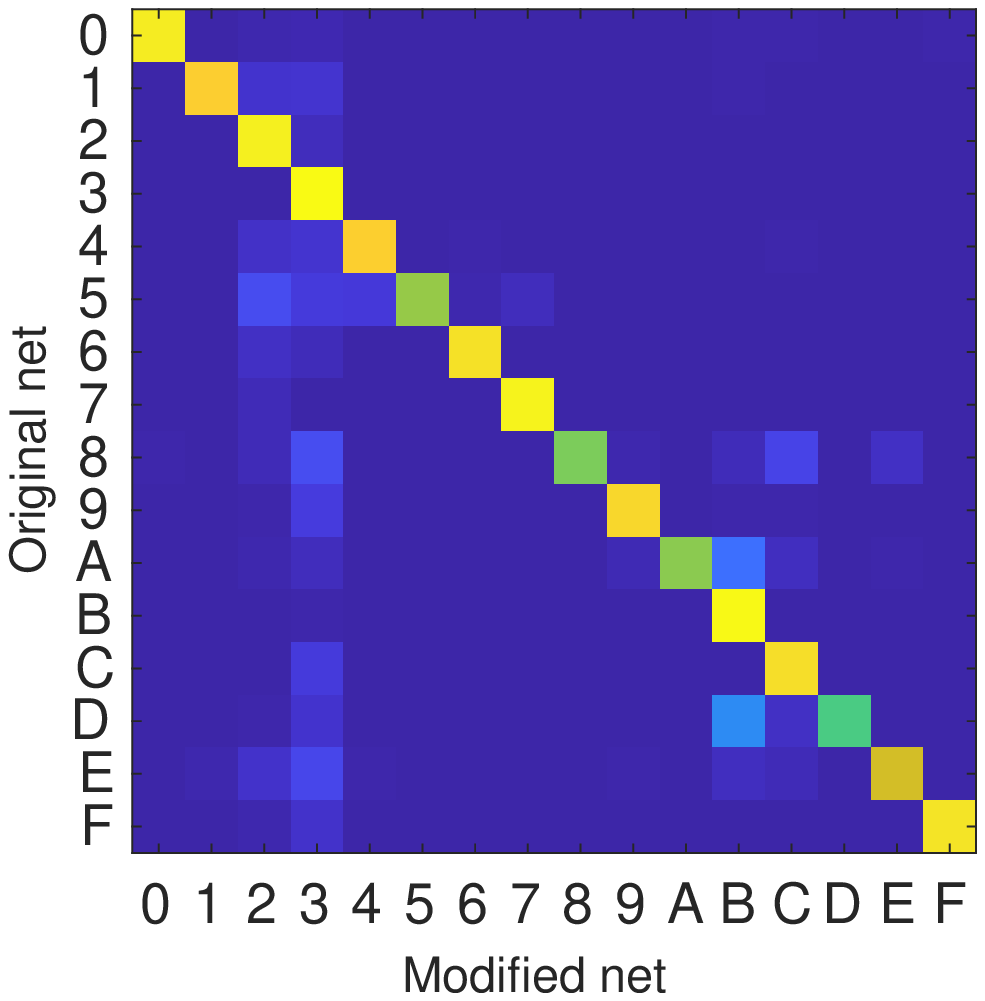}&
    \psfrag{Original net}{}
    \psfrag{Modified net}[][b][.8]{masked net}%{\caja[0.5]{c}{c}{masked \\ net}}
    \includegraphics*[width=.125\textwidth]{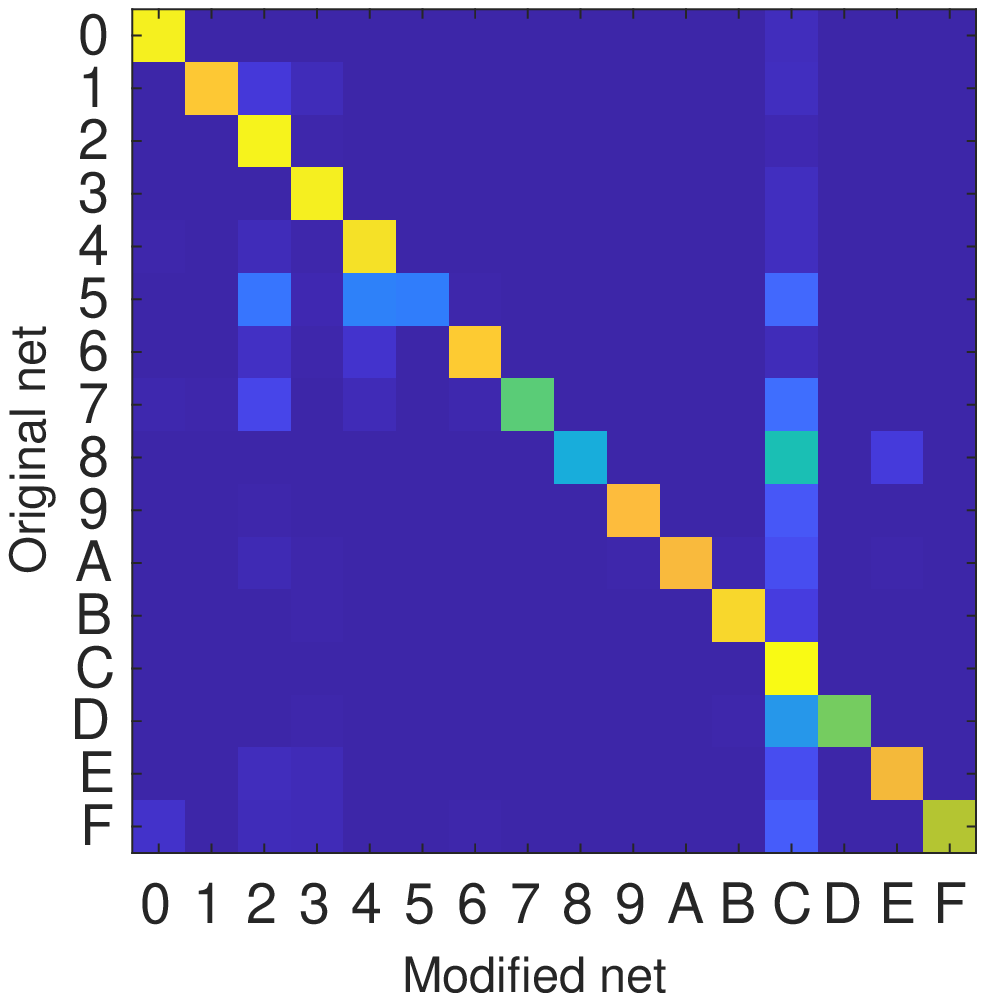}&
    \psfrag{Original net}{}
    \psfrag{Modified net}[][b][.8]{masked net}%{\caja[0.5]{c}{c}{masked \\ net}}
    \includegraphics*[width=.125\textwidth]{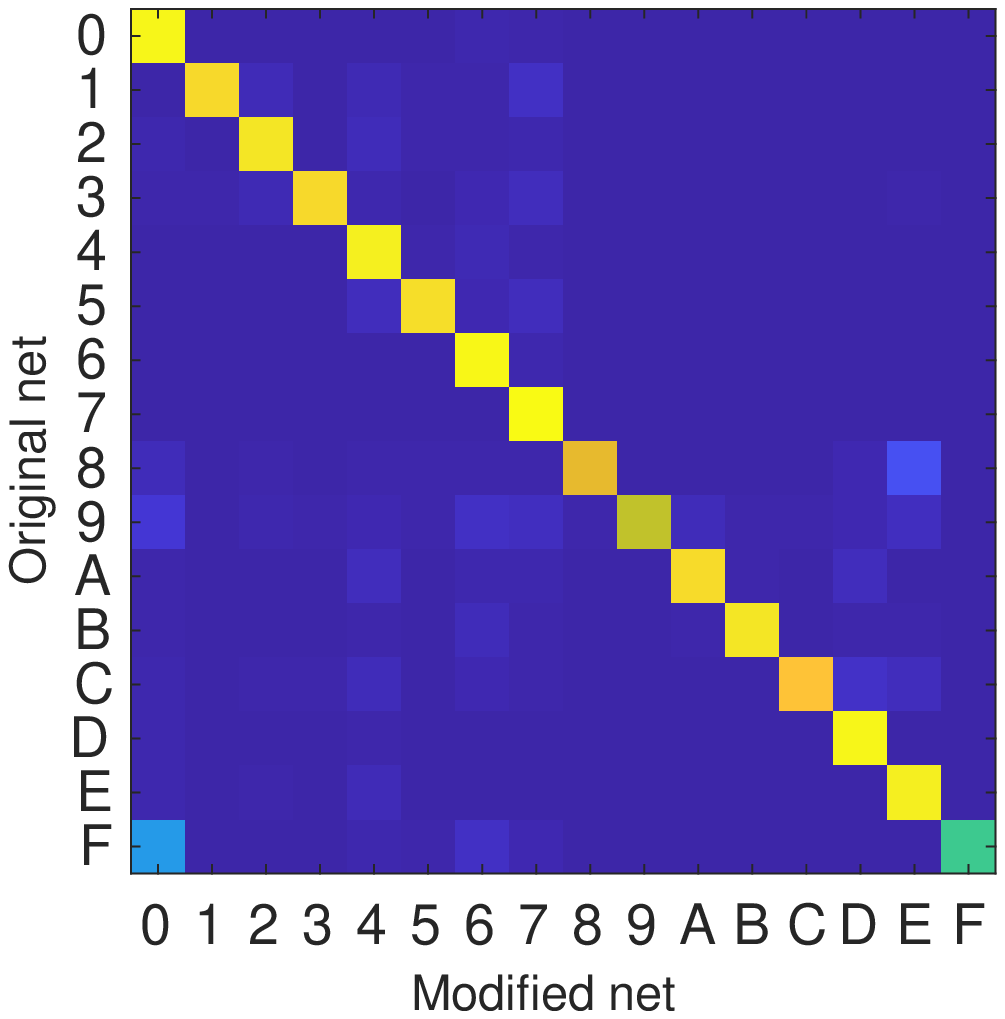}&
    \psfrag{Original net}{}
    \psfrag{Modified net}[][b][.8]{masked net}%{\caja[0.5]{c}{c}{masked \\ net}}
    \includegraphics*[width=.125\textwidth]{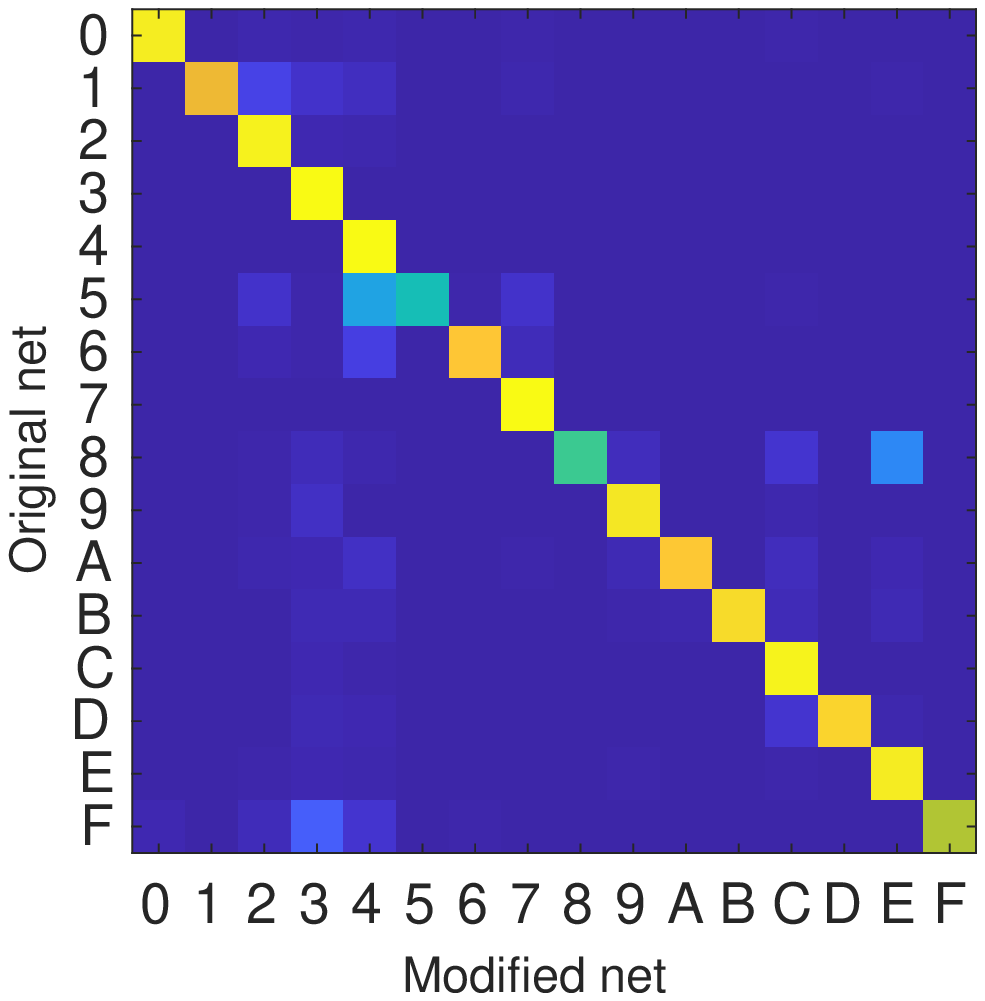}&
    \psfrag{Original net}{}
    \psfrag{Modified net}[][b][.8]{masked net}%{\caja[0.5]{c}{c}{masked \\ net}}
    \includegraphics*[width=.125\textwidth]{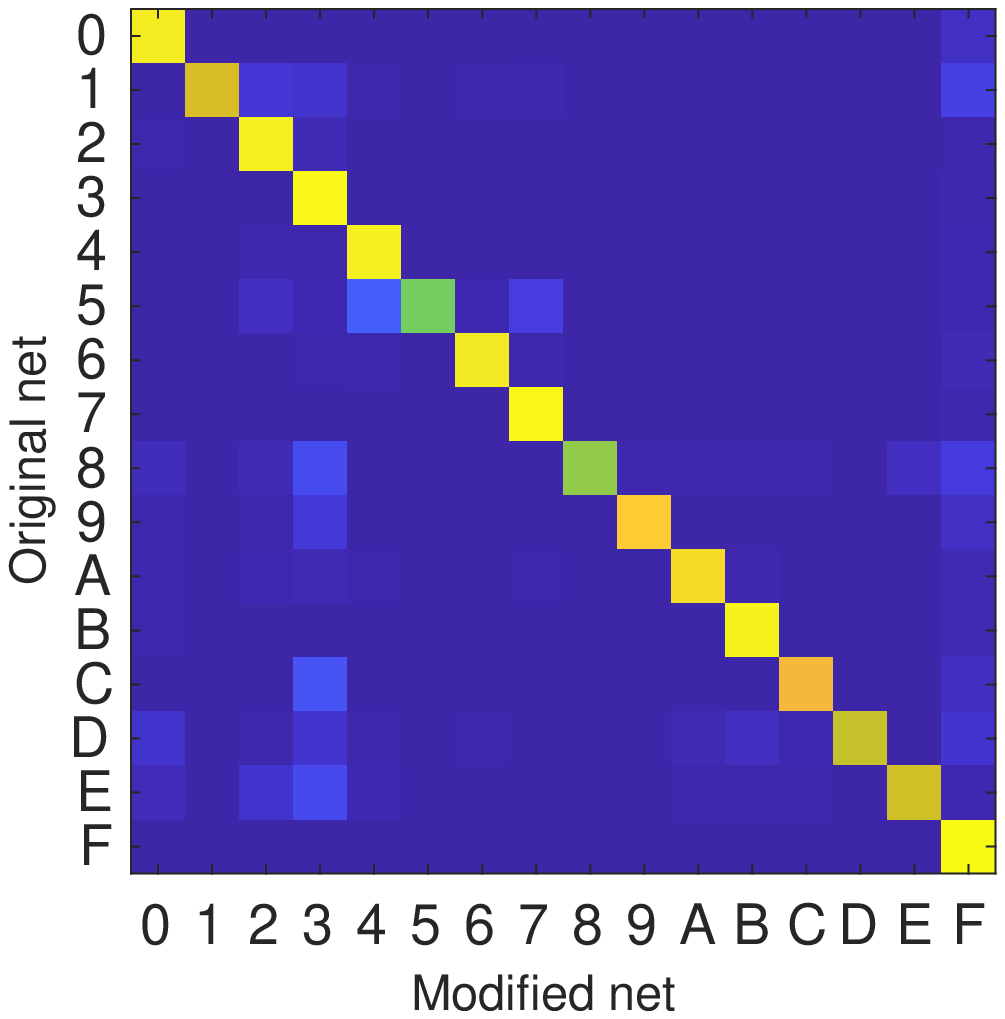}
  \end{tabular}
  \caption{Like fig.~\ref{f:VGG-masks-test-comp} but using an axis-aligned tree.}
  \label{f:VGG-masks-test-cart}
\end{figure}

\clearpage

\bibliographystyle{abbrvnat}
%\bibliography{macp,macp-xref}

\end{document}